# Quantum spectral analysis: frequency in time, with applications to signal and image processing.

**Mario Mastriani**

School of Computing & Information Sciences, Florida International University,
11200 S.W. 8th Street, Miami, FL 33199, USA

mmastria@fiu.edu

ORCID Id: 0000-0002-5627-3935

**Abstract** A *quantum time-dependent spectrum analysis*, or simply, *quantum spectral analysis* (QSA) is presented in this work, and it's based on Schrödinger equation, which is a partial differential equation that describes how the quantum state of a non-relativistic physical system changes with time. In classic world is named *frequency in time* (FIT), which is presented here in opposition and as a complement of traditional spectral analysis frequency-dependent based on Fourier theory. Besides, FIT is a metric, which assesses the impact of the flanks of a signal on its frequency spectrum, which is not taken into account by Fourier theory and even less in real time. Even more, and unlike all derived tools from Fourier Theory (i.e., continuous, discrete, fast, short-time, fractional and quantum Fourier Transform, as well as, Gabor) FIT has the following advantages: a) compact support with excellent energy output treatment, b) low computational cost, $O(N)$ for signals and $O(N^2)$ for images, c) it does not have phase uncertainties (indeterminate phase for magnitude = 0) as Discrete and Fast Fourier Transform (DFT, FFT, respectively), d) among others. In fact, FIT constitutes one side of a triangle (which from now on is closed) and it consists of the original signal in time, spectral analysis based on Fourier Theory and FIT. Thus a toolbox is completed, which it is essential for all applications of Digital Signal Processing (DSP) and Digital Image Processing (DIP); and, even, in the latter, FIT allows edge detection (which is called flank detection in case of signals), denoising, despeckling, compression, and superresolution of still images. Such applications include signals intelligence and imagery intelligence. On the other hand, we will present other DIP tools, which are also derived from the Schrödinger equation. Besides, we discuss several examples for spectral analysis, edge detection, denoising, despeckling, compression and superresolution in a set of experimental results in an important section on Applications and Simulations, respectively. Finally, we finish this work with special section dedicated to Conclusions.

## 1 Introduction

Quantum computation and quantum information is the study of the information processing tasks that can be accomplished using quantum mechanical systems. Like many simple but profound ideas it was a long time before anybody thought of doing information processing using quantum mechanical systems [1].

Quantum computation is the field that investigates the computational power and other properties of computers based on quantum-mechanical principles. An important objective is to find quantum algorithms that are significantly faster than any classical algorithm solving the same problem. The field started in the early 1980s with suggestions for analog quantum computers by Paul Benioff [2] and Richard Feynman [3, 4], and reached more digital ground when in 1985 David Deutsch defined the universal quantum Turing machine [5]. The following years saw only sparse activity, notably the development of the first algorithms by Deutsch and Jozsa [6] and by Simon [7], and the development of quantum complexity theory by Bernstein and Vazirani [8]. However, interest in the field increased tremendously after Peter Shor's very surprising discovery of efficient quantum algorithms (or simulations on a quantum computer) for the problems of integer factorization and discrete logarithms in 1994 [9].

***Quantum Information Processing (QIP)*** - The main concepts related to Quantum Information Processing may be grouped in the next topics: quantum bit (qubit, which is the elemental quantum information unity), Bloch's Sphere (geometric environment for qubit representation), Hilbert's Space (which generalizes the notion of Euclidean space), Schrödinger's Equation (which is a partial differential equation that describes how the quantum state of a physical system changes with time.), Unitary Operators, Quantum Circuits (in quantum information theory, a quantum circuit is a model for quantum computation in which a computation is a sequence of quantum gates, which are reversible transformations on a quantum mechanical analog of an $n$-bit register. This analogous structure is referred to as an $n$-qubit register.), Quantum Gates (in quantum computing and specifically the quantum circuit model of computation, a quantum gate or quantum logic gate is a basic quantum circuit operating on a small number of qubits), and Quantum Algorithms (in quantum computing, a quantum algorithm is an algorithm which runs on a realistic model of quantum computation, the most commonly used model being the quantum circuit model of computation) [1, 10-12].

Nowadays, other concepts complement our knowledge about QIP, they are:

***Quantum Signal Processing (QSP)*** - The main idea is to take a classical signal, sample it, quantify it (for example, between 0 and 255), use a classical-to-quantum interface, give an internal representation to that signal, make a processing to that quantum signal (denoising, compression, among others), measure the result, use a quantum-to-classical interface and subsequently detect the classical outcome signal. Interestingly, and as will be seen later, the quantum image processing has aroused more interest than QSP. In the words of its creator: "many new classes of signal processing algorithms have been developed by emulating the behavior of physical systems. There are also many examples in the signal processing literature in which new classes of algorithms have been developed by artificially imposing physical constraints on implementations that are not inherently subject to these constraints". Therefore, Quantum Signal Processing (QSP) is a signal processing framework [13, 14] that is aimed at developing new or modifying existing signal processing algorithms by borrowing from the principles of quantum mechanics and some of its interesting axioms and constraints. However, in contrast to such fields as quantum computing and quantum information theory, it does not inherently depend on the physics associated with quantum mechanics. Consequently, in developing the QSP framework we are free to impose quantum mechanical constraints that we find useful and to avoid those that are not. This framework provides a unifying conceptual structure for a variety of traditional processing techniques and a precise mathematical setting for developing generalizations and extensions of algorithms, leading to a potentially useful paradigm for signal processing with applications in areas including frame theory, quantization and sampling methods, detection, parameter estimation, covariance shaping, and multiuser wireless communication systems." The truth is that to date, papers on this discipline are less than half a dozen, and its practical use is practically nil. Moreover, although it is an interesting idea, developed so far, does not withstand further comment.

***Quantum Image Processing (QImP)*** - it is a young discipline and it is in training now, however, it's much more developed than QSP. QImP starts in 1997. That year, Vlasov proposed a method of using quantum computation to recognize so-called *orthogonal images* [15]. Five years later, in 2002, Schutzhold described a quantum algorithm that searches specific patterns in binary images [16]. A year later, in October 2003, Beach, Lomont, and Cohen from Cybernet Systems Corporation, (an organization with a close cooperative relationship with the US Defense Department) demonstrated the possibility that quantum algorithms (such as Grover's algorithm) can be used in image processing. In that paper, they describe a method which uses a quantum algorithm to detect the posture of certain targets. Their study implies that quantum image processing may, in future, play a valuable role during wartime [17].

Later, we can found the works of Venegas-Andraca [18], where he proposes quantum image representations such as Qubit Lattice [19, 20]; in fact, this is the first doctoral thesis in the specialty, The history continues with the quantum image representation via the Real Ket [21] of Latorre Sentís, with a special interest in image compression in a quantum context. A new stage begins with the proposal of Le et al. [22], for a flexible representation of quantum images to provide a representation for images on quantum computers in the form of a normalized state which captures information about colors and their corresponding positions in the images. History continues up to date by different authors and their innovative internal representation techniques of the image [23-44].

Very similar to the case of QSP, the idea in back of QImP is to take a classic image (captured by a digital camera or photon counter) and place it in a quantum machine through a classical-to-quantum interface, give some internal representation to the image using the procedures mentioned above, perform processing on it (denoising, compression, among others), measure the results, restore the image through another interface but this time quantum-classical, y ready. The contribution of a quantum machine over a classic machine when it comes to process images it is that the former has much more power of processing. This last advantage can handle images and algorithms of a high computational cost, which would be unmanageable in a classic machine in a practical sense.

The problem of this discipline lies in its genetic, given that QImP is the daughter of Quantum Information Processing and Digital Image Processing, thus, we fall into the old dilemma of teaching, i.e.: to teach Latin to Peter, we should know more about Latin and more about Peter? The answer is simple: we should know very well of both, but the mission becomes impossible. In other words, what is acceptable in Quantum Information Processing, and (at the same time) inadmissible in Digital Image Processing?

The mentioned problem begins with the quantum measurement, then, if after processing the image within the quantum computer, we want to retrieve the result by tomography of quantum states, we will encounter a serious obstacle, this is:

*if we make a tomography of quantum states in Quantum Information Processing (even, this can be extended to any method of quantum measurement after the tomography) with an error of 6% in our knowledge of the state, this constitutes an excellent measure of such state, but on the other hand, and this time from the standpoint of Digital Image Processing [45-48], an error of 6 % in each pixel of the outcome image constitutes a disaster, since this error becomes unmanageable and exaggerated noise. So overwhelming is the aforementioned disaster that the recovered image loses its visual intelligibility, i.e., its look and feel, and morphology, due to the destruction of edges and textures.*

This speaks clearly (and for this purpose, one need only read the papers of QImP cited above) that these works are based on computer simulations in classical machines, exclusively (in most cases in MATLAB® [49]), and do not represent test in a laboratory of Quantum Physics. In fact, if these field trials were held, the result would be the aforementioned. We just have to go to the lab and try with a single pixel of an image, then extrapolate the results to the entire image and therefore the inconvenience will be explicit. On the other hand, today there are obvious difficulties to treat a full image inside a quantum machine, however, there is no difficulty for a single pixel, since that pixel represents a single qubit, and this can be tested in any laboratory in the world, without problems. Therefore, there are no excuses.

Definitely, the problem lies in the hostile relationship between the internal representation of the image (inside quantum machine), the outcome measurement, and the recovery of the image outside of quantum machine. Therefore, the only technique of QImP that survives is QuBoIP [50]. This is because it works with CBS, exclusively, and the quantum measurement does not affect the value of states. However, it is important to clarify that both, i.e., traditional techniques QImP and QuBoIP share a common enemy, and this is the decoherence [1, 50].

***Quantum Cryptography*** - Since most of current classical cryptography is based on the assumption that these two problems are computationally hard, the ability to actually build and use a quantum computer would allow us to break most current classical cryptographic systems, notably the Rivest, Shamir y Adleman (RSA) system [51, 52]. In contrast, a quantum form of cryptography due to Bennett and Brassard [53] is unbreakable even for quantum computers. Besides, Quantum cryptography is the synthesis of quantum mechanics with the art of code-making (cryptography) [54]. The idea was first conceived in an unpublished manuscript written by Stephen Wiesner around 1970 [55]. However, the subject received little attention until its resurrection by a classic paper published by Bennett and Brassard in 1984 [53]. The goal of quantum cryptography is to perform tasks that are impossible or intractable with conventional cryptography. Quantum cryptography makes use of the subtle properties of quantum mechanics such as the quantum no-cloning theorem and the Heisenberg uncertainty principle. Unlike conventional cryptography, whose security is often based on unproven computational assumptions, quantum cryptography has an important advantage in that its security is often based on the laws of physics. Thus far, proposed applications of quantum cryptography include quantum key distribution (abbreviated QKD), quantum bit commitment and quantum coin tossing. These applications have varying degrees of success. The most successful and important application—QKD—has been proven to be unconditionally secure. Moreover, experimental QKD has now been performed over hundreds of kilometers over both standard commercial telecom optical fibers and open-air. In fact, commercial QKD systems are currently available on the market.

On a wider context, quantum cryptography is a branch of quantum information processing, which includes quantum computing, quantum measurements, and quantum teleportation. Among all branches, quantum cryptography is the branch that is closest to real-life applications. Therefore, it can be a concrete avenue for the demonstrations of concepts in quantum information processing. On a more fundamental level, quantum cryptography is deeply related to the foundations of quantum mechanics, particularly the testing of Bell-inequalities and the detection efficiency loophole. On a technological level, quantum cryptography is related to technologies such as single-photon measurements and detection and single-photon sources.

***Quantum Technology*** - Quantum technology is a new field of physics and engineering, which transitions some of the stranger features of quantum mechanics, especially quantum entanglement and most recently quantum tunneling, into practical applications such as quantum computing, quantum cryptography, quantum simulation, quantum metrology, quantum sensing, and quantum imaging.

The field of quantum technology was first outlined in a 1997 book by Gerard J. Milburn [56], which was then followed by a 2003 article by Jonathan P. Dowling and Gerard J. Milburn [57, 58], as well as a 2003 article by David Deutsch [59]. The field of quantum technology has benefited immensely from the influx of new ideas from the field of quantum information processing, particularly quantum computing. Disparate areas of quantum physics, such as quantum optics, atom optics, quantum electronics, and quantum nanomechanical devices, have been unified under the search for a quantum computer and given a common language, that of quantum information theory. In actuality, the most outstanding works in this area belong to Cappelaro's group at MIT [60-63].

***Quantum Fourier Transform (QFT)*** - In quantum computing, the QFT is a linear transformation on quantum bits and is the quantum analogue of the discrete Fourier transform. The QFT is a part of many quantum algorithms, notably Shor's algorithm for factoring and computing the discrete logarithm, the quantum phase estimation algorithm for estimating the eigenvalues of a unitary operator, and algorithms for the hidden subgroup problem.

The QFT can be performed efficiently on a quantum computer, with a particular decomposition into a product of simpler unitary matrices. Using a simple decomposition, the discrete Fourier transform can be implemented as a quantum circuit consisting of only $O(n^2)$ Hadamard gates and controlled phase shift gates, where $n$ is the number of qubits [1]. This can be compared with the classical discrete Fourier transform, which takes $O(2n^2)$ gates (where n is the number of bits), which is exponentially more than $O(n^2)$. However, the quantum Fourier transform acts on a quantum state, whereas the classical Fourier transform acts on a vector, so not every task that uses the classical Fourier transform can take advantage of this exponential speedup. The best QFT algorithms known today require only $O(n \log n)$ gates to achieve an efficient approximation [64].

***Quantum Information Theory (QIT)*** - QIT is motivated by the study of communications channels, but it has a much wider domain of application, and it is a thought-provoking challenge to capture in a verbal nutshell the goals of the field. QIT is fundamentally richer than classical information theory, because quantum mechanics includes so many more elementary classes of static and dynamic resources – not only does it support all the familiar classical types, but there are entirely new classes such as the static resource of entanglement to make life even more interesting than it is classically [1].

QIT deals with four main topics [65]:
- Transmission of classical information over quantum channels.
- The tradeoff between acquisition of information about a quantum state and disturbance of the state.
- Quantifying quantum entanglement.
- Transmission of quantum information over quantum channels.

For which, it involves four components:
- Quantum Entropy [1]: In QIT, we can talk of: quantum relative entropy, the von Neumann entropy, the joint quantum entropy, and the conditional quantum entropy,
- Quantum Channel [66],
- Quantum Cryptography (aforementioned), and
- Quantum Compression [1, 67-71].

Finally, this work is organized as follows: Prolegomenous to QSA are outlined in Section 2, where, we present the follow concepts: continuous, discrete, fast, short-time (including Gabor transform), and fractional Fourier transform, wavelets and multirresolution, smoothing of coefficients in wavelet domain in 1D and 2D, superresolution with special emphasis to still images, and edge detection. In Section 3, we show the proposed new spectral methods with its consequences. Section 4 deal with the applications of the different versions of FIT in signal and image processing, i.e., denoising, compression, edge detection, superresolution, among others. Besides, in this section, we show same numerical and graphic examples of spectral analysis for signals and images, with edge detection, denoising, despeckling, and compression of signals and images too; all in a set of experimental results. By last, Section 5 provides conclusions and a proposal for future works.

## 2 Prolegomenous to QSA

In this section, we discuss the tools, which are needs to understand the full extent to QSA. These tools are:
- Continuous, Discrete (DFT), Fast (FFT), Fractional (FRFT), Short-Time Fourier Transform (STFT), and Gabor transform (GT)
- Wavelets (denoising/despeckling and compression) in general and Haar basis in particular
- Smoothing of coefficients in wavelet domain in 1D (signals) and 2D (images)
- Superresolution in general, and for still images in particular
- Edge detection

At the end of this section it should be clear: what is the ubiQITy of QSA in the context of a much larger, modern and full spectral analysis. On the other hand, this section will allow us to better understand the role QSA as the origin of several tools used today in DSP and DIP. Finally, it will be clear why we say that QSA completes a set of tools to date incomplete.

## 2.1 Transforms from the Fourier Theory

From all existing versions of the Fourier transform [72-79], that is to say, continuous-time, discrete, fractional, short-time (and a particular case of it due to Gabor), and quantum, not forgetting those versions based on the cosine [74-79] - in this section - we discuss the main characteristics of classical versions of Fourier transforms (including Gabor, and excluding cosine versions), their strengths and weaknesses, and as the two do not QITe fill a gap in the field of spectral analysis, and in fact, any other tool it has done to date.

### 2.1.1 Fourier transform

The Fourier transform decomposes a function of time (a *signal*) into the frequencies that make it up, similarly to how a musical chord can be expressed as the amplitude (or loudness) of its constituent notes. The Fourier transform of a function of time itself is a complex-valued function of frequency, whose absolute value represents the amount of that frequency present in the original function, and whose complex argument is the phase offset of the basic sinusoid in that frequency. The Fourier transform is called the *frequency domain representation* of the original signal. The term *Fourier transform* refers to both the frequency domain representation and the mathematical operation that associates the frequency domain representation to a function of time. The Fourier transform is not limited to functions of time, but in order to have a unified language, the domain of the original function is commonly referred to as the *time domain*. For many functions of practical interest one can define an operation that reverses this: the *inverse Fourier transformation*, also called *Fourier synthesis*, of a frequency domain representation combines the contributions of all the different frequencies to recover the original function of time [72].

Linear operations performed in one domain (time or frequency) have corresponding operations in the other domain, which are sometimes easier to perform. The operation of differentiation in the time domain corresponds to multiplication by the frequency, so some differential equations are easier to analyze in the frequency domain. Also, convolution in the time domain corresponds to ordinary multiplication in the frequency domain. Concretely, this means that any linear time-invariant system, such as a filter applied to a signal, can be expressed relatively simply as an operation on frequencies. After performing the desired operations, transformation of the result can be made back to the time domain. Harmonic analysis is the systematic study of the relationship between the frequency and time domains, including the kinds of functions or operations that are "simpler" in one or the other, and has deep connections to almost all areas of modern mathematics [72].

Functions that are localized in the time domain have Fourier transforms that are spread out across the frequency domain and vice versa, a phenomenon known as the uncertainty principle. The critical case for this principle is the Gaussian function, of substantial importance in probability theory and statistics as well as in the study of physical phenomena exhibiting normal distribution (e.g., diffusion). The Fourier transform of a Gaussian function is another Gaussian function. Joseph Fourier introduced the transform in his study of heat transfer, where Gaussian functions appear as solutions of the heat equation [72].

The Fourier transform can be formally defined as an improper Riemann integral, making it an integral transform, although this definition is not suitable for many applications requiring a more sophisticated integration theory. For example, many relatively simple applications use the Dirac delta function, which can be treated formally as if it were a function, but the justification requires a mathematically more sophisticated viewpoint. The Fourier transform can also be generalized to functions of several variables on Euclidean space, sending a function of 3-dimensional space to a function of 3-dimensional momentum (or a function of space and time to a function of 4-momentum). This idea makes the spatial Fourier transform very natural in the study of waves, as well as in quantum mechanics, where it is important to be able to represent wave solutions as functions of either space or momentum and sometimes both. In general, functions to which Fourier methods are applicable are complex-valued, and possibly vector-valued. Still further generalization is possible to functions on groups, which, besides the original Fourier transform on $\mathbb{R}$ or $\mathbb{R}^n$ (viewed as groups under addition), notably includes the discrete-time Fourier transform (DTFT, group = $\mathbb{Z}$), the discrete Fourier transform (DFT, group = $\mathbb{Z}$ mod $N$) and the Fourier series or circular Fourier transform (group = $S^1$, the unit circle ≈ closed finite interval with endpoints identified). The latter is routinely employed to handle periodic functions. The Fast Fourier transform (FFT) is an algorithm for computing the DFT [72].

There are several common conventions (see, [72]) for defining the Fourier transform $\hat{f}$ of an integrable function $f : \mathbb{R} \rightarrow \mathbb{C}$. In this paper, we will use the following definition:

$$\hat{f}(\xi) = \int_{-\infty}^{\infty} f(x) e^{-2\pi i x \xi} dx, \quad \text{for any real number } \xi.$$

When the independent variable $x$ represents *time* (with SI unit of seconds), the transform variable $\xi$ represents frequency (in hertz). Under suitable conditions $f$, is determined by $\hat{f}$ via the inverse transform:

$$f(x) = \int_{-\infty}^{\infty} \hat{f}(\xi) e^{2\pi i x \xi} d\xi, \quad \text{for any real number } x.$$

The statement that $f$ can be reconstructed from $\hat{f}$ is known as the Fourier inversion theorem, and was first introduced in Fourier's *Analytical Theory of Heat*, although what would be considered a proof by modern standards was not given until much later. The functions $f$ and $\hat{f}$ often are referred to as a *Fourier integral pair* or *Fourier transform pair* [72].

For other common conventions and notations, including using the angular frequency $\omega$ instead of the frequency $\xi$, see [72]. The Fourier transform on Euclidean space is treated separately, in which the variable $x$ often represents position and $\xi$ momentum.

***Notes:***

- In practice, the continuous-time version of the cosine transform is not used. Therefore, we will omit in this work.
- The properties of the Fourier transform will see in the next subsection, that is, for Discrete Fourier Transform (DFT), although only the most relevant in terms of this work.
- We will not develop here the two-dimensional version of the Fourier transform, if we instead for subsequent versions, using the property known as separability [74-79].
- Any extension on the Fourier transform shown in [72, 73].

### 2.1.2 DFT

In mathematics, the discrete Fourier transform (DFT) converts a finite list of equally spaced samples of a function into the list of coefficients of a finite combination of complex sinusoids, ordered by their frequencies, that has those same sample values. It can be said to convert the sampled function from its original domain (often time or position along a line) to the frequency domain [74].

The input samples are complex numbers (in practice, usually real numbers), and the output coefficients are complex as well. The frequencies of the output sinusoids are integer multiples of a fundamental frequency, whose corresponding period is the length of the sampling interval. The combination of sinusoids obtained through the DFT is therefore periodic with that same period. The DFT differs from the discrete-time Fourier transform (DTFT) in that its input and output sequences are both finite; it is therefore said to be the Fourier analysis of finite-domain (or periodic) discrete-time functions [74].

The DFT is the most important discrete transform, used to perform Fourier analysis in many practical applications. In digital signal processing, the function is any quantity or signal that varies over time, such as the pressure of a sound wave, a radio signal, or daily temperature readings, sampled over a finite time interval (often defined by a window function). In image processing, the samples can be the values of pixels along a row or column of a raster image. The DFT is also used to efficiently solve partial differential equations, and to perform other operations such as convolutions or multiplying large integers [74].

Since it deals with a finite amount of data, it can be implemented in computers by numerical algorithms or even dedicated hardware. These implementations usually employ efficient fast Fourier transform (FFT) algorithms; so much so that the terms "FFT" and "DFT" are often used interchangeably. Prior to its current usage, the "FFT" initialism may have also been used for the ambiguous term "finite Fourier transform" [74].

The sequence of *N* complex numbers $x_0, x_1, \ldots, x_{N-1}$ is transformed into an *N*-periodic sequence of complex numbers:

$$X_k \triangleq \sum_{n=0}^{N-1} x_n . e^{-2\pi i k n/N}, \quad k \in \mathbb{Z} \text{ (integers)} \tag{1}$$

Each $X_k$ is a complex number that encodes both amplitude and phase of a sinusoidal component of function $x_n$. The sinusoid's frequency is *k* cycles per *N* samples. Its amplitude and phase are:

$$\begin{aligned} |X_k|/N &= \sqrt{\operatorname{Re}(X_k)^2 + \operatorname{Im}(X_k)^2}/N \\ \arg(X_k) &= \operatorname{atan2}\left(\operatorname{Im}(X_k), \operatorname{Re}(X_k)\right) = -i\ln\left(\frac{X_k}{|X_k|}\right), \end{aligned} \tag{2}$$

where atan2 is the two-argument form of the arctan function. Assuming periodicity (see Periodicity in [74]), the customary domain of *k* actually computed is [*0*, *N*-1]. That is always the case when the DFT is implemented via the Fast Fourier transform algorithm. But other common domains are [-*N*/2, *N*/2-1] (*N* even) and [-(*N*-1)/2, (*N*-1)/2] (*N* odd), as when the left and right halves of an FFT output sequence are swapped.

From all its properties, the most important for this paper are the following [74-79]:

***The unitary DFT -*** Another way of looking at the DFT is to note that in the above discussion, the DFT can be expressed as a Vandermonde matrix, introduced by Sylvester in 1867,

$$\mathrm{F} = \begin{bmatrix} \omega_N^{00} & \omega_N^{01} & \cdots & \omega_N^{0(N-1)} \\ \omega_N^{10} & \omega_N^{11} & \cdots & \omega_N^{1(N-1)} \\ \vdots & \vdots & \ddots & \vdots \\ \omega_N^{(N-1)0} & \omega_N^{(N-1)1} & \cdots & \omega_N^{(N-1)(N-1)} \end{bmatrix} \tag{3}$$

where

$$\omega_N = e^{-2\pi i/N} \tag{4}$$

is a primitive *Nth* root of unity called twiddle factor.

While for the case of discrete cosine transform (DCT), we have:

$$\omega_N = cos\left(2\pi/N\right) \tag{5}$$

The inverse transform is then given by the inverse of the above matrix,

$$F^{-1} = \frac{1}{N} F^{*} \tag{6}$$

For unitary normalization, we use a constant like $1/\sqrt{N}$, then, the DFT becomes a unitary transformation, defined by a unitary matrix:

$$\begin{aligned} &U = F/\sqrt{N} \\ &U^{-1} = U^{*} \\ &\left|\det\left(U\right)\right| = 1 \end{aligned} \tag{7}$$

where *det*(•) is the *determinant function of* (•), and $(\bullet)^{*}$ means *conjugate transpose* of (•).

All this shows that the DFT is the product of a matrix by a vector, essentially, as follows:

$$\begin{bmatrix} X_0 \\ X_1 \\ \vdots \\ X_{N-1} \end{bmatrix} = \begin{bmatrix} \omega_N^{00} & \omega_N^{01} & \cdots & \omega_N^{0(N-1)} \\ \omega_N^{10} & \omega_N^{11} & \cdots & \omega_N^{1(N-1)} \\ \vdots & \vdots & \ddots & \vdots \\ \omega_N^{(N-1)0} & \omega_N^{(N-1)1} & \cdots & \omega_N^{(N-1)(N-1)} \end{bmatrix} \begin{bmatrix} x_0 \\ x_1 \\ \vdots \\ x_{N-1} \end{bmatrix} \tag{8}$$

***No Compact Support*** – Based on Eq.(8), we can see that each element $X_k$ of output vector results from multiplying the *kth* row of the matrix by the complete input vector, that is to say, each element $X_k$ of output vector contains every element of the input vector. A direct consequence of this is that DFT scatters the energy to its output, in other words, DFT has a disastrous treatment of the output energy. Therefore, no compact support is equivalent to:

- DFT has a bad treatment of energy at the output
- DFT in not a time-varying transform, but frequency-varying transform

***Time-domain vs frequency-domain measurements*** – As we can see in Fig. 1, thanks to DFT we have a new perspective regarding to signals measurement, i.e., the spectral viewing [78,79].

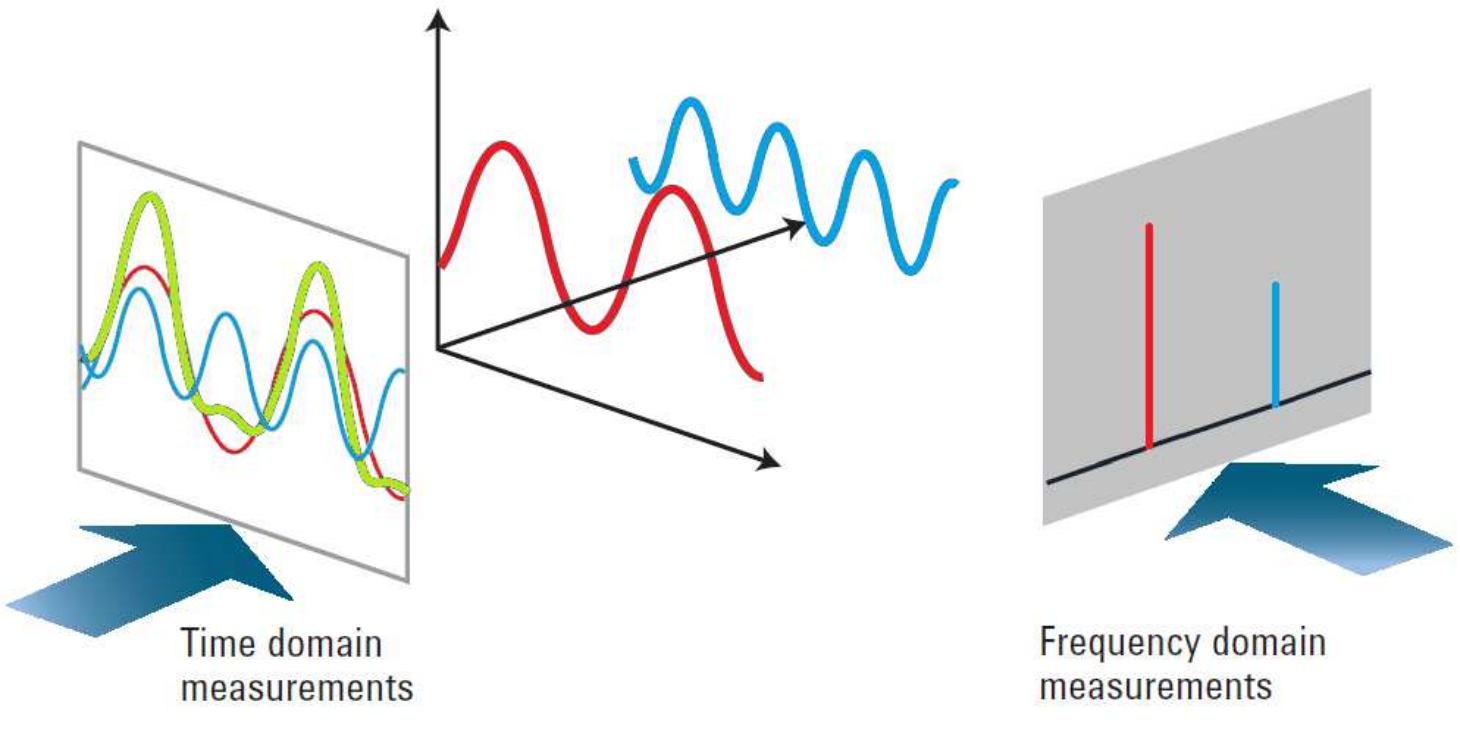

**Fig. 1** Time domain vs frequency domain measurements.

Both point of view allow us to make a nearly complete analysis of the main characteristics of the signal [74-79]. As we can see in Eq.(8), DFT consists in a product between a complex matrix by a real vector (signal).

This gives us a vector output also complex [78, 79]. Therefore, for practical reasons, it is more useful to use the Power Spectral Density (PSD) [74-79].

For example, in MATLAB®, we have

```
f = 10;                          % frequency of sine wave
overSampRate = 30;               % oversampling rate
fs = overSampRate*f;             % sampling frequency
phase = 1/3*pi;                  % desired phase shift in radians
nCyl = 5;                        % to generate five cycles of sine wave
t = 0:1/fs:nCyl*1/f;             % time base
x = sin(2*pi*f*t+phase);         % replace with cos if a cosine wave is desired
L = length(x);                   % length of sine wave
NFFT = 1024;                     % number of considered FFT points
X = fftshift(fft(x,NFFT));       % application of DFT (for practical reasons FFT, see next subsection)
PSD = X.*conj(X)/(NFFT*L);       % Power of each frequency components
fVals = fs*(0:NFFT/2-1)/NFFT;    % deleting negative frequencies
```

If we rewrite Eq.(8), we will have

$$X = \mathrm{F}\,x \qquad \text{(In MATLAB® code, X = fftshift(fft(x,NFFT));)} \tag{9}$$

where $X$ is the output vector (frequency domain), F is the DFT matrix (see Eq.3), and $x$ is the input vector (time domain), then

$$\mathrm{PSD} = \frac{X.\times conj(X)}{NFFT \times L} \qquad \text{(In MATLAB® code, PSD = X.* conj(X)/(NFFT*L);)} \tag{10}$$

In Eq.(10), "$.\times$" means infixed version of Hadamard's product of vectors [80], e.g., if we have two vectors $A = \{a_0,...,a_{N-1}\}$ and $B = \{b_0,...,b_{N-1}\}$, then $A.\times B = \{a_0\, b_0,\, a_1\, b_1,...,\, a_{N-1}\, b_{N-1}\}$, while *conj*(•) means *complex conjugate of* (•), while, "$\times$" means simply product of scalars.

In DSP, some authors work with the square root of PSD [74-77], and others - on the contrary - with the modulus (or absolute value) of $X$ [78, 79], directly.

***Spectral analysis*** **-** When the DFT is used for signal spectral analysis, the $\{x_n\}$ sequence usually represents a finite set of uniformly spaced time-samples of some signal $x(t)$, where $t$ represents time. The conversion from continuous time to samples (discrete-time) changes the underlying Fourier transform of $x(t)$ into a discrete-time Fourier transform (DTFT), which generally entails a type of distortion called aliasing. Choice of an appropriate sample-rate (see *Nyquist rate*) is the key to minimizing that distortion. Similarly, the conversion from a very long (or infinite) sequence to a manageable size entails a type of distortion called *leakage*, which is manifested as a loss of detail (a.k.a. resolution) in the DTFT. Choice of an appropriate sub-sequence length is the primary key to minimizing that effect. When the available data (and time to process it) is more than the amount needed to attain the desired frequency resolution, a standard technique is to perform multiple DFTs, for example to create a spectrogram. If the desired result is a power spectrum and noise or randomness is present in the data, averaging the magnitude components of the multiple DFTs is a useful procedure to reduce the variance of the spectrum (also called a periodogram in this context); two examples of such techniques are the Welch method and the Bartlett method; the general subject of estimating the power spectrum of a noisy signal is called spectral estimation.

A final source of distortion (or perhaps *illusion*) is the DFT itself, because it is just a discrete sampling of the DTFT, which is a function of a continuous frequency domain. That can be mitigated by increasing the resolution of the DFT. That procedure is illustrated at sampling the DTFT [78, 79].

- The procedure is sometimes referred to as *zero-padding*, which is a particular implementation used in conjunction with the fast Fourier transform (FFT) algorithm. The inefficiency of performing multiplications and additions with zero-valued *samples* is more than offset by the inherent efficiency of the FFT.
- As already noted, leakage imposes a limit on the inherent resolution of the DTFT. So there is a practical limit to the benefit that can be obtained from a fine-grained DFT.

Summing-up, we summarize the most important advantages and disadvantages of DFT.

***Disadvantages:***
- DFT fails at the edges. This is the reason why in the JPEG algorithm (employed in image compression) we use the DCT instead of DFT [45-48]. Even, discrete Hartley transform has an outperform to DFT in DSP and DIP [45, 46].
- No compact support, therefore, to arrive at the frequency domain the correspondence element by element between the two domains (time and frequency) is lost, with a lousy treatment of energy.
- As a consequence of not having compact support, it is not at time. In fact, it moves away from the time domain. For this reason, in the last decades, the scientific community has created some palliatives with better performance in both domain simultaneously, i.e., time and frequency, such tools are: STFT, GT, and wavelets.
- DFT has phase uncertainties (indeterminate phase for magnitude = 0) [78,79].
- As it arises from the product of a matrix by a vector, its computational cost is $O(N^2)$ for signals (1D), and $O(N^4)$ for images (2D).

All this would seem to indicate that it is a bad transform, however, they are its advantages that keep it afloat. Then, we describe here only some of them.

***Advantages:***
- As the decisions (relative to filtering and compression) are taken in the spectral domain, the DFT is in its element for both applications. Although as we mentioned before, given its problem with the edges, we use DCT instead DFT.
- It makes the convolutions easier when we use the fast release of DFT, i.e., FFT.
- It is separable (separability property), which is extremely useful when DFT should apply to bi and three-dimensional arrays [45-48].
- Given its internal canonical form (distribution of twiddle factors within the DFT matrix), it allows faster versions of itself, such as FFT.

### 2.1.3 FFT

***Fast Fourier Transform*** - FFT inherits all the disadvantages of the DFT, except the computational complexity of this. In fact, and unlike DFT, the computational cost of FFT is $O(N*log_2N)$ for signals (1D), and $O((N*log_2N)^2)$ for images (2D). For this, it is called fast Fourier transform.

FFT is an algorithm that computes the Discrete Fourier Transform (DFT) of a sequence, or its inverse. Fourier analysis converts a signal from its original domain (often time or space) to the frequency domain and vice versa. An FFT rapidly computes such transformations by factorizing the DFT matrix into a product of sparse (mostly zero) factors [81, 82]. As a result, it manages to reduce the complexity of computing the DFT from $O(N^2)$, which arises if one simply applies the definition of DFT, to $O(N*log_2N)$, where $N$ is the data size. The computational cost for this technique is never greater than the conventional approach and usually significantly less. Further, the computational cost as a function of $n$ is highly continuous, so that linear convolutions of sizes somewhat larger than a power of two.

FFT are widely used for many applications in engineering, science, and mathematics. The basic ideas were popularized in 1965, but some algorithms had been derived as early as 1805 [83]. In 1994 Gilbert Strang described the fast Fourier transform as *the most important numerical algorithm of our lifetime* [84] and it was included in Top 10 Algorithms of 20th Century by the IEEE journal Computing in Science & Engineering [85].

***Overview*** - There are many different FFT algorithms involving a wide range of mathematics, from simple complex-number arithmetic to group theory and number theory; this article gives an overview of the available techniques and some of their general properties, while the specific algorithms are described in subsidiary articles linked below.

The DFT is obtained by decomposing a sequence of values into components of different frequencies. This operation is useful in many fields (see discrete Fourier transform for properties and applications of the transform) but computing it directly from the definition is often too slow to be practical. An FFT is a way to compute the same result more quickly: computing the DFT of $N$ points in the naive way, using the definition, takes $O(N^2)$ arithmetical operations, while an FFT can compute the same DFT in only $O(N*log_2N)$ operations. The difference in speed can be enormous, especially for long data sets where $N$ may be in the thousands or millions. In practice, the computation time can be reduced by several orders of magnitude in such cases, and the improvement is roughly proportional to $N/log_2N$. This huge improvement made the calculation of the DFT practical; FFTs are of great importance to a wide variety of applications, from digital signal processing and solving partial differential equations to algorithms for quick multiplication of large integers.

The best-known FFT algorithms depend upon the factorization of $N$, but there are FFTs with $O(N*log_2N)$ complexity for all $N$, even for prime $N$. Many FFT algorithms only depend on the fact that $e^{-2\pi i/N}$ is an $N$-th primitive root of unity, and thus can be applied to analogous transforms over any finite field, such as number-theoretic transforms. Since the inverse DFT is the same as the DFT, but with the opposite sign in the exponent and a $1/N$ factor, any FFT algorithm can easily be adapted for it.

***History*** - The development of fast algorithms for DFT can be traced to Gauss's unpublished work in 1805 when he needed it to interpolate the orbit of asteroids Pallas and Juno from sample observations [86]. His method was very similar to the one published in 1965 by Cooley and Tukey, who are generally credited for the invention of the modern generic FFT algorithm. While Gauss's work predated even Fourier's results in 1822, he did not analyze the computation time and eventually used other methods to achieve his goal.

Between 1805 and 1965, some versions of FFT were published by other authors. Yates in 1932 published his version called *interaction algorithm*, which provided efficient computation of Hadamard and Walsh transforms [87]. Yates' algorithm is still used in the field of statistical design and analysis of experiments. In 1942, Danielson and Lanczos published their version to compute DFT for x-ray crystallography, a field where calculation of Fourier transforms presented a formidable bottleneck [88]. While many methods in the past had focused on reducing the constant factor for $O(N^2)$ computation by taking advantage of *symmetries*, Danielson and Lanczos realized that one could use the *periodicity* and apply a "doubling trick" to get $O(N*log_2N)$ runtime [89].

Cooley and Tukey published a more general version of FFT in 1965 that is applicable when N is composite and not necessarily a power of 2 [90]. Tukey came up with the idea during a meeting of President Kennedy's Science Advisory Committee where a discussion topic involved detecting nuclear tests by the Soviet Union by setting up sensors to surround the country from outside. To analyze the output of these sensors, a fast Fourier transform algorithm would be needed. Tukey's idea was taken by Richard Garwin and given to Cooley (both worked at IBM's Watson labs) for implementation while hiding the original purpose from him for security reasons. The pair published the paper in a relatively short six months [91]. As Tukey didn't work at IBM, the patentability of the idea was doubted and the algorithm went into the public domain, which, through the computing revolution of the next decade, made FFT one of the indispensable algorithms in digital signal processing.

***Fourier Uncertainty Principle*** - In quantum mechanics, the uncertainty principle [92], also known as Heisenberg's uncertainty principle, is any of a variety of mathematical inequalities asserting a fundamental limit to the precision with which certain pairs of physical properties of a particle, known as complementary variables, such as energy $E$ and time $t$, can be known simultaneously, although $p$ and $x$ are other important, i.e., position and momentum, respectively. They cannot be simultaneously measured with arbitrarily high precision. There is a minimum for the product of the uncertainties of these two measurements.

Introduced first in 1927, by the German physicist Werner Heisenberg, it states that the more precisely the position of some particle is determined, the less precisely its momentum can be known, and vice versa. The formal inequality relating the uncertainty of energy $\Delta E$ and the uncertainty of time $\Delta t$ was derived by Earle Hesse Kennard later that year and by Hermann Weyl in 1928:

$$\Delta E \; \Delta t \geq \hbar/2 \tag{11}$$

where $\hbar$ is the reduced Planck constant, $h / 2\pi$. The energy associated to such system is

$E = \hbar\omega$ (where $\omega = 2\pi f$, being $f$ the frequency, and $\omega$ the angular frequency) (12)

Then, any uncertainty about $\omega$ is transferred to the energy, that is to say,

$$\Delta E = \hbar \, \Delta\omega \tag{13}$$

Replacing Eq.(13) into (11), we will have,

$$\hbar \, \Delta\omega \, \Delta t \geq \hbar/2 \tag{14}$$

Finally, simplifying Eq.(14), we will have,

$$\Delta\omega \, \Delta t \geq 1/2 \tag{15}$$

Eq.(15) say us that a simultaneous decimation in time and frequency is impossible for FFT. Therefore, we must make do with decimate in time or frequency, but not both at once. The four following transforms (STFT, GT, FrFT, and WT) represent a futile effort -to date- to link more closely (individually) each sample in time with its counterpart in frequency in a biunivocal correspondence. That is to say, they are transforms without compact support.

### 2.1.4 STFT

The ***short-time Fourier transform*** (**STFT**), or alternatively *short-term Fourier transform*, is a Fourier-related transform used to determine the sinusoidal frequency and phase content of local sections of a signal as it changes over time [93, 94]. In practice, the procedure for computing STFTs is to divide a longer time signal into shorter segments of equal length and then compute the Fourier transform separately on each shorter segment. This reveals the Fourier spectrum on each shorter segment. One then usually plots the changing spectra as a function of time.

***Continuous-time STFT* -** Simply, in the continuous-time case, the function to be transformed is multiplied by a window function which is nonzero for only a short period of time. The Fourier transform (a one-dimensional function) of the resulting signal is taken as the window is slid along the time axis, resulting in a two-dimensional representation of the signal. Mathematically, this is written as:

$$\mathrm{STFT}\{x(t)\}(\tau,\omega) \equiv X(\tau,\omega) = \int_{-\infty}^{\infty} x(t)\, w(t-\tau)\, e^{-j\omega t}\, dt \tag{16}$$

where $w(t)$ is the window function, commonly a Hann window or Gaussian window bell centered around zero, and $x(t)$ is the signal to be transformed. (Note the difference between $w$ and $\omega$.) $X(\tau,\omega)$ is essentially the Fourier Transform of $x(t)w(t-\tau)$, a complex function representing the phase and magnitude of the signal over time and frequency. Often phase unwrapping is employed along either or both the time axis, $\tau$, and frequency axis, $\omega$, to suppress any jump discontinuity of the phase result of the STFT. The time index $\tau$ is normally considered to be "*slow*" time and usually not expressed in as high resolution as time $t$. The STFT represents an effort to try to fix the spectral components almost instantaneously, i.e., linked wings temporary signal samples.

***Discrete-time STFT* -** In the discrete time case, the data to be transformed could be broken up into chunks or frames (which usually overlap each other, to reduce artifacts at the boundary). Each chunk is Fourier transformed, and the complex result is added to a matrix, which records magnitude and phase for each point in time and frequency. This can be expressed as:

$$\mathrm{STFT}\{x[n]\}(m,\omega) \equiv X(m,\omega) = \sum_{n=-\infty}^{\infty} x[n]w[n-m]e^{-j\omega n} \tag{17}$$

likewise, with signal $x[n]$ and window $w[n]$. In this case, $m$ is discrete and ω is continuous, but in most typical applications the STFT is performed on a computer using the Fast Fourier Transform, so both variables are discrete and quantized.

The magnitude squared of the STFT yields the spectrogram of the function:

$$\mathrm{spectrogram}\{x(t)\}(\tau,\omega) \equiv |X(\tau,\omega)|^2 \tag{18}$$

See also the modified discrete cosine transform (MDCT), which is also a Fourier-related transform that uses overlapping windows.

***Sliding DFT* -** If only a small number of ω are desired, or if the STFT is desired to be evaluated for every shift *m* of the window, then the STFT may be more efficiently evaluated using a sliding DFT algorithm [95].

***Inverse STFT* -** The STFT is invertible, that is, the original signal can be recovered from the transform by the Inverse STFT. The most widely accepted way of inverting the STFT is by using the overlap-add (OLA) method, which also allows for modifications to the STFT complex spectrum. This makes for a versatile signal processing method [96], referred to as the *overlap and add with modifications* method.

***Continuous-time STFT* -** Given the width and definition of the window function $w(t)$, we initially require the area of the window function to be scaled so that

$$\int_{-\infty}^{\infty} w(\tau)\,d\tau = 1. \tag{19}$$

It easily follows that

$$\int_{-\infty}^{\infty} w(t-\tau)\,d\tau = 1 \quad \forall t \tag{20}$$

and

$$x(t) = x(t)\int_{-\infty}^{\infty} w(t-\tau)\,d\tau = \int_{-\infty}^{\infty} x(t)w(t-\tau)\,d\tau. \tag{21}$$

The continuous Fourier Transform is

$$X(\omega) = \int_{-\infty}^{\infty} x(t)e^{-j\omega t}\,dt. \tag{22}$$

Substituting $x(t)$ from above:

$$\begin{aligned} X(\omega) &= \int_{-\infty}^{\infty}\left[\int_{-\infty}^{\infty} x(t)w(t-\tau)\,d\tau\right]e^{-j\omega t}\,dt \\ &= \int_{-\infty}^{\infty}\int_{-\infty}^{\infty} x(t)w(t-\tau)e^{-j\omega t}\,d\tau\,dt. \end{aligned} \tag{23}$$

Swapping order of integration:

$$X(\omega) = \int_{-\infty}^{\infty}\int_{-\infty}^{\infty} x(t)\,w(t-\tau)\,e^{-j\omega t}dt\,d\tau$$
$$= \int_{-\infty}^{\infty}\left[\int_{-\infty}^{\infty} x(t)\,w(t-\tau)\,e^{-j\omega t}dt\right]d\tau \tag{24}$$
$$= \int_{-\infty}^{\infty} X(\tau,\omega)\,d\tau.$$

So the Fourier Transform can be seen as a sort of phase coherent sum of all of the STFTs of $x(t)$. Since the inverse Fourier transform is

$$x(t) = \frac{1}{2\pi}\int_{-\infty}^{\infty} X(\omega)\,e^{+j\omega t}\,d\,\omega, \tag{25}$$

then $x(t)$ can be recovered from $X(\tau,\omega)$ as

$$x(t) = \frac{1}{2\pi}\int_{-\infty}^{\infty}\int_{-\infty}^{\infty} X(\tau,\omega)\,e^{+j\omega t}\,d\,\tau\,d\,\omega, \tag{26}$$

or

$$x(t) = \int_{-\infty}^{\infty}\left[\frac{1}{2\pi}\int_{-\infty}^{\infty} X(\tau,\omega)\,e^{+j\omega t}\,d\,\omega\right]d\,\tau. \tag{27}$$

It can be seen, comparing to above that windowed "grain" or "wavelet" of $x(t)$ is

$$x(t)\,w(t-\tau) = \frac{1}{2\pi}\int_{-\infty}^{\infty} X(\tau,\omega)\,e^{+j\omega t}\,d\,\omega. \tag{28}$$

the inverse Fourier transform of $X(\tau,\omega)$ for $\tau$ fixed.

***Resolution issues*** **-** One of the pitfalls of the STFT is that it has a fixed resolution. The width of the windowing function relates to how the signal is represented—it determines whether there is good frequency resolution (frequency components close together can be separated) or good time resolution (the time at which frequencies change). A wide window gives better frequency resolution but poor time resolution. A narrower window gives good time resolution but poor frequency resolution. These are called narrowband and wideband transforms, respectively.

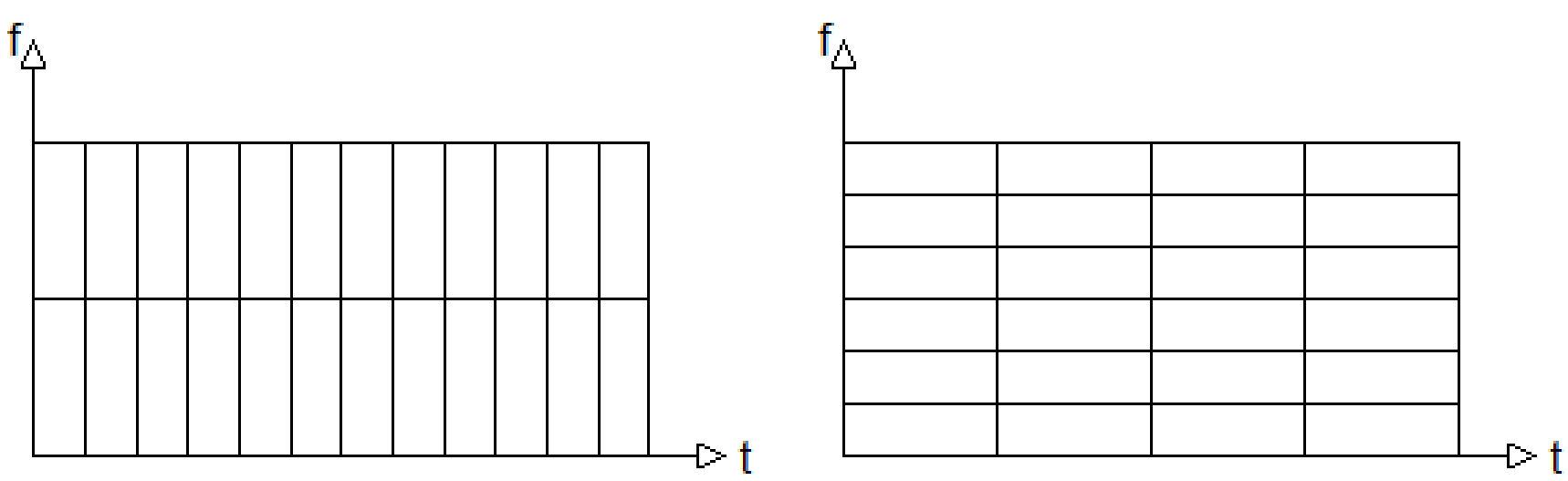

**Fig. 2** Comparison of STFT resolution. Left has better time resolution, and right has better frequency resolution.

This is one of the reasons for the creation of the wavelet transform and multiresolution analysis, which can give good time resolution for high-frequency events and good frequency resolution for low-frequency events, the combination best suited for many real signals. This property is related to the Heisenberg uncertainty principle, but not directly – see Gabor limit for discussion. The product of the standard deviation in time and frequency is limited. The boundary of the uncertainty principle (best simultaneous resolution of both) is reached with a Gaussian window function, as the Gaussian minimizes the Fourier uncertainty principle. This is called the Gabor transform (and with modifications for multiresolution becomes the Morlet wavelet transform). One can consider the STFT for varying window size as a two-dimensional domain (time and frequency), as illustrated in the example below, which can be calculated by varying the window size. However, this is no longer a strictly time–frequency representation – the kernel is not constant over the entire signal.

***Example* -** Using the following sample signal $x(t)$ that is composed of a set of four sinusoidal waveforms joined together in sequence. Each waveform is only composed of one of four frequencies (10, 25, 50, 100 Hz). The definition of $x(t)$ is:

$$x(t)=\begin{cases}\cos(2\pi 10t) & 0\,\text{s}\le t<5\,\text{s}\\ \cos(2\pi 25t) & 5\,\text{s}\le t<10\,\text{s}\\ \cos(2\pi 50t) & 10\,\text{s}\le t<15\,\text{s}\\ \cos(2\pi 100t) & 15\,\text{s}\le t<20\,\text{s}\end{cases} \tag{29}$$

Then it is sampled at 400...Hz. The following spectrograms were produced.

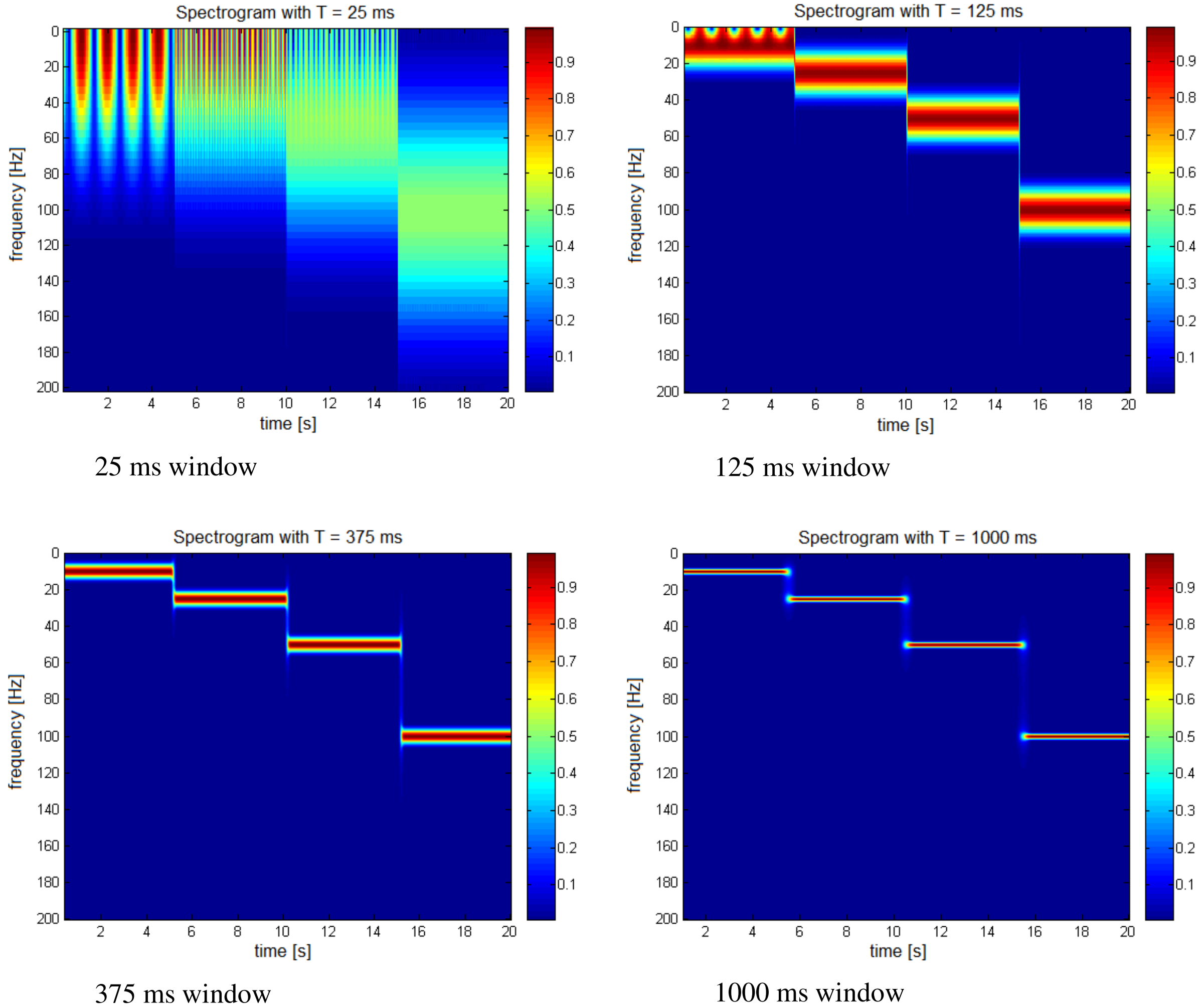

**Fig.3** The 25 ms window allows us to identify a precise time at which the signals change but the precise frequencies are difficult to identify. At the other end of the scale, the 1000 ms window allows the frequencies to be precisely seen but the time between frequency changes is blurred.

***Application* -** STFTs as well as standard Fourier transforms and other tools are frequently used to analyze music. spectrogram can, for example, show frequency on the horizontal axis, with the lowest frequencies at left, and the highest at the right. The height of each bar (augmented by color) represents the amplitude of the frequencies within that band. The depth dimension represents time, where each new bar was a separate distinct transform. Audio engineers use this kind of visual to gain information about an audio sample, for example, to locate the frequencies of specific noises (especially when used with greater frequency resolution) or to find frequencies which may be more or less resonant in the space where the signal was recorded. This information can be used for equalization or tuning other audio effects.

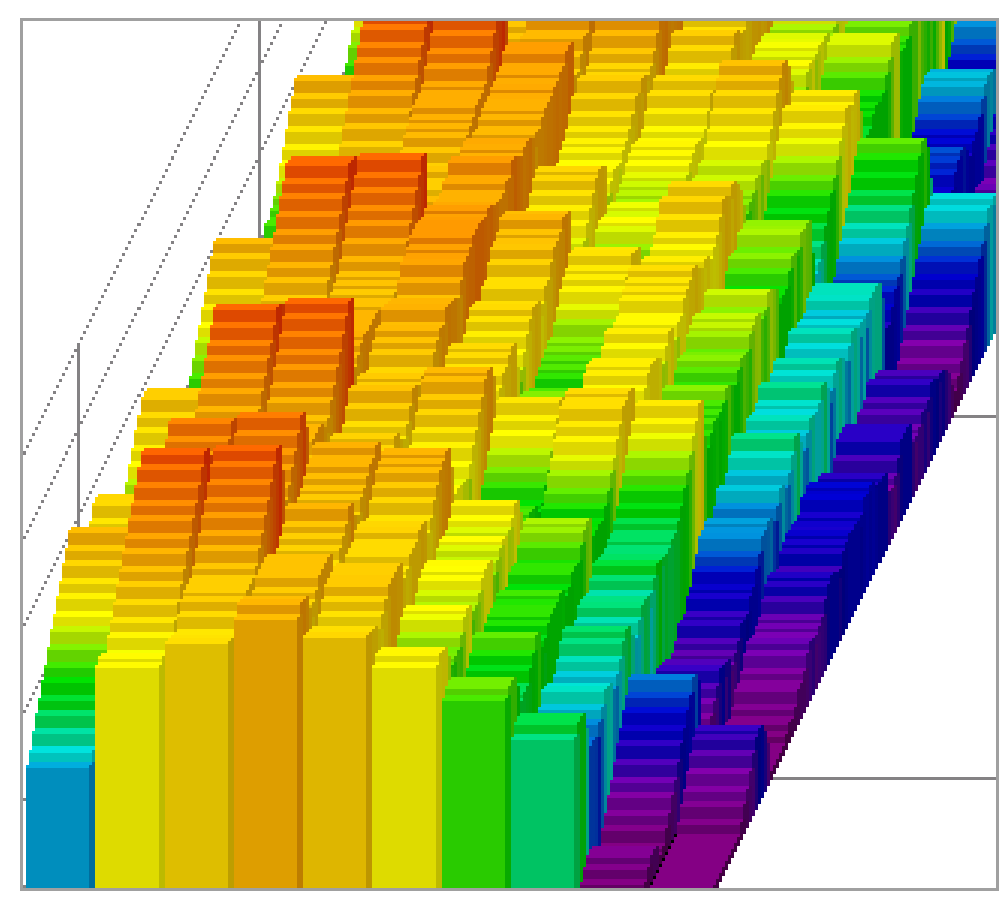

**Fig.4** An STFT being used to analyze an audio signal across time.

The properties of the STFT can be seen in full in [93].

### 2.1.5 GT

The ***Gabor transform***, named after Dennis Gabor, is a special case of the short-time Fourier transform. It is used to determine the sinusoidal frequency and phase content of local sections of a signal as it changes over time. The function to be transformed is first multiplied by a Gaussian function, which can be regarded as a window function, and the resulting function is then transformed with a Fourier transform to derive the time-frequency analysis [94, 97]. The window function means that the signal near the time being analyzed will have higher weight. The Gabor transform of a signal $x(t)$ is defined by this formula:

$$G_x(t,f) = \int_{-\infty}^{\infty} e^{-\pi(\tau-t)^2} e^{-j2\pi f\tau} x(\tau)\, d\tau \qquad (30)$$

The Gaussian function has infinite range and it is impractical for implementation. However, a level of significance can be chosen (for instance 0.00001) for the distribution of the Gaussian function.

$$\begin{cases} e^{-\pi a^2} \geq 0.00001; & |a| \leq 1.9143 \\ e^{-\pi a^2} < 0.00001; & |a| > 1.9143 \end{cases} \qquad (31)$$

Outside these limits of integration ($|a| > 1.9143$) the Gaussian function is small enough to be ignored. Thus the Gabor transform can be satisfactorily approximated as

$$G_x(t,f) = \int_{-1.9143+t}^{1.9143+t} e^{-\pi(\tau-t)^2} e^{-j2\pi f\tau} x(\tau)\, d\tau \qquad (32)$$

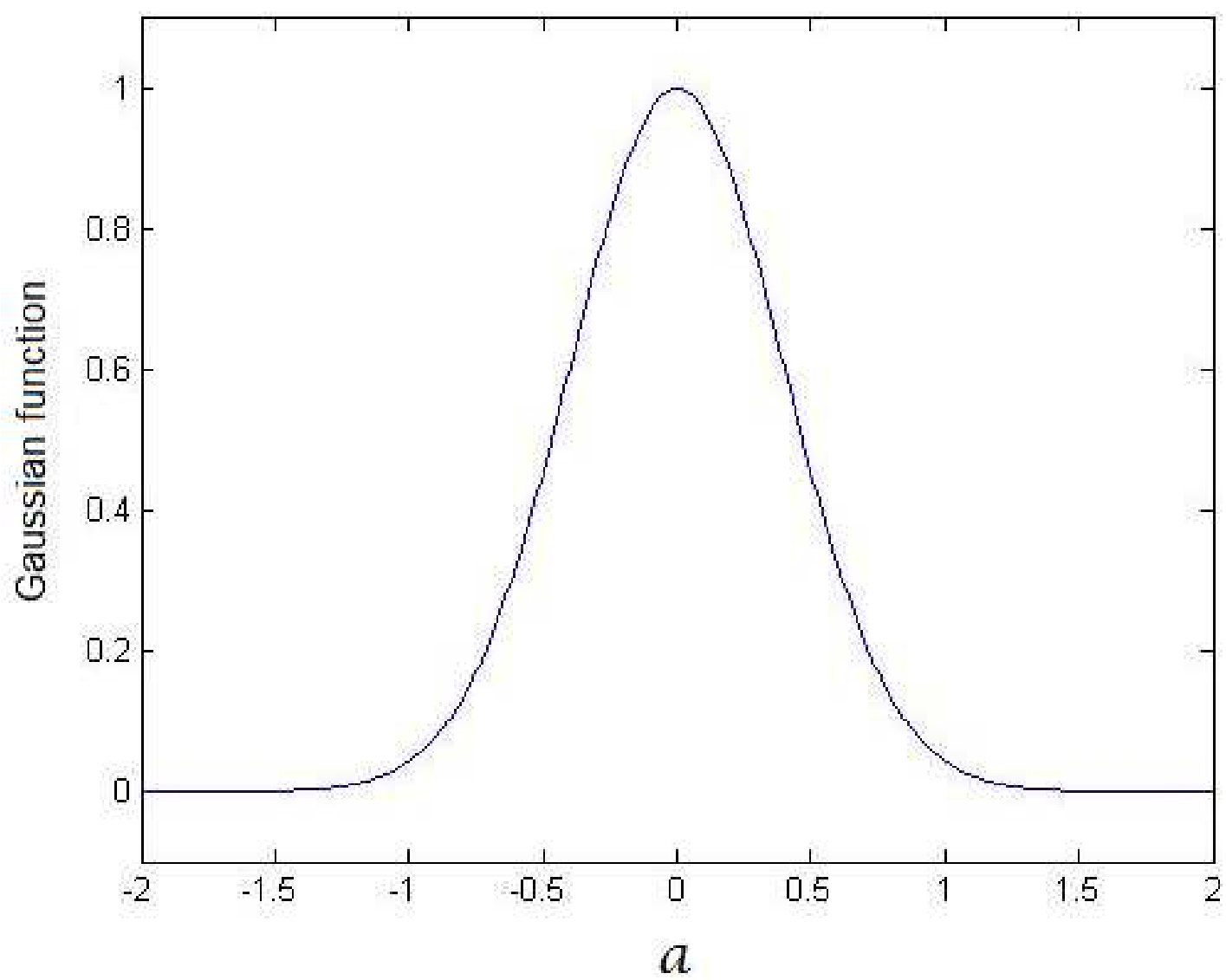

**Fig.5** Magnitude of Gaussian function.

This simplification makes the Gabor transform practical and realizable, and with very important applications, such as: face recognition, texture features and classification, facial expression classification, face reconstruction, fingerprint recognition, facial landmark location, and iris recognition [45-48], etc.

The window function width can also be varied to optimize the time-frequency resolution tradeoff for a particular application by replacing the $-\pi(\tau-t)^2$ with $-\pi\alpha(\tau-t)^2$ for some chosen alpha.

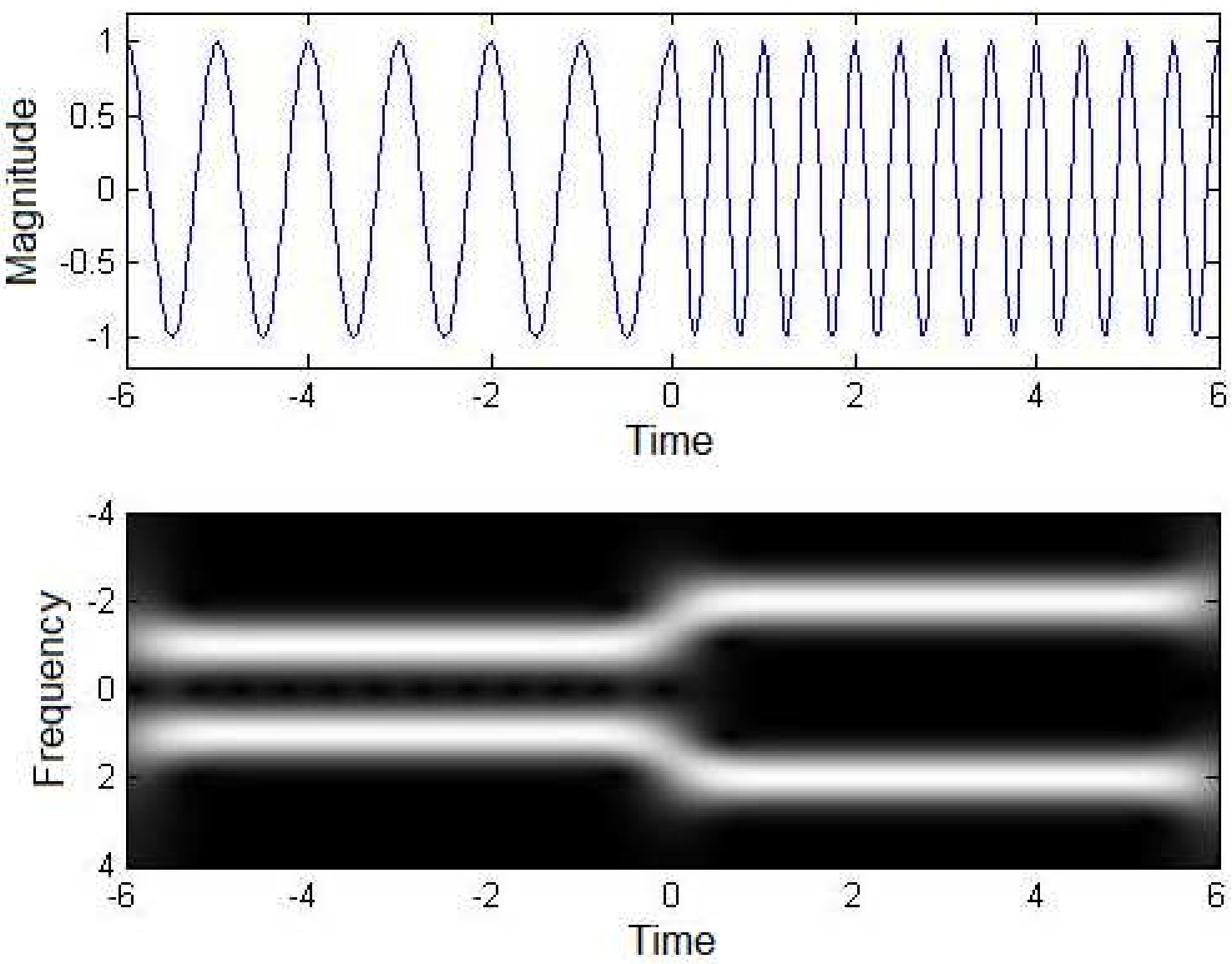

**Fig.6** Time/frequency distribution.

***Application and example*** - The main application of the Gabor transform is used in time frequency analysis. Take the following equation as an example. The input signal has 1 Hz frequency component when $t \leq 0$ and has 2 Hz frequency component when $t > 0$

$$x(t) = \begin{cases} \cos(2\pi t) & \text{for } t \le 0 \\ \cos(4\pi t) & \text{for } t > 0 \end{cases} \tag{33}$$

But if the total bandwidth available is 5 Hz, other frequency bands except $x(t)$ are wasted. Through time frequency analysis by applying the Gabor transform, the available bandwidth can be known and those frequency bands can be used for other applications and bandwidth is saved. The right side picture show the input signal $x(t)$ and the output of the Gabor transform. As was our expectation, the frequency distribution can be separated into two parts. One is $t \le 0$ and the other is $t > 0$. The white part is the frequency band occupied by $x(t)$ and the black part is not used [98].

***Discrete Gabor-transformation* -** A discrete version of Gabor representation

$$y(t) = \sum_{m=-\infty}^{\infty} \sum_{n=-\infty}^{\infty} C_{nm} \cdot g_{nm}(t) \tag{34}$$

with

$$g_{nm}(t) = s(t - m\tau_0).e^{j\Omega nt} \tag{35}$$

can be derived easily by discretizing the Gabor-basis-function in these equations. Hereby the continuous parameter $t$ is replaced by the discrete time $k$. Furthermore the now finite summation limit in Gabor representation has to be considered. In this way, the sampled signal $y(k)$ is split into $M$ time frames of length $N$. According to $\Omega \le 2\pi/\tau_0$, the factor $\Omega$ for critical sampling is $\Omega = 2\pi/N$ .

Similar to the DFT (discrete Fourier transformation) a frequency domain split into $N$ discrete partitions is obtained. An inverse transformation of these $N$ spectral partitions then leads to $N$ values $y(k)$ for the time window, which consists of $N$ sample values. For overall $M$ time windows with $N$ sample values, each signal $y(k)$ contains K = N . M sample values: (the discrete Gabor representation)

$$y(k) = \sum_{m=0}^{M-1} \sum_{n=0}^{N-1} C_{nm} \cdot g_{nm}(k) \tag{36}$$

with

$$g_{nm}(k) = s(k - mN).e^{j\Omega nk} \tag{37}$$

According to the equation above, the $N$x$M$ coefficients $C_{nm}$ correspond to the number of sample values $K$ of the signal.

For over-sampling $\Omega$ is set to $\Omega \le 2\pi/N = 2\pi/N'$ with N' > N, which results in N' > N summation coefficients in the second sum of the discrete Gabor representation. In this case, the number of obtained Gabor-coefficients would be MxN' > K . Hence, more coefficients than sample values are available and therefore a redundant representation would be achieved [98].

The properties of the GT can be seen in full in [97].

### 2.1.6 FRFT

In mathematics, in the area of harmonic analysis, the fractional Fourier transform (FRFT) is a family of linear transformations generalizing the Fourier transform. It can be thought of as the Fourier transform to the $n$-th power, where $n$ need not be an integer — thus, it can transform a function to any *intermediate* domain between time and frequency. Its applications range from filter design and signal analysis to phase retrieval and pattern recognition.

The FRFT can be used to define fractional convolution, correlation, and other operations, and can also be further generalized into the linear canonical transformation (LCT). An early definition of the FRFT was introduced by Condon [99, 100], by solving for the Green's function for phase-space rotations, and also by Namias [101], generalizing work of Wiener [102] on Hermite polynomials. However, it was not widely recognized in signal processing until it was independently reintroduced around 1993 by several groups [103]. Since then, there has been a surge of interest in extending Shannon's sampling theorem [104, 105] for signals which are band-limited in the Fractional Fourier domain.

A completely different meaning for "fractional Fourier transform" was introduced –after that– by Bailey and Swartztrauber [106] as essentially another name for a z-transform, and in particular for the case that corresponds to a discrete Fourier transform shifted by a fractional amount in frequency space (multiplying the input by a linear chirp) and evaluating at a fractional set of frequency points (e.g. considering only a small portion of the spectrum). Such transforms can be evaluated efficiently by Bluestein's FFT algorithm. This terminology has fallen out of use in most of the technical literature, however, in preference to the FRFT.

***Introduction*** **-** As we can see in Subsection 2.1.1, the Fourier transform (or, continuous Fourier transform) $\mathbb{F}$ of a function $f$: $\mathbb{R} \to \mathbb{C}$ is a unitary operator of $L^2$ that maps the function $f$ to its frequential version $\hat{f}$ :

$$\hat{f}(\xi) = \int_{-\infty}^{\infty} f(x) e^{-2\pi i x \xi} dx, \quad \text{for any real number } \xi. \tag{38}$$

And $f$ is determined by $\hat{f}$ via the inverse transform $\mathbb{F}^{-1}$

$$f(x) = \int_{-\infty}^{\infty} \hat{f}(\xi) e^{2\pi i x \xi} d\xi, \quad \text{for any real number } x. \tag{39}$$

Let us study its $n$-th iterated $\mathbb{F}^n$ defined by $\mathbb{F}^n[f] = \mathbb{F}\left[\mathbb{F}^{n-1}[f]\right]$ and $\mathbb{F}^{-n} = \left(\mathbb{F}^{-1}\right)^n$ when $n$ is a non-negative integer, and $\mathbb{F}^0[f] = f$. Their sequence is finite since $\mathbb{F}$ is a 4-periodic automorphism: for every function $f$, $\mathbb{F}^4[f] = f$ .

More precisely, let us introduce the *parity operator* $\mathbb{P}$ that inverts time, $\mathbb{P}[f]: t \to f(-t)$. Then the following properties hold:

$$\begin{aligned} &\mathbb{F}^0 = \mathrm{Id}, \quad \mathbb{F}^1 = \mathbb{F}, \quad \mathbb{F}^2 = \mathbb{P}, \quad \mathbb{F}^4 = \mathrm{Id}, \\ &\mathbb{F}^3 = \mathbb{F}^{-1} = \mathbb{P} \circ \mathbb{F} = \mathbb{F} \circ \mathbb{P}. \end{aligned} \tag{40}$$

The FrFT provides a family of linear transforms that further extends this definition to handle non-integer powers $n = 2\alpha/\pi$ of the FT.

***Definition*** **-** For any real $\alpha$, the $\alpha$-angle fractional Fourier transform of a function $f$ is denoted by $\mathbb{F}_\alpha(u)$ and defined by

$$\mathbb{F}_\alpha[f](u) = \sqrt{1 - i\cot(\alpha)}\, e^{i\pi\cot(\alpha)u^2} \int_{-\infty}^{\infty} e^{-i2\pi\left(\csc(\alpha)ux - \frac{\cot(\alpha)}{2}x^2\right)} f(x)\,\mathrm{dx}. \tag{41}$$

(the square root is defined such that the argument of result lies in the interval $[-\pi/2, \pi/2]$ )

If $\alpha$ is an integer multiple of π, then the cotangent and cosecant functions above diverge. However, this can be handled by taking the limit, and leads to a Dirac delta function in the integrand. More directly, since $\mathbb{F}^2(f)=f(-t)$, $\mathbb{F}_\alpha(f)$ must be simply $f(t)$ or $f(-t)$ for $\alpha$ an even or odd multiple of $\pi$, respectively.

For α = π/2, this becomes precisely the definition of the continuous Fourier transform, and for α = −π/2 it is the definition of the inverse continuous Fourier transform.

The FrFT argument $u$ is neither a spatial one $x$ nor a frequency $\xi$. We will see why it can be interpreted as linear combination of both coordinates $(x, \xi)$. When we want to distinguish the $\alpha$-angular fractional domain, we will let $x_a$ denote the argument of $\mathbb{F}_\alpha$.

***Remark:*** with the angular frequency ω convention instead of the frequency one, the FrFT formula is the Mehler kernel,

$$\mathbb{F}_\alpha(f)(\omega)=\sqrt{\frac{1-i\cot(\alpha)}{2\pi}}\,e^{i\cot(\alpha)\omega^2/2}\int_{-\infty}^{\infty}e^{-i\csc(\alpha)\omega t+\cot(\alpha)t^2/2}f(t)\,\mathrm{d}t. \tag{42}$$

***Interpretation of the fractional Fourier transform*** **-** The usual interpretation of the Fourier transform is as a transformation of a time domain signal into a frequency domain signal. On the other hand, the interpretation of the inverse Fourier transform is as a transformation of a frequency domain signal into a time domain signal. Apparently, fractional Fourier transforms can transform a signal (either in the time domain or frequency domain) into the domain between time and frequency: it is a rotation in the time-frequency domain. This perspective is generalized by the linear canonical transformation, which generalizes the fractional Fourier transform and allows linear transforms of the time-frequency domain other than rotation.

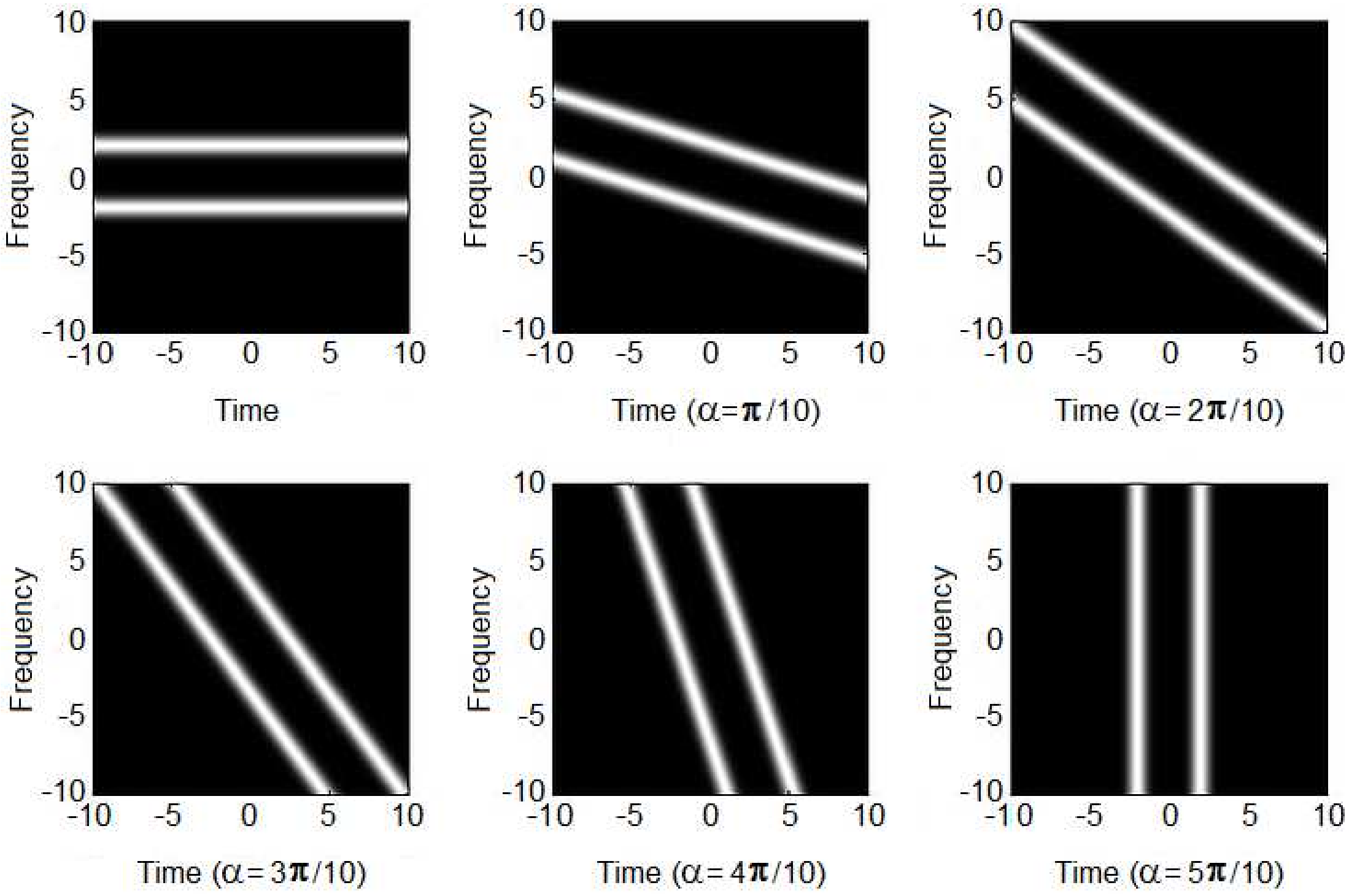

**Fig.7** Time/frequency distribution of fractional Fourier transform.

Take the below figure as an example. If the signal in the time domain is rectangular (as below), it will become a sinc function in the frequency domain. But if we apply the fractional Fourier transform to the rectangular signal, the transformation output will be in the domain between time and frequency.

Actually, fractional Fourier transform is a rotation operation on the time frequency distribution. From the definition above, for $\alpha = 0$, there will be no change after applying fractional Fourier transform, and for $\alpha = \pi/2$, fractional Fourier transform becomes a Fourier transform, which rotates the time frequency distribution with $\pi/2$. For other value of $\alpha$, fractional Fourier transform rotates the time frequency distribution according to $\alpha$. The following figure shows the results of the fractional Fourier transform with different values of $\alpha$.

***Application*** **-** Fractional Fourier transform can be used in time frequency analysis and DSP [107-109]. It is useful to filter noise, but with the condition that it does not overlap with the desired signal in the time frequency domain. Consider the following example. We cannot apply a filter directly to eliminate the noise, but with the help of the fractional Fourier transform, we can rotate the signal (including the desired signal and noise) first. We then apply a specific filter which will allow only the desired signal to pass. Thus the noise will be removed completely. Then we use the fractional Fourier transform again to rotate the signal back and we can get the desired signal.

Fractional Fourier transforms are also used to design optical systems and optimize holographic storage efficiency [110-112].

Thus, using just truncation in the time domain, or equivalently low-pass filters in the frequency domain, one can cut out any convex set in time-frequency space; just using time domain or frequency domain methods without fractional Fourier transforms only allow cutting out rectangles parallel to the axes.

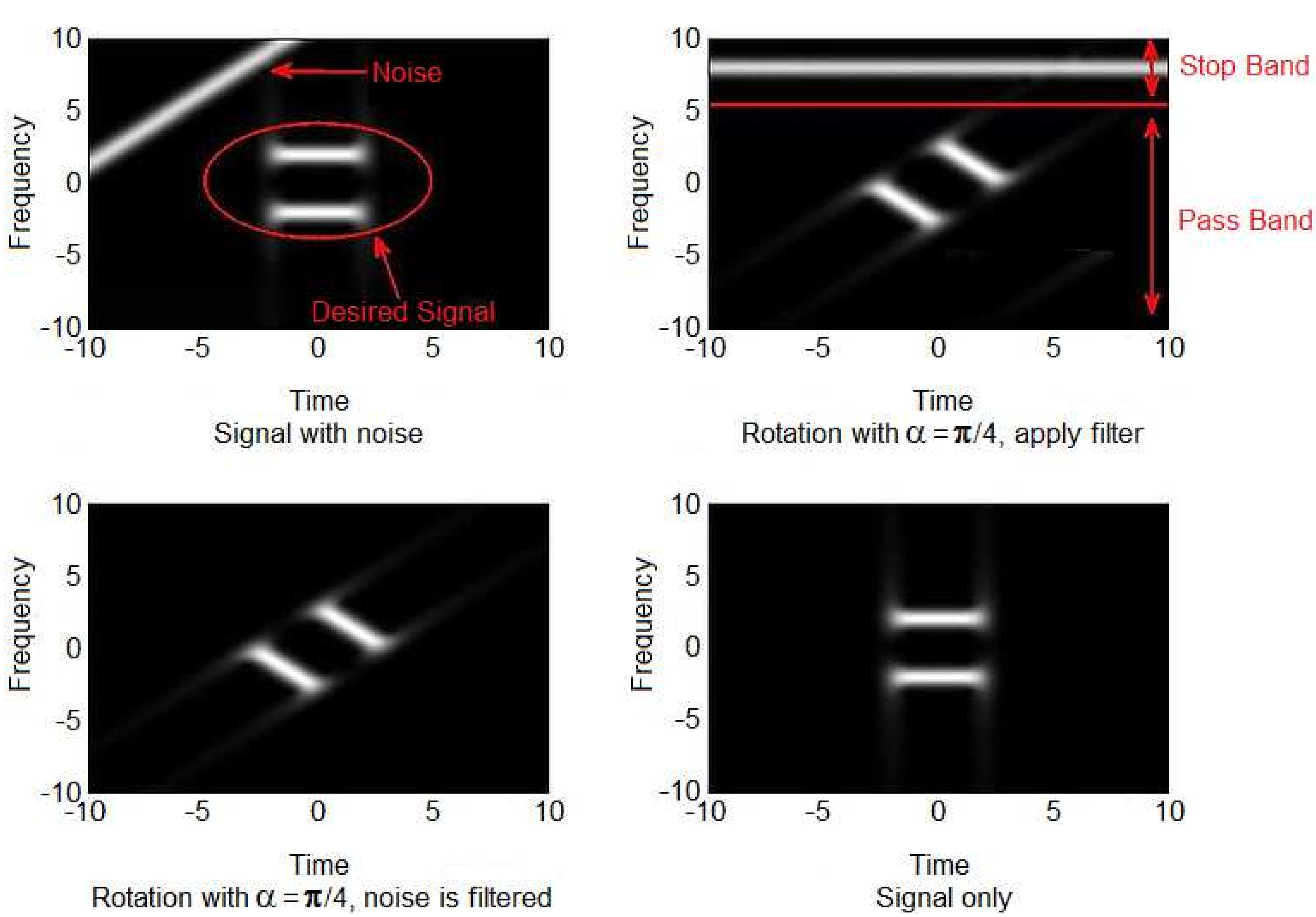

**Fig.8** Fractional Fourier transform in DSP.

The properties of the FrFT can be seen in full in [99].

## 2.2 Wavelets in general, and Haar basis in particular

### 2.2.1 Wavelets in general

In mathematics, a *wavelet series* is a representation of a square-integrable (real- or complex-valued) function by a certain orthonormal series generated by a wavelet. Nowadays, wavelet transformation is one of the most popular candidates of the time-frequency-transformations [113]. This article provides a formal, mathematical definition of an *orthonormal wavelet* and of the *integral wavelet transform*.

***Definition*** **-** A function $\psi \in L^2(\mathbb{R})$ is called an *orthonormal wavelet* if it can be used to define a Hilbert basis [113-116], that is a complete orthonormal system, for the Hilbert space $L^2(\mathbb{R})$ of square integrable functions. The Hilbert basis is constructed as the family of functions $\psi_{jk} : j,k \in \mathbb{Z}$ by means of dyadic translations and dilations of $\psi$,

$$\psi_{jk}(x) = 2^{\frac{j}{2}} \psi\left(2^j x - k\right) \tag{43}$$

for integers $j,k \in \mathbb{Z}$.

If under the standard inner product on $L^2(\mathbb{R})$,

$$\langle f, g \rangle = \int_{-\infty}^{\infty} f(x) \overline{g(x)} dx \tag{44}$$

this family is orthonormal, it is an orthonormal system:

$$\begin{aligned} \langle \psi_{jk}, \psi_{lm} \rangle &= \int_{-\infty}^{\infty} \psi_{jk}(x) \overline{\psi_{lm}(x)} dx \\ &= \delta_{jl}\, \delta_{km} \end{aligned} \tag{45}$$

where $\delta_{jl}$ is the Kronecker delta.

Completeness is satisfied if every function $h \in L^2(\mathbb{R})$ may be expanded in the basis as

$$h(x) = \sum_{j,k=-\infty}^{\infty} c_{jk}\, \psi_{jk}(x) \tag{46}$$

with convergence of the series understood to be convergence in norm. Such a representation of a function *f* is known as a *wavelet series*. This implies that an orthonormal wavelet is self-dual [117, 118].

***Wavelet transform*** **-** The *integral wavelet transform* [119] is the integral transform defined as

$$\left[W_\psi f\right](a,b) = \frac{1}{\sqrt{|a|}} \int_{-\infty}^{\infty} \overline{\psi\left(\frac{x-b}{a}\right)} f(x) dx \tag{47}$$

The *wavelet coefficients* $c_{jk}$ are then given by

$$c_{jk} = \left[W_\psi f\right]\left(2^{-j}, k2^{-j}\right) \tag{48}$$

Here, $a = 2^{-j}$ is called the *binary dilation* or *dyadic dilation*, and $b = k\,2^{-j}$ is the *binary* or *dyadic position*.

***Basic idea*** **-** The fundamental idea of wavelet transforms is that the transformation should allow only changes in time extension, but not shape [120]. This is effected by choosing suitable basis functions that allow for this. Changes in the time extension are expected to conform to the corresponding analysis frequency of the basis function. Based on the uncertainty principle of signal processing,

$$\Delta t\, \Delta\omega \geq \frac{1}{2} \tag{49}$$

Equation (49) results be Eq.(15) of Subsection 2.1.3.

The higher the required resolution in time, the lower the resolution in frequency has to be. The larger the extension of the analysis windows is chosen, the larger is the value of $\Delta t$ .

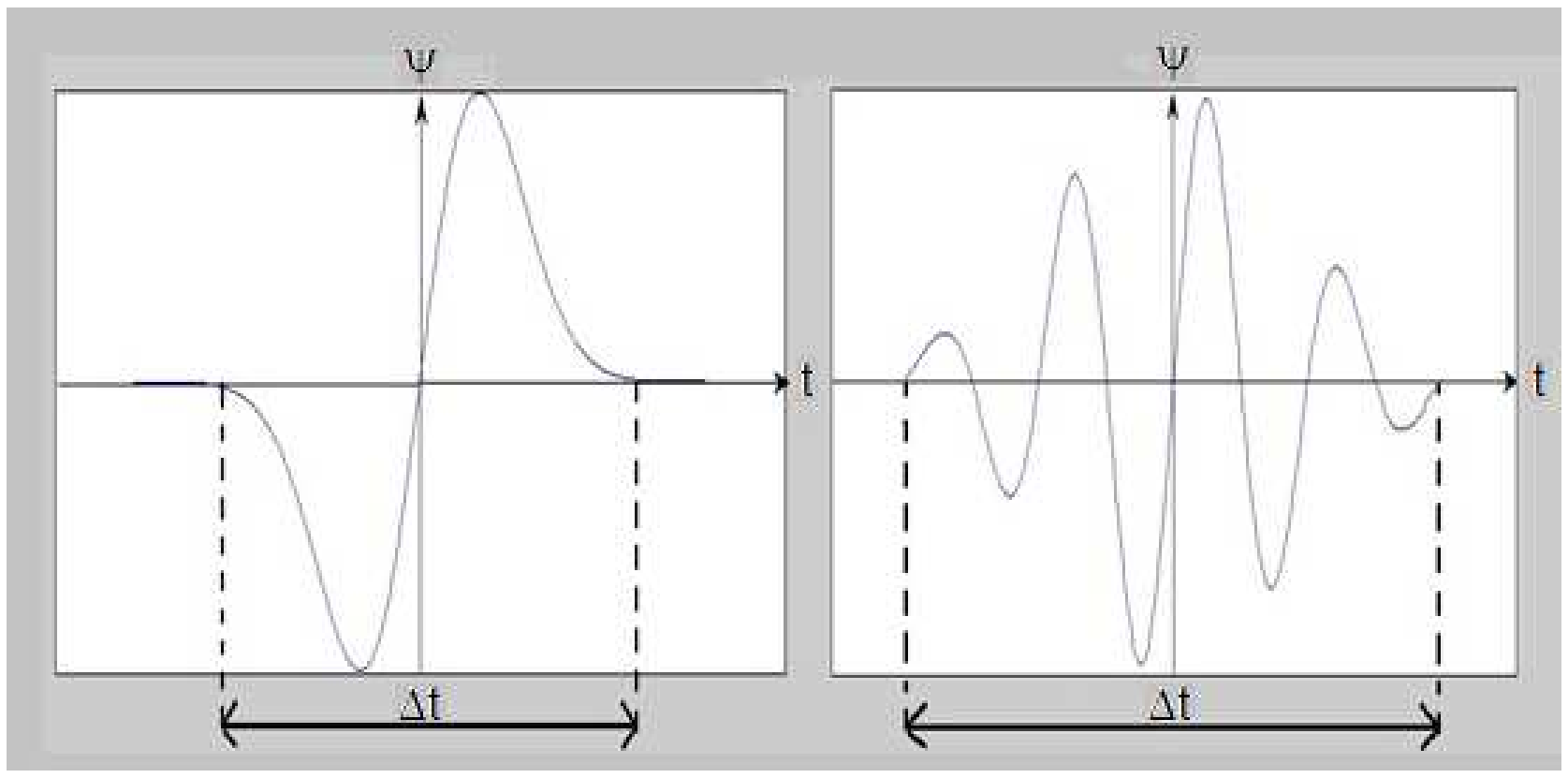

**Fig.9** Wavelet with small and large $\Delta t$, respectively.

When $\Delta t$ is large [113],

1. Bad time resolution
2. Good frequency resolution
3. Low frequency, large scaling factor

When $\Delta t$ is small [113]

1. Good time resolution
2. Bad frequency resolution
3. High frequency, small scaling factor

In other words, the basis function $\psi$ can be regarded as an impulse response of a system with which the function $x(t)$ has been filtered [121, 122]. The transformed signal provides information about the time and the frequency. Therefore, wavelet-transformation contains information similar to the short-time-Fourier-transformation, but with additional special properties of the wavelets, which show up at the resolution in time at higher analysis frequencies of the basis function. The difference in time resolution at ascending frequencies for the Fourier transform and the wavelet transform is shown below [123, 124].

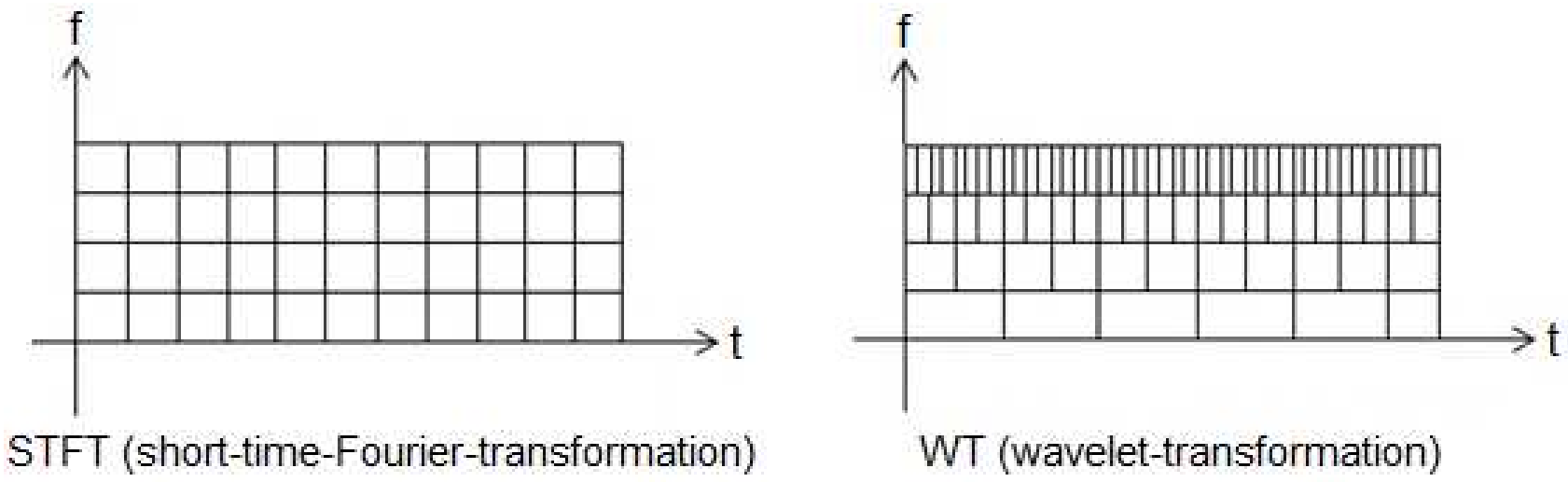

**Fig.10** STFT vs WT, decimation in time and frequency simultaneously.

This shows that wavelet transformation is good in time resolution of high frequencies, while for slowly varying functions, the frequency resolution is remarkable.

***Another example*** - The analysis of three superposed sinusoidal signals

$y(t) = \sin(2\pi f_0 t) + \sin(4\pi f_0 t) + \sin(8\pi f_0 t)$ with STFT and wavelet-transformation.

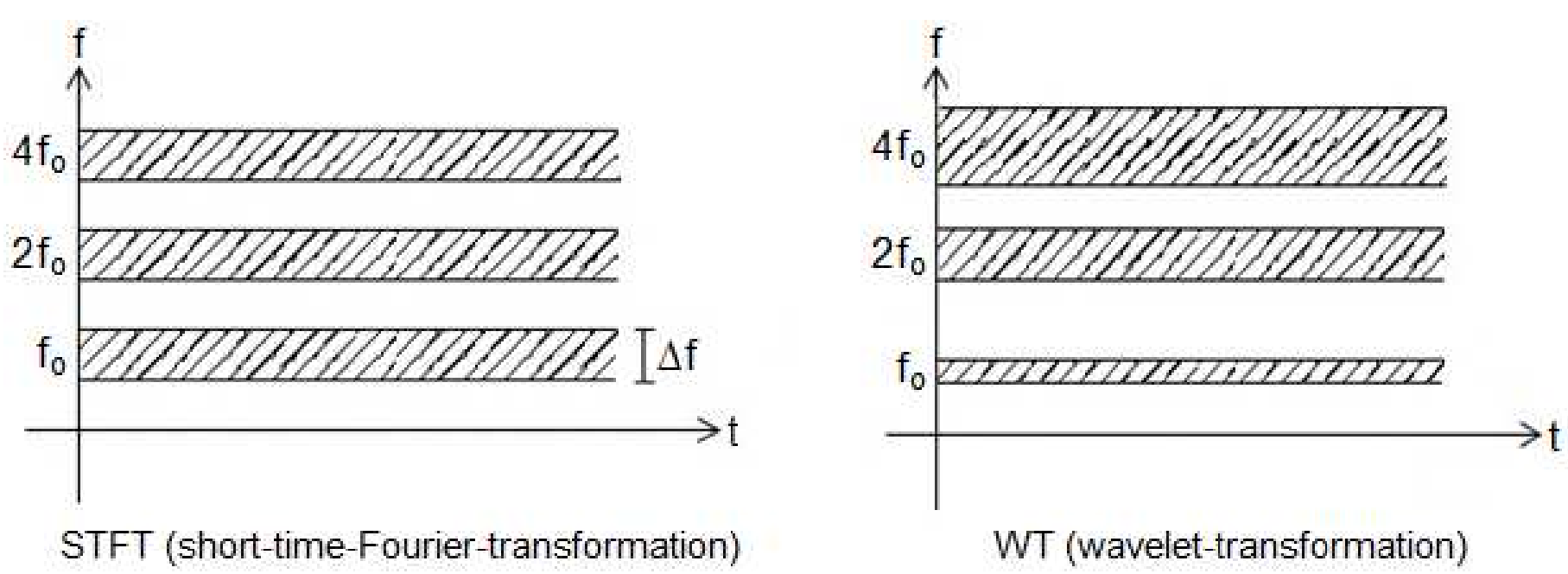

**Fig.11** STFT vs WT, decimation in frequency.

The properties of the WT can be seen in full in [113].

### 2.2.2 Haar basis in particular

We will use Haar basis (without loss of generality regarding another wavelet basis) first for denoising and compression, and then –in Section 2.4– for superresolution and additional compression (generally called supercompression). Besides, although we begin with two-dimensional discrete wavelet transform for images, immediately after, we will explain the one-dimensional version of the discrete wavelet transform for signals, as well as, the corresponding Haar basis in each case.

Finally, it is important to mention in here that inside wavelets context, denoising and compression are the same, because the procedure is the same, and all the denoising wavelet techniques inevitably reduce the details of the treated image (or signal). Such details reduction directly affects the volume of the final image (or signal) file.

***Denoising/compression of images*** - An image is affected by noise in its acquisition and processing. The denoising techniques are used to remove the additive noise while retaining as much as possible the important image features. In the recent years there has been an important amount of research on wavelet thresholding and threshold selection for images denoising [125-170], because wavelet provides an appropriate basis for separating noisy signal from the image signal. The motivation is that as the wavelet transform is good at energy compaction, the small coefficients are more likely due to noise and large coefficient due to important signal features [125-127]. These small coefficients can be thresholded without affecting the significant features of the image.

In fact, the thresholding technique is the last approach based on wavelet theory to provide an enhanced approach for eliminating such noise source [128, 129] and ensure better image quality [130, 131]. Thresholding is a simple non-linear technique, which operates on one wavelet coefficient at a time. In its basic form, each coefficient is thresholded by comparing against threshold, if the coefficient is smaller than threshold, set to zero; otherwise it is kept or modified. Replacing the small noisy coefficients by zero and inverse wavelet transform on the result may lead to reconstruction with the essential signal characteristics and with less noise. Since the work of Donoho & Johnstone [125-127], there has been much research on finding thresholds, however few are specifically designed for images [138-170].

***Two-dimensional Discrete Wavelet Transform (DWT-2D)*** - The DWT-2D [130, 131, 136-170] corresponds to multiresolution approximation expressions. In practice, mutiresolution analysis is carried out using 4 channel filter banks (for each level of decomposition) composed of a low-pass and a high-pass filter and each filter bank is then sampled at a half rate (1/2 down sampling) of the previous frequency. By repeating this procedure, it is possible to obtain wavelet transform of any order. The down sampling procedure keeps the scaling parameter constant (equal to ½) throughout successive wavelet transforms so that is benefits for simple computer implementation. In the case of an image, the filtering is implemented in a separable way be filtering the lines and columns. Note that the DWT of an image consists of four frequency channels for each level of decomposition [130, 131]. For example, for *i*-level (superscript) of decomposition for noisy (subscript) image, we have:

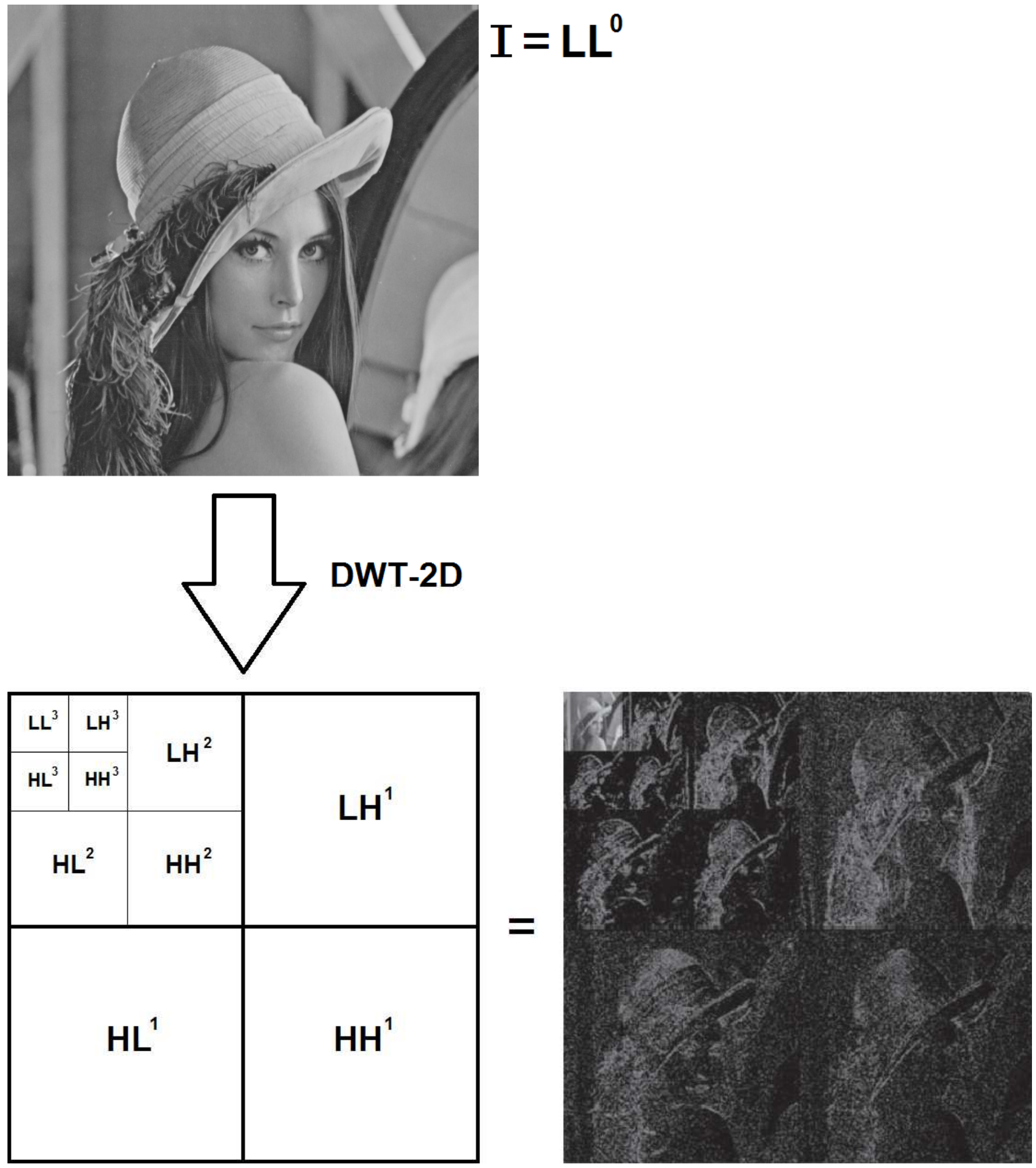

**Fig.12** Data preparation of the image. Recursive decomposition of LL parts.

$LL_n^i$ : Noisy Coefficients of Approximation.

$LH_n^i$ : Noisy Coefficients of Horizontal Detail,

$HL_n^i$ : Noisy Coefficients of Vertical Detail, and

$HH_n^i$ : Noisy Coefficients of Diagonal Detail.

The LL part at each scale is decomposed recursively, as illustrated in Fig. 12 [130, 131].

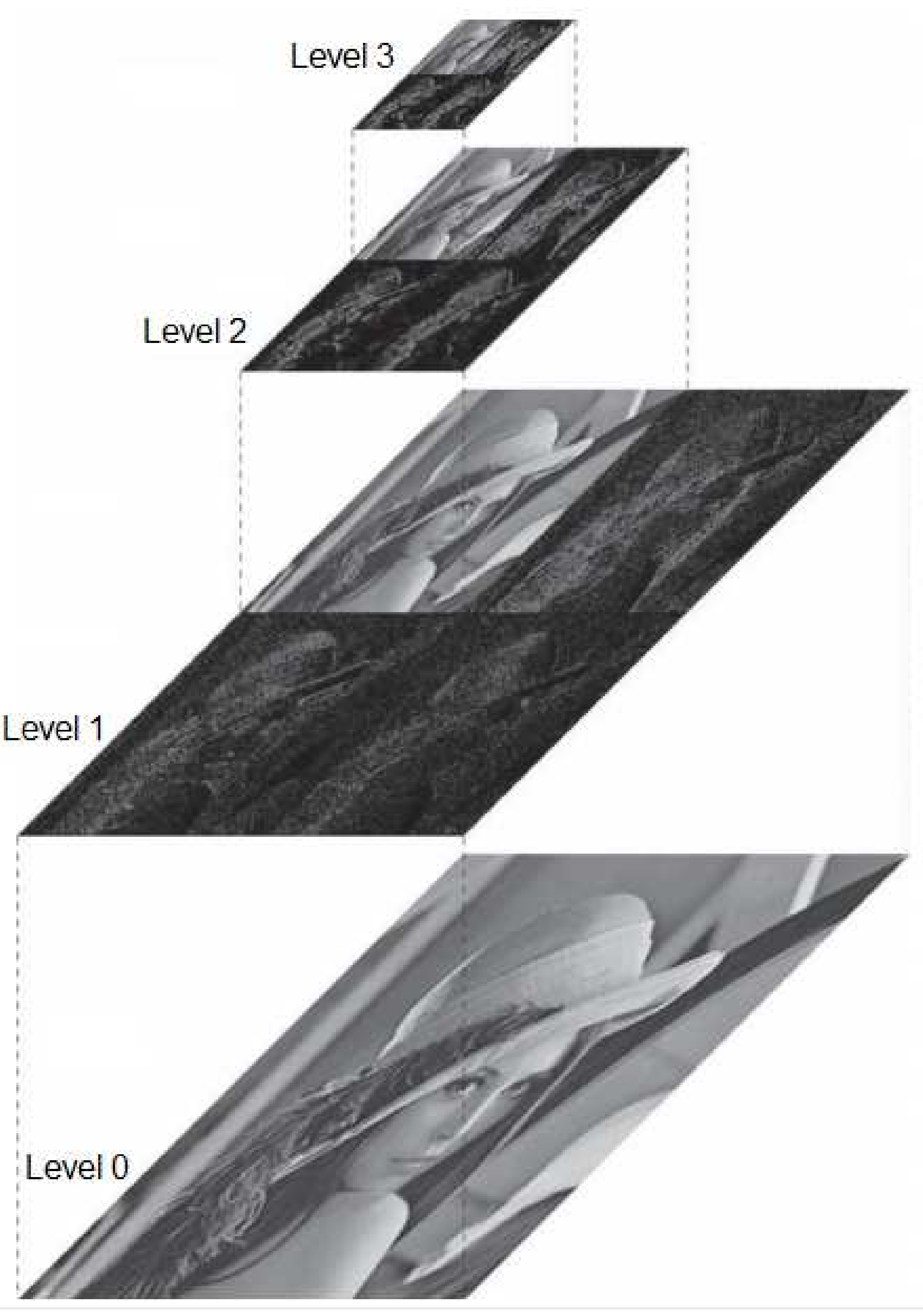

**Fig.13** Detail of level decomposition for down-right image of Fig.12.

Figure 13 shows –in detail– three levels of decomposition for gray version of Lena. In this figure, we can see that the splitting occurs from the subband of approximation coefficients, always. Each application of DWT-2D provides four subbands, which, every $LL^i$ will have less noise and size as the previous one, i.e., $LL^{i-1}$.

To achieve space-scale adaptive noise reduction, we need to prepare the 1-D coefficient data stream which contains the space-scale information of 2-D images. This is somewhat similar to the “zigzag” arrangement of the DCT (Discrete Cosine Transform) coefficients in image coding applications [169]. In this data preparation step, the DWT-2D coefficients are rearranged as a 1-D coefficient series in spatial order so that the adjacent samples represent the same local areas in the original image [165].

Figure 14 shows inside of DWT-2D with the four subbands of the transformed image [170], while Fig.15 shows inside of IDWT-2D (which is the inverse of DWT-2D), both, i.e., DWT-2D and IDWT-2d will be used in Fig.16. Each output of Fig. 14 represents a subband of splitting process of the 2-D coefficient matrix corresponding to Fig. 12. More split levels are not shown to avoid complicating the figures 14 and 15.

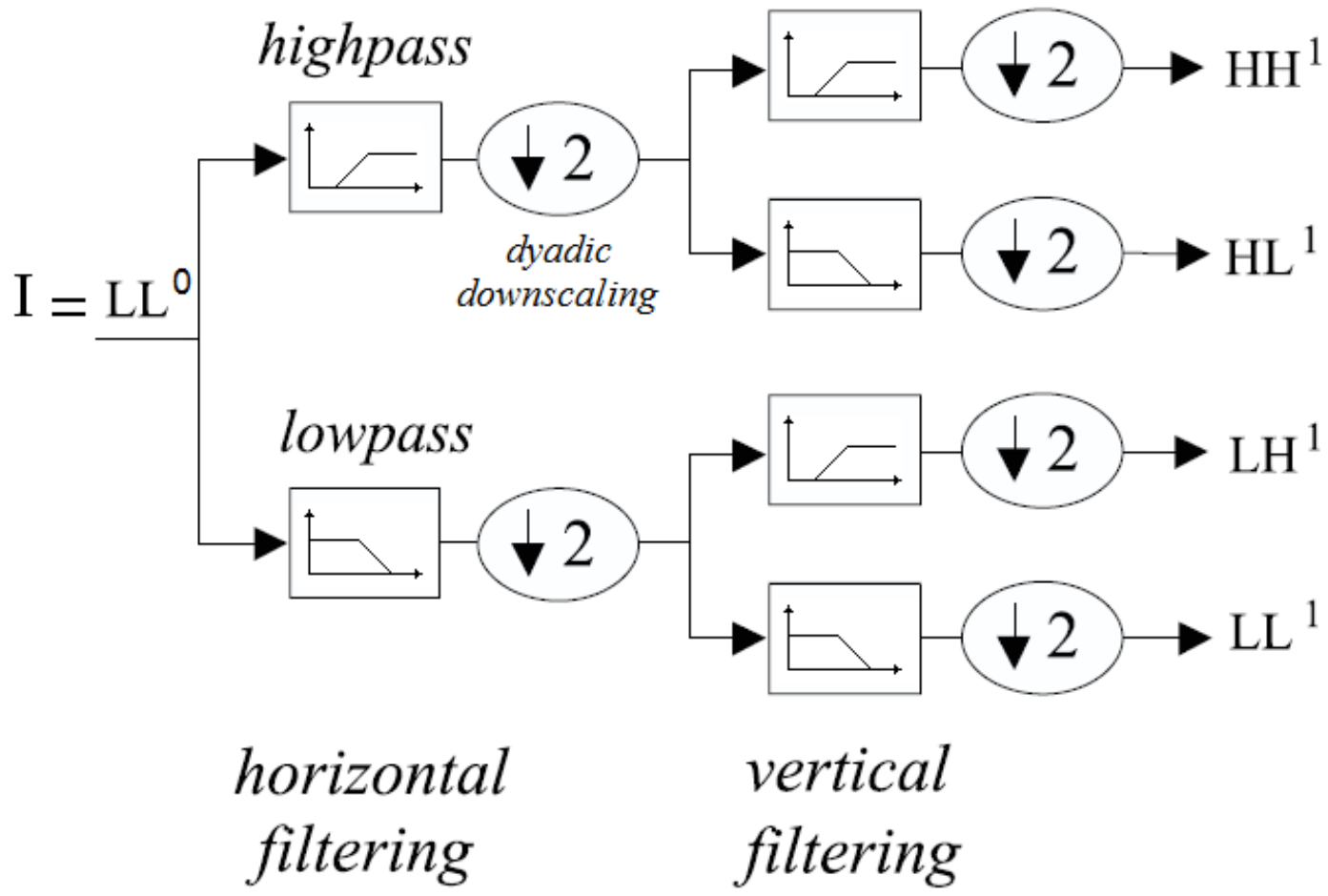

**Fig.14** DWT-2D. A decomposition step. Usual splitting of the subbands.

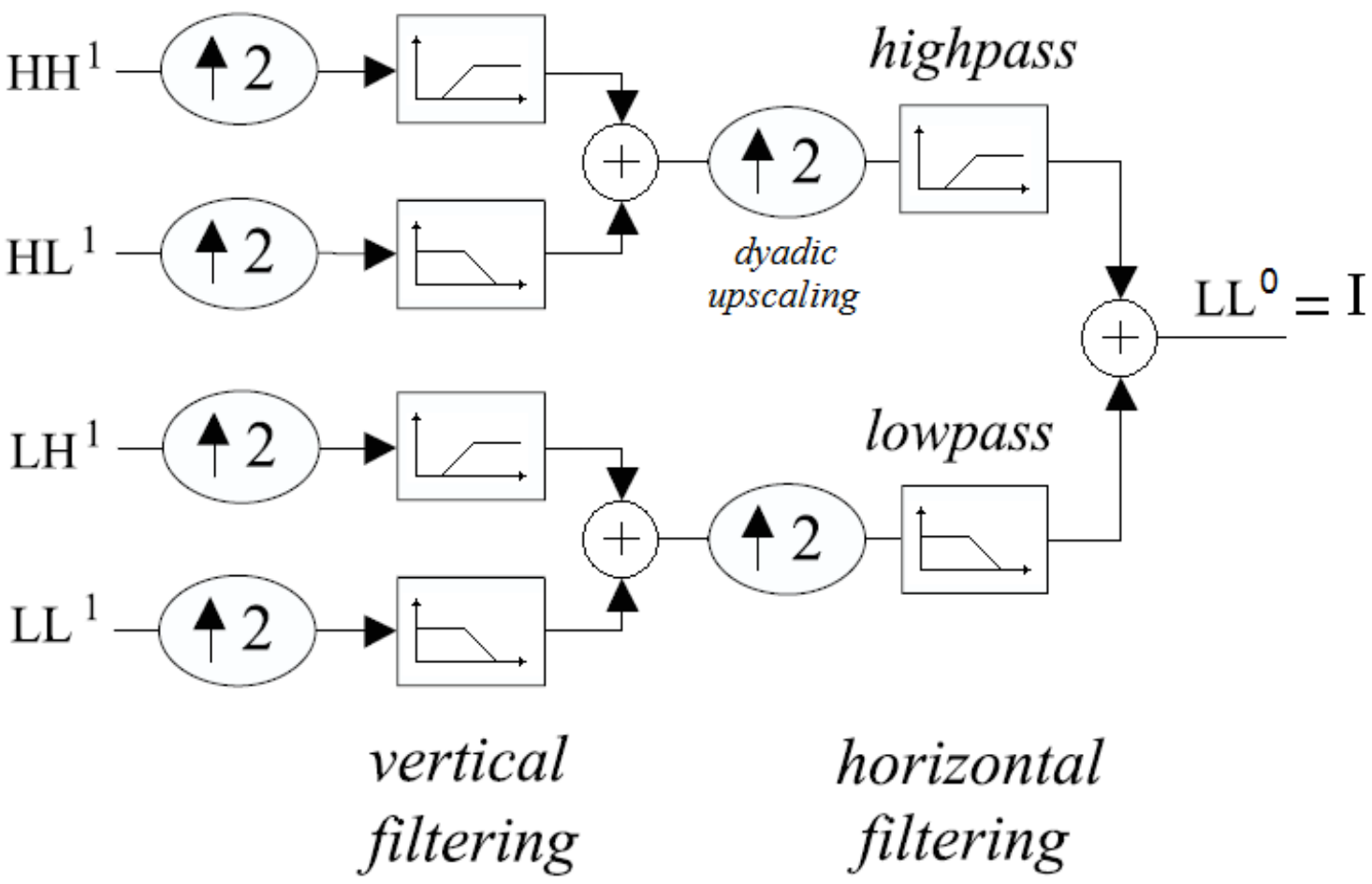

**Fig.15** Inverse DWT-2D (IDWT-2D). A recomposition step. Usual integration of the subbands.

This stage does not do much except for splitting the image into four disjoint sets of pixels. In our case one group consists of the low frequency indexed pixels in one subband and the other group consists of three high frequency subbands of indexed pixels. Each subband contains a quarter as many pixels as the original image. The splitting into low and high frequency is called the Lazy wavelet transform.

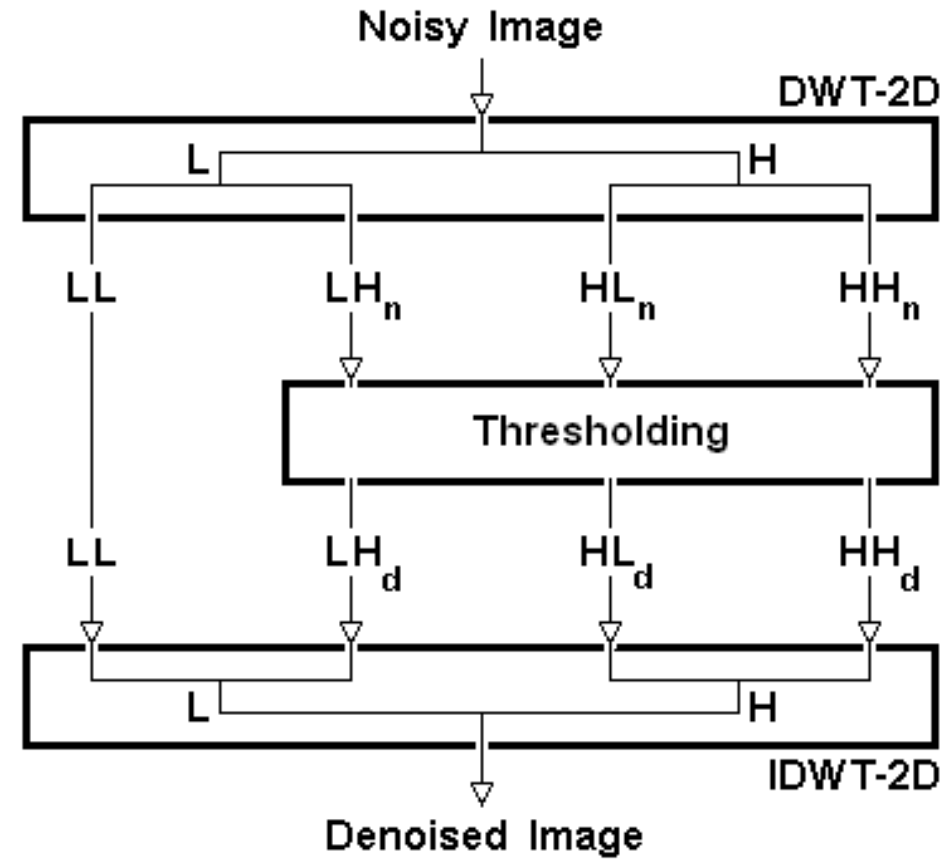

**Fig.16** Thresholding Techniques.

***Wavelet Noise Thresholding*** - The wavelet coefficients calculated by a wavelet transform represent change in the image at a particular resolution. By looking at the image in various resolutions it should be possible to filter out noise, at least in theory. However, the definition of noise is a difficult one. In fact, "one person's noise is another's signal". In part this depends on the resolution one is looking at. One algorithm to remove Gaussian white noise is summarized by D.L. Donoho and I.M. Johnstone [125-127], and synthesized in Fig. 16.

The algorithm is:

1. Calculate a wavelet transform and order the coefficients by increasing frequency. This will result in an array containing the image average plus a set of coefficients of length 1, 2, 4, 8, etc. The noise threshold will be calculated on the highest frequency coefficient spectrum (this is the largest spectrum).

2. Calculate the *median absolute deviation* (mad) on the largest coefficient spectrum. The median is calculated from the absolute value of the coefficients. The equation for the median absolute deviation is shown below:

$$\delta_{mad} = \frac{median\left(|\,C_n^i\,|\right)}{0.6745} \tag{50}$$

   where $C_n^i$ may be $LH_n^i$, $HL_n^i$, or $HH_n^i$ for *i*-level of decomposition. The factor 0.6745 in the denominator rescales the numerator so that $\delta_{mad}$ is also a suitable estimator for the standard deviation for Gaussian white noise [5, 165, 169].

3. For calculating the noise threshold $\lambda$ we have used a modified version of the equation that has been discussed in papers by D.L. Donoho and I.M. Johnstone. The equation is:

$$\lambda = \delta_{mad}\sqrt{2\,log\,[N]} \tag{51}$$

   where N is the number of pixels in the subimage, i.e., LH, HL or HH.

4. Apply a thresholding algorithm to the coefficients. There are two popular versions:

   4.1 *Soft thresholding* sets any coefficient less than or equal to the threshold to zero, see Fig.17(a). The space between $-\lambda$ and $+\lambda$ is called *dead zone*.

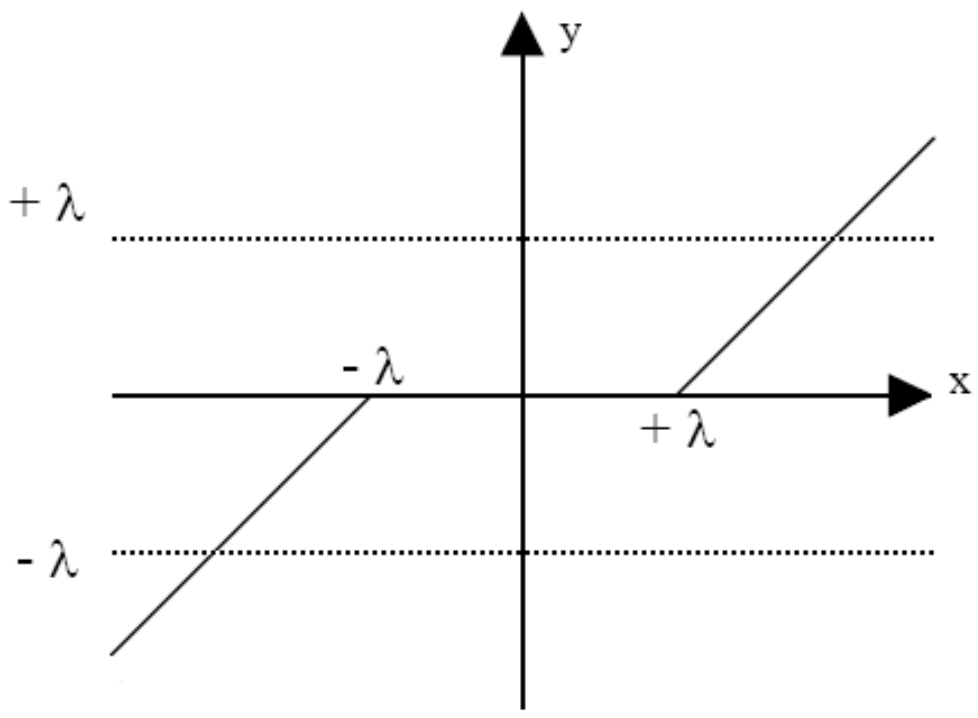

**Fig. 17(a)** Soft-Thresholfing

where ***x*** may be $LH_n^i$, $HL_n^i$, or $HH_n^i$, ***y*** may be

$LH_d^i$ : Denoised Coefficients of Horizontal Detail,

$HL_d^i$ : Denoised Coefficients of Vertical Detail, and

$HH_d^i$ : Denoised Coefficients of Diagonal Detail, for *i*-level of decomposition.

The respective code is:

```
for row = 1:N^(1/2)
   for column = 1:N^(1/2)
      if |C_n^i [row][column]| ≤ λ,
         C_n^i [row][column] = 0.0;
      end
   end
end
```

4.2 *Hard thresholding* sets any coefficient less than or equal to the threshold to zero, see Fig.17(b). The threshold is subtracted from any coefficient that is greater than the threshold. This moves the image coefficients toward zero.

Here too, the space between –λ and + λ is called *dead zone*.

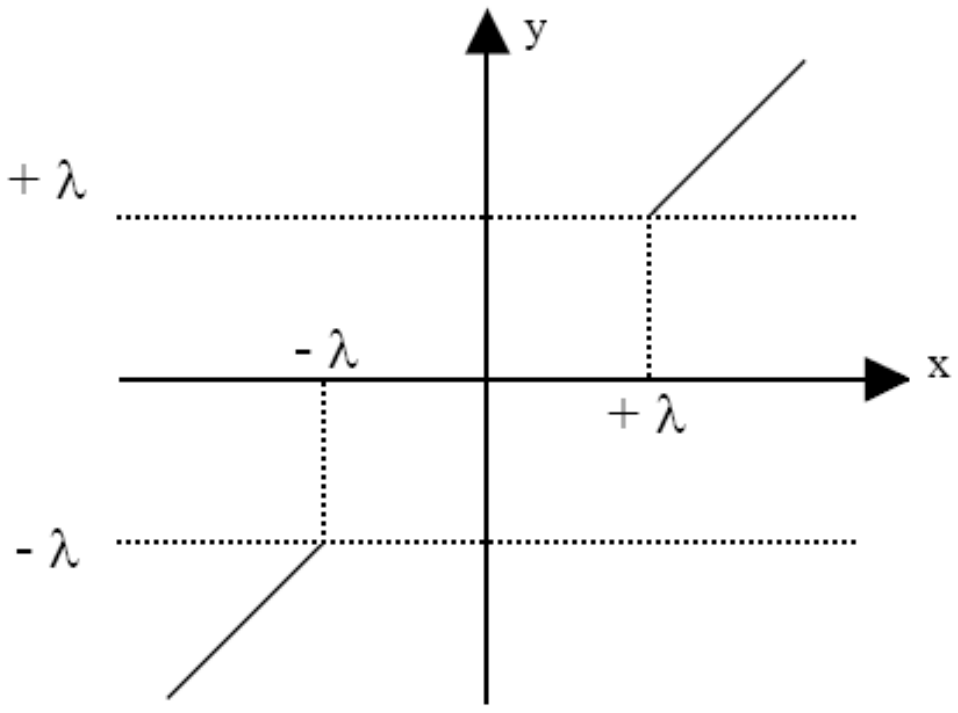

**Fig. 17(b)** Hard-Thresholfing

The respective code is:

```
for row = 1:N^{1/2}
   for column = 1:N^{1/2}
      if |C_n^i[row][column]| ≤ λ,
         C_n^i[row][column] = 0.0;
      else
         C_n^i[row][column] = C_n^i[row][column] − λ;
      end
   end
end
```

A much more efficient denoising/compression technique based on detail subbands filtering of DWT-2D is due to Mastriani [162, 166], which is called *Smoothing of Coefficients in Wavelet Domain*. This technique is notably superior to thresholding methods above seen [162, 166], and then it will be explained. However, before, we will explain the basic principles of a two-dimensional convolution mask [45-48].

***Two-dimensional convolution mask*** – It consists in a mask, which makes a horizontal rafter (see Fig.18) on that image (or detail wavelet subband) to which we must obtain, e.g., an image or subband denoising [45-48] via two-dimensional convolution between the mask and a similarly dimensional portion of the image or subband.

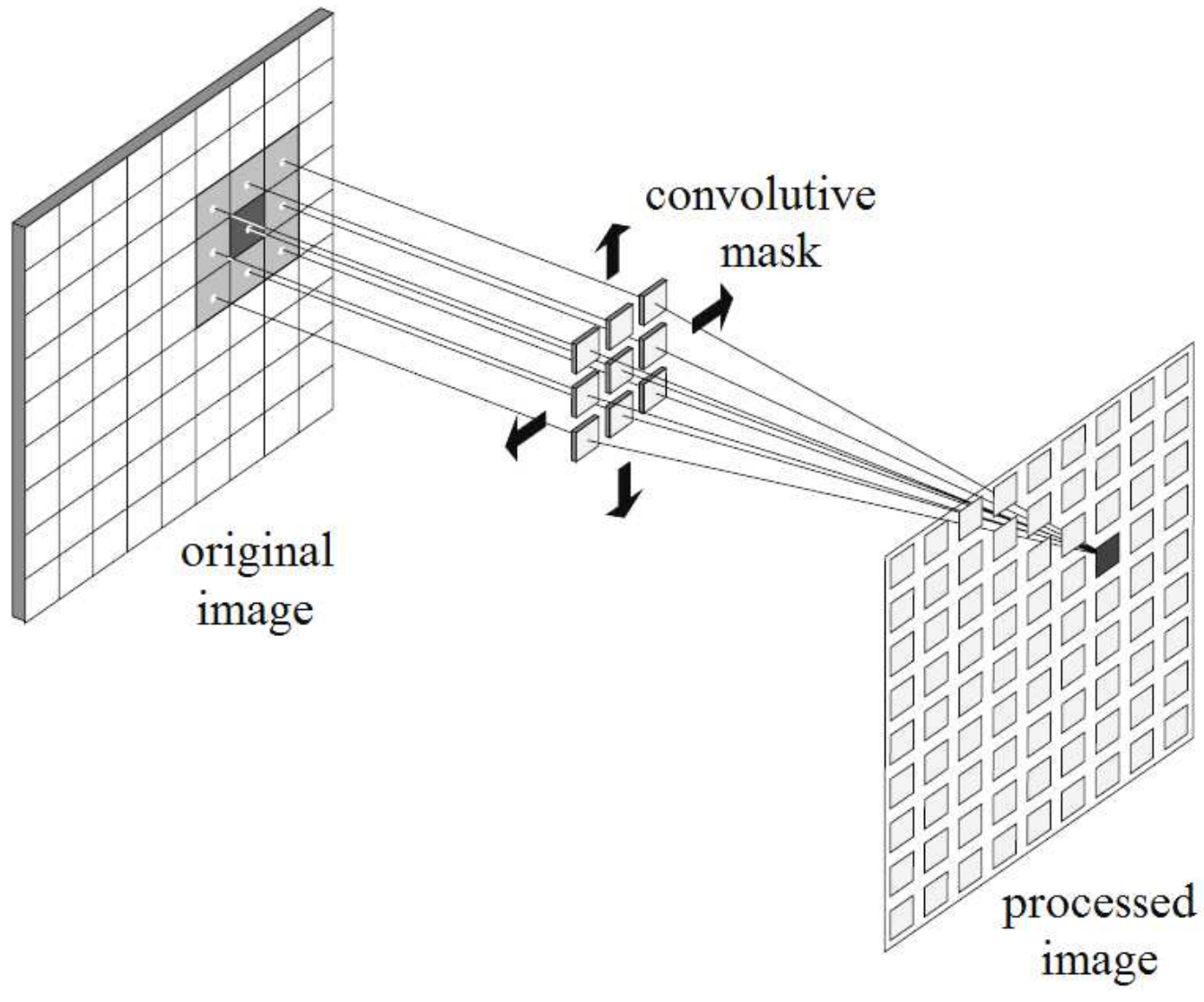

**Fig. 18** The convolution between the mask and the original image (or subband) in a horizontal rafter produce the processed image (or subband).

The main idea is to make an interaction between the mask and a portion of the image to be processed (with the same dimension as the mask) and that the result of said interaction to replace central pixel value of the image portion affected by the mask [45-48]. Based on Fig.19, we take a mask of 3×3 (often called *kernel,* which should be of any size, that is, not only 3×3, provided it has the same number of rows and columns and the dimension is an odd number) which is applied in a horizontal rafter way.

Various types of convolution masks are used in Digital Image Processing for filtering, enhancement, edge detection, among others applications [45-48].

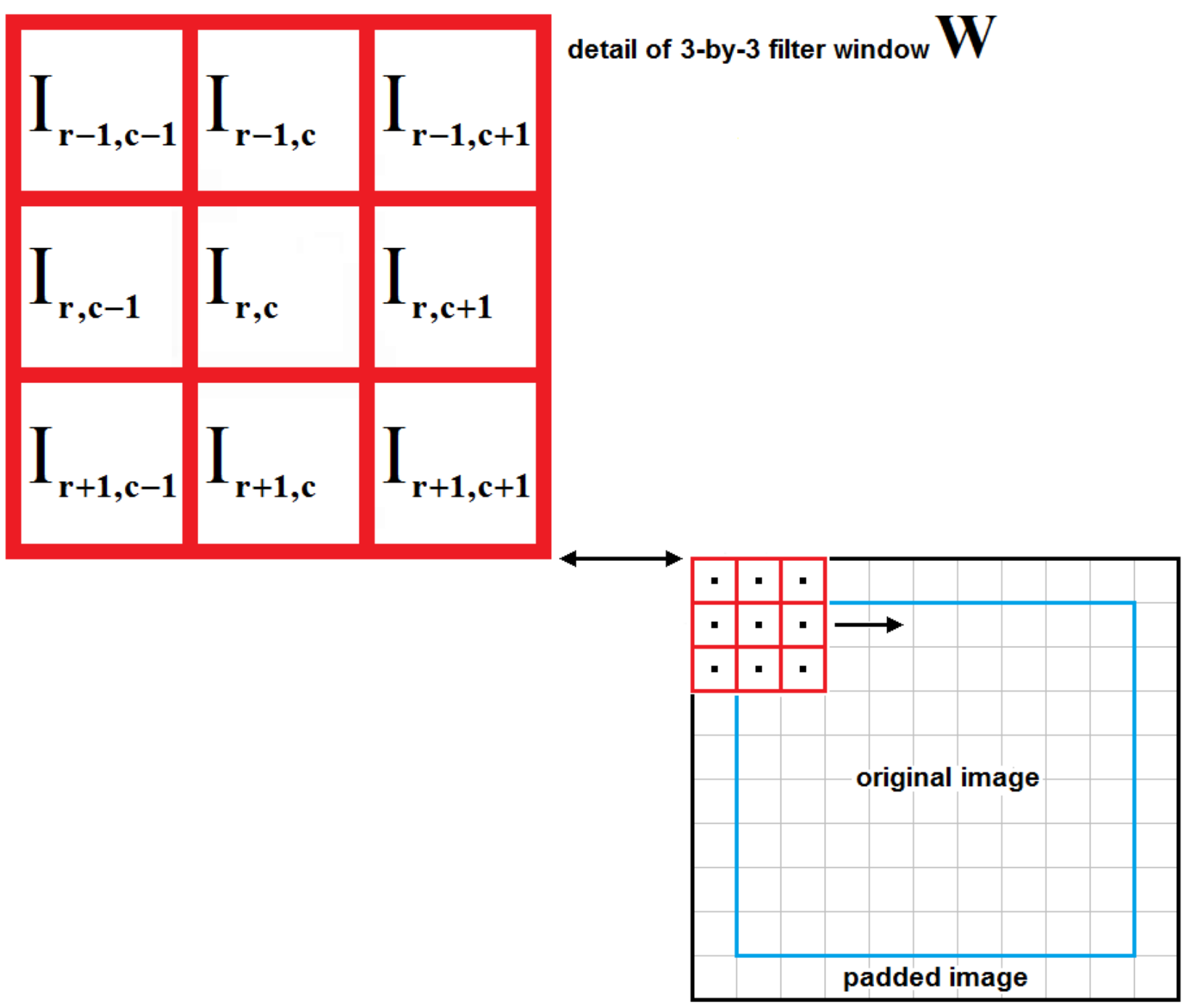

**Fig. 19** An example of 3x3 filter window for new edge-detection algorithm on an image (or subband).

***Smoothing of coefficients (SC) in wavelet domain*** – Like to last methods, we decompose the noisy image into four wavelet subbands, and then, we apply a two-dimensional smoothing within each highest subband, and reconstruct an image from the modified wavelet coefficients, that is to say, denoised coefficients, as shown in Fig. 20.

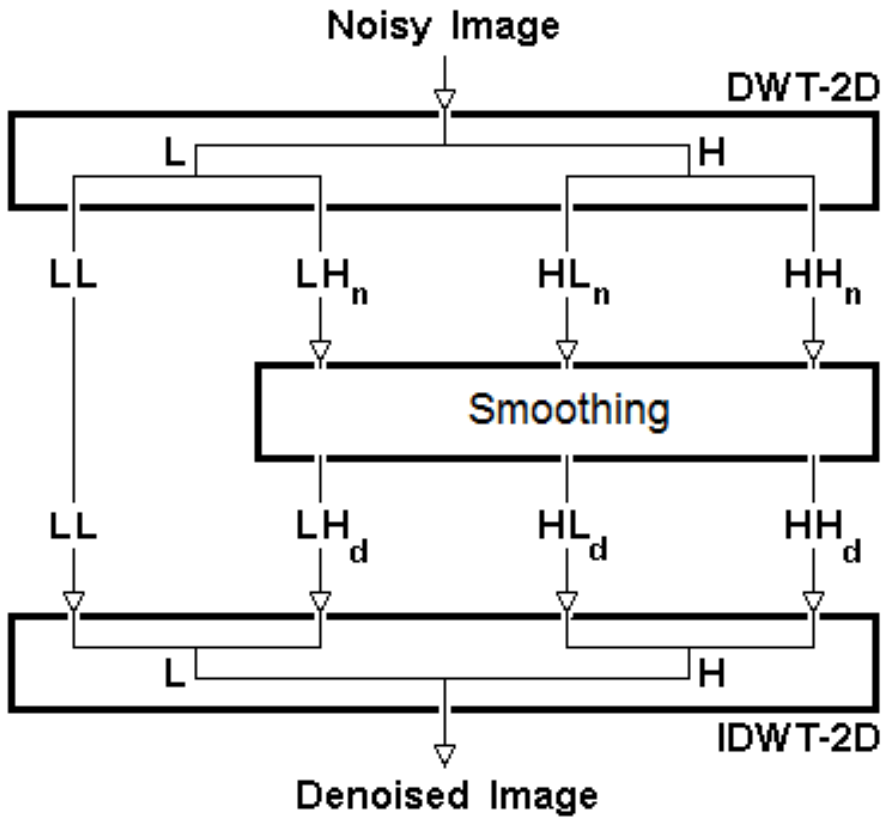

**Fig. 20** Smoothing of coefficients in wavelet domain.

If we use an original image of *R-by-C* pixels, then each subbands will have (*R/*2)-*by*-(*C/*2) coefficients. The SC process is applied - in principle - a single time, and exclusively to the first level of decomposition.

***Directional Smoothing*** - To protect the edges from blurring while smoothing the respective coefficients of subband, an appropriate filter must be applied. The most of statistical filters have a noise reduction approach that performs spatial filtering in a square-moving window defined as kernel, and is based on the statistical relationship between the central pixel and its surrounding pixels as shown in Fig. 21.

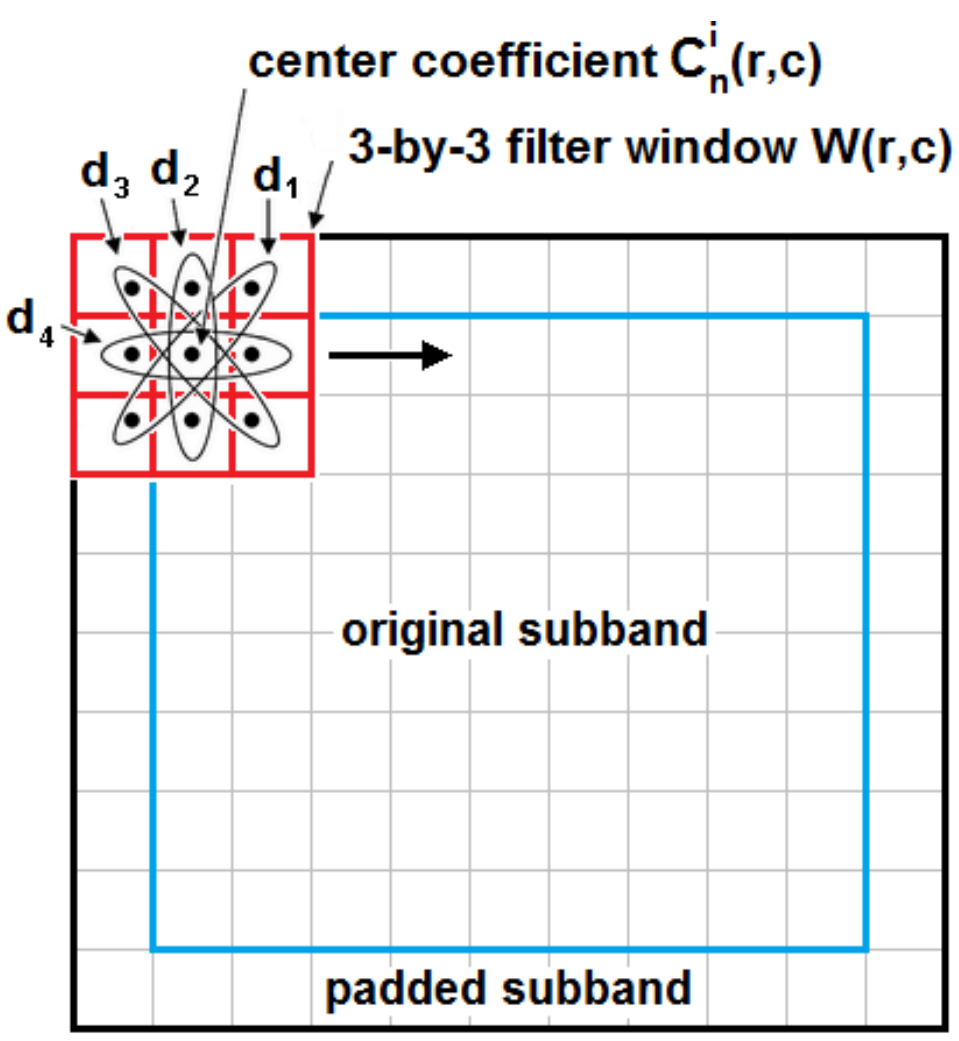

**Fig. 21** 3-by-3 filter window for noise smoothing over each highest subband using directional smoothing.

The size of the filter window can range from 3-by-3 to 33-by-33, with an odd number of cells in both directions. A larger filter window means that a larger area of the subband will be used for calculation and requires more computation time depending on the complexity of the filter's algorithm. If the size of filter window is too large, the important details will be lost due to over smoothing.

On the other hand, if the size of the filter window is too small, noise reduction may not be very effective. In practice, a 3-by-3 or a 7-by-7 filter window usually yields good results in the cases under study [171, 172].

For example, if the statistical filter used inside SC method (that is to say, over each highest subbands) is Directional Smoothing (DS), then, let $x[r, c]$ denote the value of the corresponding noisy detail coefficient at location $(r, c)$. Let $W[r, c]$ represent the group of coefficients contained in a filtering window with the size of 3-by-3 pixels and centered at location $(r, c)$ of the corresponding noisy detail coefficient, as shown in Fig.21:

$$W(r,c) = \left\{ C_n^i(r+p,c+q) \,/\, (p,q) \in [-1,0,1] \right\} \tag{52}$$

where $p$ and $q$ are integer indices each individually ranging from $-1$ to 1. Here, $r$ and $c$ are the row and the column indices, respectively, with $2 \leq r \leq (R/2)\text{-}1$ and $2 \leq c \leq (C/2)\text{-}1$.

The noisy input subband is processed by sliding a 3-by-3 filtering window on the subband. The window is started from the upper-left corner of the subband and moved sideways and progressively downwards in a raster scanning fashion. Meanwhile, the directional averaging filter (selective with respect to direction) examine the average based on several directionally oriented masks, as shown in Fig.21, and it compute averages in d(1), d(2), d(3) and d(4) directions. Such as, I(r,c) = d(k) where d(k) is the one closest in amplitude to I(r,c), i.e. | I(r,c)- d(k) | minimum, and $k \in [1,4]$. That is to say, it has a tendency not to destroy boundaries. On the other hand, the directional analysis can also be used to check if a coefficient belongs to a directional edge (leave unchanged) or is noise (remove noise).

Directional smoothing method in much finer than the mean filtering, reason by which, the first does not require a post-filtering enhancement (based on a realce mask) as in the case of the second one. That is, in the directional smoothing the edges are not injured as mean filtering. See next subsection.

Being $I(r,c) = C_n^i(r,c)$, the following MATLAB® [49] code represents the Directional Smoothing (DS) function for four directions and a 3x3 kernel [166, 171].

```
function Id = ds(I)
[ROW,COL] = size(I);
Id = [];
for r = 2:ROW-1      % padded image
  for c = 2:COL-1    % padded image
     d(1) = (I(r+1,c-1)+I(r,c)+I(r-1,c+1))/3;
     d(2) = (I(r-1,c)+I(r,c)+I(r+1,c))/3;
     d(3) = (I(r-1,c-1)+I(r,c)+I(r+1,c+1))/3;
     d(4) = (I(r,c-1)+I(r,c)+I(r,c+1))/3;
     for k = 1:4
        aux(k) = abs(d(k)-I(r,c));
     end
     [min_aux,allo_min_aux] = min(aux);
     Id(r,c) = round(d(allo_min_aux));
   end
end
```

where I and Id represent the noisy and denoised image or subband, respectively, while, ROW and COL, are the number of rows, and columns, respectively, the size of I is ROWxCOL, while for Id is ROW-2xCOL-2.

On the other hand, this method can be passed more than once, therefore, said method becomes in a multi-pass directional smoothing [166-171].

Finally, if the image or subband is in color, thus, we must apply in each bitmap matrix, that is to say, red, green, and blue, individually.

***Mean Filtering*** – The idea of classical mean filtering is simply to replace each pixel value in an image with the mean (`average') value of its neighbors, including itself. This has the effect of eliminating pixel values which are unrepresentative of their surroundings. Mean filtering is usually thought of as a convolution filter. Like other convolutions it is based around a kernel (or mask), which represents the shape and size of the neighborhood to be sampled when calculating the mean. Often a 3×3 square kernel is used, as shown in Fig.22, although larger kernels (*e.g.* 5×5 squares) can be used for more severe smoothing. (Note that a small kernel can be applied more than once in order to produce a similar but not identical effect as a single pass with a large kernel.)

| $\frac{1}{9}$ | $\frac{1}{9}$ | $\frac{1}{9}$ |
|---|---|---|
| $\frac{1}{9}$ | $\frac{1}{9}$ | $\frac{1}{9}$ |
| $\frac{1}{9}$ | $\frac{1}{9}$ | $\frac{1}{9}$ |

**Fig. 22** 3×3 averaging kernel often used in mean filtering.

The mean filtering is the biggest noise remover of all employees in the Digital Image Processing [45-48]. However, it is the heat that injures edges and image texture to a greater degree. Undoubtedly, its main advantage is the simplicity of coding. However, and as in the case of directional smoothing, the image must be padded before use, i.e., if the image has ROWxCOL pixels, and the mask has M-by-M elements, the padded image should have (ROW-(M-1))x(COL-(M-1)) pixels.

The following MATLAB® [49] code represents the Mean Filtering (MF) function for a 3x3 kernel [45-48].

```
function Id = mf(I)
[ROW,COL] = size(I);
mask = ones(3,3)/9;
Id = [];
for r = 2:ROW-1      % padded image
  for c = 2:COL-1    % padded image
     for p = -1:1:1
        for q = -1:1:1
           W(p+2,q+2) = I(r+p,c+q);
        end
     end
     Id(r,c) = sum(sum(W.*mask));
   end
end
```

Here also, the size of I is ROWxCOL, and the size of mask is 3x3, while for Id is ROW-2xCOL-2.

***Finally*** - Any used filter performs the filtering based on either local statistical data given in the filter window to determine the noise variance within the filter window, or estimating the local noise variance of the subband under study, e.g., Lee, Kuan, Gamma-Map, Enhanced Lee, Frost, Enhanced Frost [171, 173], Wiener [171], DS [171, 172], Enhanced DS (EDS) [171], Savitzky-Golay 2D (see Subsection 2.3), and a new filter from QSA (see Subsection 3.3.1 for signals, and 3.3.2 for images, Figures 18 to 22, and Subsection 4.1.3).

***Haar basis for DWT-2D*** – without loss of generality, we show the first time for application of Haar basis, considering:

$$I = LL^{0} \quad \text{(original image)} \tag{53}$$

$$\forall\, r \in [1,R] \wedge c \in [1,C]$$

$$\begin{aligned}
LL^{1}_{r,c} &= \left(\ \ LL^{0}_{r,c} + LL^{0}_{r,c+1} + LL^{0}_{r+1,c} + LL^{0}_{r+1,c+1}\right)/4 && \textit{scaling function}\\
LH^{1}_{r,c} &= \left(-LL^{0}_{r,c} + LL^{0}_{r,c+1} - LL^{0}_{r+1,c} + LL^{0}_{r+1,c+1}\right)/4 && \textit{wavelet function}\\
HL^{1}_{r,c} &= \left(-LL^{0}_{r,c} - LL^{0}_{r,c+1} + LL^{0}_{r+1,c} + LL^{0}_{r+1,c+1}\right)/4 && \textit{wavelet function}\\
HH^{1}_{r,c} &= \left(\ \ LL^{0}_{r,c} - LL^{0}_{r,c+1} - LL^{0}_{r+1,c} + LL^{0}_{r+1,c+1}\right)/4 && \textit{wavelet function}
\end{aligned} \tag{54}$$

and so on [125-127].

***One-dimensional Discrete Wavelet Transform (DWT-1D)*** - For DWT-1D [125-127], we will take similar considerations to DWT-2D, that is to say,

$L^{i}_{n}$ : Noisy Coefficients of Approximation, and

$H^{i}_{n}$ : Noisy Coefficients of Detail.

The L part at each scale is decomposed recursively like two-dimensional version, as illustrated in Fig.23 [125-127], e.g., for a typical electrocardiographic signal [174].

Figure 24 shows the interior of the DWT-1D with the two subbands of the transformed signal [125-127], which will be used in Fig.25, where IDWT-1D is the inverse of DWT-1D. Each output of Fig.24 represents a subband of splitting process of the 1-D coefficient vector corresponding to Fig.23.

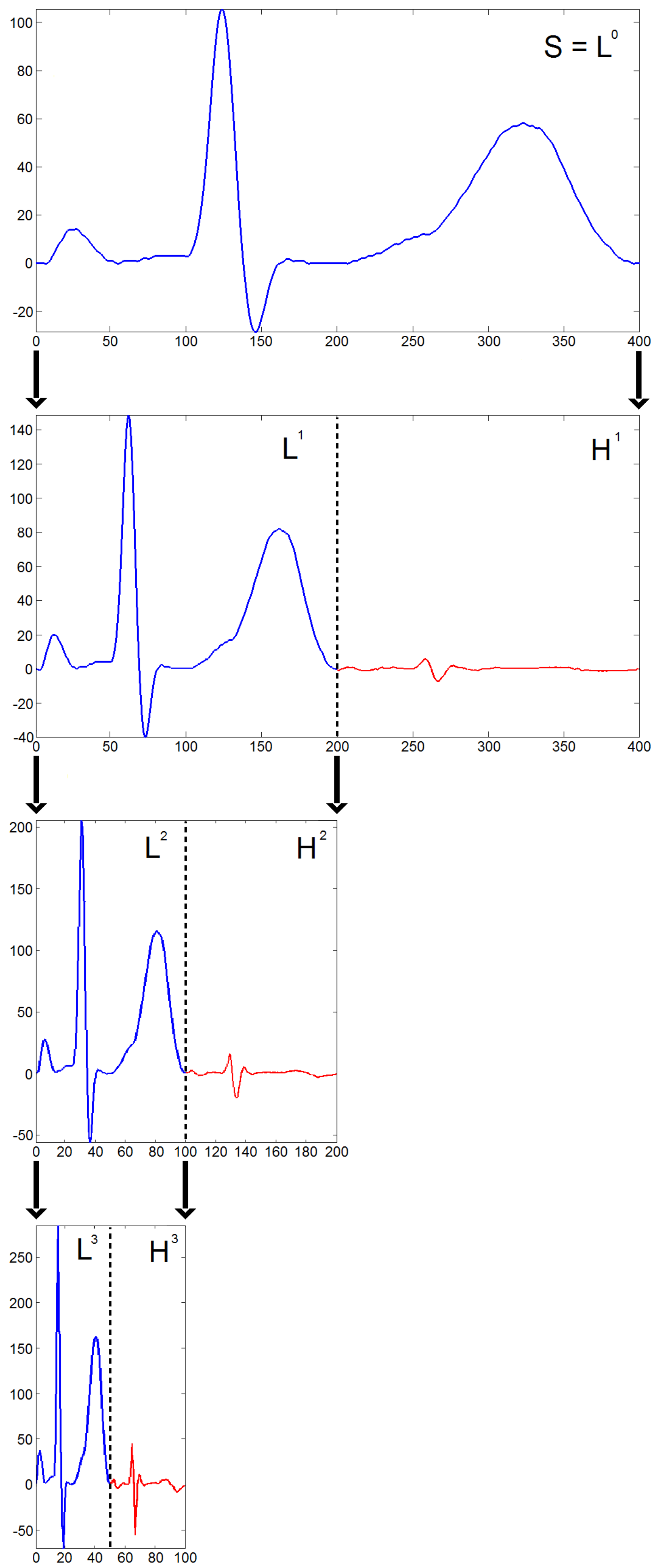

**Fig.23** Data preparation of the signal. Recursive decomposition of L parts.

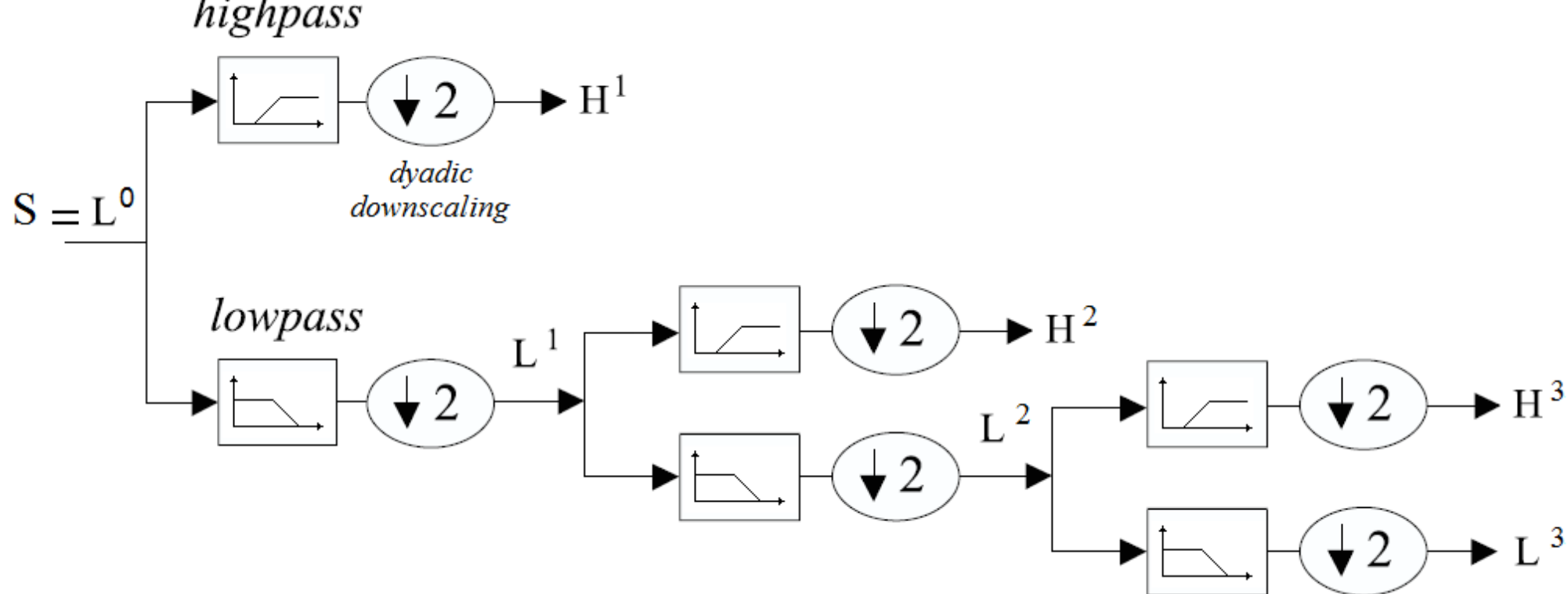

**Fig.24** DWT-1D. Three decomposition steps. Usual splitting of the subbands.

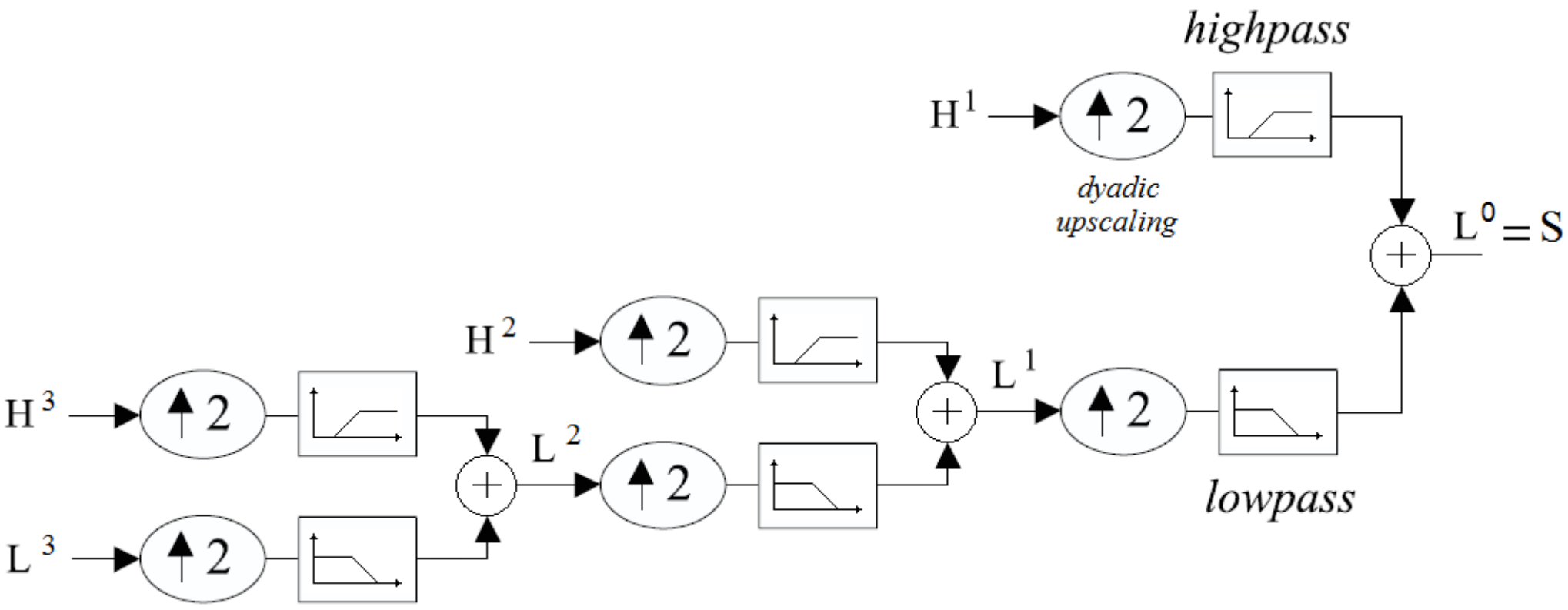

**Fig.25** Inverse DWT-1D (IDWT-1D). Three recomposition steps. Usual integration of the subbands.

***Wavelet Noise Thresholding*** –It is similar to the two-dimensional case, with the same algorithm, see Fig. 26.

***Smoothing of coefficients (SC) in wavelet domain*** – Like to two-dimensional case, we decompose the noisy signal into wavelet subbands (only two in here), and then, we apply a smoothing (e.g., Savitzky-Golay, see Subsection 2.3, and a new filter from QSA, see Subsections 3.3.1 for signals, and 3.3.2 for images, Figures 18 to 22, and Subsection 4.1.3) within each highest subband, and reconstruct a signal from the modified wavelet coefficients, that is to say, denoised coefficients, as shown in Fig.27.

***Haar basis for DWT-1D*** – without loss of generality, we show the first time for application of Haar basis, considering:

$$S = L^0 \text{ (original signal)} \tag{55}$$

$\forall\, t \in [1,T]$, where t is time.

$$\begin{aligned} L_t^1 &= \left( L_t^0 + L_{t+1}^0 \right) / 2 \qquad \textit{scaling function} \\ H_t^1 &= \left( -L_t^0 + L_{t+1}^0 \right) / 2 \qquad \textit{wavelet function} \end{aligned} \tag{56}$$

and so on [125-127].

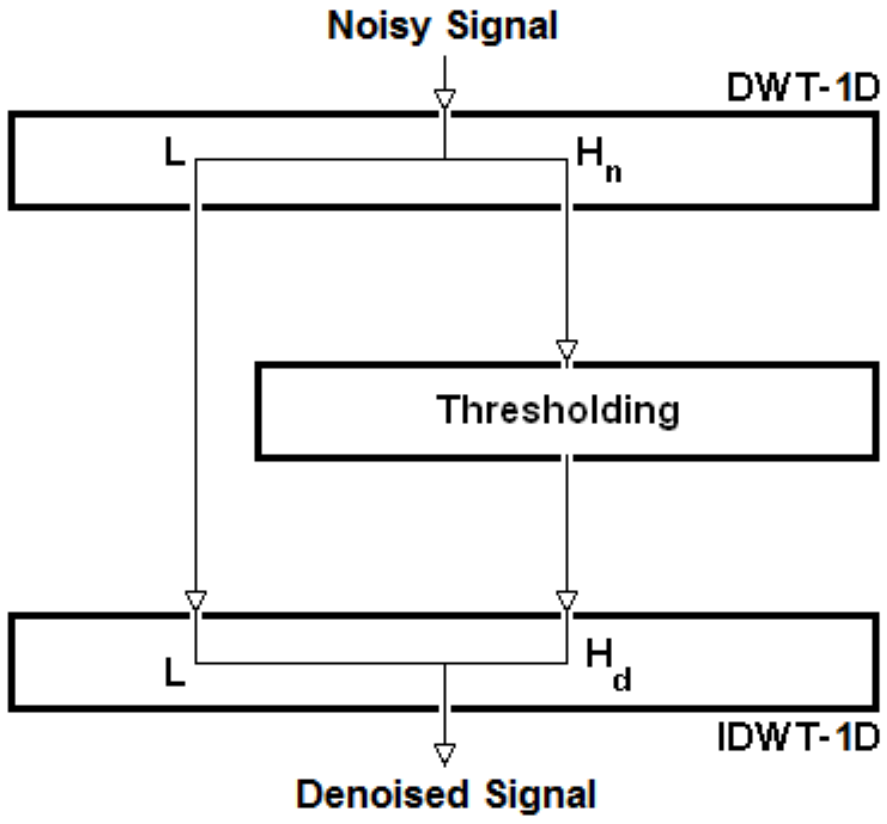

**Fig.26** Thresholding Techniques.

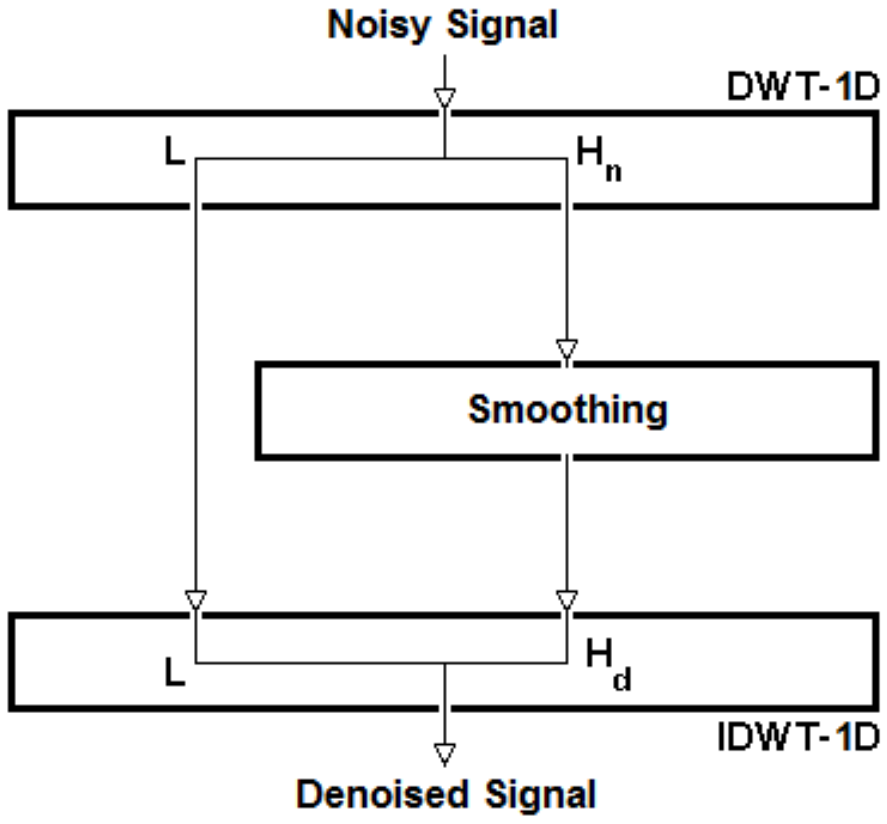

**Fig.27** Smoothing Technique.

Finally, we can apply smoothing of coefficients in wavelet domain via directional smoothing over highest subband of a signal, as in two-dimensional case, but adapted to one-dimensional case, thanks to a new technique. Coming up next.

***Directional Smoothing to a Signal*** – The procedure starts from the construction of a mask (kernel, or window), as in the two-dimensional case, although in this case it applies to a signal instead of an image.

Given a signal *S*, as a sequence of time samples, and assuming that it is integrable and differentiable, we have:

$$S = \left[ \mathrm{S}\left(\mathrm{t\text{-}N\Delta t}\right) \ldots \mathrm{S}\left(\mathrm{t\text{-}2\Delta t}\right) \mathrm{S}\left(\mathrm{t\text{-}\Delta t}\right) \mathrm{S}\left(\mathrm{t}\right) \mathrm{S}\left(\mathrm{t+\Delta t}\right) \mathrm{S}\left(\mathrm{t+2\Delta t}\right) \ldots \mathrm{S}\left(\mathrm{t+N\Delta t}\right) \right] \tag{57}$$

In every $\Delta$t we have a sample, altogether 2N+1 samples. On the other hand, we will call *U* (up) to the integral of said signal in time, so that:

$$U = \int S\,\mathrm{dt} = \left[ \mathrm{U}\left(\mathrm{t\text{-}N\Delta t}\right) \ldots \mathrm{U}\left(\mathrm{t\text{-}2\Delta t}\right) \mathrm{U}\left(\mathrm{t\text{-}\Delta t}\right) \mathrm{U}\left(\mathrm{t}\right) \mathrm{U}\left(\mathrm{t+\Delta t}\right) \mathrm{U}\left(\mathrm{t+2\Delta t}\right) \ldots \mathrm{U}\left(\mathrm{t+N\Delta t}\right) \right] \tag{58}$$

While, $D$ (down) will be the time derivative of the signal $S$:

$$D = \frac{\partial S}{\partial \mathrm{t}} = \left[ \mathrm{D}\left(\mathrm{t\text{-}N\Delta t}\right) \ldots \ \mathrm{D}\left(\mathrm{t\text{-}2\Delta t}\right) \ \mathrm{D}\left(\mathrm{t\text{-}\Delta t}\right) \ \mathrm{D}\left(\mathrm{t}\right) \ \mathrm{D}\left(\mathrm{t+\Delta t}\right) \ \mathrm{D}\left(\mathrm{t+2\Delta t}\right) \ldots \ \mathrm{D}\left(\mathrm{t+N\Delta t}\right) \right] \tag{59}$$

Figure 28 shows the original signal in green, its integral in red, and its derivative in blue.

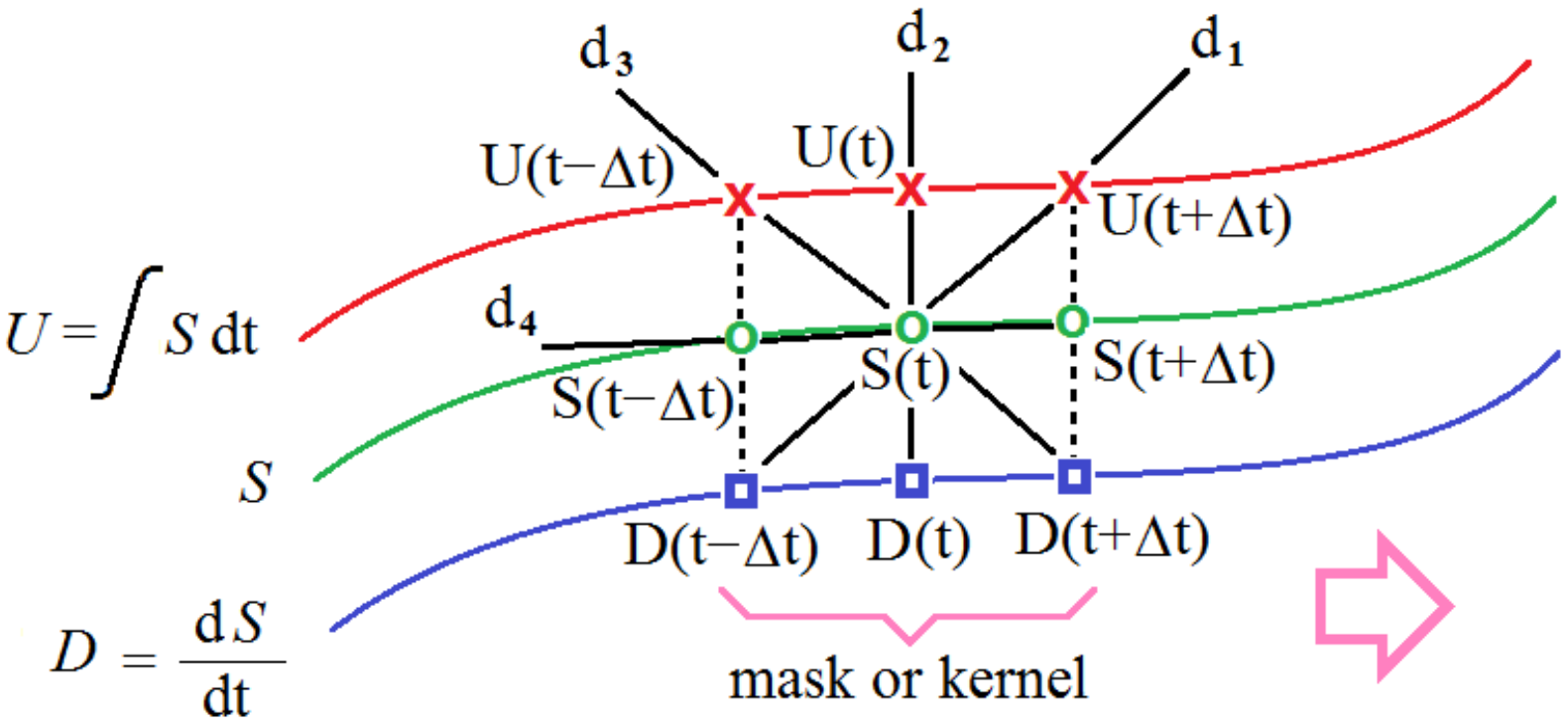

**Fig.28** 3-by-3 filter window for noisy smoothing over the image or highest subband using directional smoothing.

For example, the window will be constituted (at time instant t) as follow, see Fig.28:

$$\mathrm{W}\left(\mathrm{t}\right) = \begin{bmatrix} \mathrm{U}\left(\mathrm{t\text{-}\Delta t}\right) & \mathrm{U}\left(\mathrm{t}\right) & \mathrm{U}\left(\mathrm{t+\Delta t}\right) \\ \mathrm{S}\left(\mathrm{t\text{-}\Delta t}\right) & \mathrm{S}\left(\mathrm{t}\right) & \mathrm{S}\left(\mathrm{t+\Delta t}\right) \\ \mathrm{D}\left(\mathrm{t\text{-}\Delta t}\right) & \mathrm{D}\left(\mathrm{t}\right) & \mathrm{D}\left(\mathrm{t+\Delta t}\right) \end{bmatrix} \tag{60}$$

As the window moves from left to right (i.e., it progresses over time) the coefficients of the same change.

The size of the filter window can range from 3-by-3 to 33-by-33, with an odd number of cells in both directions. For example, for a mask of 5-by-5, we have the following elements constituting the top row of the filter window (in each instant t):

$$\begin{aligned} U^2 &= \int U \, \mathrm{dt} = \iint S \, \mathrm{dt}^2 \\ &= \left[ \mathrm{U}^2\left(\mathrm{t\text{-}N\Delta t}\right) \ldots \ \mathrm{U}^2\left(\mathrm{t\text{-}2\Delta t}\right) \ \mathrm{U}^2\left(\mathrm{t\text{-}\Delta t}\right) \ \mathrm{U}^2\left(\mathrm{t}\right) \ \mathrm{U}^2\left(\mathrm{t+\Delta t}\right) \ \mathrm{U}^2\left(\mathrm{t+2\Delta t}\right) \ldots \ \mathrm{U}^2\left(\mathrm{t+N\Delta t}\right) \right] \end{aligned} \tag{61}$$

where, superscript 2 means “second integral”.

Similarly, the down row of the filter window (in each instant t), we have:

$$\begin{aligned} D^2 &= \frac{\partial D}{\partial \mathrm{t}} = \frac{\partial^2 S}{\partial \mathrm{t}^2} \\ &= \left[ \mathrm{D}^2\left(\mathrm{t\text{-}N\Delta t}\right) \ldots \ \mathrm{D}^2\left(\mathrm{t\text{-}2\Delta t}\right) \ \mathrm{D}^2\left(\mathrm{t\text{-}\Delta t}\right) \ \mathrm{D}^2\left(\mathrm{t}\right) \ \mathrm{D}^2\left(\mathrm{t+\Delta t}\right) \ \mathrm{D}^2\left(\mathrm{t+2\Delta t}\right) \ldots \ \mathrm{D}^2\left(\mathrm{t+N\Delta t}\right) \right] \end{aligned} \tag{62}$$

where, superscript 2 means “second derivative”.

Then, the filter window in this case will be:

$$W(t) = \begin{bmatrix} U^2(t\text{-}2\Delta t) & U^2(t\text{-}\Delta t) & U^2(t) & U^2(t+\Delta t) & U^2(t+2\Delta t) \\ U(t\text{-}2\Delta t) & U(t\text{-}\Delta t) & U(t) & U(t+\Delta t) & U(t+2\Delta t) \\ S(t\text{-}2\Delta t) & S(t\text{-}\Delta t) & S(t) & S(t+\Delta t) & S(t+2\Delta t) \\ D(t\text{-}2\Delta t) & D(t\text{-}\Delta t) & D(t) & D(t+\Delta t) & D(t+2\Delta t) \\ D^2(t\text{-}2\Delta t) & D^2(t\text{-}\Delta t) & D^2(t) & D^2(t+\Delta t) & D^2(t+2\Delta t) \end{bmatrix} \tag{63}$$

and similarly for larger window.

From this point, the procedure is similar to the two-dimensional case, thus, it only remains to adapt the corresponding algorithm to the one-dimensional case environment. Therefore, the following MATLAB® [49] code represents the Directional Smoothing (DS) function for four directions and a 3x3 kernel for signal denoising.

```
function Sd = ds(S)
T = length(S);
% Padding signal
% aux_S = [ 0 S 0 ];         % for non-cyclical signals
   aux_S = [ S(T) S S(1) ]; % for cyclical signals
D = [];
for n = -N+1:N-1
  t = n-(-N+1)+3;
  D(t-2) = (aux_S(t)-aux_S(t-2))/2;
end
U = cumsum(S);
Sd = [];
for n = -N+1:N-1
  t = n-(-N+1)+2;
  d(1) = (D(t-1)+S(t)+U(t+1))/3;
  d(2) = (D(t)+S(t)+U(t))/3;
  d(3) = (D(t+1)+S(t)+U(t-1))/3;
  d(4) = (S(t-1)+S(t)+S(t+1))/3;
     for k = 1:4
        aux(k) = abs(d(k)-S(t));
     end
     [min_aux,allo_min_aux] = min(aux);
     Sd(t-1) = d(allo_aux_min);
   end
end
```

where T = 2N+1, with $-N+1 \leq t \leq N-1$, Sd is the denoised signal or subband. While, the size of *S*, *U* and *D* is T, the size of Sd is T-2 = 2N-1 (padded signal or subband).

*Note:* this tool was created by the author for this paper.

***Mean Filtering to a Signal*** – Similar to two-dimensional case, the idea of classical mean filtering applied to a signal is simply to replace each sample S(t) of this with the mean (`average') value of its neighbors (in window), including itself, see Fig. 29.

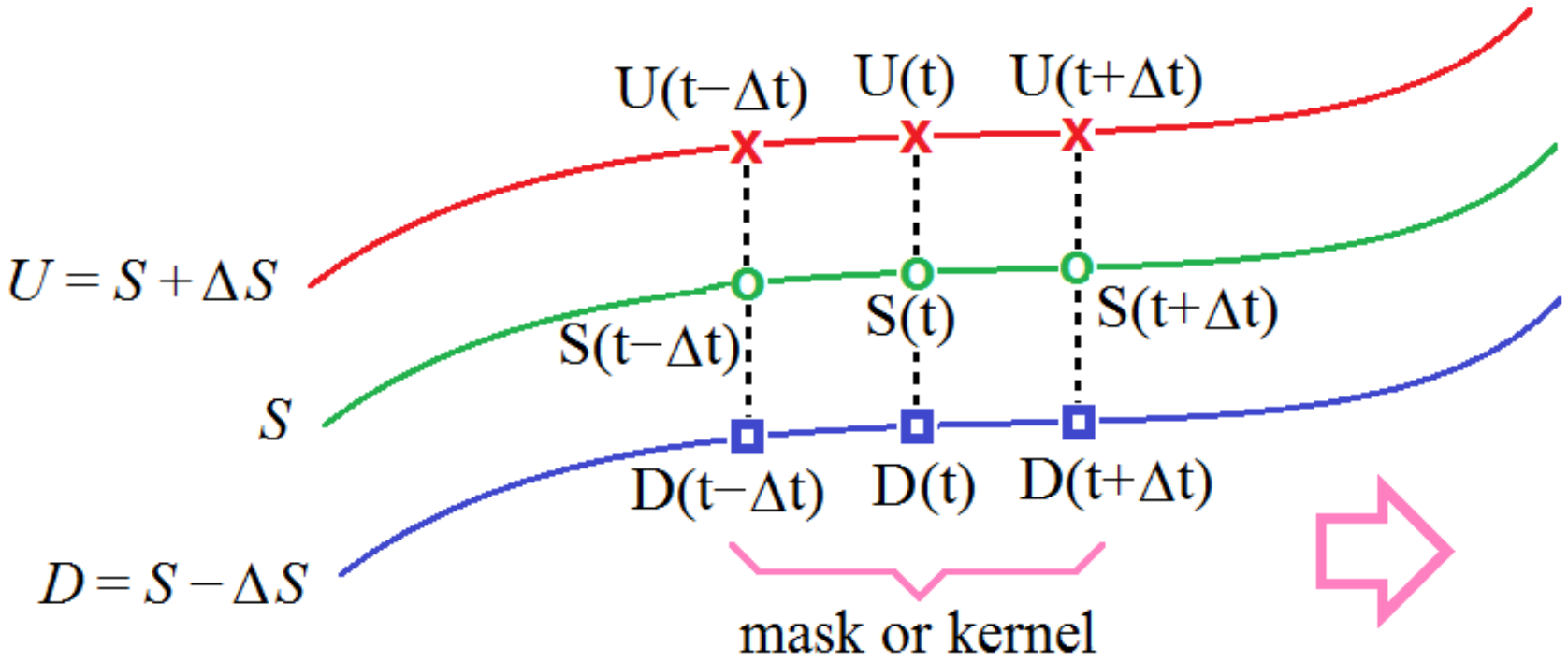

**Fig.29** 3-by-3 filter window for noisy smoothing over the image or highest subband using mean filtering.

This has the effect of eliminating signal sample which are unrepresentative of their surroundings.

Give a signal like Eq.(57), here too, in every $\Delta$t we have a sample, altogether 2N+1 samples. However, *U* (up) results from a positive increase on the signal, so that:

$$U = S + \Delta S = \left[ \mathrm{U}\left(\text{t-N}\Delta\text{t}\right) \ldots\ \mathrm{U}\left(\text{t-2}\Delta\text{t}\right)\ \mathrm{U}\left(\text{t-}\Delta\text{t}\right)\ \mathrm{U}\left(\text{t}\right)\ \mathrm{U}\left(\text{t+}\Delta\text{t}\right)\ \mathrm{U}\left(\text{t+2}\Delta\text{t}\right) \ldots\ \mathrm{U}\left(\text{t+N}\Delta\text{t}\right) \right] \quad (64)$$

While, *D* (down) will result from a negative increase on the signal *S*:

$$D = S + \Delta S = \left[ \mathrm{D}\left(\text{t-N}\Delta\text{t}\right) \ldots\ \mathrm{D}\left(\text{t-2}\Delta\text{t}\right)\ \mathrm{D}\left(\text{t-}\Delta\text{t}\right)\ \mathrm{D}\left(\text{t}\right)\ \mathrm{D}\left(\text{t+}\Delta\text{t}\right)\ \mathrm{D}\left(\text{t+2}\Delta\text{t}\right) \ldots\ \mathrm{D}\left(\text{t+N}\Delta\text{t}\right) \right] \quad (65)$$

As usual, $\Delta S = .1S$. Then, the filter windows is the same of Eq.(60).

Here too, the size of the filter window can range from 3-by-3 to 33-by-33, with an odd number of cells in both directions. For example, for a mask of 5-by-5, we have the following elements constituting the top row of the filter window (in each instant t):

$$\begin{aligned} U^2 &= S + 2\Delta S \\ &= \left[ \mathrm{U}^2\left(\text{t-N}\Delta\text{t}\right) \ldots\ \mathrm{U}^2\left(\text{t-2}\Delta\text{t}\right)\ \mathrm{U}^2\left(\text{t-}\Delta\text{t}\right)\ \mathrm{U}^2\left(\text{t}\right)\ \mathrm{U}^2\left(\text{t+}\Delta\text{t}\right)\ \mathrm{U}^2\left(\text{t+2}\Delta\text{t}\right) \ldots\ \mathrm{U}^2\left(\text{t+N}\Delta\text{t}\right) \right] \end{aligned} \quad (66)$$

Similarly, the down row of the filter window (in each instant t), we have:

$$\begin{aligned} D^2 &= S - 2\Delta S \\ &= \left[ \mathrm{D}^2\left(\text{t-N}\Delta\text{t}\right) \ldots\ \mathrm{D}^2\left(\text{t-2}\Delta\text{t}\right)\ \mathrm{D}^2\left(\text{t-}\Delta\text{t}\right)\ \mathrm{D}^2\left(\text{t}\right)\ \mathrm{D}^2\left(\text{t+}\Delta\text{t}\right)\ \mathrm{D}^2\left(\text{t+2}\Delta\text{t}\right) \ldots\ \mathrm{D}^2\left(\text{t+N}\Delta\text{t}\right) \right] \end{aligned} \quad (67)$$

Then, the filter windows is the same of Eq.(63).

The following MATLAB® [49] code represents the Mean Filtering (MF) function for a 3x3 kernel for signal denoising.

```
function Sd = mf(S)
T = length(S);
U = 1.1*S;
D = 0.9*S;
Sd = [];
for n = -N+1:N-1
   t = n-(-N+1)+2;
   acuD = 0;
   acuS = 0;
   acuU = 0;
   for aux = -1:1
     acuD = acuD + D(t+aux);
     acuS = acuS + S(t+aux);
     acuU = acuU + U(t+aux);
  end
  Sd(t-1) = (acuD+acuS+acuU)/9;
end
```

With similar considerations about Directional Smoothing regarding to names and sizes.

*Note:* this tool was created by the author for this paper.

## 2.3 Savitzky-Golay Filtering

A ***Savitzky–Golay filter*** **(SGF)** [175] is a digital filter that can be applied to a set of digital data points for the purpose of smoothing the data, that is, to increase the signal-to-noise ratio without greatly distorting the signal. This is achieved, in a process known as convolution, by fitting successive sub-sets of adjacent data points with a low-degree polynomial by the method of linear least squares. When the data points are equally spaced an analytical solution to the least-squares equations can be found, in the form of a single set of "convolution coefficients" that can be applied to all data sub-sets, to give estimates of the smoothed signal, (or derivatives of the smoothed signal) at the central point of each sub-set. The method, based on established mathematical procedures [176, 177], was popularized by Abraham Savitzky and Marcel J. E. Golay who published tables of convolution coefficients for various polynomials and sub-set sizes in 1964 [178, 179]. Some errors in the tables have been corrected [180]. The method has been extended for the treatment of 2- and 3-dimensional data.

Savitzky and Golay's paper is one of the most widely cited papers in the journal *Analytical Chemistry* [181] and is classed by that journal as one of its "10 seminal papers" saying "it can be argued that the dawn of the computer-controlled analytical instrument can be traced to this article" [182].

SGF [183] is a particular type of low-pass filter, well-adapted for data smoothing, and termed variously *Savitzky-Golay* [178], *least-squares* [184], or *DISPO* (Digital Smoothing Polynomial) [185] filters. Rather than having their properties defined in the Fourier domain, and then translated to the time domain, Savitzky-Golay filters derive directly from a particular formulation of the data smoothing problem in the time domain, as we will now see. SGFs were initially (and are still often) used to render visible the relative widths and heights of spectral lines in noisy spectrometric data.

Recall that a digital filter is applied to a series of equally spaced data values $f_i \equiv f(t_i)$, where $t_i \equiv t_0 + i\Delta$ for some constant sample spacing $\Delta$ and $i = \ldots -2, -1, 0, 1, 2, \ldots$. We have seen [183] that the simplest type of digital filter (the nonrecursive or finite impulse response filter) replaces each data value $f_i$ by a linear combination $g_i$ of itself and some number of nearby neighbors,

$$g_i = \sum_{n=-n_L}^{n_R} c_n \, f_{i+n} \tag{68}$$

Here $n_L$ is the number of points used "to the left" of a data point $i$, i.e., earlier than it, while $n_R$ is the number used to the right, i.e., later. A so-called *causal* filter would have $n_R = 0$.

As a starting point for understanding SGFs, consider the simplest possible averaging procedure: For some fixed $n_L = n_R$, compute each $g_i$ as the average of the data points from $f_{i-n_L}$ to $f_{i+n_R}$. This is sometimes called *moving window averaging* and corresponds to equation (68) with constant $c_n = 1/(n_L + n_R + 1)$. If the underlying function is constant, or is changing linearly with time (increasing or decreasing), then no bias is introduced into the result. Higher points at one end of the averaging interval are on the average balanced by lower points at the other end. A bias is introduced, however, if the underlying function has a nonzero second derivative. At a local maximum, for example, moving window averaging always reduces the function value. In the spectrometric application, a narrow spectral line has its height reduced and its width increased. Since these parameters are themselves of physical interest, the bias introduced is distinctly undesirable.

Note, however, that moving window averaging does preserve the area under a spectral line, which is its zeroth moment, and also (if the window is symmetric with $n_L = n_R$) its mean position in time, which is its first moment. What is violated is the second moment, equivalent to the line width.

The idea of Savitzky-Golay filtering is to find filter coefficients $c_n$ that preserve higher moments. Equivalently, the idea is to approximate the underlying function within the moving window not by a constant (whose estimate is the average), but by a polynomial of higher order, typically quadratic or quartic: For each point $f_i$, we least-squares fit a polynomial to all $n_L + n_R + 1$ points in the moving window, and then set $g_i$ to be the value of that polynomial at position $i$. We make no use of the value of the polynomial at any other point. When we move on to the next point $f_{i+1}$, we do a whole new least-squares fit using a shifted window.

All these least-squares fits would be laborious if done as described. Luckily, since the process of least-squares fitting involves only a linear matrix inversion, the coefficients of a fitted polynomial are themselves linear in the values of the data. That means that we can do all the fitting in advance, for fictitious data consisting of all zeros except for a single 1, and then do the fits on the real data just by taking linear combinations. This is the key point, then: There are particular sets of filter coefficients $c_n$ for which equation (68) "automatically" accomplishes the process of polynomial least-squares fitting inside a moving window.

To derive such coefficients, consider how $g_0$ might be obtained: We want to fit a polynomial of degree $M$ in $i$, namely $a_0 + a_1 i + \cdot \cdot \cdot + a_M i^M$ to the values $f_{-n_L}, \ldots, f_{n_R}$. Then $g_0$ will be the value of that polynomial at $i = 0$, namely $a_0$. The design matrix for this problem is

$$A_{ij} = i^j \qquad i = -n_L, \ldots, n_R, \qquad j = 0, \ldots, M \tag{69}$$

and the normal equations for the vector of $a_j$'s in terms of the vector of $f_i$'s is in matrix notation

$$\left(\mathrm{A^T A}\right)\mathrm{a} = \mathrm{A\,f} \qquad \text{or} \qquad \mathrm{a} = \left(\mathrm{A^T A}\right)^{-1}\left(\mathrm{A^T f}\right) \tag{70}$$

We also have the specific forms

$$\left\{\mathrm{A^T A}\right\}_{\mathrm{ij}} = \sum_{k=-n_L}^{n_R} A_{ki} A_{kj} = \sum_{k=-n_L}^{n_R} k^{i+j} \tag{71}$$

and

$$\left\{\mathrm{A^T f}\right\}_{\mathrm{j}} = \sum_{k=-n_L}^{n_R} A_{kj} f_k = \sum_{k=-n_L}^{n_R} k^j f_k \tag{72}$$

Since the coefficient $c_n$ is the component $a_0$ when $\mathbf{f}$ is replaced by the unit vector $\mathbf{e}_n$, $-n_L \leq n < n_R$, we have

$$c_n = \left\{ \left( \mathrm{A}^{\mathrm{T}} \mathrm{A} \right)^{-1} \left( \mathrm{A}^{\mathrm{T}} \mathrm{e}_n \right) \right\}_0 = \sum_{m=0}^{M} \left\{ \left( \mathrm{A}^{\mathrm{T}} \mathrm{A} \right)^{-1} \right\}_{0m} n^m \tag{73}$$

Note that equation (73) says that we need only one row of the inverse matrix. Numerically we can get this by *LU* decomposition with only a single backsubstitution [183].

The accompanying Table I shows some typical Savitzky-Golay coefficients [183]. For orders 2 and 4, the coefficients of SGFs with several choices of $n_L$ and $n_R$ are shown. The central column is the coefficient applied to the data $f_i$ in obtaining the smoothed $g_i$. Coefficients to the left are applied to earlier data; to the right, to later. The coefficients always add (within roundoff error) to unity. One sees that, as befits a smoothing operator, the coefficients always have a central positive lobe, but with smaller, outlying corrections of both positive and negative sign. In practice, the SGFs are most useful for much larger values of $n_L$ and $n_R$, since these few-point formulas can accomplish only a relatively small amount of smoothing.

TABLE I
SAMPLE SAVITZKY-GOLAY COEFFICIENTS.

| $M$ | $n_L$ | $n_R$ | Coefficients | | | | | | | | | | |
|---|---|---|---|---|---|---|---|---|---|---|---|---|---|
| 2 | 2 | 2 | | | | -0.086 | 0.343 | 0.486 | 0.343 | -0.086 | | | |
| 2 | 3 | 1 | | | -0.143 | 0.171 | 0.343 | 0.371 | 0.257 | | | | |
| 2 | 4 | 0 | | 0.086 | -0.143 | -0.086 | 0.257 | 0.886 | | | | | |
| 2 | 5 | 5 | -0.084 | 0.021 | 0.103 | 0.161 | 0.196 | 0.207 | 0.196 | 0.161 | 0.103 | 0.021 | -0.084 |
| 4 | 4 | 4 | | 0.035 | -0.128 | 0.070 | 0.315 | 0.417 | 0.315 | 0.070 | -0.128 | 0.035 | |
| 4 | 5 | 5 | 0.042 | -0.105 | -0.023 | 0.140 | 0.280 | 0.333 | 0.280 | 0.140 | -0.023 | -0.105 | 0.042 |

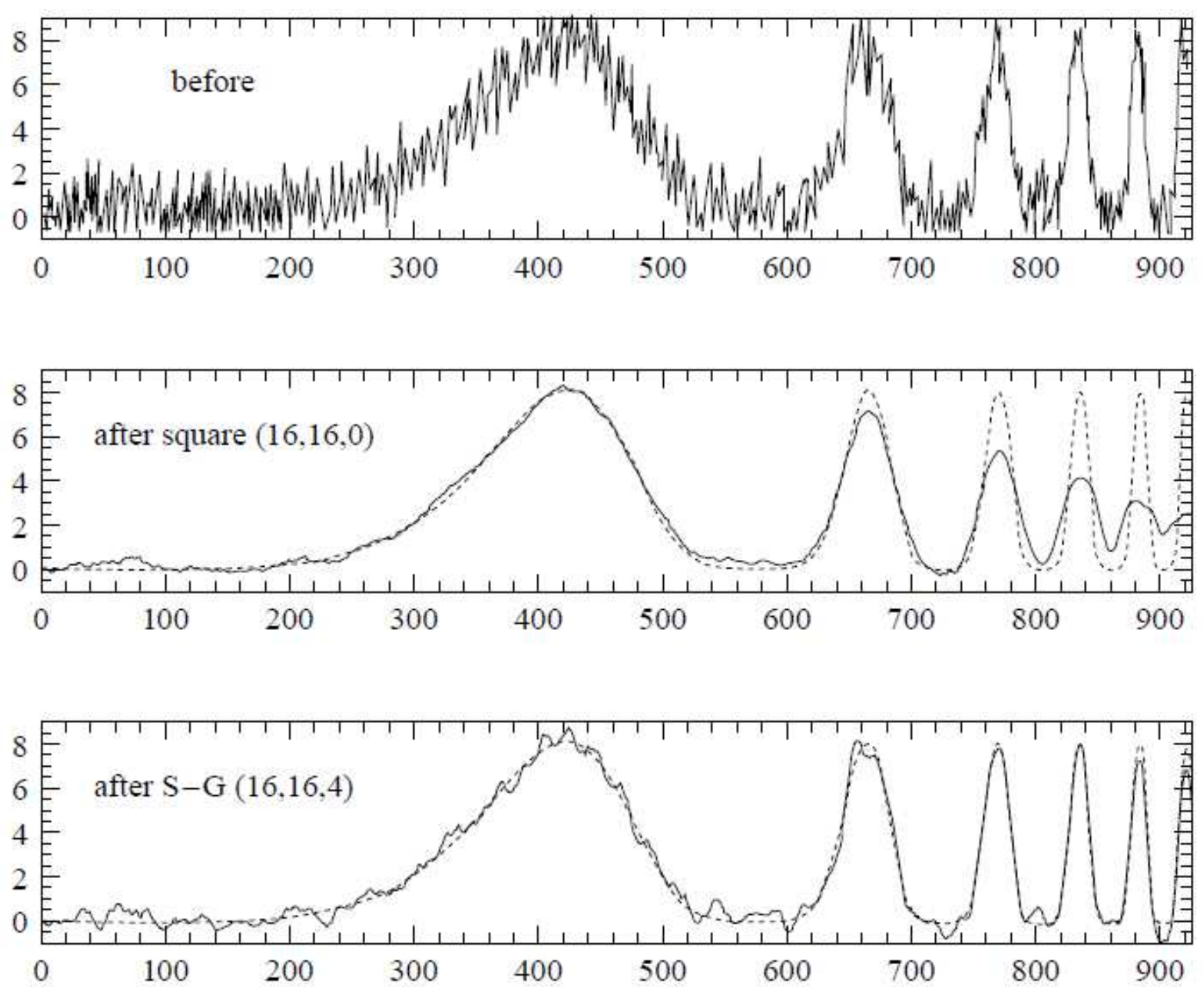

**Fig.30** Top: Synthetic noisy data consisting of a sequence of progressively narrower bumps, and additive Gaussian white noise. Center: Result of smoothing the data by a simple moving window average. The window extends 16 points leftward and rightward, for a total of 33 points. Note that narrow features are broadened and suffer corresponding loss of amplitude. The dotted curve is the underlying function used to generate the synthetic data. Bottom: Result of smoothing the data by a Savitzky-Golay smoothing filter (of degree 4) using the same 33 points. While there is less smoothing of the broadest feature, narrower features have their heights and widths preserved.

Figure 30 shows a numerical experiment using a 33 point smoothing filter, that is, $n_L = n_R = 16$. The upper panel shows a test function, constructed to have six “bumps” of varying widths, all of height 8 units. To this function Gaussian white noise of unit variance has been added. The test function without noise is shown as the dotted curves in the center and lower panels. The widths of the bumps (full width at half of maximum, or FWHM) are 140, 43, 24, 17, 13, and 10, respectively. The middle panel of Figure 30 shows the result of smoothing by a moving window average. One sees that the window of width 33 does QITe a nice job of smoothing the broadest bump, but that the narrower bumps suffer considerable loss of height and increase of width. The underlying signal (dotted) is very badly represented. The lower panel shows the result of smoothing with a SGF of the identical width, and degree $M = 4$. One sees that the heights and widths of the bumps are QITe extraordinarily preserved. A trade-off is that the broadest bump is less smoothed. That is because the central positive lobe of the SGF coefficients fills only a fraction of the full 33 point width. As a rough guideline, best results are obtained when the full width of the degree 4 SGF is between 1 and 2 times the FWHM of desired features in the data [185, 186]. Figure 31 shows the result of smoothing the same noisy “data” with broader SGFs of 3 different orders. Here we have $n_L = n_R = 32$ (65 point filter) and $M = 2$, 4, 6. One sees that, when the bumps are too narrow with respect to the filter size, then even the SGF must at some point give out. The higher order filter manages to track narrower features, but at the cost of less smoothing on broad features. The accompanying Table II shows some typical Savitzky-Golay coefficients for Fig.31.

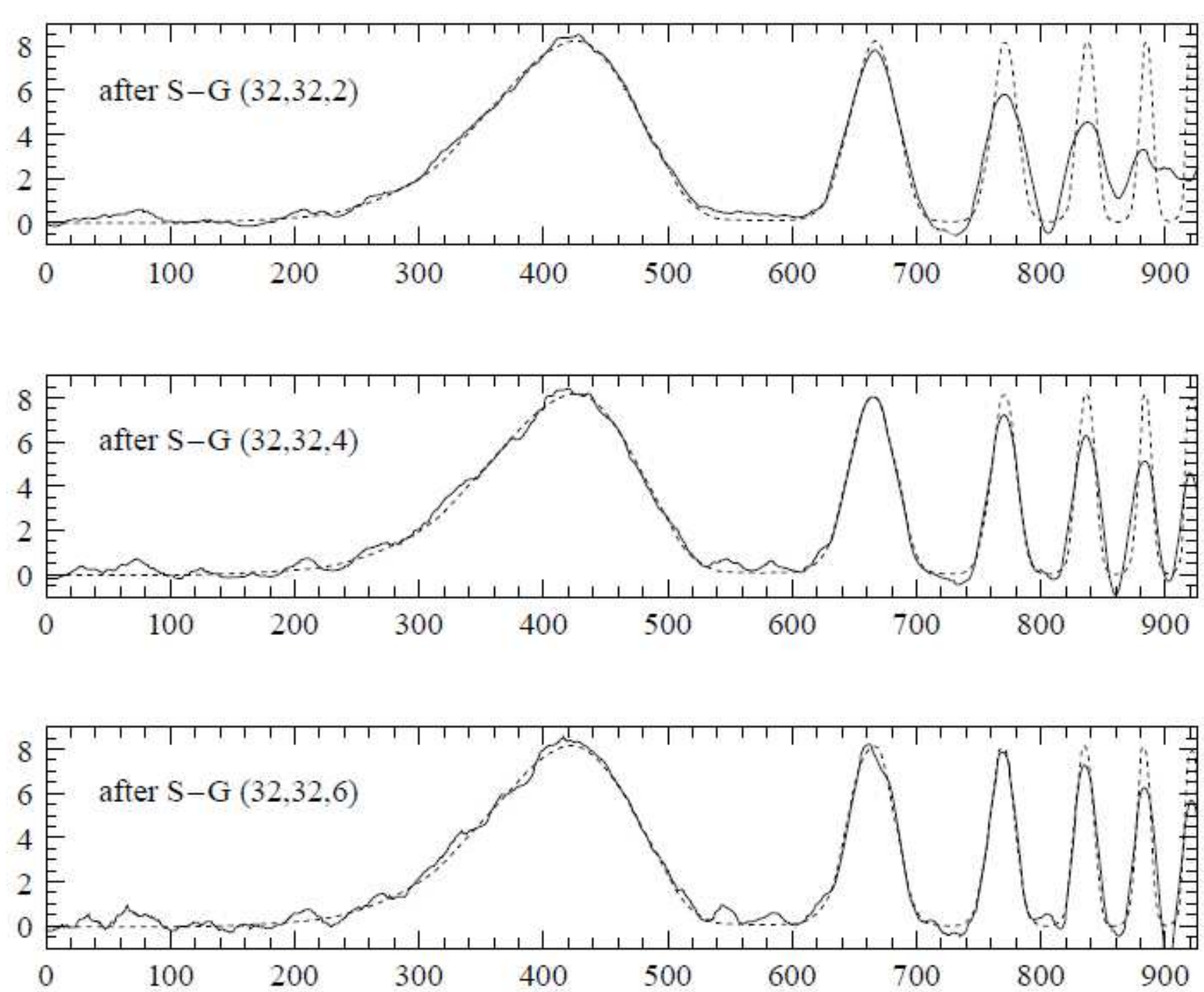

**Fig.31** Result of applying wider 65 point Savitzky-Golay filters to the same data set as in Figure 30. Top: degree 2. Center: degree 4. Bottom: degree 6. All of these filters are inoptimally broad for the resolution of the narrow features. Higher-order filters do best at preserving feature heights and widths, but do less smoothing on broader features.

TABLE II

PATCH OF IMAGE DATA f(i) WITH ITS LOCAL COORDINATE SYSTEM.

| | | $x_i$ | | | | |
|---|---|---|---|---|---|---|
| | | -2 | -1 | 0 | 1 | 2 |
| $y_i$ | -2 | f(0) | f(1) | f(2) | f(3) | f(4) |
| | -1 | f(5) | f(6) | f(7) | f(8) | f(9) |
| | 0 | f(10) | f(11) | f(12) | f(13) | f(14) |
| | 1 | f(15) | f(16) | f(17) | f(18) | f(19) |
| | 2 | f(20) | f(21) | f(22) | f(23) | f(24) |

***Example of How to Compute 2D SGF*** - As an example, suppose we want to smooth or differentiate an image based on 5x5 image patches [187]. The image patch is laid out as follows:

f(i) is the pixel value, and the column vector $\underline{f}$ represents all the image data, i.e.

$$f = \left[ f(0)\; f(1)\; \ldots\; f(24) \right]^T \tag{74}$$

We want to fit a 3[rd] order, two-dimensional polynomial to this data. The polynomial is

$$\begin{aligned} f(i) &\approx F(x_i, y_i) \\ &= a_{00} + a_{10}x_i + a_{01}y_i + a_{20}x_i^2 + a_{11}x_i y_i + a_{02}y_i^2 + a_{30}x_i^3 + a_{21}x_i^2 y_i + a_{12}x_i y_i^2 + a_{03}y_i^3 \end{aligned} \tag{75}$$

Note that the coefficient of $x^i y^j$ is $a_{ij}$ . $(x_i, y_i)$ is the pixel coordinate of f(i). To compute the coefficients from the data we set up a matrix equation:

$$A\, a = f \tag{76}$$

where

$$A = \begin{bmatrix} 1 & x_0 & y_0 & x_0^2 & x_0 y_0 & y_0^2 & x_0^3 & x_0^2 y_0 & x_0 y_0^2 & y_0^3 \\ 1 & x_1 & y_1 & x_1^2 & x_1 y_1 & y_1^2 & x_1^3 & x_1^2 y_1 & x_0 y_1^2 & y_1^3 \\ \vdots & \vdots & \vdots & \vdots & \vdots & \vdots & \vdots & \vdots & \vdots & \vdots \\ 1 & x_{24} & y_{24} & x_{24}^2 & x_{24} y_{24} & y_{24}^2 & x_{24}^3 & x_{24}^2 y_{24} & x_0 y_{24}^2 & y_{24}^3 \end{bmatrix} \tag{77}$$

and $\underline{a}$ is the vector of polynomial coefficients:

$$a = \left[ a_{00}\; a_{10}\; a_{01}\; a_{20}\; a_{11}\; a_{02}\; a_{30}\; a_{21}\; a_{12}\; a_{03} \right]^T \tag{78}$$

Equation (76) simply reproduces the polynomial for each pixel in the image patch. We solve for the polynomial coefficients using least squares:

$$a = \left( A^T A \right)^{-1} A^T f \qquad \text{(identical to Eq.70)} \tag{79}$$

$C = \left( A^T A \right)^{-1} A^T$ is the pseudo-inverse of A, and it is independent of the image data. Each polynomial coefficient is computed as the inner product of one row of C and the column of pixel values $\underline{f}$. This is the surprising part about SGFs: the polynomial coefficients are computed using a linear filter on the data. Just as one can reassemble $\underline{f}$ back into a rectangular patch of pixels, one can also assemble each row of C into the same size rectangle to get a traditional-looking image filter. To complete this example, here are the filters for the first three polynomial coefficients. We will use the naming convention that coefficient $a_{ij}$ is computed from rectangular filter $C_{ij}$. The first three filters are:

$$C_{00} = \begin{bmatrix} -0.0743 & 0.0114 & 0.0400 & 0.0114 & -0.0743 \\ 0.0114 & 0.0971 & 0.1257 & 0.0971 & 0.0114 \\ 0.0400 & 0.1257 & 0.1543 & 0.1257 & 0.0400 \\ 0.0114 & 0.0971 & 0.1257 & 0.0971 & 0.0114 \\ -0.0743 & 0.0114 & 0.0400 & 0.0114 & -0.0743 \end{bmatrix} \tag{80}$$

$$C_{10} = \begin{bmatrix} 0.0738 & -0.0119 & -0.0405 & -0.0119 & 0.0738 \\ -0.1048 & -0.1476 & -0.1619 & -0.1476 & -0.1048 \\ 0 & 0 & 0 & 0 & 0 \\ 0.1048 & 0.1476 & 0.1619 & 0.1476 & 0.1048 \\ -0.0738 & 0.0119 & 0.0405 & 0.0119 & -0.0738 \end{bmatrix} \tag{81}$$

$$C_{01} = \begin{bmatrix} 0.0738 & -0.1048 & 0 & 0.1048 & -0.0738 \\ -0.0119 & -0.1476 & 0 & 0.1476 & 0.0119 \\ -0.0405 & -0.1619 & 0 & 0.1619 & 0.0405 \\ -0.0119 & -0.1476 & 0 & 0.1476 & 0.0119 \\ 0.0738 & -0.1048 & 0 & 0.1048 & -0.0738 \end{bmatrix} \tag{82}$$

***Example of How to Use 2D SGF*** - Suppose we want to smooth an image. Using a SGFs, we are conceptually fitting a two-dimensional polynomial to the image patch surrounding each pixel and then evaluating this polynomial [187]. The local coordinate system that we use for the image patch (see Table II) has (x,y) = (0,0) at the pixel of interest in the middle of the patch. Thus to compute the smoothed value of the pixel, we just evaluate the polynomial at (x,y) = (0,0). This turns out to be merely $a_{00}$, which we can compute by applying filter $C_{00}$ to the image patch.

Suppose we want to compute partial derivatives on the patch. The two partial derivatives of the fitted polynomial are

$$\begin{aligned} F_x(x_i, y_i) &= a_{10} + 2a_{20}x_i + a_{11}y_i + 3a_{30}x_i^2 + 2a_{21}x_iy_i + a_{12}y_i^2 \\ F_y(x_i, y_i) &= a_{01} + a_{11}x_i + 2a_{02}y_i + a_{21}x_i^2 + 2a_{12}x_iy_i + 3a_{03}y_i^2 \end{aligned} \tag{83}$$

Evaluating at (x,y) = (0,0), the results are simply $F_x(0,0) = a_{10}$ and $F_y(0,0) = a_{01}$, which are computed with filters $C_{10}$ and $C_{01}$ above.

### 2.4 Superresolution in general, and for still images in particular

***Introduction*** - Digital image capture produces discrete representations of continuous scenes [188]. This discretisation in both space and intensity is a sampling process that creates aliasing, and information at frequencies above the Nyquist rate is lost. It is common to wish to construct a higher resolution image from a template image or a set of images, but the aliasing and loss of frequency information makes this an ill-posed (inverse) problem.

The typical solution to this problem (known in the literature as image super-resolution reconstruction, or simply superresolution) is to use an ensemble of related lower-resolution images. As each of these images has aliased the higher frequency information slightly differently, under certain conditions it is possible to "unwrap" some of the aliasing and reconstruct the lost higher frequencies.

There are numerous methods [189-193] of performing super-resolution. Many of them are computationally expensive in nature, but allow for complicated motion models, significant noise and image degradation, and other aspects that are not considered in this work. Given assumptions of global translational motion, low noise and linear space and time invariant blur due to the imaging sensor point spread function (PSF), image superresolution reconstruction can be split into there distinct steps:

- Registration
- Reconstruction/Interpolation
- Deblurring

Image registration is a technique that can be used to determine the relative translations between the input images. Generally, the desire is to do this from the contents of the images alone, without any prior knowledge. There are many different methods for performing registration [189]; however, in the context of image superresolution, image registration is required to determine the offsets between the images with accuracy down to a small fraction of a pixel [189].

Once the images have been registered, all the pixels from the ensemble can be combined to form a composite image. The resultant image is no longer sampled on a uniform rectangular grid, but due to global translational motion, it has a semi-uniform structure, as can be seen in Figure 32. Reconstructing the image data at all points on a high resolution grid requires that the semi-regular data is interpolated and resampled. It is this interpolation problem that is addressed in this paper.

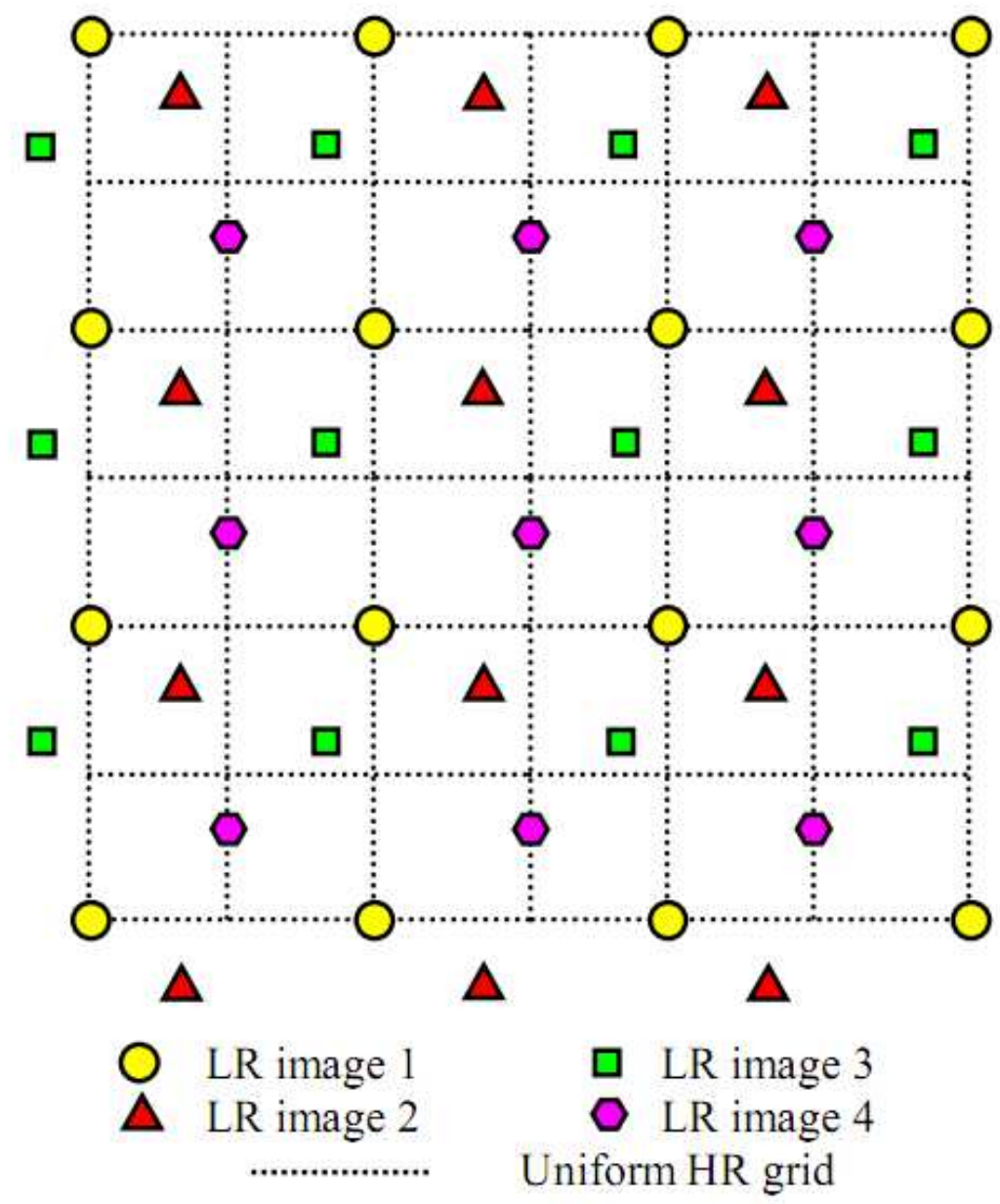

**Fig.32** Composite image exhibits a semi-uniform structure.

For the full super-resolution approach, a deblurring procedure can now be applied that restores the high frequencies that have been suppressed by the low-resolution imaging process. In this paper we perform this deblurring after the interpolated process via a convolution between the up-sampled image and a convolution mask deduced specifically.

However, the main goal of Super-Resolution (SR) methods –in practice– is to recover a high-resolution image from one or more low-resolution input images. Methods for SR can be broadly classified into two families of methods:

- The classical multi-image super-resolution, and
- Example-Based super-resolution.

In the classical multi-image SR (e.g., [190] to name just a few) asset of low-resolution images of the same scene are taken (at subpixel misalignments). Each low-resolution image imposes asset of linear constraints on the unknown high-resolution intensity values. If enough low-resolution images are available (at subpixel shifts), then the set of equations becomes determined and can be solved to recover the high-resolution image. Practically, however, this approach is numerically limited only to small increases in resolution [190] (by factors smaller than 2).

These limitations have lead to the development of "Example-Based Super-Resolution" also termed "image hallucination" (see references in [190]). In example-based SR, correspondences between low and high-resolution image patches are learned from a data base of low and high-resolution image pairs (usually with a relative scale factor of 2), and then applied to a new low-resolution image to recover its most likely high-resolution version. Higher SR factors have often been obtained by repeated applications of this process. Example-based SR has been shown to exceed the limits of classical SR. However, unlike classical SR, the high resolution details reconstructed ("hallucinated") by example-based SR are not guaranteed to provide the true (unknown) high-resolution details.

Sophisticated methods for image up-scaling based on learning edge models have also been proposed (e.g., [191-193]). The goal of these methods is to magnify (up-scale) an image while maintaining the sharpness of the edges and the details in the image. In contrast, in SR (example-based as well as classical) the goal is to recover new *missing high-resolution details* that are not explicitly found in any individual low-resolution image (details beyond the Nyquist frequency of the low-resolution image). In the classical SR, this high-frequency information is assumed to be split across multiple low-resolution images, implicitly found there in aliased form. In example-based SR, this missing high-resolution information is assumed to be available in the high-resolution data base patches, and learned from the low-res/high-res pairs of examples in the database. However beyond what we have said above, in this paper, we consider the interpolation of a single image.

On the other hand, the rendering of lower resolution image data on higher resolution displays has become a very common task, in particular because of the increasing popularity of webcams, camera phones, and low-bandwidth video streaming. Thus, there is a strong demand for real-time, high-quality image magnification. In this work, we suggest to exploit the high performance of general-purpose computation on programmable graphics processing units (GPGPUs) for an original image magnification method. To this end, we propose a GPGPU-friendly algorithm for image up-sampling with edge restoration image interpolation, which avoids ringing artifacts, excessive blurring, and stair-casing of oblique edges. At the same time it features gray-scale invariance, is applicable to color images, and allows for real-time processing of full-screen images on today's GPGPUs [194-197].

***Two-dimensional Interpolation*** - A digital image is not an exact snapshot of reality, it is only a discrete approximation. This fact should be apparent to the average web surfer, as images commonly become blocky or jagged after being resized to fit the browser. The goal of image interpolation is to produce acceptable images at different resolutions from a single low-resolution image. The actual resolution of an image is defined as the number of pixels, but the effective resolution is a much harder quantity to define as it depends on subjective human judgment and perception. The goal of this section is to explore different mathematical formulations of this essentially aesthetic quantity.

The image interpolation problem goes by many names, depending on the application: image resizing, image up-sampling/down-sampling, digital zooming, image magnification, resolution enhancement, etc. The term super-resolution is sometimes used, although in the literature this generally refers to producing a high-resolution image from multiple images such as a video sequence. If we define interpolation as "filling in the pixels in between," the image interpolation problem can be viewed as a subset of the inpainting problem (see Figure 33).

The applications of image interpolation range from the common place viewing of online images to the more sophisticated magnification of satellite images. With the rise of consumer-based digital photography, users expect to have a greater control over their digital images. Digital zooming has a role in picking up clues and details in surveillance images and video. As high-definition television (HDTV) technology enters the marketplace, engineers are interested in fast interpolation algorithms for viewing traditional low-definition programs on HDTV.

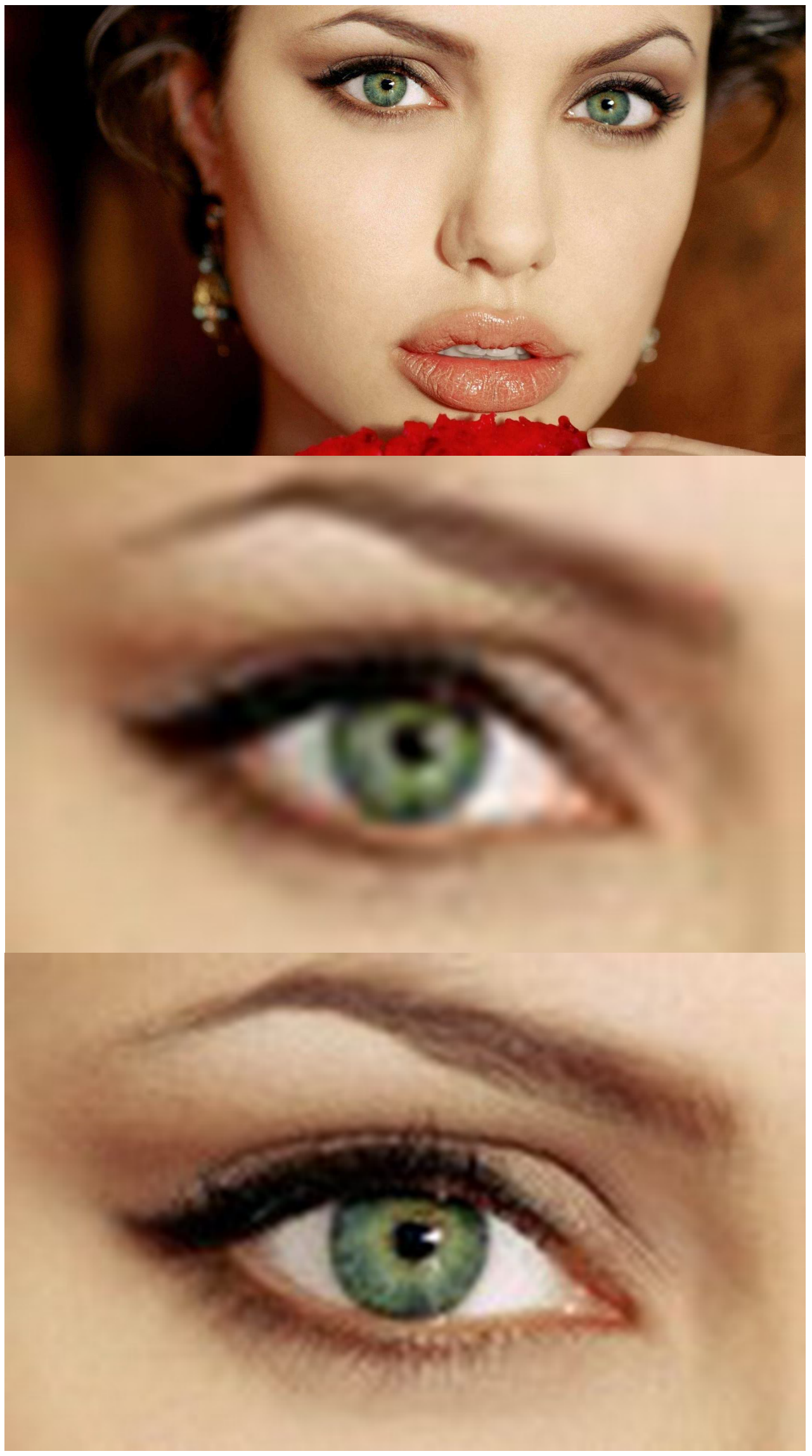

**Fig.33** Image interpolation using linear method of interp2 built-in MATLAB® function. Top: original image. Medium: close-up of eye in image. Down: interpolated image.

Astronomical images from rovers and probes are received at an extremely low transmission rate (about 40 bytes per second), making the transmission of high-resolution data infeasible [198]. In medical imaging, neurologists would like to have the ability to zoom in on specific parts of brain tomography images. This is just a short list of applications, but the wide variety cautions us that our desired interpolation result could vary depending on the application and user.

***The Image Interpolation Problem*** - In this section, we will establish the notation for image interpolation used throughout the paper. Suppose our image is defined over some rectangle $\Omega \subset \mathbb{R}^2$. Let the function $f : \Omega \rightarrow \mathbb{R}$ be our ideal continuous image. In an abstract sense, we can think of f as being "reality" and $\Omega$ as our "viewing window". Our observed image $u_0$ is a discrete sampling of *f* at equally spaced points in the plane. If we suppose the resolution of $u_0$ is $\delta x \times \delta y$, we can express $u_0$ by

$$u_0(x, y) = C_{\delta x, \delta y}(x, y) f(x, y), \qquad (x, y) \in \Omega \tag{84}$$

where $C$ denotes the Dirac comb:

$$C_{\delta x,\delta y}(x,y)=\sum_{k,l\in Z}\delta(k\delta x,l\delta y),\qquad (x,y)\in\mathbb{R}^2. \tag{85}$$

The goal of image interpolation is to produce an image $u$ at a different resolution $\delta x'\times\delta y'$. For simplicity, we will assume that the Euclidean coordinates are scaled by the same factor $K$:

$$u(x,y)=C_{\frac{\delta x}{K},\frac{\delta y}{K}}(x,y)f(x,y),\qquad (x,y)\in\Omega. \tag{86}$$

Given only the image $u_0$, we will have to devise some reconstruction of $f$ at the pixel values specified by this new resolution. We will refer to $K$ as our zoom or magnification factor. Obviously, if $K = 1$ we trivially recover $u_0$ .The image $u_0$ is upsampled if $K\uparrow 1$ and downsampled if $K\downarrow 1$. In this paper, we will focus on the upsampling case when $K\uparrow 1$ is an integer.

Let $\Omega_K\subset\Omega$ denote the lattice induced by (86) for a fixed zoom $K$. Note that the lattice of the original image $u_0$ in (85) is $\Omega_1$. Also note that for infinite magnification we obtain $\Omega_K\subset\Omega$ as $K\to\infty$. For computation purposes, we can shift the lattices to the positive integers. So if the observed image $u_0$ is an $m$ x $n$ image,

$$\Omega_K=[1,2,\ldots,Km]\times[1,2,\ldots,Kn],\qquad K\in Z_+. \tag{87}$$

Many interpolation techniques impose the constraint $\Omega_1\subseteq\Omega_K$. In this case, only a subset of the pixels in $\Omega_K$ needs to be determined and the interpolation problem becomes a version of the inpainting problem.

Given the notation above, we can state the image interpolation problem succinctly: Given a low-resolution image $u_0:\Omega_1\to\mathbb{R}$ and a zoom $K\uparrow 1$, find a high-resolution image $u:\Omega_K\to\mathbb{R}$. Obviously, this is an ill-posed problem. We need to impose assumptions on the reconstruction of $f$ in equation (86). The choice of interpolation technique depends on the choice of assumptions. In other words, we need a mathematical understanding of what constitutes our perception of "reality" $f$. Interpolation methods differ in their mathematical description of a "good" interpolated image. Although it is difficult to compare methods and judge their output, [198] proposes 9 basic criteria for a good interpolation method. The first 8 are visual properties of the interpolated image, the last is a computational property of the interpolation method.

1) Geometric Invariance: The interpolation method should preserve the geometry and relative sizes of objects in an image. That is, the subject matter should not change under interpolation.
2) Contrast Invariance: The method should preserve the luminance values of objects in an image and the overall contrast of the image.
3) Noise: The method should not add noise or other artifacts to the image, such as ringing artifacts near the boundaries.
4) Edge Preservation: The method should preserve edges and boundaries, sharpening them where possible.
5) Aliasing: The method should not produce jagged or "staircase" edges.
6) Texture Preservation: The method should not blur or smooth textured regions.
7) Over-smoothing: The method should not produce undesirable piecewise constant or blocky regions.
8) Application Awareness: The method should produce results appropriate to the type of image and order of resolution. For example, the interpolated results should appear realistic for photographic images, but for medical images the results should have crisp edges and high contrast. If the interpolation is for general images, the method should be independent of the type of image.
9) Sensitivity to Parameters: The method should not be too sensitive to internal parameters that may vary from image to image.

Of course, these are qualitative and somewhat subjective criteria. We unlike [198], do not hope to develop a mathematical model of image interpolation and error analysis, but simply apply the most efficient method for our development. In a sense, the method employed in this paper presents a mathematical model of these visual criteria.

***Linear Interpolation Filters*** - The simplest approach is to assume that $f$ in equation (86) is reconstructed by a convolution kernel $\phi : \mathbb{R}^2 \to \mathbb{R}$ where $\int \varphi(x, y)\, dy\, dx = 1$. Then we can approximate $f$ by

$$f \approx u_0 * \varphi. \tag{88}$$

Substituting this into (86) gives rise to a general linear interpolation filter

$$u(x, y) = C_{\frac{\delta x}{K}, \frac{\delta y}{K}}(x, y).(u_0 * \varphi)(x, y), \qquad (x, y) \in \Omega. \tag{89}$$

The simplest linear filters are the bilinear and bicubic interpolation, which assume the pixel values can be fit locally to linear and cubic functions, respectively [198]. Along with simple nearest neighbor interpolation, these two filters are the most common interpolation schemes in commercial software. These methods are easy to code as matrix multiplications of $u_0$. However, an image contains edges and texture, in other words discontinuities. So the assumptions that pixel values locally fit a polynomial function will produce undesirable results. The bilinear and bicubic interpolation methods [198] may introduced blurring, create ringing artifacts, and produce a jagged aliasing effect along edges (see Fig.34). The blurring effects arise from the fact that the methods compute a weighted average of nearby pixels, just as in Gaussian blurring. The aliasing effects arise because the linear filters do not take into consideration the presence of edges or how to reconstruct them.

Other linear interpolation filters include quadratic zoom, the B-spline method, and zero-padding. But these schemes produce the same undesirable effects as the bilinear and bicubic methods, as mentioned in [198]. Linear filters differ in the choice of $\varphi$, which essentially determine show to compute the weighted average of nearby pixels. While this is a natural interpolation scheme for general data sets, this is not necessarily appropriate for *visual* data. In order to improve upon these linear filters, we need to consider interpolation methods that some how quantify and preserve visual information.

***Which Methods to Consider?*** - Generally speaking, mathematical approaches to image processing can be divided into five categories:

1) Partial-Differential Equation (PDE)-Based Methods (e.g. heat diffusion, Perona-Malik, Navier-Stokes, and mean curvature).
2) Variations of Energy (e.g. Total Variation, Mumford-Shah, active contours)
3) Multiscale Analysis (e.g. wavelets, Fourier analysis, Gabor analysis, Laplacian pyramids)
4) Machine Learning (e.g. unsupervised learning, data mining, Markov networks)
5) Statistical / Probabilistic Methods (e.g. Bayesian inference, Natural Scene Statistics, pattern theory)

We are trying to describe the field in broad terms, but not to rank or pigeonhole work in computer vision. Indeed, many techniques such as TV-wavelets inpainting certainly do not fit into one category. Also, these methods differ at the mathematical level, but not necessarily at the conceptual level. For example, some versions of the TV energy can be minimized by solving a PDE or by optimizing a variation of energy.

In our attempt to survey recent work in image interpolation and also display the variety of mathematics used, we will highlight one method from each of the five categories [198]

1) A PDE-Based Approach: anisotropic heat diffusion
2) A Variation of Energy Approach: Mumford-Shah inpainting
3) A Multiscale Approach: wavelet-based interpolation
4) A Machine Learning Approach: LLE-based neighbor embeddings
5) A Statistical Approach: NL-means interpolation

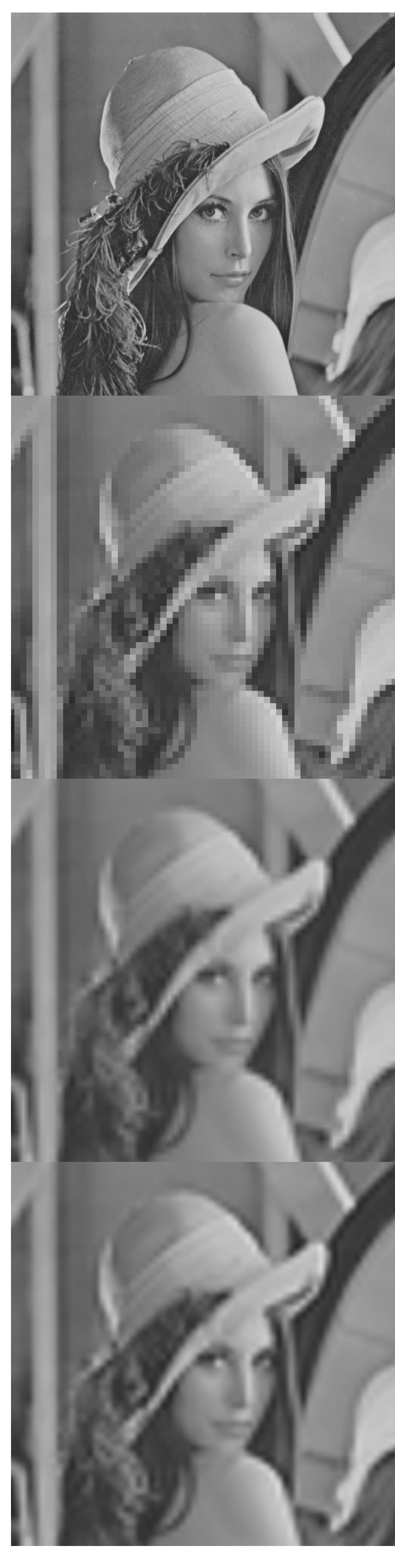

**Fig.34** Part of Lena image down-sampled and then up-sampled by factor K = 16. Top: Original, Second: Nearest Neighbor, Third: Bilinear, and Down: Bicubic.

These methods are, in some sense, representative of the mathematical approaches to the image interpolation problem and, in a larger sense, to the field of image processing. For example, the heat equation is the most studied PDE in image processing and the Mumford-Shah energy has generated dozens, if not hundreds, of research papers. However, these methods have a number of disadvantages to its implementation, reason, we choose a configuration based on linear interpolation. The main disadvantages are:

1) Their hard coding
2) They depend heavily on initial conditions
3) Their high computational complexity
4) Their visual quality is not superior to linear interpolation, except for high levels of downsampling/upsampling, with automatically means a high rate of compression/decompression.

In the latter case, we use a convolution mask [46, 162, 166, 171] to enhance the edges, as discussed in the next section.

***Super-resolution Scheme for Compression*** - This subsection is organized into four parts, for a better understanding of the concepts:

1) Super-resolution vs Deblurring,
2) Compression vs Super-compression,
3) Deduction of the mask
4) Applications

The latter two can be seen in [188].

***Super-resolution vs Deblurring*** - As we saw in the Introduction of this section, there is much confusion between the concepts of superresolution and deblurring in Digital Image Processing [45, 46]. We are going to establish here two rigorous definitions for the purpose of eliminating this confusion.

*We say that a process is super-resolution if it restores the sharpness of an image involving an increase in the resolution of the same* [188-193, 199-201].

*We say that a process is deblurring if it restores the sharpness of an image not involving an increase in the resolution of the same. This process is applied when the image sharpness suffers an aberration called blur* [45, 46]*, which comes from a high relative speed of the object in focus in relation to the camera, fast opening and closing the shutter, etc.*

We consider important to mention that both processes can involve each other as part of the process of improving the sharpness of the image. In fact, we can understand the superresolution as a process of increasing the resolution followed by a restoration of the edges by a deblurring process. On the other hand, previously established definitions are fundamental to understanding what follows.

***Compression vs Super-compression*** - We define compression as the process reduces the average number of bit-per-pixel (bpp) of an image. In Fig. 35, we represent the set of bit-planes in which decomposes a gray or color image. As seen in Fig. 35, the compression process does not alter the image size [45, 46, 188].

Instead, we define supercompression as the process reduces the average number of bit-per-pixel (bpp) of an image after downsizing. The size reduction process is performed by down-sampling, which takes shrinkage in rows and columns, without obligation to respect the aspect ratio [202]. When we say, *we increase the standard compression 5 times*, this means that we move from a resolution of ROWxCOL (high-resolution) to another 5 times lower of original, i.e., ROWxCOL/5 (low-resolution). The standard image compression, e.g., Join Picture Expert Group (JPEG, [188, 45-48]) is not affected by the supercompression. As discussed in [188], supercompression requires minimal eQImPment at the encoder and the reverse procedure to supercompression in the decoder [188]. In Fig. 36, we represent the set of bit-planes in which decomposes a gray or color image.

As discussed in [188], our supercompression procedure consists in two parts spread in encoder and decoder.

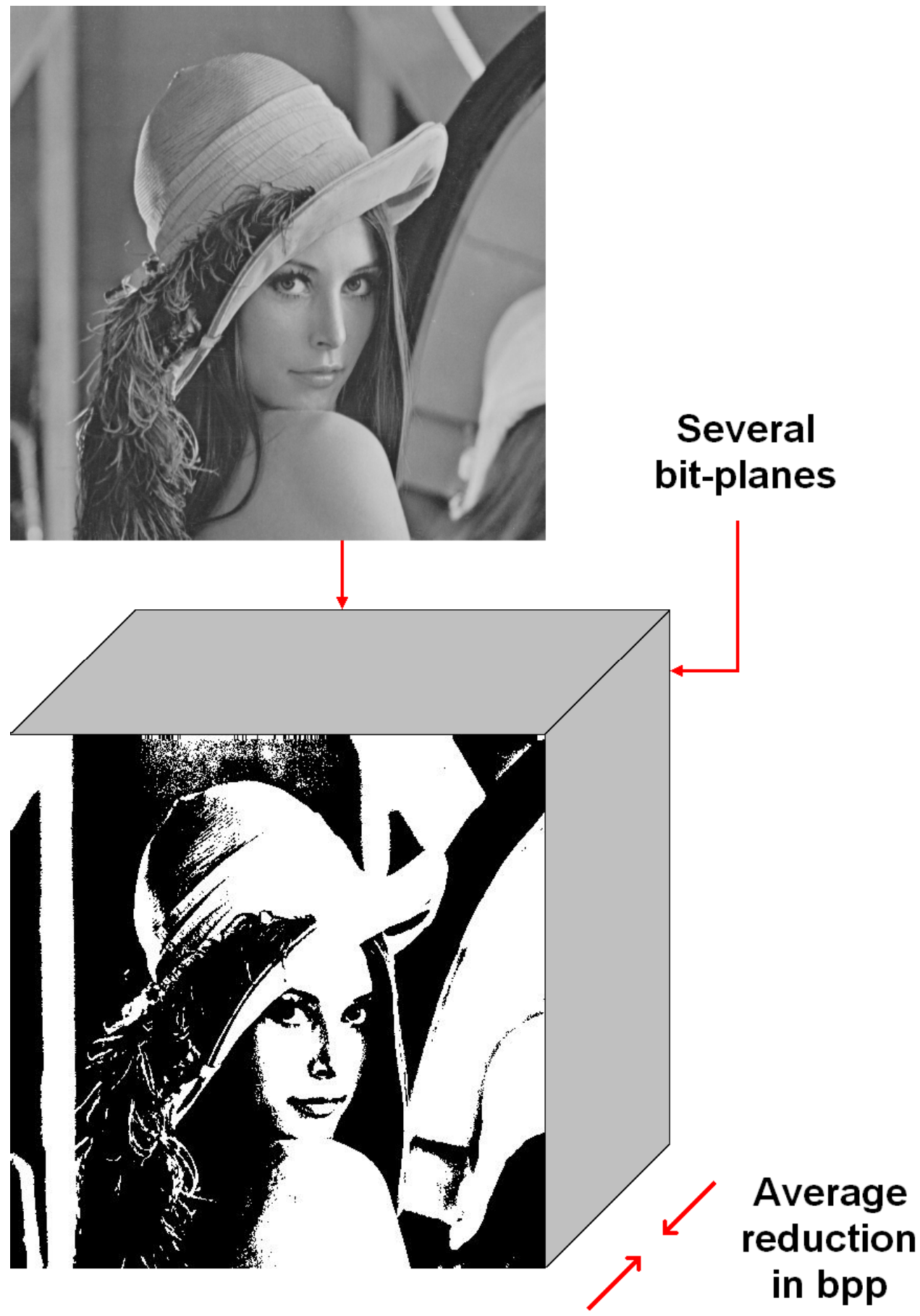

**Fig.35** Compression.

In encoder we have two steps:

1) Downsampling
2) Encoding via JPEG

and in decoder we have four steps:

1) Decoding via JPEG$^{-1}$
2) Upsampling
3) Deblurring

In our case, the downsampling and upsampling is done with the techniques of Subsection 4.1.3, while the deblurring is done by a two-dimensional convolution mask of NxN pixels, which makes a rafter over the upsampled (blurred) image [188]. The parameters of this squared mask (where N is odd) are criticals, therefore, such parameters must be calculated and adjusted with total accuracy. Moreover, in [188], we proceed to deduct the mask and set the optimal relationship between its parameters. Later we proceed to adjust them via a Genetic Algorithm [203], which gives us the best combination of value of a unsharp masking whose center is different from the elements around it (which are all the same) and added all give zero.

Summing-up, since the theory of Fourier has all the difficulties mentioned throughout this section, it is imperative to a new tool that allows us instant spectral analysis of the signal and so complete the time-frequency toolbox available to the scientific community.

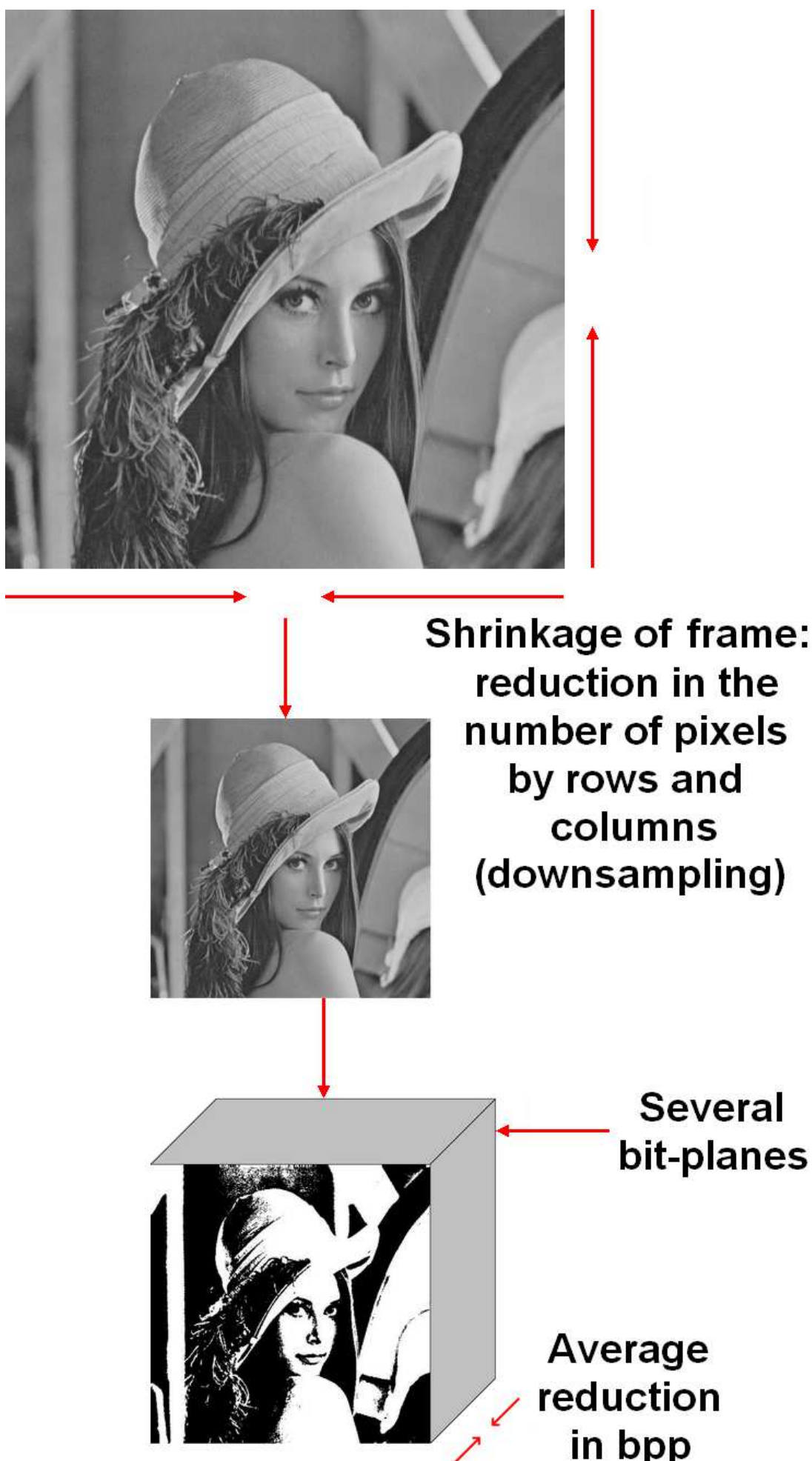

**Fig.36** Supercompression.

## 2.5 Edge detection

***Introduction –*** An edge in an image is a significant local change in the image intensity, usually associated with a discontinuity in either the image intensity or the first derivative of the image intensity. Discontinuities in the image intensity can be either (1) *step* discontinuities, where the image intensity abruptly changes from one value on one side of the discontinuity to a different value on the opposite side, or (2) *line* discontinuities, where the image intensity abruptly changes value but then returns to the starting value within some short distance. However, step and line edges are rare in real images. Because of low-frequency components or the smoothing introduced by most sensing devices, sharp discontinuities rarely exist in real signals. Step edges become *ramp* edges and line edges become *roof* edges, where intensity changes are not instantaneous but occur over a finite distance.

Illustrations of these edge profiles are shown in Fig.37.

***Gradient –*** Edge detection is essentially the operation of detecting significant local changes in an image. In one dimension, a step edge is associated with a local peak in the first derivative. The gradient is a measure of change in a function, and an image can be considered to be an array of samples of some continuous function of image intensity. By analogy, significant changes in the gray values in an image can be detected by using a discrete approximation to the gradient. The gradient is the two-dimensional equivalent of the first derivative and is defined as the *vector*

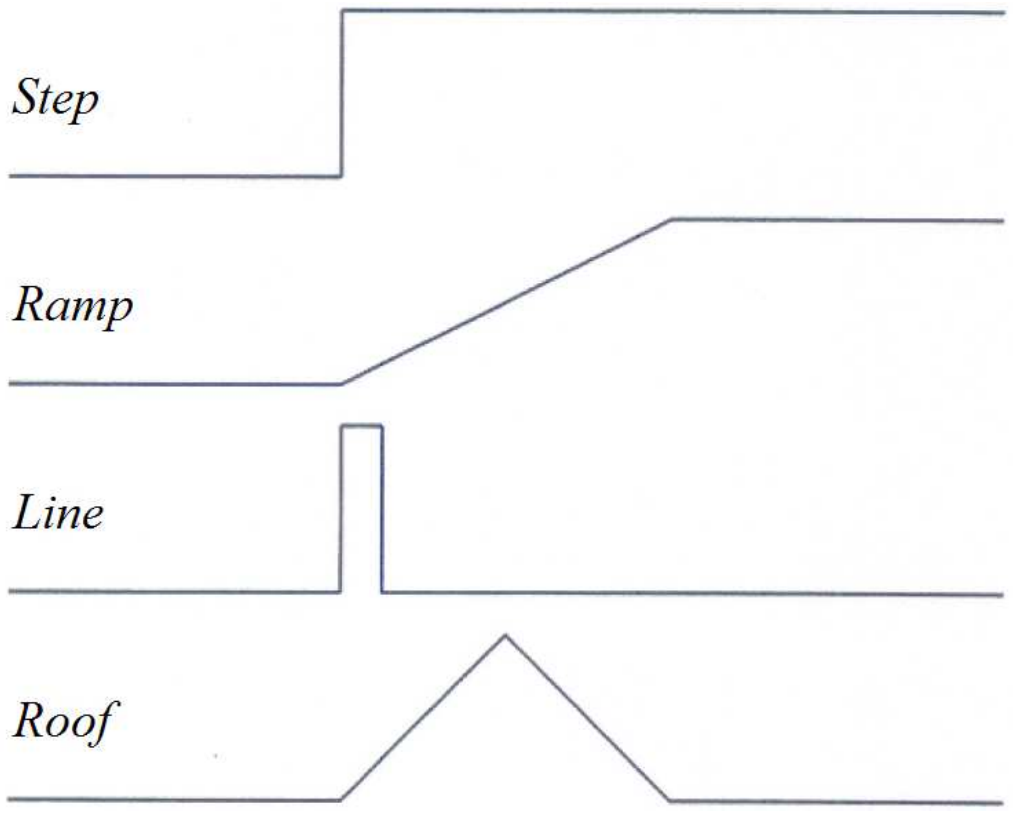

**Fig.37** One-dimensional edge profiles.

$$\mathrm{G}\left[f(x,y)\right]=\begin{bmatrix}G_x\\G_y\end{bmatrix}=\begin{bmatrix}\dfrac{\partial f}{\partial x}\\\dfrac{\partial f}{\partial y}\end{bmatrix} \tag{90}$$

There are two important properties associated with the gradient: (1) the vector G[$f(x, y)$] points in the direction of the maximum rate of increase of the function $f(x, y)$, and (2) the magnitude of the gradient, given by

$$\mathrm{G}\left[f(x,y)\right]=\sqrt{G_x^2+G_y^2} \tag{91}$$

equals the maximum rate of increase of $f(x, y)$ per unit distance in the direction $G$. It is common practice, however, to approximate the gradient magnitude by absolute values:

$$\mathrm{G}\left[f(x,y)\right]\approx\left|G_x\right|+\left|G_y\right| \tag{92}$$

or

$$\mathrm{G}\left[f(x,y)\right]\approx max\left(\left|G_x\right|,\left|G_y\right|\right) \tag{93}$$

From vector analysis, the *direction* of the gradient is defined as:

$$\alpha(x,y)=tan^{-1}\left(\frac{G_y}{G_x}\right) \tag{94}$$

where the angle $\alpha$ is measured with respect to the $x$ axis.

Note that the magnitude of the gradient is actually independent of the direction of the edge. Such operators are called *isotropic operators.*

***Numerical Approximation –*** For digital images, the derivatives in Eq.(90) are approximated by differences. The simplest gradient approximation is

$$G_x\cong f\left[i,j+1\right]-f\left[i,j\right] \tag{95}$$

$$G_y \cong f[i,j] - f[i+1,j] \tag{96}$$

Remember that $j$ corresponds to the $x$ direction and $i$ to the negative $y$ direction. These can be implemented with simple convolution masks as shown below:

$$G_x = \begin{bmatrix} -1 & 1 \end{bmatrix} \qquad G_y = \begin{bmatrix} 1 \\ -1 \end{bmatrix} \tag{97}$$

When computing an approximation to the gradient, it is critical that the $x$ and $y$ partial derivatives be computed at exactly the same position in space. However, using the above approximations, $G_x$ is actually the approximation to the gradient at the interpolated point $[i, j+1/2]$ and $G_y$ at $[i+1/2, j]$. For this reason, 2 x 2 first differences, rather than 2 x 1 and 1 x 2 masks, are often used for the $x$ and $y$ partial derivatives:

$$G_x = \begin{bmatrix} -1 & 1 \\ -1 & 1 \end{bmatrix} \qquad G_y = \begin{bmatrix} 1 & 1 \\ -1 & -1 \end{bmatrix} \tag{98}$$

Now, the positions about which the gradients in the $x$ and $y$ directions are calculated are the same. This point lies between all four pixels in the 2 x 2 neighborhood at the interpolated point $[i+1/2, j+1/2]$. This fact may lead to some confusion. Therefore, an alternative approach is to use a 3 x 3 neighborhood and calculate the gradient about the center pixel. These methods are discussed next.

***Steps in edge detection* –** Algorithms for edge detection contain three steps:

- **Filtering:** Since gradient computation based on intensity values of only two points are susceptible to noise and other vagaries in discrete computations, filtering is commonly used to improve the performance of an edge detector with respect to noise. However, there is a trade-off between edge strength and noise reduction. More filtering to reduce noise results in a loss of edge strength.

- **Enhancement:** In order to facilitate the detection of edges, it is essential to determine changes in intensity in the neighborhood of a point. Enhancement emphasizes pixels where there is a significant change in local intensity values and is usually performed by computing the gradient magnitude.

- **Detection:** We only want points with strong edge content. However, many points in an image have a nonzero value for the gradient, and not all of these points are edges for a particular application. Therefore, some method should be used to determine which points are edge points. Frequently, thresholding provides the criterion used for detection.

- **Localization:** The location of the edge can be estimated with subpixel resolution if required for the application. The edge orientation can also be estimated.

***Roberts Operator* –** The Roberts cross operator provides a simple approximation to the gradient magnitude (above view):

$$\mathrm{G}\left[f[i,j]\right] = \left|f[i,j] - f[i+1,j+1]\right| + \left|f[i+1,j] - f[i,j+1]\right| \tag{99}$$

Using convolution masks, this becomes

$$\mathrm{G}\left[f[i,j]\right] = \left|G_x\right| + \left|G_y\right| \tag{100}$$

where $G_x$ and $G_y$ are calculated using the following masks:

$$G_x = \begin{bmatrix} 1 & 0 \\ 0 & -1 \end{bmatrix} \qquad G_y = \begin{bmatrix} 0 & -1 \\ 1 & 0 \end{bmatrix} \tag{101}$$

As with the previous 2 x 2 gradient operator, the differences are computed at the interpolated point [*i*+1/2, *j*+1/2]. The Roberts operator is an approximation to the continuous gradient at that point and not at the point [*i*, *j*] *as* might be expected. The results of Roberts edge detector are shown in the figures at the end of this section.

***Sobel Operator –*** As mentioned previously, a way to avoid having the gradient calculated about an interpolated point between pixels is to use a 3 x 3 neighborhood for the gradient calculations. Consider the arrangement of pixels about the pixel [*i*, *j*] shown in Fig.38. The Sobel operator is the magnitude of the gradient computed by

$$M = \sqrt{s_x^2 + s_y^2} \tag{102}$$

where the partial derivatives are computed by

$$s_x = \left(a_2 + ca_3 + a_4\right) - \left(a_0 + ca_7 + a_6\right) \tag{103}$$
$$s_y = \left(a_0 + ca_1 + a_2\right) - \left(a_6 + ca_5 + a_4\right) \tag{104}$$

with the constant c = 2.

Like the other gradient operators, $s_x$ and $s_y$ can be implemented using convolution masks:

$$s_x = \begin{bmatrix} -1 & 0 & 1 \\ -2 & 0 & 2 \\ -1 & 0 & 1 \end{bmatrix} \qquad s_y = \begin{bmatrix} 1 & 2 & 1 \\ 0 & 0 & 0 \\ -1 & -2 & -1 \end{bmatrix} \tag{105}$$

Note that this operator places an emphasis on pixels that are closer to the center of the *mask.* The figures at the end of this section show the performance of this operator. The Sobel operator is one of the most commonly used edge detectors.

| $a_0$ | $a_1$ | $a_2$ |
|---|---|---|
| $a_7$ | [i, j] | $a_3$ |
| $a_6$ | $a_5$ | $a_4$ |

**Fig.38** The labeling of neighborhood pixels used to explain the Sobel and Prewitt operators [45-48].

***Prewitt Operator –*** The Prewitt operator uses the same equations as the Sobel operator, except that the constant c = 1. Therefore:

$$s_x = \begin{bmatrix} -1 & 0 & 1 \\ -1 & 0 & 1 \\ -1 & 0 & 1 \end{bmatrix} \qquad s_y = \begin{bmatrix} 1 & 1 & 1 \\ 0 & 0 & 0 \\ -1 & -1 & -1 \end{bmatrix} \tag{106}$$

Note that, unlike the Sobel operator, this operator does not place any emphasis on pixels that are closer to the center of the masks. The performance of this edge detector is also shown in the figures at the end of this section.

***Canny Edge Detector –*** The Canny edge detector is the first derivative of a Gaussian and closely approximates the operator that optimizes the product of signal-to-noise ratio and localization. The Canny edge detection algorithm is summarized by the following notation. Let $I[i, j]$ denote the image. The result from convolving the image with a Gaussian smoothing filter using separable filtering is an array of smoothed data,

$$S[i,j] = G[i,j;\sigma] * I[i,j] \tag{107}$$

where $\sigma$ is the spread of the Gaussian and controls the degree of smoothing. The gradient of the smoothed array $S[i, j]$ can be computed using the 2 x 2 first-difference approximations (Subsection relating to Gradient) to produce two arrays $P[i, j]$ and $Q[i, j]$ for the $x$ and $y$ partial derivatives:

$$P[i,j] \approx \left(S[i,j+1] - S[i,j] + S[i+1,j+1] - S[i+1,j]\right)/2 \tag{108}$$

$$Q[i,j] \approx \left(S[i,j] - S[i+1,j] + S[i,j+1] - S[i+1,j+1]\right)/2 \tag{109}$$

The finite differences are averaged over the 2 x 2 square so that the $x$ and $y$ partial derivatives are computed at the same point in the image. The magnitude and orientation of the gradient can be computed from the standard formulas for rectangular-to-polar conversion:

$$M[i,j] = \sqrt{P[i,j]^2 + Q[i,j]^2} \tag{110}$$

$$\theta[i,j] = arctan\left(Q[i,j], P[i,j]\right) \tag{111}$$

where the *arctan* function takes two arguments and generates an angle over the entire circle of possible directions. These functions must be computed efficiently, preferably without using floating-point arithmetic [45-48].
Finally, the Canny Edge Detection Algorithm is:

- *Smooth the image with a Gaussian filter.*
- *Compute the gradient magnitude and orientation using finite-difference approximations for the partial derivatives.*
- *Apply nonmaxima suppression to the gradient magnitude.*
- *Use the double thresholding algorithm to detect and link edges.*

To see in detail the items of the algorithm, we recommend reading Chapter 5 of [204].

In Fig.39, we can see original image of *Agus in Miami* (top), Roberts (middle-left), Sobel (middle-right), Prewitt (down-left), and finally, Canny (down-right).

In [204], the authors compare the different edge detectors discussed so far. The comparisons are presented according to the first three steps described at the beginning of this section: filtering, enhancement, and detection. The estimation step is not be shown there. In addition, They give results of edge detection on noisy images for two specific cases-one utilizing the filtering step and one omitting the filtering step. Results of edge detection using varying amounts of filtering are also be given. In fact, shows results of all the edge detection methods discussed so far, from the simple 1 x 2 gradient approximation up to the Prewitt operator.

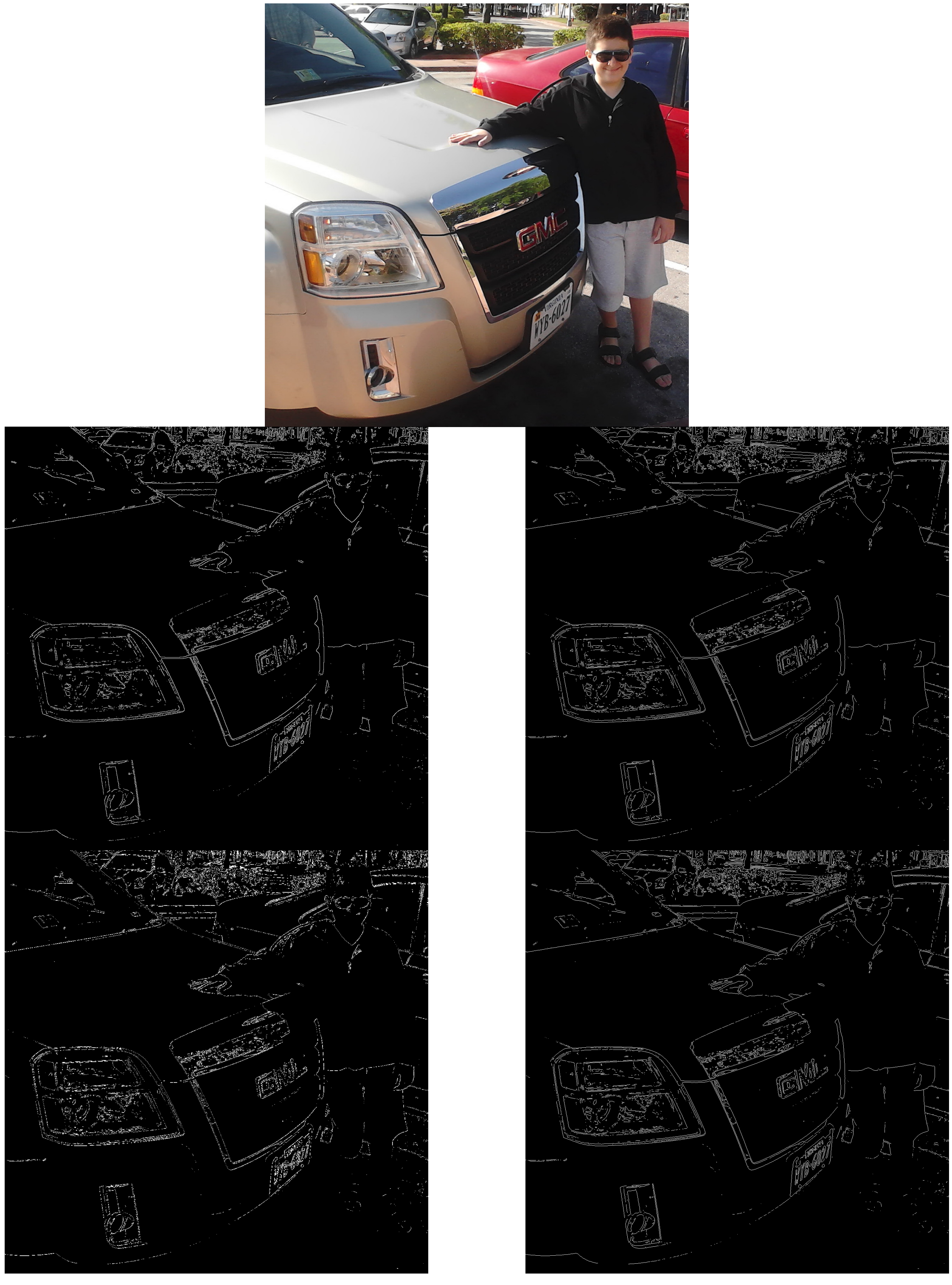

**Fig.39** Original image (top) *Agus in Miami*, and the edge detectors: Roberts (middle-left), Sobel (middle-right), Prewitt (down-left), and Canny (down-right).

Such figure shows the results of the edge detectors when the filtering step is omitted. Note the many false edges detected as a result of the noise.

## 3 Quantum Spectral Analysis (QSA)

### 3.1 In the beginning … Schrödinger's equation

#### 3.1.1 Qubits and Bloch's sphere

The bit is the fundamental concept of classical computation and classical information. Quantum computation and quantum information are built upon an analogous concept, the quantum bit, or qubit for short. In this section we introduce the properties of single and multiple qubits, comparing and contrasting their properties to those of classical bits [1]. The difference between bits and qubits is that a qubit can be in a state other than $|0\rangle$ or $|1\rangle$ [51, 52]. It is also possible to form linear combinations of states, often called superpositions:

$$|\psi\rangle = \alpha|0\rangle + \beta|1\rangle, \tag{112}$$

where $|\psi\rangle$ is called *wave function*, $|\alpha|^2 + |\beta|^2 = 1$, with the states $|0\rangle$ and $|1\rangle$ are understood as different polarization states of light. Besides, a column vector $|\psi\rangle$ is called a *ket* vector $[\alpha\ \beta]^T$, where, $(\bullet)^T$ means transpose of (•), while a row vector $\langle\psi|$ is called a *bra* vector $[\alpha\ \beta]$. The numbers $\alpha$ and $\beta$ are complex numbers, although for many purposes not much is lost by thinking of them as real numbers. Put another way, the state of a qubit is a vector in a two-dimensional complex vector space. The special states $|0\rangle$ and $|1\rangle$ are known as Computational Basis States (CBS), and form an orthonormal basis for this vector space, being

$$|0\rangle = \begin{bmatrix} 1 \\ 0 \end{bmatrix} \quad \text{and} \quad |1\rangle = \begin{bmatrix} 0 \\ 1 \end{bmatrix}$$

One picture useful in thinking about qubits is the following geometric representation.

Because $|\alpha|^2 + |\beta|^2 = 1$, we may rewrite Eq.(112) as

$$|\psi\rangle = e^{i\gamma}\left( cos\frac{\theta}{2}|0\rangle + e^{i\phi}\, sin\frac{\theta}{2}|1\rangle \right) = e^{i\gamma}\left( cos\frac{\theta}{2}|0\rangle + \left(cos\,\phi + i\, sin\,\phi\right) sin\frac{\theta}{2}|1\rangle \right) \tag{113}$$

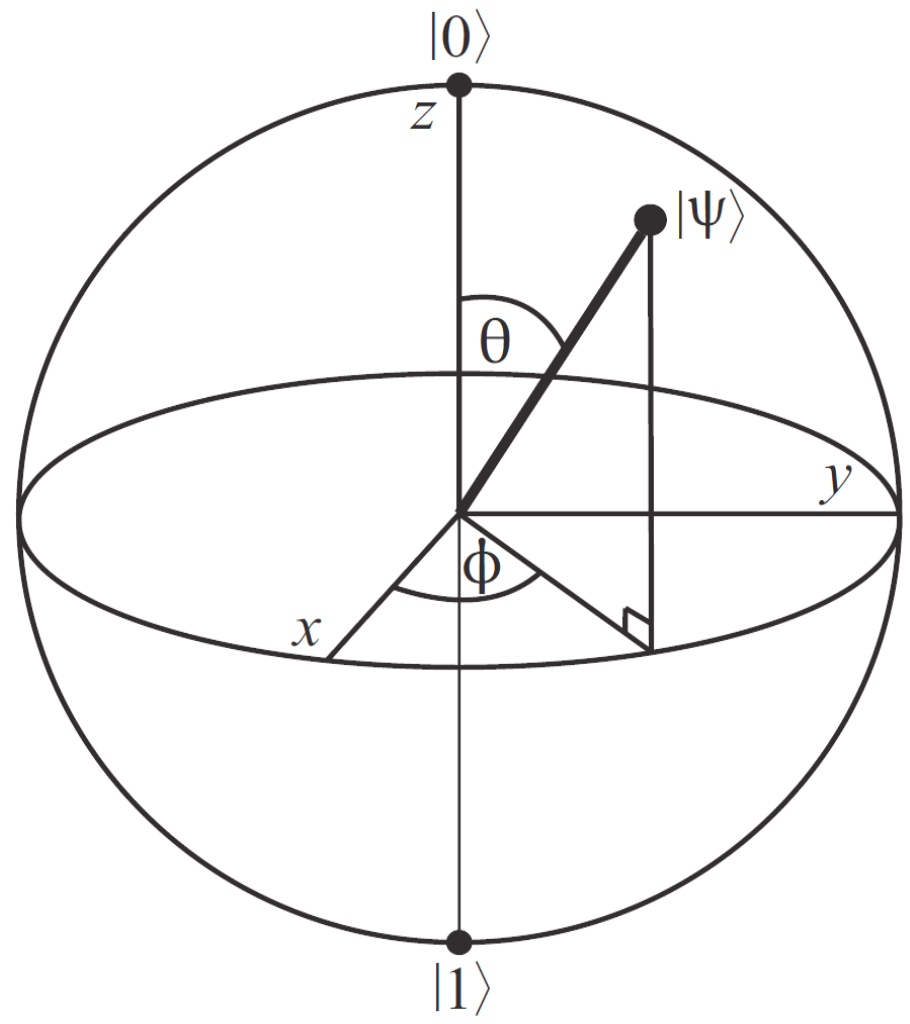

**Fig. 40** Bloch's Sphere.

where $0 \leq \theta \leq \pi$, $0 \leq \phi < 2\pi$. We can ignore the factor of $e^{i\gamma}$ out the front, because it has no observable effects [1], and for that reason we can effectively write

$$|\psi\rangle = cos\frac{\theta}{2}|0\rangle + e^{i\phi}\, sin\frac{\theta}{2}|1\rangle \tag{114}$$

The numbers $\theta$ and $\phi$ define a point on the unit three-dimensional sphere, as shown in Fig.40.

Quantum mechanics is mathematically formulated in Hilbert space or projective Hilbert space. The space of pure states of a quantum system is given by the one-dimensional subspaces of the corresponding Hilbert space (or the "points" of the projective Hilbert space). In a two-dimensional Hilbert space this is simply the complex projective line, which is a geometrical sphere.

This sphere is often called the Bloch's sphere; it provides a useful means of visualizing the state of a single qubit, and often serves as an excellent testbed for ideas about quantum computation and quantum information. Many of the operations on single qubits which can be seen in [1] are neatly described within the Bloch's sphere picture. However, it must be kept in mind that this intuition is limited because there is no simple generalization of the Bloch's sphere known for multiple qubits [1, 51, 52].

Except in the case where $|\psi\rangle$ is one of the ket vectors $|0\rangle$ or $|1\rangle$ the representation is unique. The parameters $\theta$ and $\phi$, re-interpreted as spherical coordinates, specify a point $\vec{a} = (sin\,\theta\, cos\,\phi\, ,\, sin\,\theta\, sin\,\phi\, ,\, cos\,\theta)$ on the unit sphere in $\mathbb{R}^3$ (according to Eq.113).

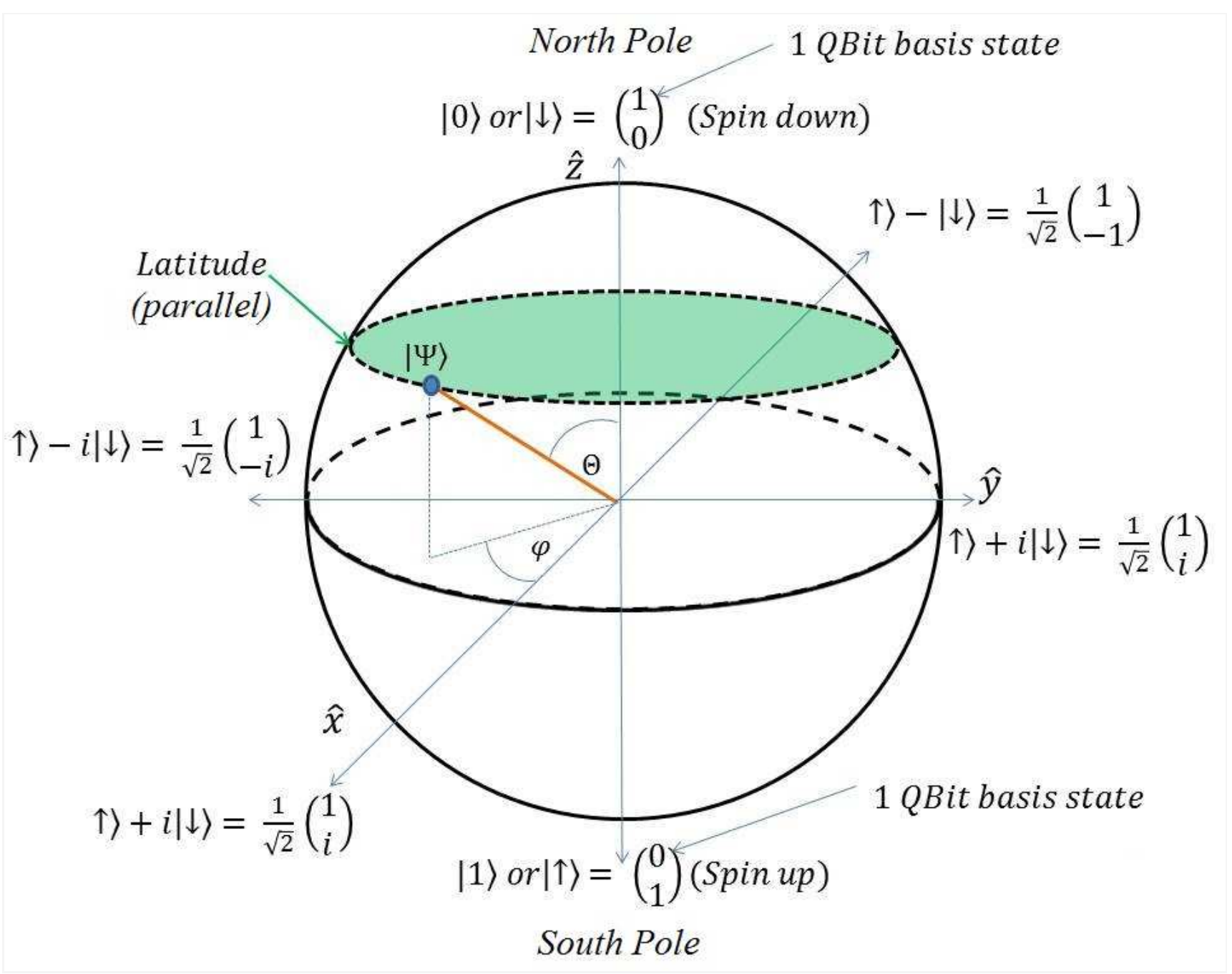

**Fig. 41** Details of the poles, as well as an example of parallel and several qubit states on the sphere.

Figure 41 highlights all components (details) concerning the Bloch's sphere, namely

$$Spin\ down = |\downarrow\rangle = |0\rangle = \begin{bmatrix} 1 \\ 0 \end{bmatrix} = qubit\ basis\ state = North\ Pole$$

and

$$Spin\ up = \left|\uparrow\right\rangle = \left|1\right\rangle = \begin{bmatrix} 0 \\ 1 \end{bmatrix} = qubit\ basis\ state = South\ Pole$$

Both poles play a fundamental role in the development of the quantum computing [1]. Besides, a very important concept to the affections of the development quantum information processing, in general, i.e., the notion of latitude (parallel) on the Bloch's sphere is hinted. Such parallel as shown in green in Fig.41, where we can see the complete coexistence of poles, parallels and meridians on the sphere, including computational basis states ($\left|0\right\rangle, \left|1\right\rangle$). The poles and the parallels form the geometric bases of criteria and logic needed to implement any quantum gate or circuit.

### 3.1.2 Schrödinger's equation and unitary operators

A quantum state can be transformed into another state by a unitary operator, symbolized as $U$ ($U : H \rightarrow H$ on a Hilbert space $H$, being called an unitary operator if it satisfies $U^{\dagger}U = UU^{\dagger} = I$, where $(\bullet)^{\dagger}$ is the adjoint of (•), and $I$ is the identity matrix), which is required to preserve inner products: If we transform $\left|\chi\right\rangle$ and $\left|\psi\right\rangle$ to $U\left|\chi\right\rangle$ and $U\left|\psi\right\rangle$, then $\left\langle\chi\right|U^{\dagger}U\left|\psi\right\rangle = \left\langle\chi|\psi\right\rangle$. In particular, unitary operators preserve lengths:

$$\left\langle\psi\right|U^{\dagger}U\left|\psi\right\rangle = \left\langle\psi|\psi\right\rangle = 1, \text{ if } \left|\psi\right\rangle \text{ is on the Bloch's sphere (i.e., it is a pure state).} \tag{115}$$

On the other hand, the unitary operator satisfies the following differential equation known as the Schrödinger equation [1, 10-12]:

$$\frac{d}{dt}U(t) = \frac{-i\,\hat{H}}{\hbar}U(t) \tag{116}$$

where $\hat{H}$ represents the Hamiltonian matrix of the Schrödinger equation, $i = \sqrt[2]{-1}$, and $\hbar$ is the reduced Planck constant, i.e., $\hbar = h/2\pi$. Multiplying both sides of Eq.(116) by $\left|\psi(0)\right\rangle$ and setting

$$\left|\psi(t)\right\rangle = U(t)\left|\psi(0)\right\rangle \tag{117}$$

yields

$$\frac{d}{dt}\left|\psi(t)\right\rangle = \frac{-i\,\hat{H}}{\hbar}\left|\psi(t)\right\rangle \tag{118}$$

The solution to the Schrödinger equation is given by the matrix exponential of the Hamiltonian matrix:

$$U(t) = e^{\frac{-i\hat{H}t}{\hbar}} \qquad \text{(if Hamiltonian is not time dependent)} \tag{119}$$

and

$$U(t) = e^{\frac{-i}{\hbar}\int_0^t \hat{H}\,dt} \qquad \text{(if Hamiltonian is time dependent)} \tag{120}$$

Thus the probability amplitudes evolve across time according to the following equation:

$$\left|\psi(t)\right\rangle = e^{\frac{-i\hat{H}t}{\hbar}}\left|\psi(0)\right\rangle \quad \text{(if Hamiltonian is not time dependent)} \tag{121}$$

or

$$\left|\psi(t)\right\rangle = e^{\frac{-i}{\hbar}\int_0^t \hat{H}\,dt}\left|\psi(0)\right\rangle \quad \text{(if Hamiltonian is time dependent)} \tag{122}$$

The Eq.(121) is the main piece in building circuits, gates and quantum algorithms, being $U$ who represents such elements [1].

Finally, the discrete version of Eq.(118) is

$$\left|\psi_{n+1}\right\rangle = \frac{-i\hat{H}}{\hbar}\left|\psi_n\right\rangle, \tag{123}$$

for a time dependent (or not) Hamiltonian, being $n$ the discrete time.

### 3.1.3 Quantum Circuits, Gates, and Algorithms; Reversibility and Quantum Measurement

As we can see in Fig.42, and remember Eq.(117), the quantum algorithm (identical case to circuits and gates) viewed as a transfer (or mapping input-to-output) has two types on output:

a) the result of algorithm (circuit of gate), i.e., $\left|\psi_n\right\rangle$
b) part of the input $\left|\psi_0\right\rangle$, i.e., $\left|\underline{\psi}_0\right\rangle$ (underlined $\left|\psi_0\right\rangle$), in order to impart reversibility to the circuit, which is a critical need in quantum computing [1].

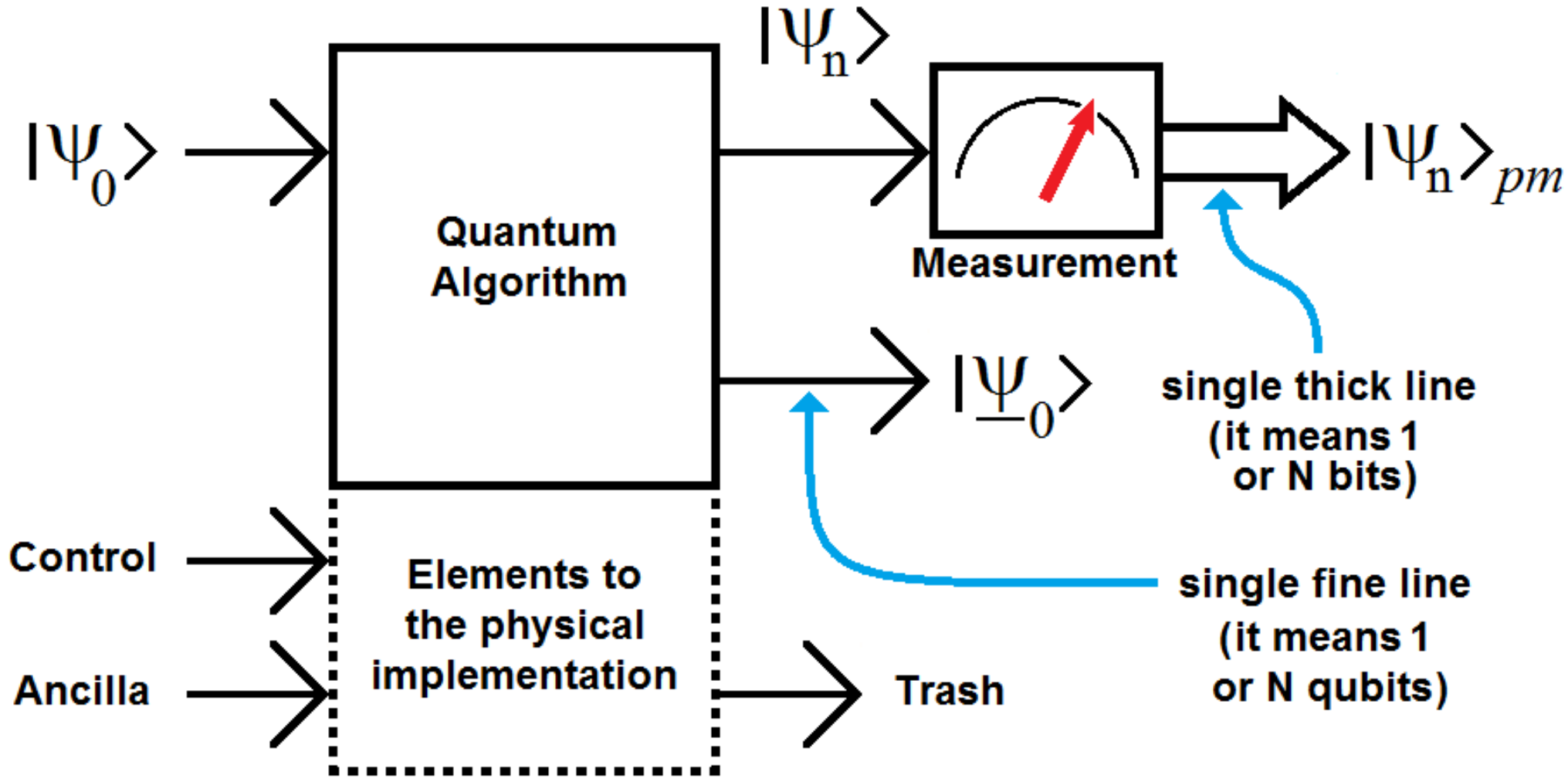

**Fig. 42** Module to measuring, quantum algorithm and the elements needs to its physical implementation.

Besides, we can see clearly a module for measuring $\left|\psi_n\right\rangle$ with their respective output, i.e., $\left|\psi_n\right\rangle_{pm}$ (or, $\left|\psi_n\right\rangle$ post-measurement), and a number of elements needed for the physical implementation of the quantum algorithm (circuit or gate), namely: control, ancilla and trash [1].

In Fig.42 as well as in the rest of them (unlike [1]) a single fine line represents a wire carrying *1* qubit or *N*

qubits (qudit), interchangeably, while a single thick line represents a wire carrying *1* or *N* classical bits, interchangeably too. However, the mentioned concept of reversibility is closely related to energy consumption, and hence to the Landauer's Principle [1].

On the other hand, computational complexity studies the amount of time and space required to solve a computational problem. Another important computational resource is energy. In [1], the authors show the energy requirements for computation. Surprisingly, it turns out that computation, both classical and quantum, can in principle be done without expending any energy! Energy consumption in computation turns out to be deeply linked to the reversibility of the computation. In other words, it is inexcusable the need of the $|\underline{\Psi}_0\rangle$ presence to the output of quantum gate [1].

On the other hand, in quantum mechanics, measurement is a non-trivial and highly counter-intuitive process. Firstly, because measurement outcomes are inherently probabilistic, i.e. regardless of the carefulness in the preparation of a measurement procedure, the possible outcomes of such measurement will be distributed according to a certain probability distribution. Secondly, once the measurement has been performed, a quantum system in unavoidably altered due to the interaction with the measurement apparatus. Consequently, for an arbitrary quantum system, pre-measurement and post-measurement quantum states are different in general [1].

***Postulate.*** Quantum measurements are described by a set of measurement operators $\{\hat{M}_m\}$, index *m* labels the different measurement outcomes, which act on the state space of the system being measured. Measurement outcomes correspond to values of *observables*, such as position, energy and momentum, which are Hermitian operators [1] corresponding to physically measurable quantities.

Let $|\psi\rangle$ be the state of the quantum system immediately before the measurement. Then, the probability that result *m* occurs is given by

$$p(m) = \langle\psi|\hat{M}_m^{\dagger}\hat{M}_m|\psi\rangle \tag{124}$$

and the post-measurement quantum state is

$$|\psi\rangle_{pm} = \frac{\hat{M}_m|\psi\rangle}{\sqrt{\langle\psi|\hat{M}_m^{\dagger}\hat{M}_m|\psi\rangle}} \tag{125}$$

Operators $\hat{M}_m$ must satisfy the completeness relation of Eq.(126), because that guarantees that probabilities will sum to one; see Eq.(127) [1]:

$$\sum_m \hat{M}_m^{\dagger}\hat{M}_m = I \tag{126}$$

$$\sum_m \langle\psi|\hat{M}_m^{\dagger}\hat{M}_m|\psi\rangle = \sum_m p(m) = 1 \tag{127}$$

Let us work out a simple example. Assume we have a polarized photon with associated polarization orientations 'horizontal' and 'vertical'. The horizontal polarization direction is denoted by $|0\rangle$ and the vertical polarization direction is denoted by $|1\rangle$. Thus, an arbitrary initial state for our photon can be described by the quantum state $|\psi\rangle = \alpha|0\rangle + \beta|1\rangle$ (remembering Subsection 3.1.1), where $\alpha$ and $\beta$ are complex numbers constrained by the normalization condition $|\alpha|^2 + |\beta|^2 = 1$, and $\{|0\rangle, |1\rangle\}$ is the computational basis spanning $\mathrm{H}^2$.

Now, we construct two measurement operators $\hat{M}_0 = |0\rangle\langle 0|$ and $\hat{M}_1 = |1\rangle\langle 1|$ and two measurement outcomes $a_0$, $a_1$. Then, the full observable used for measurement in this experiment is $\hat{M} = a_0 |0\rangle\langle 0| + a_1 |1\rangle\langle 1|$. According to Postulate, the probabilities of obtaining outcome $a_0$ or outcome $a_1$ are given by $\mathrm{p}(a_0) = |\alpha|^2$ and $\mathrm{p}(a_1) = |\beta|^2$. Corresponding post-measurement quantum states are as follows: if outcome = $a_0$, then $|\psi\rangle_{pm} = |0\rangle$; if outcome = $a_1$ then $|\psi\rangle_{pm} = |1\rangle$.

### 3.2 QSA properly speaking

The Eq.(118) represents the Schrödinger equation, which we are going to write it in a better way, so as to simplify notation

$$|\dot{\psi}(t)\rangle = -i\,\Omega(t)\,|\psi(t)\rangle \tag{128}$$

where $|\dot{\psi}(t)\rangle = \frac{d}{dt}|\psi(t)\rangle$ and $\Omega(t) = \frac{\hat{H}(t)}{\hbar}$, being Ω the angular frequency matrix, and $\hat{H}$ the Hamiltonian matrix. Both time dependents, simultaneously, i.e., at each instant, we will have a matrix.

On the other hand, Ω depends on the respective –non relativistic– system, that is to say, where the most general form for one qubit is

$$\Omega(t) = \begin{bmatrix} \omega_{11}(t) & \omega_{12}(t) \\ \omega_{21}(t) & \omega_{22}(t) \end{bmatrix} \tag{129}$$

Two interesting particular cases are represented by

$$\Omega(t) = \begin{bmatrix} \omega_1(t) & 0 \\ 0 & \omega_2(t) \end{bmatrix} \tag{130}$$

and

$$\Omega(t) = \begin{bmatrix} \omega(t) & 0 \\ 0 & \omega(t) \end{bmatrix} = \omega(t)\begin{bmatrix} 1 & 0 \\ 0 & 1 \end{bmatrix} = \omega(t)\,I = \omega(t) \tag{131}$$

being *I* the identity matrix. Thus, replacing Eq.(131) in Eq.(128), we will have,

$$|\dot{\psi}(t)\rangle = -i\,\omega(t)\,|\psi(t)\rangle. \tag{132}$$

Now, we multiply both sides (by left) of Eq.(132) by $\langle\psi(t)|$,

$$\langle\psi(t)|\dot{\psi}(t)\rangle = -i\,\omega(t)\,\langle\psi(t)|\psi(t)\rangle. \tag{133}$$

Finally, $\omega(t)$ results,

$$\omega(t) = i\,\frac{\langle\psi(t)|\dot{\psi}(t)\rangle}{\langle\psi(t)|\psi(t)\rangle}. \tag{134}$$

Equation (134) represents QSA for the monotone case. Now, and considering Equations (128) and (129), where Ω represents an irreducible matrix, then, we are going to multiply both sides (by right) of Eq.(128) by

$\langle \psi(t) |$, therefore,

$$|\dot{\psi}(t)\rangle\langle\psi(t)| = -i\,\Omega(t)\,|\psi(t)\rangle\langle\psi(t)| \tag{135}$$

Finally, $\Omega(t)$ results,

$$\begin{aligned}\Omega(t) &= i\,|\dot{\psi}(t)\rangle\langle\psi(t)|\left[|\psi(t)\rangle\langle\psi(t)|\right]^{-1} \\ &= i\,|\dot{\psi}(t)\rangle\langle\psi(t)|^{\dagger}\end{aligned} \tag{136}$$

where

$$\langle\psi(t)|^{\dagger} = \langle\psi(t)|\left[|\psi(t)\rangle\langle\psi(t)|\right]^{-1} \text{ (is the pseudoinverse of } |\dot{\psi}(t)\rangle \text{)} \tag{137}$$

Equation (136) represents QSA for the multitone case.

### 3.3 Frequency in time (FIT)

Once we have arrived at the classical world, we can then apply an adaptation QSA to signals and images, with and without overlap of samples or pixels, respectively. Such adaptation is called *frequency in time* (FIT).

#### 3.3.1 FIT for signals

***Case with overlap*** – In the classical version of Eq.(134) we are going to replace qubits by samples of a real signal, therefore, inner products disappear, and the classical version of Eq.(134) in a symbolic form is

$$\Delta\varpi(t) = i\frac{1}{S(t)}\frac{\partial S(t)}{\partial t} = i\,\frac{\dot{S}(t)}{S(t)}, \tag{138}$$

where $\dot{S}(t) = \dfrac{\partial S(t)}{\partial t}$, and $S$ is a signal defined in $\mathbb{R}^N$, being $N$ the size of the signal, and $\Delta\varpi$ the frequency (in here, while in QSA is the imaginary angular frequency).

Appealing (for simplicity) to the discrete version of $\Delta\varpi$, we will have,

$$\Delta\varpi = i\,\dot{S}./S, \tag{139}$$

where "$./$" represents the infixed version of Hadamard's quotient of vectors [80], $S = [s_0\ s_1\ s_2\ \ldots\ s_{N-1}]$ is a signal of $N$ samples, $\dot{S} = [\dot{s}_0\ \dot{s}_1\ \dot{s}_2\ \ldots\ \dot{s}_{N-1}]$ is its derivative, and $\Delta\varpi = [\Delta\varpi_0\ \Delta\varpi_1\ \Delta\varpi_2\ \ldots\ \Delta\varpi_{N-1}]$. That is to say, for each sample, we will have,

$$\Delta\varpi_n = i\,\dot{s}_n / s_n \qquad \forall n \in [0, N-1], \text{ being: } \dot{s}_n = (s_{n+1} - s_{n-1})/2 \text{, and } n \text{ the discrete time.} \tag{140}$$

Equation (140) is the discrete version of $\Delta\varpi$ in its most inapplicable form, given that this is not applicable in cases where the denominator is zero (although unlike the FFT, $\Delta\varpi$ has solution), without mentioning that $\Delta\varpi$ is an imaginary operator to be applied to real signals. Therefore, this form is called raw version. To overcome this drawback, we use equalized and/or averaged versions, as the following,

$$\Delta\varpi_{eq} = i\,\dot{S}_{eq}./S_{eq}, \tag{141}$$

where subscript "*eq*" means *equalized*. In general, $S$ is equalized inside [1, 2] as follows:

- calculate maximum of $S$, $max(S)$
- calculate minimum of $S$, $min(S)$
- $S_{eq} = \left( \left( S - min\left( S \right) \right) / \left( max\left( S \right) - min\left( S \right) \right) \right) + 1$
- calculate $\dot{S}_{eq}$
- calculate $\Delta\varpi_{eq}$

While the averaged form will be,

$$\Delta\varpi_{av} = i\ \dot{S} / avg\left( S \right), \tag{142}$$

where $avg\left( S \right)$ is the average of $S$. In both cases the signal must be enlarged (padded). For example, we see this in more detail in the third version,

$$\Delta\varpi_{um} = i\ \dot{S}_{um} ./ S_{um}, \tag{143}$$

where the subscript *um* means *one-dimensional mask*. That is to say, in order, we will have,

$$S = \left[ s_0\ s_1\ s_2\ \ldots\ s_{N-1} \right] \tag{144}$$

of length $N$, with masks of length $M$ (odd number), and side surplus (padding) of length $L = (M-1)/2$, finally, we define four new elements, such as the padded $S$

$$\begin{aligned} S_{padding} &= \left[ s_{-2L}\ \ldots\ s_{-1}\ , s_0\ s_1\ s_2\ \ldots\ s_{N-1}\ , s_N\ \ldots\ s_{N-1+2L} \right], \\ &= \left[ s_{-2L}\ \ldots\ s_{-1}\ , S\ , s_N\ \ldots\ s_{N-1+2L} \right] \end{aligned} \tag{145}$$

of length $N' = N + 2L = N + M - 1$, and for each $n \in \left[ -2L, N + 2L \right]$ generic sample, we will have

$$\Delta s_{n,L} = \left[ s_{n-L}\ s_{n-L+1}\ \ldots\ s_n\ \ldots\ s_{n+L-1}\ s_{n+L} \right], \tag{146}$$

of length $M$. Finally, we are going to define two additional masks of length $M$

$$D = \left[\ 1\ 1\ \ldots\ 1\ \ldots\ 1\ 1\ \right] / \mathrm{M}, \quad \text{(scaling operator)} \tag{147}$$

and

$$\Pi = \left[ -1\ -1\ \ldots\ -1\ \ 0\ \ 1\ \ldots\ 1\ \ 1\ \right] / \left( \mathrm{M}\text{-}1 \right), \quad \text{(wavelet operator)} \tag{148}$$

with,

$$s_{um,n} = \Delta s_{n,L}\ D^T \quad \text{(approximation band)} \tag{149}$$

where superscript $T$ means *transpose*, and

$$\dot{s}_{um,n} = \Delta s_{n,L}\ \Pi^T \quad \text{(detail band)} \tag{150}$$

If $M = 3$, then $\dot{s}_{um,n}$ will be the traditional discrete derivative case (see the same row of Eq.140), i.e., $\dot{s}_{um,n} = \left( s_{n+1} - s_{n-1} \right) / 2$, while, $s_{um,n} = \left( s_{n+1} + s_n + s_{n-1} \right) / 3$. These are a special case of what we have seen above. Finally, for each sample $n$, we will have, the instantaneous version of Eq.(143) in term of Equations (149) and (150),

$$\Delta\varpi_{um,n} = i \;\; \dot{s}_{um,n} \,/\, s_{um,n}\,, \tag{151}$$

$\forall n \in [0, N-1]$. On the other hand, and to save the fact that $\Delta\varpi$ is an imaginary operator to be applied to real signals (in all four versions), we will use (based on Eq.139) a more pure and useful version of *frequency in time* (FIT) for all mentioned versions:

$$\begin{aligned}
\Delta\omega &= \sqrt{\Delta\varpi \;.\!\times conj\left(\Delta\varpi\right)} \\
&= \sqrt{\left(i \;\; \dot{S}./\,S\right).\!\times conj\left(i \;\; \dot{S}./\,S\right)} \\
&= \left|\dot{S}./\,S\right| = \left|\dot{S}\right|./\left|S\right| = \frac{1}{\left|S\right|}\frac{\left|\Delta S\right|}{\Delta t}
\end{aligned} \tag{152}$$

Being $\Delta f = \Delta\omega / 2\pi = [\, \Delta f_0 \;\, \Delta f_1 \;\, \Delta f_2 \ldots \Delta f_{N-1}]$, the frequencies in hertz. Besides, $\Delta\omega$ is now a real operator to be applied to real signals. Remember that, this version (the original) depends on a possible denominator equal to zero, therefore, we will use the next versions directly dependent on the frequency:

$$\begin{aligned}
\Delta f_{eq} &= \frac{1}{2\pi}\sqrt{\Delta\varpi_{eq} \;.\!\times conj\left(\Delta\varpi_{eq}\right)} \\
&= \frac{1}{2\pi}\sqrt{\left(i \;\; \dot{S}_{eq}./\,S_{eq}\right).\!\times conj\left(i \;\; \dot{S}_{eq}./\,S_{eq}\right)} \qquad \text{for Eq.(141)} \\
&= \frac{1}{2\pi}\left|\dot{S}_{eq}./\,S_{eq}\right| = \frac{1}{2\pi}\left|\dot{S}_{eq}\right|./\,S_{eq}
\end{aligned} \tag{153}$$

$$\begin{aligned}
\Delta f_{av} &= \frac{1}{2\pi}\sqrt{\Delta\varpi_{av} \;.\!\times conj\left(\Delta\varpi_{av}\right)} \\
&= \frac{1}{2\pi}\frac{\sqrt{\left(i \;\; \dot{S}\right).\!\times conj\left(i \;\; \dot{S}\right)}}{avg\left(S\right)} \qquad \text{for Eq.(142)} \\
&= \frac{1}{2\pi}\left|\dot{S}\right| / \left|avg\left(S\right)\right|
\end{aligned} \tag{154}$$

and finally, one version more based on Eq.(143), that is to say,

$$\begin{aligned}
\Delta f_{um} &= \frac{1}{2\pi}\sqrt{\Delta\varpi_{um} \;.\!\times conj\left(\Delta\varpi_{um}\right)} \\
&= \frac{1}{2\pi}\sqrt{\left(i \;\; \dot{S}_{um}./\,S_{um}\right).\!\times conj\left(i \;\; \dot{S}_{um}./\,S_{um}\right)} \\
&= \frac{1}{2\pi}\left|\dot{S}_{um}./\,S_{um}\right| = \frac{1}{2\pi}\left|\dot{S}_{um}\right|./\left|S_{um}\right|
\end{aligned} \tag{155}$$

Because usually a (or more) denominator of Eq.(151) is zero, then here also it is necessary to use a procedure of equalization or normalization as in the previous cases.

***Examples -*** Next, we will implement the four seen cases of FIT to a signal as shown in Fig.43, which is an electrocardiographic (ECG) signal of 80 pulses per second. Top of Fig.43 shows the ECG, while its down shows the waterfall of ECG signal, where the positive peaks are clear, the negative peaks are dark, and the intermediates are gray.

On the other hand, Fig.44 shows the same signal of Fig.43, i.e., ECG of 80 cycles per second, with 256 samples per cycle, however, in this case, the six cells (sub-figures) contain: ECG signal in blue, and FIT in red with their respective scales, i.e., ECG scale in blue to the left and FIT scale in red to the right. It is important to mention that the bottom of each figure shows a sequence of witness bars. The distribution of such witness bars is in each case the FIT itself, that is to say,the accumulation of said bars has to do with the flanks of the original signal, in other words, the most pronounced flanks accumulate more bars, while less steep flanks accumulate less bars. This indicates us that the bars are witnessing an indirect flank detection, and thanks to FIT, and the existing spectral components thanks to the steep flank. Finally, *top-left*: equa-lized version of Eq.(153); *top-right*: averaged version of Eq.(154); *middle-left*: equalized one-dimensional mask with size equal to 3; *middle-right*: averaged one-dimensional mask with size equal to 3; *down-left*: equalized one-dimensional mask with size equal to 33; and, *down-right*: averaged one-dimensional mask with size equal to 33. The last four mentioned sub-figures are based on Eq.(155).

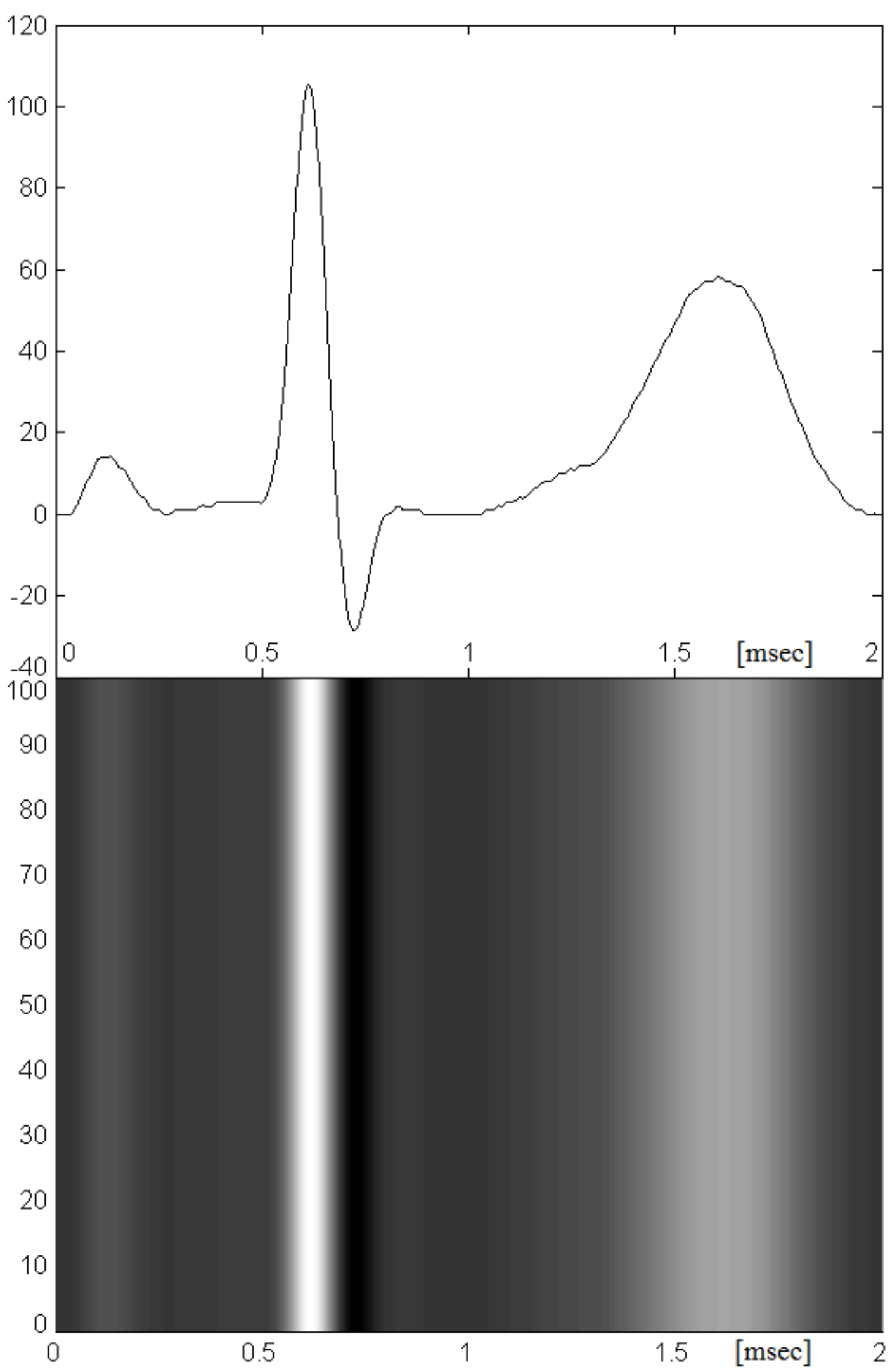

**Fig. 43** Top: electrocardiographic signal. Down: its waterfall.

As we can see in Fig.44, for a size of mask equal to 3, images of the first and second rows are identical, with the same profilometry. The change occurs when we increase the size of the one-dimensional mask. This increase is distinguished in the transition between the second and third row of the Fig.44, when the size changes from 3 to 33. In this last case, the FIT loses resolution and consequently for this reason it loss texture details of signal, and therefore, the focus is on grosser aspects relating to the flanks (low frequencies or approximations). In other words, the increase in mask size desensitizes the process regarding the details of the signal (high frequencies). In consequence, we say that FIT is a flank detector, with adjustable sensitivity by the size of the mask. This sensitivity is directly proportional to the resolution, i.e., it reacts with the spectral components which are represented by the degree of inclination of flanks in time. Finally, in Fig.44, we can notice that the FIT reaches the maximum where the signal has more pronounced flanks.

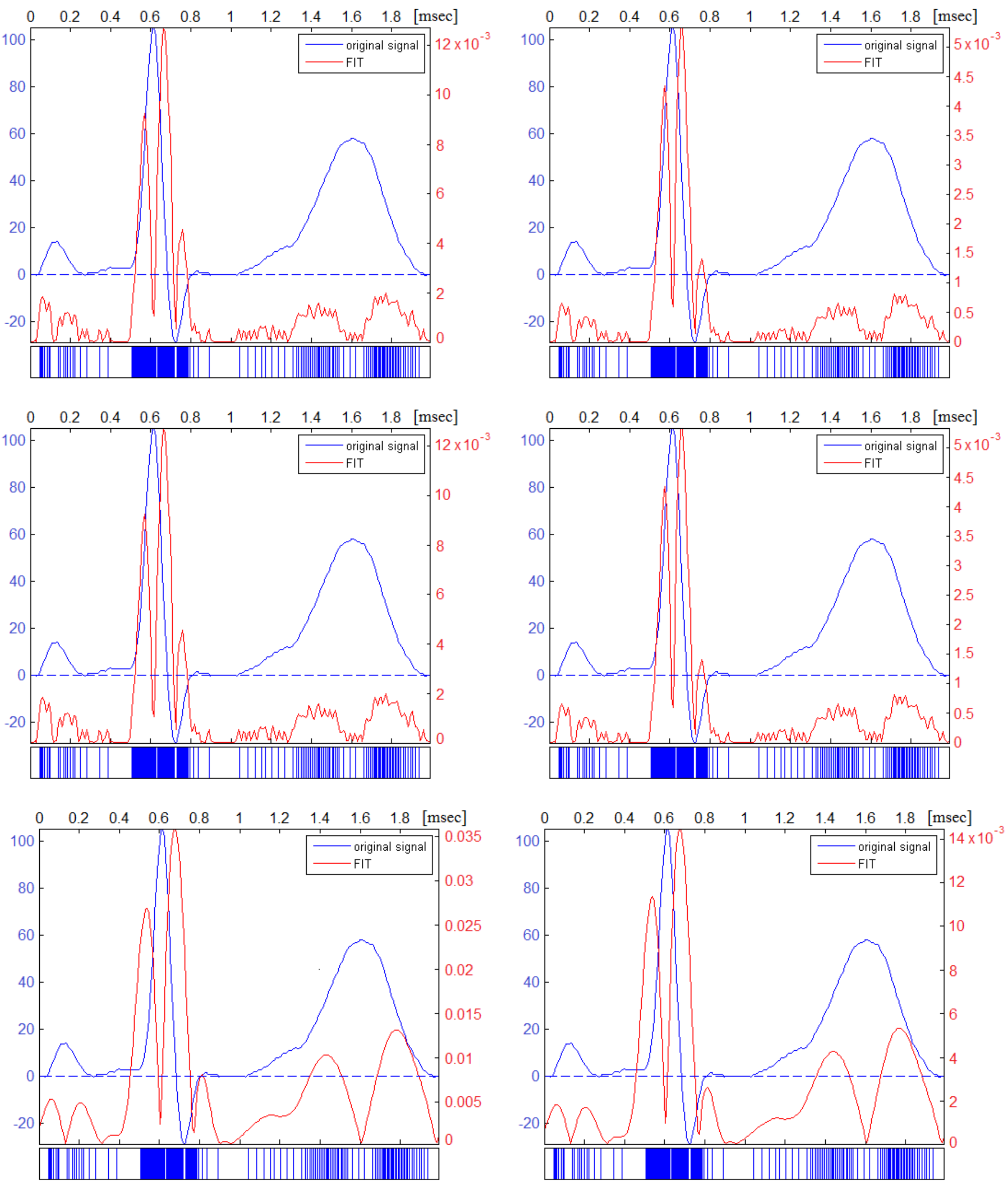

**Fig. 44** In all of sub-figures, we have: ECG signal in blue, and FIT in red with their respective scales, i.e., ECG scale in blue to the left and FIT scale in red to the right. Besides, ECG signal has 80 cycles per second, with 256 samples. It is important to mention that the bottom of each figure shows a sequence of witness bars. The distribution of such witness bars is each case the FIT itself, that is to say, the accumulation of said bars has to do with the flanks of the original signal, in other words, the most pronounced flanks accumulate more bars, while less steep flanks accumulate less bars. This indicates us that the bars are witnessing an indirect flank detection, and thanks to FIT, and the existing spectral components thanks to the steep flank. Finally, *top-left*: equalized version of Eq.(153); *top-right*: averaged version of Eq.(154); *middle-left*: equalized one-dimensional mask with size equal to 3; *middle-right*: averaged one-dimensional mask with size equal to 3; *down-left*: equalized one-dimensional mask with size equal to 33; and, *down-right*: averaged one-dimensional mask with size equal to 33.

A fourth version is due to a change in the Schrödinger equation, i.e., Eq.(118). We may reexpress this equation, to first order in the infinitesimal quantity *dt*, as

$$\begin{aligned} \left|\psi\left(t+dt\right)\right\rangle &= \left(I - \frac{i\,\hat{H}\left(t\right)dt}{\hbar}\right)\left|\psi\left(t\right)\right\rangle \\ &= \left(I - i\,\Omega\left(t\right)dt\right)\left|\psi\left(t\right)\right\rangle \end{aligned} \tag{156}$$

The operator $\mathrm{U}\left(t+dt,t\right) \equiv I - i\,\Omega\left(t\right)dt$ is unitary [205]; because $\Omega$ is self-adjoint it satisfies $U^{\dagger}U = 1$ to linear order in *dt*.

On the other hand, we can reexpress Eq.(140) as follow

$$\Delta\varpi_n = i\;\frac{s_{n+1}-s_{n-1}}{2s_n} \qquad \forall n \in \left[0, N-1\right], \text{ being } n \text{ the discrete time,} \tag{157}$$

where the signal should be padded.

With similar considerations to those carried out from Eq.(128) to (157), we arrive at

$$\Delta\varpi_n = i\;\frac{s_{n+1}-s_n}{s_n} \qquad \forall n \in \left[0, N-1\right], \text{ being } n \text{ the discrete time,} \tag{158}$$

from Eq.(156). Here, too, the signal must be padded.

The general case for the complete signal

$$\Delta\varpi_{diff} = i\;\; diff\left(S\right)/\,S\;, \tag{159}$$

being *diff*(•) the imaginary part of numerator of Eq.(158). This function is the same built-in function *diff*(•) of MATLAB® [49].

Therefore, the original FIT for this equation will be

$$\begin{aligned} \Delta f_{diff} &= \frac{1}{2\pi}\sqrt{\Delta\varpi_{diff}\;.\!\times conj\left(\Delta\varpi_{diff}\right)} \\ &= \frac{1}{2\pi}\sqrt{\left(i\; diff\left(S\right)./\,S\right).\!\times conj\left(i\; diff\left(S\right)./\,S\right)} \\ &= \frac{1}{2\pi}\left|diff\left(S\right)./\,S\right| = \frac{1}{2\pi}\left|diff\left(S\right)\right|./\left|S\right| \end{aligned} \qquad \text{for Eq.(159)} \tag{160}$$

Here, too, this version (the original) depends on a possible denominator equal to zero, therefore, we will use the next versions:

$$\begin{aligned} \Delta f_{diff,eq} &= \frac{1}{2\pi}\sqrt{\Delta\varpi_{diff,eq}\;.\!\times conj\left(\Delta\varpi_{diff,eq}\right)} \\ &= \frac{1}{2\pi}\sqrt{\left(i\;\; diff\left(S_{eq}\right)./\,S_{eq}\right).\!\times conj\left(i\;\; diff\left(S_{eq}\right)./\,S_{eq}\right)} \\ &= \frac{1}{2\pi}\left|diff\left(S_{eq}\right)./\,S_{eq}\right| = \frac{1}{2\pi}\left|diff\left(S_{eq}\right)\right|./\,S_{eq} \end{aligned} \qquad \text{(It is equalized like Eq.141)} \tag{161}$$

$$
\begin{aligned}
\Delta f_{diff,av} &= \frac{1}{2\pi}\sqrt{\Delta\varpi_{diff,av} \,.\!\times conj\left(\Delta\varpi_{diff,av}\right)} \\
&= \frac{1}{2\pi}\frac{\sqrt{\left(i \;\; diff\left(S\right)\right).\!\times conj\left(i \;\; diff\left(S\right)\right)}}{avg\left(S\right)} \qquad \text{(averaged)} \qquad (162) \\
&= \frac{1}{2\pi}\left|diff\left(S\right)\right| / \left|avg\left(S\right)\right|
\end{aligned}
$$

Figure 45 shows both version, i.e., FIT for Eq.(161) and (162) for the ECG signal of Fig.44. We can see that both sides of Fig.45 has the same profilometry of first and second row of Fig.44.

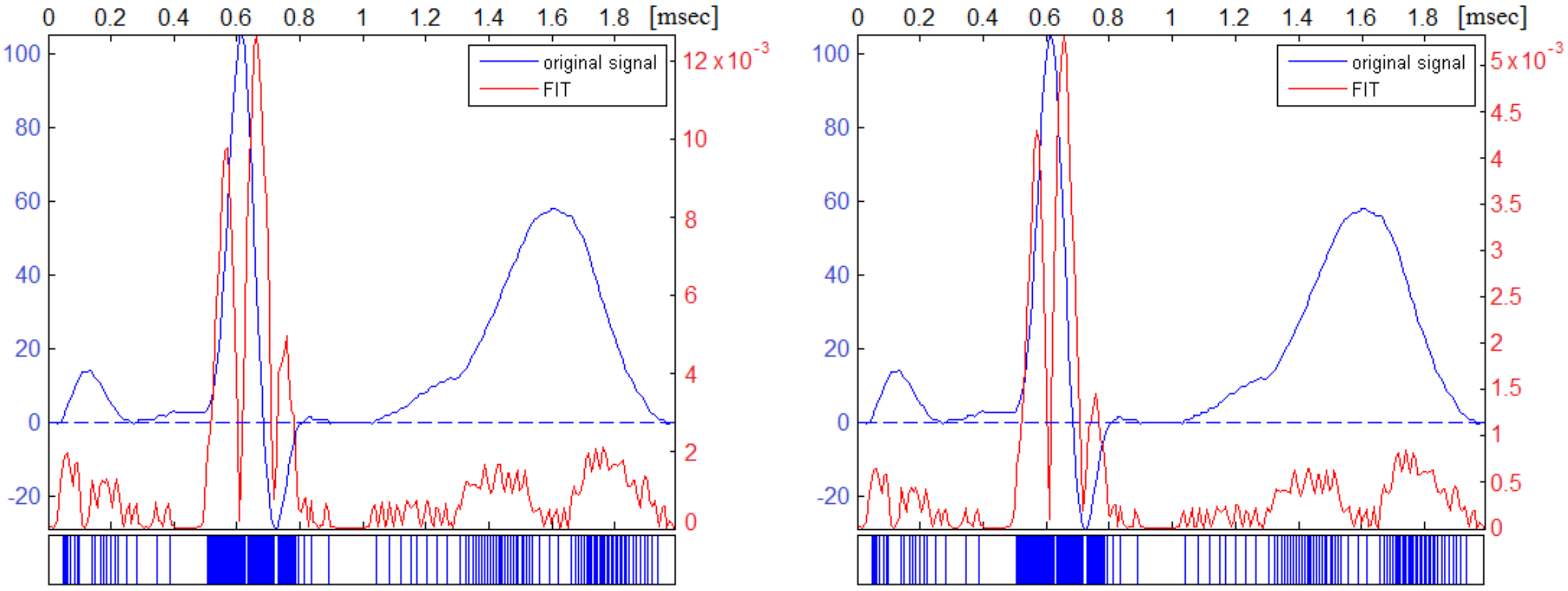

**Fig. 45** *Left*: equalized version, and *right*: averaged version.

If we had to perform a non-linear sampling of the original signal, FIT tell us what is the optimal number of samples and where to put them, all through the witness bars. It is like a pulse-position modulation (PPM) [206]. This procedure is much more efficient than a linear sampling. Summing-up, FIT represents the distribution of the witness bars.

***Case without overlap*** – Being $S = \left[s_0 \; s_1 \; s_2 \ldots , \; s_n , \ldots \; s_{N-1}\right]$, $\forall n \in \left[0, N-1\right]$, being $n$ the discrete time. Thus, we will have two components

$$
\begin{aligned}
L^1 &= \left[l_0 , l_1 , \ldots , l_k , \ldots , l_{\frac{N}{2}-1}\right] \\
&= \left[\frac{s_0 + s_1}{2}, \frac{s_2 + s_3}{2}, \ldots , \frac{s_n + s_{n+1}}{2}, \ldots , \frac{s_{N-2} + s_{N-1}}{2}\right]
\end{aligned}
\qquad \text{(approximation subband)} \qquad (163)
$$

and

$$
\begin{aligned}
H^1 &= \left[h_0 , h_1 , \ldots , h_k , \ldots , h_{\frac{N}{2}-1}\right] \\
&= \left[\frac{-s_0 + s_1}{2}, \frac{-s_2 + s_3}{2}, \ldots , \frac{-s_n + s_{n+1}}{2}, \ldots , \frac{-s_{N-2} + s_{N-1}}{2}\right]
\end{aligned}
\qquad \text{(detail subband)} \qquad (164)
$$

$\forall k \in \left[0, \frac{N}{2}-1\right]$, being $l_k = \frac{s_n + s_{n+1}}{2}$ and $h_k = \frac{-s_n + s_{n+1}}{2}$, with

$$
L^0 = S \qquad (165)
$$

$L^1$ and $H^1$ are the first level subbands of Haar basis wavelet. $L^1$ is the approximation subband (low frequency) and $H^1$ is the detail subband (high frequency). $L^0$ is the basal level or original signal. Then, QSA will be,

$$\Delta\varpi = i \;\; H^1 ./ L^1 , \tag{166}$$

while, FIT is,

$$\begin{aligned} \Delta f &= \frac{1}{2\pi}\sqrt{\Delta\varpi \;.\times conj\left(\Delta\varpi\right)} \\ &= \frac{1}{2\pi}\sqrt{\left(i\; H^1 ./ L^1\right) .\times conj\left(i\; H^1 ./ L^1\right)} \\ &= \frac{1}{2\pi}\left|H^1 ./ L^1\right| = \frac{1}{2\pi}\left|H^1\right| ./ L^1 \end{aligned} \tag{167}$$

As we see in the next section, it is very useful also another version of FIT for the non-overlapping case,

$$\begin{aligned} \Delta f &= \frac{1}{2\pi} abs\left(\Delta\varpi\right) \\ &= \frac{1}{2\pi}\left|H^1\right| ./ L^1 \end{aligned} \tag{168}$$

where, *abs*(•) means absolute value of (•). On the other hand, both of them, have the same problem, i.e., they depends on a possible denominator equal to zero, therefore, we will use the next versions:

1) *Equalization:* we are going to equalize $S$ between 1 and 2 ($S_{eq}$), like Eq.(141), and then, we will built $\left[H^1_{eq}, L^1_{eq}\right]$ with it, i.e., $\Delta\varpi_{eq}$, and hence,

$$\begin{aligned} \Delta f_{eq} &= \frac{1}{2\pi}\sqrt{\Delta\varpi_{eq} \;.\times conj\left(\Delta\varpi_{eq}\right)} \\ &= \frac{1}{2\pi}\sqrt{\left(i \;\; H^1_{eq} ./ L^1_{eq}\right) .\times conj\left(i \;\; H^1_{eq} ./ L^1_{eq}\right)} \\ &= \frac{1}{2\pi}\left|H^1_{eq} ./ L^1_{eq}\right| = \frac{1}{2\pi}\left|H^1_{eq}\right| ./ L^1_{eq} \end{aligned} \tag{169}$$

and

$$\begin{aligned} \Delta f_{eq} &= \frac{1}{2\pi} abs\left(\Delta\varpi_{eq}\right) \\ &= \frac{1}{2\pi}\left|H^1_{eq}\right| ./ L^1_{eq} \end{aligned} \tag{170}$$

2) *Averaging:* we will proceed to show this version directly starting with QSA,

$$\Delta\varpi_{av} = i \;\; H^1 / avg\left(L^1\right), \tag{171}$$

Then, FIT will be,

$$\begin{aligned} \Delta f_{av} &= \frac{1}{2\pi}\sqrt{\Delta\varpi_{av} \;.\times conj\left(\Delta\varpi_{av}\right)} \\ &= \frac{1}{2\pi}\frac{\sqrt{\left(i \;\; H^1\right) .\times conj\left(i \;\; H^1\right)}}{avg\left(L^1\right)} \\ &= \frac{1}{2\pi}\left|H^1\right| / \left|avg\left(L^1\right)\right| \end{aligned} \tag{172}$$

or

$$
\begin{aligned}
\Delta f_{av} &= \frac{1}{2\pi} abs\left(\Delta\varpi_{av}\right) \\
&= \frac{1}{2\pi}\left|H^{1}\right| / \left|avg\left(L^{1}\right)\right|
\end{aligned}
\tag{173}
$$

Finally, none of non-overlapping version needs padding.

***Examples -*** Next, we will implement the four seen cases of FIT on an ECG signal of 256 samples. As we can see in Fig.46, the first row show us in red both versions of FIT for the equalized case. However, the second row show us $L^{1}_{eq}$ in blue and $H^{1}_{eq}$ in red. The second version of FIT will be very useful in Section of Applications, for signals as well as for images. Apparently, the second version of FIT and $H^{1}_{eq}$ are identical, however, they have the same profilometry with the same scale. $L^{1}_{eq}$ has the same profilometry of the original signal with different scales. Besides, all of them have the half number of samples that original signal, i.e., 128 instead of 256.

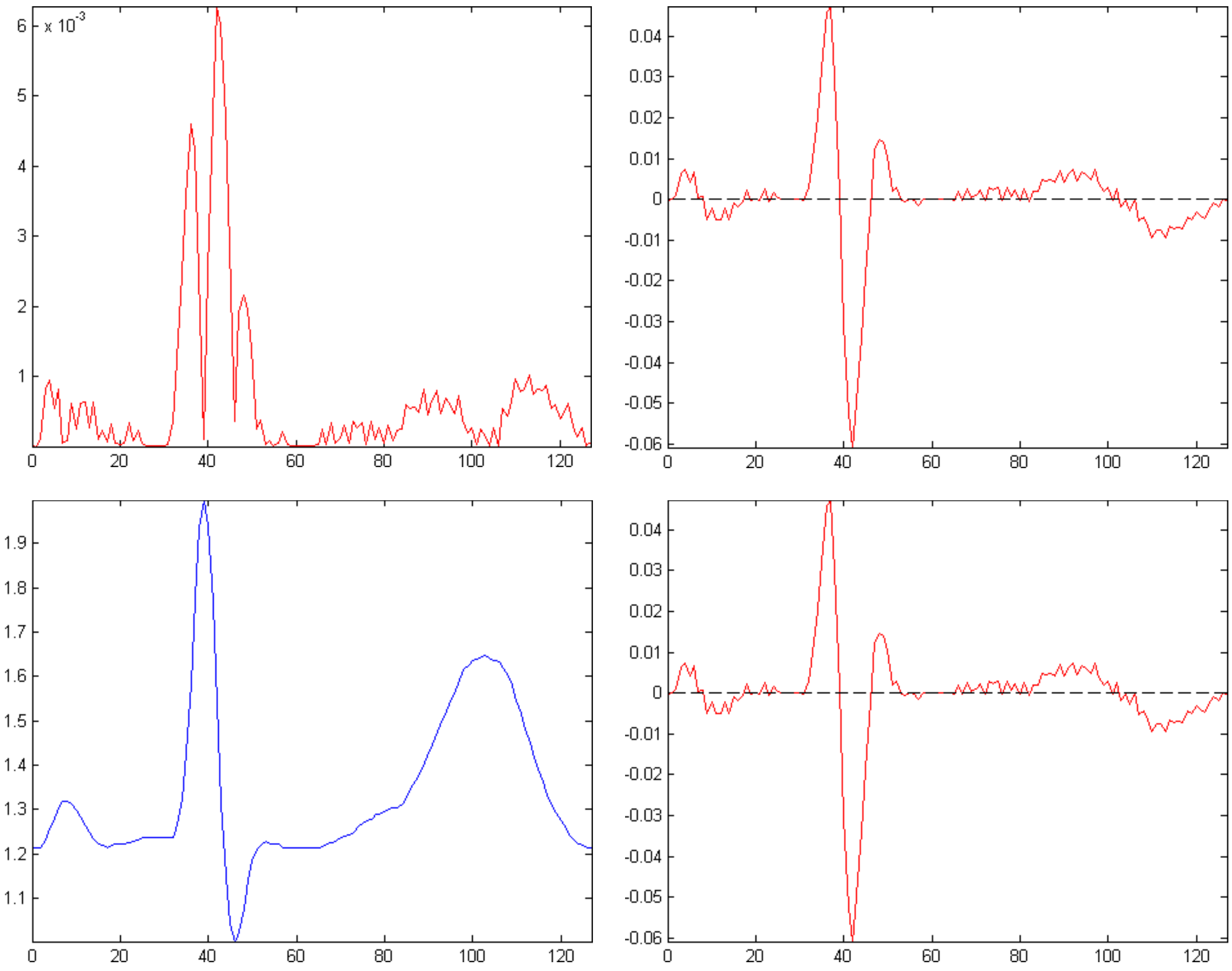

**Fig. 46 Equalized versions**. *Top-Left*: first version of FIT (Eq.169), *top-right*: second version of FIT (Eq.170), *down-left*: $L^{1}_{eq}$, and, *down-right*: $H^{1}_{eq}$. Second version of FIT and $H^{1}_{eq}$ are apparently identical, however, they have the same profilometry with the same scale. $L^{1}_{eq}$ has the same profilometry of the original signal with different scales. Besides, all of them have the half number of samples that original signal, i.e., 128 instead of 256.

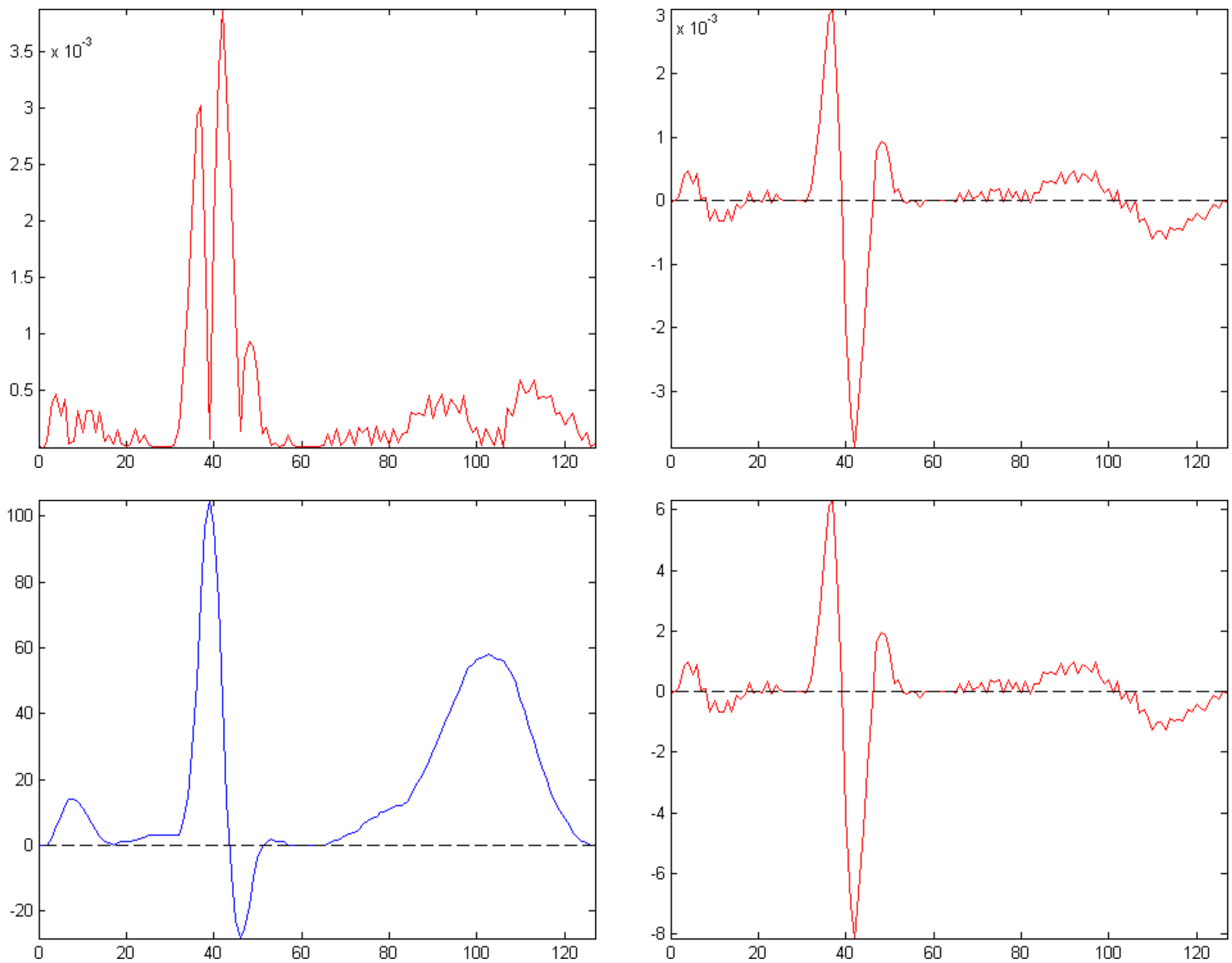

**Fig. 47 Averaged versions**. *Top-Left*: first version of FIT (Eq.172), *top-right*: second version of FIT (Eq.173), *down-left*: $L^1_{no}$, and, *down-right*: $H^1_{no}$. Second version of FIT and $H^1_{no}$ have different scales, however, they have the same profilometry. $L^1_{no}$ has the same profilometry of the original signal with the same scale. Besides, here too, all of them have the half number of samples that original signal.

On the other hand, the first row of Fig.47 show us in red both versions of FIT for the averaged case. However, the second row show us $L^1_{no}$ in blue and $H^1_{no}$ in red. Here too, the second version of FIT will be very useful in Section of Applications, for signals as well as for images. Second version of FIT and $H^1_{no}$ have different scales, however, they have the same profilometry. Besides, $L^1_{no}$ has the same profilometry of the original signal with the same scale. Here too, all of them have the half number of samples that original signal.

### 3.3.2 FIT for images

***Case with overlap*** – In the classical version of Eq.(134) but in the 2D case, and for each color (i.e., RGB), we are going to replace qubits by pixels of a real image, therefore, the classical version of Eq.(134) is represented by three directional components, depending on the direction of each derivative for each color. In this context, we will have four possibilities to make FIT analysis over an image with overlap, i.e., with:

a) segmental mask and $M = 3$,
b) square mask and $M = 3$,
c) segmental mask and $M > 3$, and
d) square mask and $M > 3$

In all cases, the image is padded depending on the value of M, i.e., if the image (e.g., red channel: $I_R$) has a ROW-by-COL size, then, $I_{R,P}$ (padded $I_R$) will have a (ROW+2L)-by-(COL+2L) size, where L = (M-1)/2. Therefore, the original image $I_R$ will be in the middle of the padded image $I_{R,P}$, which will have four lateral margins of L size to each side of $I_R$ composed exclusively by zeros.

<u>*Segmental mask and M = 3*</u>*:* in this case, we will have two mask, namely:

$$N_H = \frac{[-1\ 0\ 1]}{2}, \quad \text{(horizontal mask), and} \tag{174}$$

$$N_V = N_H^T, \quad \text{(vertical mask).} \tag{175}$$

The procedure begins with a two-dimensional convolution (first horizontal and then, vertical rafters) between $N_H$ and $I_{R,P}$, i.e.,

$$I_H = N_H * I_{R,P} \tag{176}$$

After that, we continue with another two-dimensional convolution (first vertical and then, horizontal rafters) between $N_V$ and $I_{R,P}$, i.e.,

$$I_V = N_V * I_{R,P} \tag{177}$$

Finally, $\dot{I}$ is obtained via Pythagoras between $I_H$, and $I_V$, that is to say,

$$\dot{I} = \sqrt{I_H^2 + I_V^2} \tag{178}$$

Then, we obtain the two-dimensional version of Eq.(139), that is,

$$\Delta\varpi = i \;\; \dot{I} ./ I, \tag{179}$$

While, for each pixel, we will have,

$$\Delta\varpi_{r,c} = i \;\; \dot{I}_{r,c} / I_{r,c} \qquad \forall r \in [1, ROW], and\ c \in [1, COL] \tag{180}$$

Equation (180) is the discrete version of $\Delta\omega$ in its most inapplicable form, given that this is not applicable in cases where the denominator is zero (although unlike the FFT, $\Delta\omega$ has solution), without mentioning that ω is an imaginary operator to be applied to real images. Therefore, this form is called raw version. To overcome this drawback, we use equalized and/or averaged versions (like signal case), as the following,

$$\Delta\varpi_{eq} = i \;\; \dot{I}_{eq} ./ I_{eq}, \tag{181}$$

where subscript "*eq*" means *equalized*. In general, we will pass each pixel of each channel of color of *I* from $[0,2^8-1]$ to $[1, 2^8]$.

While the averaged form will be,

$$\Delta\varpi_{av} = i \;\; \dot{I} / avg(I), \tag{182}$$

where $avg(I)$ is the average of *I*. In both cases the image must be enlarged (padded) as we explained before. For example, we see this in more detail in the third version,

$$\Delta\varpi_{bm} = i \;\; \dot{I} / I_{bm}, \tag{183}$$

where the subscript *bm* means *two-dimensional mask*.

Here too, we define two segmental mask,

$$D_H = \frac{[1\,1\,1]}{3}, \quad \text{(horizontal mask), and} \tag{184}$$

$$D_V = D_H^T, \quad \text{(vertical mask).} \tag{185}$$

The procedure begins with a two-dimensional convolution (first horizontal and then, vertical rafters) between $D_H$ and $I_{R,P}$, i.e.,

$$I_H = D_H * I_{R,P} \tag{186}$$

After that, we continue with another two-dimensional convolution (first vertical and then, horizontal rafters) between $S_V$ and $I_{R,P}$, i.e.,

$$I_V = D_V * I_{R,P} \tag{187}$$

Finally, $I_{bm}$ is obtained via Pythagoras between $I_H$, and $I_V$, that is to say,

$$I_{bm} = \sqrt{I_H^2 + I_V^2} \tag{188}$$

On the other hand, and to save the fact that $\Delta\varpi$ is an imaginary operator to be applied to real images (in all four versions), we will use another version known as *frequency in time* (FIT) for all mentioned versions:

$$\begin{aligned} \Delta f &= \frac{1}{2\pi}\sqrt{\Delta\varpi \;.\!\times conj\left(\Delta\varpi\right)} \\ &= \frac{1}{2\pi}\sqrt{\left(i\;\dot{I}./I\right).\!\times conj\left(i\;\dot{I}./I\right)} \qquad \text{for Eq.(179)} \\ &= \frac{1}{2\pi}\left|\dot{I}./I\right| = \frac{1}{2\pi}\left|\dot{I}\right|./I \end{aligned} \tag{189}$$

Being $\Delta f = \Delta\omega / 2\pi$ a matrix of ROW-by-COL frequency in hertz. Remember that, this version (the original) depends on a possible denominator equal to zero, therefore, we will use the next versions:

$$\begin{aligned} \Delta f_{eq} &= \frac{1}{2\pi}\sqrt{\Delta\varpi_{eq} \;.\!\times conj\left(\Delta\varpi_{eq}\right)} \\ &= \frac{1}{2\pi}\sqrt{\left(i\;\dot{I}_{eq}./I_{eq}\right).\!\times conj\left(i\;\dot{I}_{eq}./I_{eq}\right)} \qquad \text{for Eq.(181)} \\ &= \frac{1}{2\pi}\left|\dot{I}_{eq}./I_{eq}\right| = \frac{1}{2\pi}\left|\dot{I}_{eq}\right|./I_{eq} \end{aligned} \tag{190}$$

$$\begin{aligned} \Delta f_{av} &= \frac{1}{2\pi}\sqrt{\Delta\varpi_{av} \;.\!\times conj\left(\Delta\varpi_{av}\right)} \\ &= \frac{1}{2\pi}\frac{\sqrt{\left(i\;\dot{I}\right).\!\times conj\left(i\;\dot{I}\right)}}{avg\left(I\right)} \qquad \text{for Eq.(182)} \\ &= \frac{1}{2\pi}\left|\dot{I}\right|/avg\left(I\right) \end{aligned} \tag{191}$$

and finally, one version more based on Eq.(183), that is to say,

$$
\begin{aligned}
\Delta f_{um} &= \frac{1}{2\pi}\sqrt{\Delta\varpi_{um} \; .\times conj\left(\Delta\varpi_{um}\right)} \\
&= \frac{1}{2\pi}\sqrt{\left(i \;\; \dot{I}./\, I_{bm}\right).\times conj\left(i \;\; \dot{I}./\, I_{bm}\right)} \\
&= \frac{1}{2\pi}\left|\dot{I}./\, I_{bm}\right| = \frac{1}{2\pi}\left|\dot{I}\right|./\, I_{bm}
\end{aligned} \tag{192}
$$

Because usually a (or more) denominator of Eq.(183) is zero, then here also it is necessary to use a procedure of equalization or normalization as in the previous cases.

***Examples -*** Next, we will implement the seen case. First, we begin with the selected color image, that is to say: Angelina, a picture of 1920-by-1080 pixel with 24 bpp. See Fig.48.

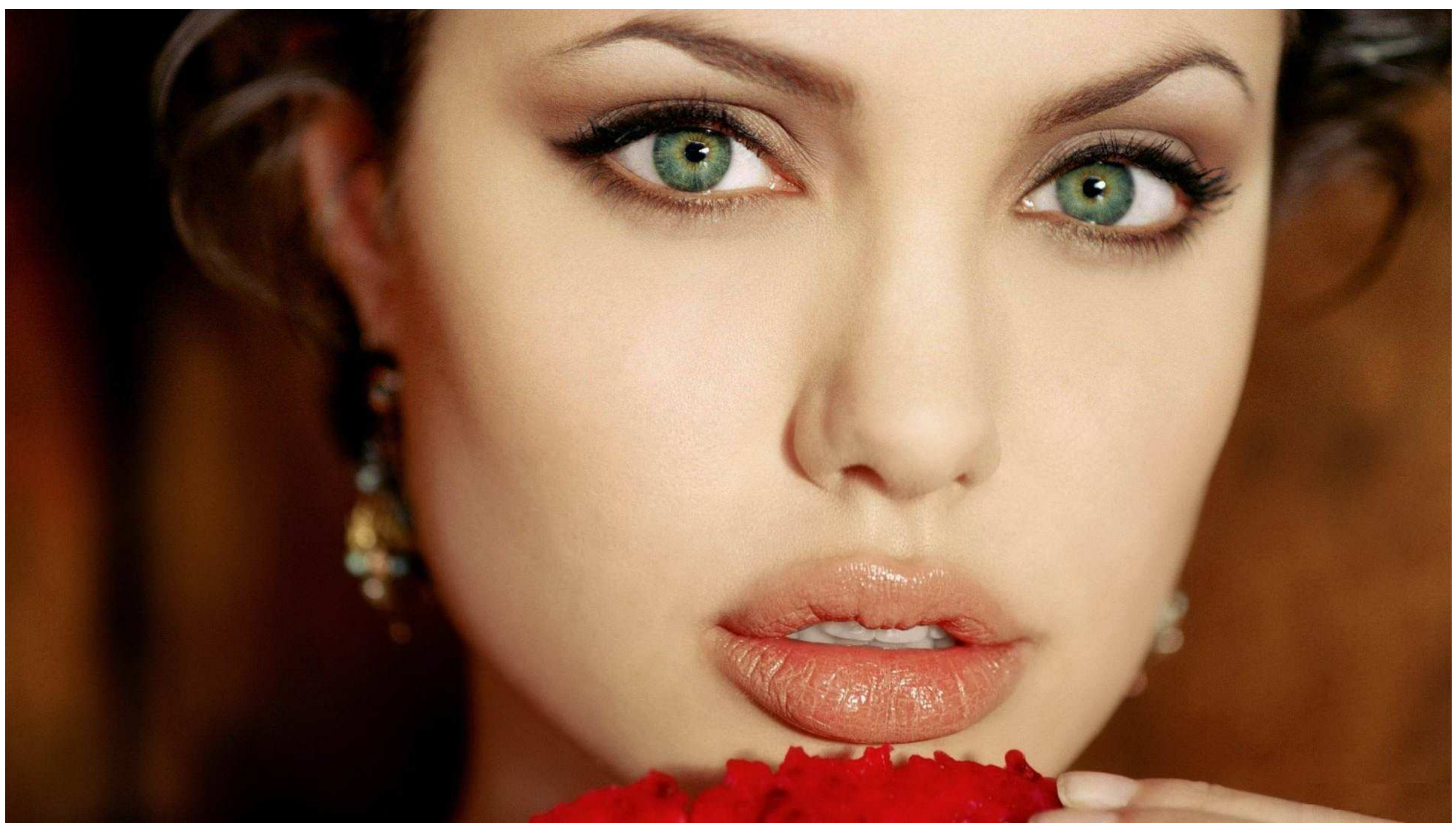

**Fig. 48 Angelina:** 1920-by-1080 pixels, with 24 bpp.

Figure 49 show us the FIT over Angelina for segmental mask with M = 3 and overlap, where the first column is for red channel, second column for green channel, and third column for blue channel. Besides, first row represents the raw case, second row the equalized case, third row the averaged case, and fourth row the two-dimensional mask case.

In Fig.49, we can see the texture and edges of the different versions of FIT. The same set of images show us Regions of Interest (ROIs), which include ergodic areas with a notable impact in the filtering (denoising) and compression contexts. On the other hand, the FIT by each color indicate us the weight of this one over the main morphological characteristics of the image.

Although the scale is different, the twelve images of Fig.49 have the same resolution of Fig.48, that is to say, FIL-by-COL. Besides, in all these images we have manipulated the brightness and contrast for better display scroll of them. Finally, FIT permit us to observe spectral components per pixel by color with a particular emphasis in texture and edges, which are notably important in applications such as visual intelligence for computer vision, image compression, superresolution, forensic analysis of images, filtering (denoising), image restoration and enhancement [45-48].

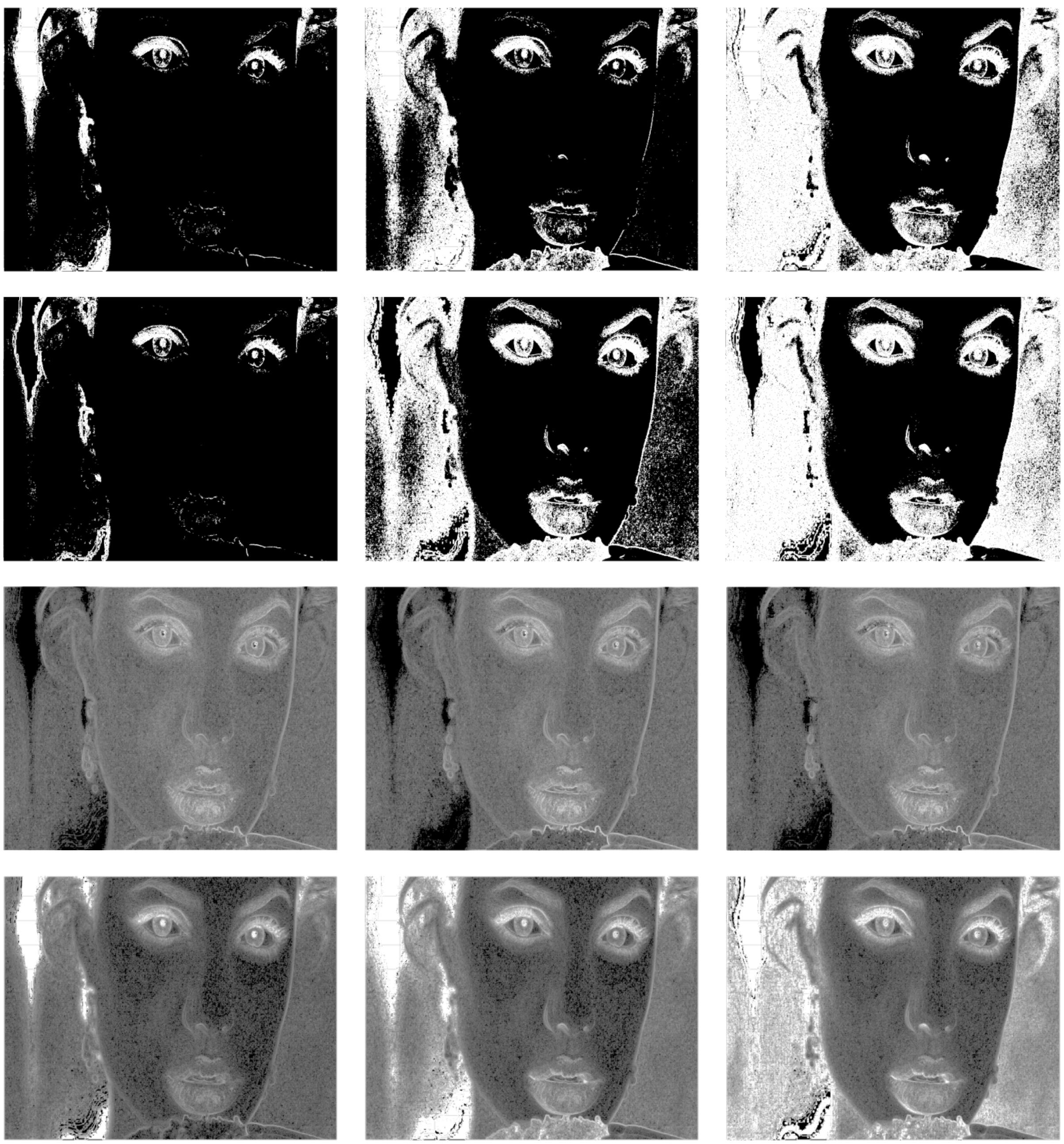

**Fig. 49 FIT over Angelina, segmental mask with M = 3 and overlap:** first column for red channel, second column for green channel, and third column for blue channel. Besides, first row for raw case, second row for equalized case, third row for averaged case, and fourth row for two-dimensional mask case.

<u>*Square mask and M = 3*</u>*:* in this case, we will have one mask alone, namely:

$$N = \begin{bmatrix} -1 & -1 & 0 \\ -1 & 0 & 1 \\ 0 & 1 & 1 \end{bmatrix} / 6 \tag{193}$$

The procedure begins with a two-dimensional convolution (first horizontal and then, vertical rafters) between $N$ and $I_{R,P}$, i.e.,

$$\dot{I} = N * I_{R,P} \tag{194}$$

Then, we obtain the two-dimensional version of Eq.(139) newly, that is,

$$\Delta\varpi = i \;\; \dot{I} ./ I \,, \tag{195}$$

While, for each pixel, we will have,

$$\Delta\varpi_{r,c} = i \;\; \dot{I}_{r,c} / I_{r,c} \qquad \forall r \in [1, ROW],\ and\ c \in [1, COL] \tag{196}$$

Equation 196 has the same problem of Eq.(180), i.e., eventual cases where denominator is zero. Thus, to overcome this drawback, we use equalized and/or normalized versions like Equations (181) and (182) with the same criterion.

A third version based on Eq.(183) can be implemented via a square mask, where

$$D = \begin{bmatrix} 1 & 1 & 1 \\ 1 & 1 & 1 \\ 1 & 1 & 1 \end{bmatrix} / 9 \tag{197}$$

The procedure consists in a two-dimensional convolution (first horizontal and then, vertical rafters) between $D$ and $I_{R,P}$, i.e.,

$$I_{bm} = D * I_{R,P} \tag{198}$$

Here too, and to save the fact that $\Delta\varpi$ is an imaginary operator to be applied to real images (in all four versions), we will use another version known as *frequency in time* (FIT, i.e., its two-dimensional version) for all mentioned versions:

$$\begin{aligned} \Delta f &= \frac{1}{2\pi}\sqrt{\Delta\varpi \,.\!\times conj\left(\Delta\varpi\right)} \\ &= \frac{1}{2\pi}\sqrt{\left(i \;\; \dot{I}./ I\right).\!\times conj\left(i \;\; \dot{I}./ I\right)} \qquad \text{for Eq.(179)} \\ &= \frac{1}{2\pi}\left|\dot{I}./ I\right| = \frac{1}{2\pi}\left|\dot{I}\right|./ I \end{aligned} \tag{199}$$

Being $\Delta f = \Delta\omega / 2\pi$ (newly) a matrix of ROW-by-COL frequencies in hertz. Remember that, this version (the original) depends on a possible denominator equal to zero, therefore, we will use the next versions:

$$\begin{aligned} \Delta f_{eq} &= \frac{1}{2\pi}\sqrt{\Delta\varpi_{eq} \,.\!\times conj\left(\Delta\varpi_{eq}\right)} \\ &= \frac{1}{2\pi}\sqrt{\left(i \;\; \dot{I}_{eq}./ I_{eq}\right).\!\times conj\left(i \;\; \dot{I}_{eq}./ I_{eq}\right)} \qquad \text{for Eq.(181)} \\ &= \frac{1}{2\pi}\left|\dot{I}_{eq}./ I_{eq}\right| = \frac{1}{2\pi}\left|\dot{I}_{eq}\right|./ I_{eq} \end{aligned} \tag{200}$$

$$\begin{aligned} \Delta f_{av} &= \frac{1}{2\pi}\sqrt{\Delta\varpi_{av} \,.\!\times conj\left(\Delta\varpi_{av}\right)} \\ &= \frac{1}{2\pi}\frac{\sqrt{\left(i \;\; \dot{I}\right).\!\times conj\left(i \;\; \dot{I}\right)}}{avg\left(I\right)} \qquad \text{for Eq.(182)} \\ &= \frac{1}{2\pi}\left|\dot{I}\right| / avg\left(I\right) \end{aligned} \tag{201}$$

and finally, one version more based on Eq.(183), in which the denominator arises from the convolution of a mask with the original image, that is to say,

$$
\begin{aligned}
\Delta f_{um} &= \frac{1}{2\pi}\sqrt{\Delta\varpi_{um} \;.\times conj\left(\Delta\varpi_{um}\right)} \\
&= \frac{1}{2\pi}\sqrt{\left(i \;\; \dot{I}./ I_{bm}\right).\times conj\left(i \;\; \dot{I}./ I_{bm}\right)} \\
&= \frac{1}{2\pi}\left|\dot{I}./ I_{bm}\right| = \frac{1}{2\pi}\left|\dot{I}\right|./ I_{bm}
\end{aligned}
\tag{202}
$$

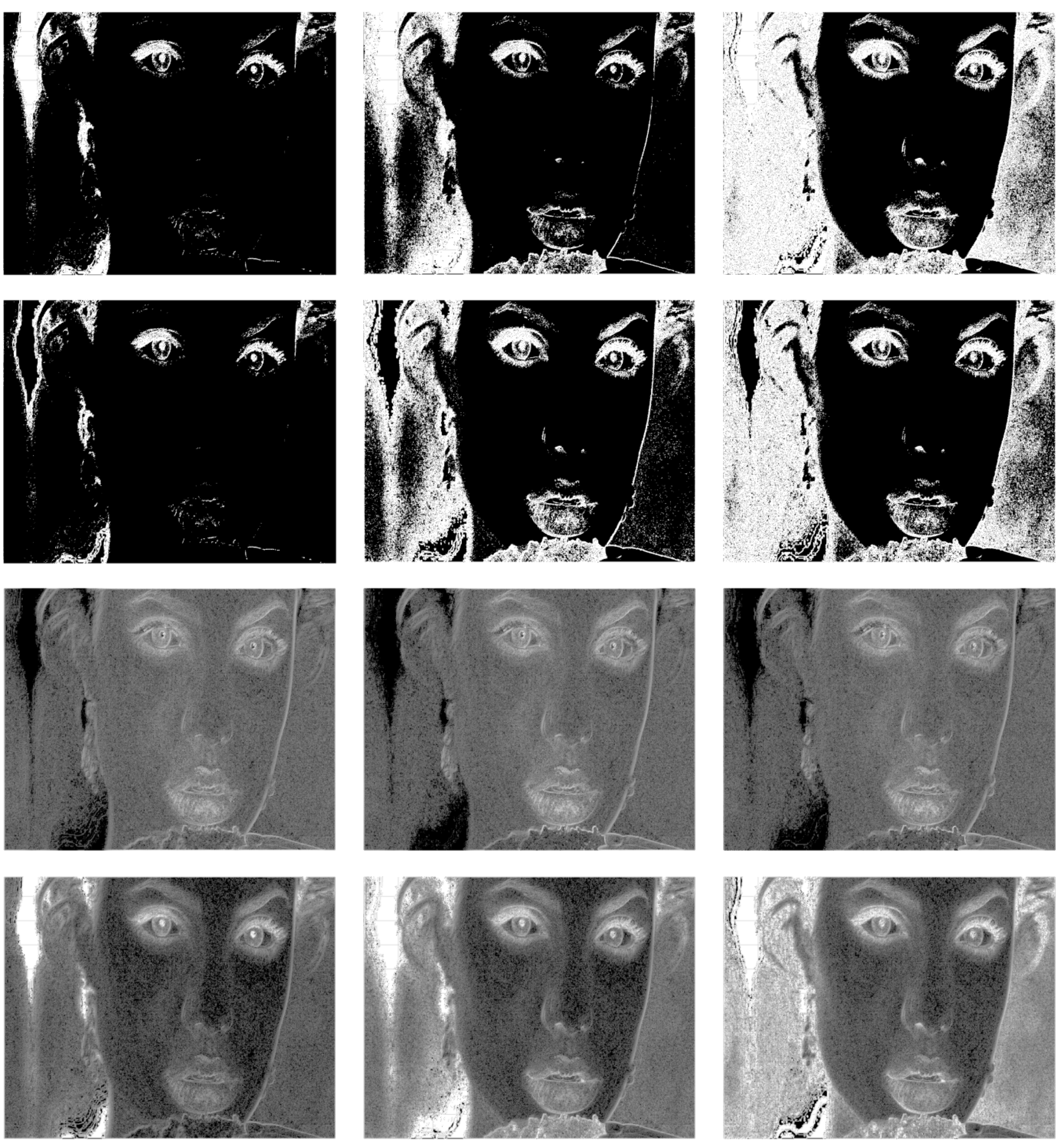

**Fig. 50 FIT over Angelina, square mask with M = 3 and overlap:** first column for red channel, second column for green channel, and third column for blue channel. Besides, first row for raw case, second row for equalized case, third row for averaged case, and fourth row for two-dimensional mask case.

Here, we can make the same considerations that we make after Eq.(192) regarding to Eq.(183) and their possible denominators equal to zero in it.

Figure 50 show us the FIT over Angelina for square mask with M = 3 and overlap, where the first column is for red channel, second column for green channel, and third column for blue channel. Besides, first row represents the raw case, second row the equalized case, third row the averaged case, and fourth row the two-dimensional mask case.

Otherwise, considerations about the Fig.50 are similar to the previous case.

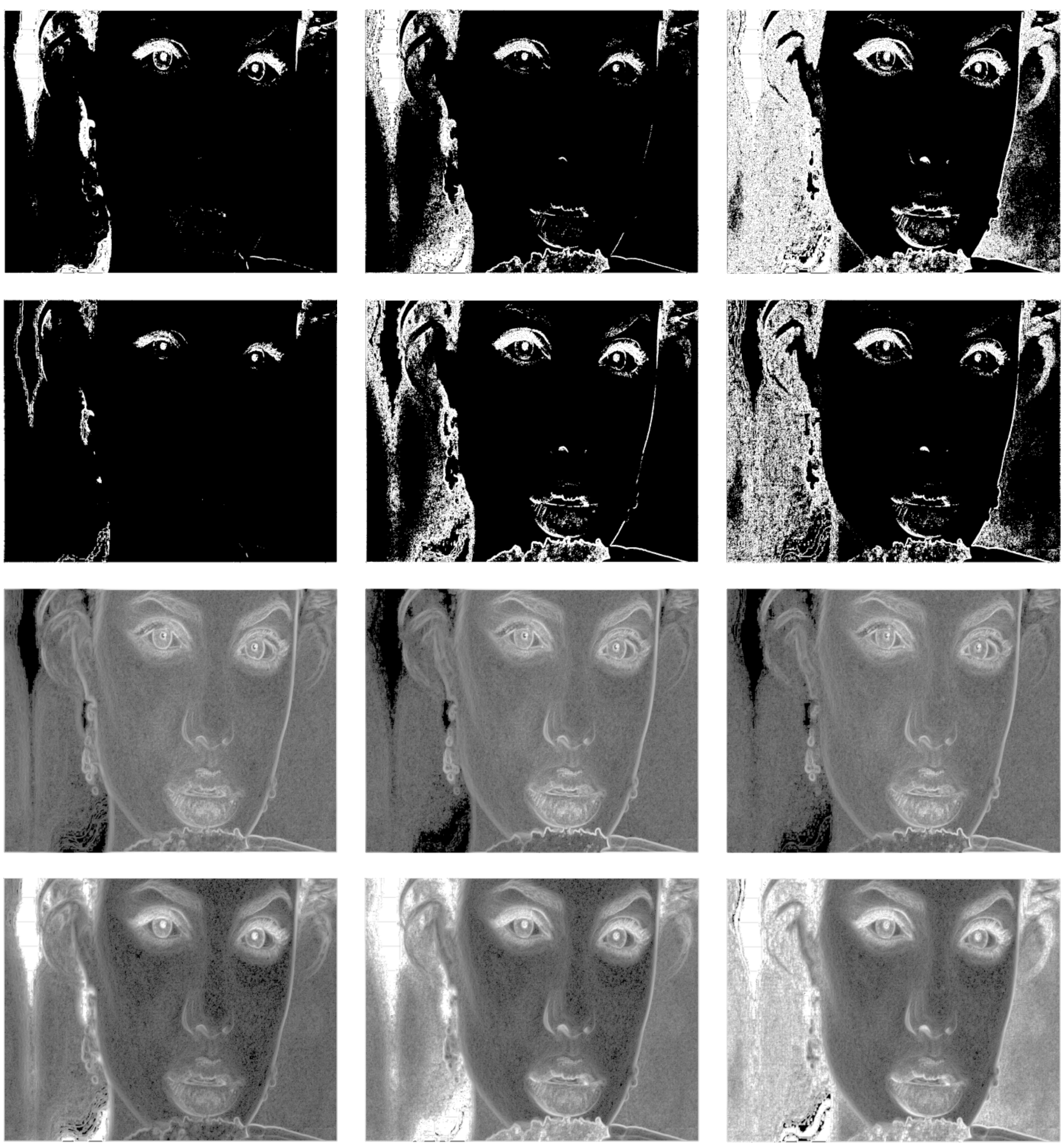

**Fig. 51 FIT over Angelina, segmental mask with M > 3 (in fact, M = 7) and overlap:** first column for red channel, second column for green channel, and third column for blue channel. Besides, first row for raw case, second row for equalized case, third row for averaged case, and fourth row for two-dimensional mask case.

*Segmental mask and $M > 3$:* in this case, only it is necessary correct Equations (174) and (184), i.e.:

$$N_H = \frac{[-1 \ldots -1\,0\,1 \ldots 1]}{M-1}, \quad \text{(horizontal mask)} \tag{203}$$

with ($M$-1)/2 repetitions of -1, only one 0, and ($M$-1)/2 repetitions of 1, and

$$D_H = \frac{[1 \ldots 1]}{M}, \quad \text{(horizontal mask),} \tag{204}$$

with $M$ repetitions of 1.

Otherwise, it's all the same to the case of $M = 3$.

Figure 51 show us the FIT over Angelina for segmental mask with $M > 3$ (in fact, $M = 7$) and overlap, where the first column is for red channel, second column for green channel, and third column for blue channel. Besides, first row represents the raw case, second row the equalized case, third row the averaged case, and fourth row the two-dimensional mask case.

Figure 51 shows us similar characteristics to previous case.

*Square mask and $M > 3$:* in this case, only it is necessary correct Equations (193) and (197), i.e.:

$$N = \begin{bmatrix} & -1 & & & 0 \\ & & & \diagup & \\ & & 0 & & \\ & \diagup & & 1 & \\ 0 & & & & \end{bmatrix} \Big/ \left[ (M-1)M \right] \tag{205}$$

with $\frac{M(M-1)}{2}$ elements equal to -1, $M$ elements equal to 0, and $\frac{M(M-1)}{2}$ elements equal to 1, and

$$D = \begin{bmatrix} 1 & \cdots & 1 & \cdots & 1 \\ \vdots & \ddots & \vdots & ⋰ & \vdots \\ 1 & \cdots & 1 & \cdots & 1 \\ \vdots & ⋰ & \vdots & \ddots & \vdots \\ 1 & \cdots & 1 & \cdots & 1 \end{bmatrix} / M^2 \tag{206}$$

with $M^2$ elements equal to 1.

Otherwise, it's all the same to the case of $M = 3$.

Figure 52 show us the FIT over Angelina for square mask with $M > 3$ (in fact, $M = 7$) and overlap, where the first column is for red channel, second column for green channel, and third column for blue channel. Besides, first row represents the raw case, second row the equalized case, third row the averaged case, and fourth row the two-dimensional mask case.

As in the previous cases, Fig.52 shows us similar characteristics to previous cases. This shows us the consistency and coherence of the FIT as a method for spectral analysis.

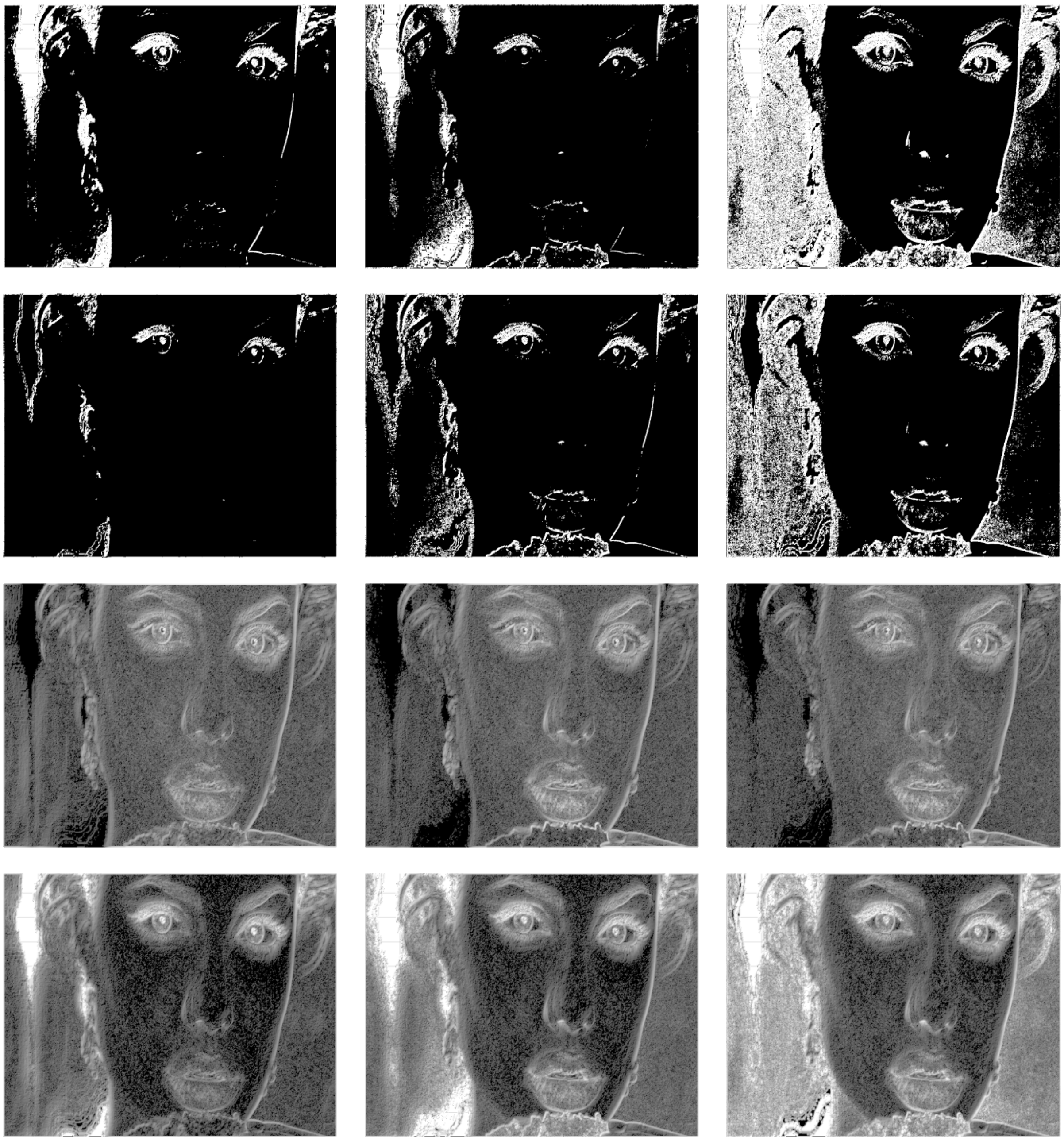

**Fig. 52 FIT over Angelina, square mask with M > 3 (in fact, M = 7) and overlap:** first column for red channel, second column for green channel, and third column for blue channel. Besides, first row for raw case, second row for equalized case, third row for averaged case, and fourth row for two-dimensional mask case.

***Case without overlap* –** Being $I = [I_R , I_G , I_B ]$, that is to say, the three color channels of the original image I, if we choose a color channel at time, e.g., $I_{R_r_0,c_0} \forall r_0 \in [1, ROW], \wedge c_0 \in [1, COL]$, thus, we will have four components (subbands)

$$LL^0_{R_r_0,c_0} = I_{R_r_0,c_0} \tag{207}$$

$$LL^1_{R_r_1,c_1} = \frac{LL^0_{R_2r_0-1,2c_0-1} + LL^0_{R_2r_0-1,2c_0} + LL^0_{R_2r_0,2c_0-1} + LL^0_{R_2r_0,2c_0}}{4} \quad \text{(approximation subband)} \tag{208}$$

$$LH^{1}_{R_r_1,c_1} = \frac{-LL^{0}_{R_2r_0-1,2c_0-1} + LL^{0}_{R_2r_0-1,2c_0} - LL^{0}_{R_2r_0,2c_0-1} + LL^{0}_{R_2r_0,2c_0}}{4} \text{ (horizontal detail subband)} \quad (209)$$

$$HL^{1}_{R_r_1,c_1} = \frac{-LL^{0}_{R_2r_0-1,2c_0-1} - LL^{0}_{R_2r_0-1,2c_0} + LL^{0}_{R_2r_0,2c_0-1} + LL^{0}_{R_2r_0,2c_0}}{4} \text{ (vertical detail subband)} \quad (210)$$

$$HH^{1}_{R_r_1,c_1} = \frac{LL^{0}_{R_2r_0-1,2c_0-1} - LL^{0}_{R_2r_0-1,2c_0} - LL^{0}_{R_2r_0,2c_0-1} + LL^{0}_{R_2r_0,2c_0}}{4} \text{ (diagonal detail subband)} \quad (211)$$

$$\forall r_1 \in [1, ROW\ /\ 2], \wedge c_1 \in [1, COL\ /\ 2]$$

$LL^1$, $LH^1$, $HL^1$, and $HH^1$ are the first level subbands of Haar basis wavelet for two dimensions. $LL^1$ is the approximation subband (low frequency), and $LH^1$, $HL^1$, and $HH^1$ are the detail subbands (high frequency). $LL^0$ is the basal level or original image. Then QSA will be,

$$\Delta\varpi^{1}_{R_H} = i \ \ LH^{1}_{R} ./ LL^{1}_{R}, \quad (212)$$

$$\Delta\varpi^{1}_{R_V} = i \ \ HL^{1}_{R} ./ LL^{1}_{R}, \quad (213)$$

$$\Delta\varpi^{1}_{R_D} = i \ \ HH^{1}_{R} ./ LL^{1}_{R}, \quad (214)$$

while, FIT is,

$$\begin{aligned} \Delta f^{1}_{R_H} &= \frac{1}{2\pi}\sqrt{\Delta\varpi^{1}_{R_H} \ .\times conj\left(\Delta\varpi^{1}_{R_H}\right)} \\ &= \frac{1}{2\pi}\sqrt{\left(i\ LH^{1}_{R} ./ LL^{1}_{R}\right) .\times conj\left(i\ LH^{1}_{R} ./ LL^{1}_{R}\right)} \\ &= \frac{1}{2\pi}\left|LH^{1}_{R} ./ LL^{1}_{R}\right| = \frac{1}{2\pi}\left|LH^{1}_{R}\right| ./ LL^{1}_{R} \end{aligned} \quad (215)$$

$$\begin{aligned} \Delta f^{1}_{R_V} &= \frac{1}{2\pi}\sqrt{\Delta\varpi^{1}_{R_V} \ .\times conj\left(\Delta\varpi^{1}_{R_V}\right)} \\ &= \frac{1}{2\pi}\sqrt{\left(i\ HL^{1}_{R} ./ LL^{1}_{R}\right) .\times conj\left(i\ HL^{1}_{R} ./ LL^{1}_{R}\right)} \\ &= \frac{1}{2\pi}\left|HL^{1}_{R} ./ LL^{1}_{R}\right| = \frac{1}{2\pi}\left|HL^{1}_{R}\right| ./ LL^{1}_{R} \end{aligned} \quad (216)$$

$$\begin{aligned} \Delta f^{1}_{R_D} &= \frac{1}{2\pi}\sqrt{\Delta\varpi^{1}_{R_D} \ .\times conj\left(\Delta\varpi^{1}_{R_D}\right)} \\ &= \frac{1}{2\pi}\sqrt{\left(i\ HH^{1}_{R} ./ LL^{1}_{R}\right) .\times conj\left(i\ HH^{1}_{R} ./ LL^{1}_{R}\right)} \\ &= \frac{1}{2\pi}\left|HH^{1}_{R} ./ LL^{1}_{R}\right| = \frac{1}{2\pi}\left|HH^{1}_{R}\right| ./ LL^{1}_{R} \end{aligned} \quad (217)$$

As we shall see in the next section, it is very useful also another version of FIT for the non-overlapping case, with a subsampling to the half resolution, which is very useful for applications such as compression and superresolution,

$$\begin{aligned} \Delta f^{1}_{R_H} &= \frac{1}{2\pi} abs\left(\Delta\varpi^{1}_{R_H}\right) \\ &= \frac{1}{2\pi}\left|LH^{1}_{R}\right| ./ LL^{1}_{R} \end{aligned} \quad (218)$$

$$\begin{aligned}\Delta f^1_{R_V} &= \frac{1}{2\pi} abs\left(\Delta\varpi^1_{R_V}\right) \\ &= \frac{1}{2\pi}\left|HL^1_R\right| ./\ LL^1_R\end{aligned} \tag{219}$$

$$\begin{aligned}\Delta f^1_{R_D} &= \frac{1}{2\pi} abs\left(\Delta\varpi^1_{R_D}\right) \\ &= \frac{1}{2\pi}\left|HH^1_R\right| ./\ LL^1_R\end{aligned} \tag{220}$$

On the other hand, all of them, have the same problem, i.e., they depends on a possible denominator equal to zero, therefore, we will use the next versions:

*Equalization:* we are going to equalize $I_R$ from $[0,2^8\text{-}1]$ to $[1,256]$ ($I_{R_eq}$), like Eq.(141), and then, we will built $\left[LL^1_{R_eq}, LH^1_{R_eq}, HL^1_{R_eq}, HH^1_{R_eq}\right]$ with it, i.e., $\left[\Delta\varpi^1_{R_H_eq}, \Delta\varpi^1_{R_V_eq}, \Delta\varpi^1_{R_D_eq}\right]$, and hence,

$$\begin{aligned}\Delta f^1_{R_H_eq} &= \frac{1}{2\pi}\sqrt{\Delta\varpi^1_{R_H_eq} .\times conj\left(\Delta\varpi^1_{R_H_eq}\right)} \\ &= \frac{1}{2\pi}\sqrt{\left(i\ \ LH^1_{R_eq} ./\ LL^1_{R_eq}\right) .\times conj\left(i\ \ LH^1_{R_eq} ./\ LL^1_{R_eq}\right)} \\ &= \frac{1}{2\pi}\left|LH^1_{R_eq} ./\ LL^1_{R_eq}\right| = \frac{1}{2\pi}\left|LH^1_{R_eq}\right| ./\ LL^1_{R_eq}\end{aligned} \tag{221}$$

$$\begin{aligned}\Delta f^1_{R_V_eq} &= \frac{1}{2\pi}\sqrt{\Delta\varpi^1_{R_V_eq} .\times conj\left(\Delta\varpi^1_{R_V_eq}\right)} \\ &= \frac{1}{2\pi}\sqrt{\left(i\ \ HL^1_{R_eq} ./\ LL^1_{R_eq}\right) .\times conj\left(i\ \ HL^1_{R_eq} ./\ LL^1_{R_eq}\right)} \\ &= \frac{1}{2\pi}\left|HL^1_{R_eq} ./\ LL^1_{R_eq}\right| = \frac{1}{2\pi}\left|HL^1_{R_eq}\right| ./\ LL^1_{R_eq}\end{aligned} \tag{222}$$

$$\begin{aligned}\Delta f^1_{R_D_eq} &= \frac{1}{2\pi}\sqrt{\Delta\varpi^1_{R_D_eq} .\times conj\left(\Delta\varpi^1_{R_D_eq}\right)} \\ &= \frac{1}{2\pi}\sqrt{\left(i\ \ HH^1_{R_eq} ./\ LL^1_{R_eq}\right) .\times conj\left(i\ \ HH^1_{R_eq} ./\ LL^1_{R_eq}\right)} \\ &= \frac{1}{2\pi}\left|HH^1_{R_eq} ./\ LL^1_{R_eq}\right| = \frac{1}{2\pi}\left|HH^1_{R_eq}\right| ./\ LL^1_{R_eq}\end{aligned} \tag{223}$$

and

$$\begin{aligned}\Delta f^1_{R_H_eq} &= \frac{1}{2\pi} abs\left(\Delta\varpi^1_{R_H_eq}\right) \\ &= \frac{1}{2\pi}\left|LH^1_{R_eq}\right| ./\ LL^1_{R_eq}\end{aligned} \tag{224}$$

$$\begin{aligned}\Delta f^1_{R_V_eq} &= \frac{1}{2\pi} abs\left(\Delta\varpi^1_{R_V_eq}\right) \\ &= \frac{1}{2\pi}\left|HL^1_{R_eq}\right| ./\ LL^1_{R_eq}\end{aligned} \tag{225}$$

$$\begin{aligned}\Delta f^1_{R_D_eq} &= \frac{1}{2\pi} abs\left(\Delta\varpi^1_{R_D_eq}\right) \\ &= \frac{1}{2\pi}\left|HH^1_{R_eq}\right| ./\ LL^1_{R_eq}\end{aligned} \tag{226}$$

*Averaging:* we will proceed to show this version directly starting with QSA,

$$\Delta\varpi^{1}_{R_H_av} = i \;\; LH^{1}_{R} \,/\, avg\left(LL^{1}_{R}\right), \tag{227}$$

$$\Delta\varpi^{1}_{R_V_av} = i \;\; HL^{1}_{R} \,/\, avg\left(LL^{1}_{R}\right), \tag{228}$$

$$\Delta\varpi^{1}_{R_D_av} = i \;\; HH^{1}_{R} \,/\, avg\left(LL^{1}_{R}\right), \tag{229}$$

Then, FIT will be,

$$\begin{aligned} \Delta f^{1}_{R_H_av} &= \frac{1}{2\pi}\sqrt{\Delta\varpi^{1}_{R_H_av} \;.\!\times conj\left(\Delta\varpi^{1}_{R_H_av}\right)} \\ &= \frac{1}{2\pi}\frac{\sqrt{\left(i \;\; LH^{1}_{R}\right).\!\times conj\left(i \;\; LH^{1}_{R}\right)}}{avg\left(LL^{1}_{R}\right)} \\ &= \frac{1}{2\pi}\left|LH^{1}_{R}\right| / \, avg\left(LL^{1}_{R}\right) \end{aligned} \tag{230}$$

$$\begin{aligned} \Delta f^{1}_{R_V_av} &= \frac{1}{2\pi}\sqrt{\Delta\varpi^{1}_{R_V_av} \;.\!\times conj\left(\Delta\varpi^{1}_{R_V_av}\right)} \\ &= \frac{1}{2\pi}\frac{\sqrt{\left(i \;\; HL^{1}_{R}\right).\!\times conj\left(i \;\; HL^{1}_{R}\right)}}{avg\left(LL^{1}_{R}\right)} \\ &= \frac{1}{2\pi}\left|HL^{1}_{R}\right| / \, avg\left(LL^{1}_{R}\right) \end{aligned} \tag{231}$$

$$\begin{aligned} \Delta f^{1}_{R_D_av} &= \frac{1}{2\pi}\sqrt{\Delta\varpi^{1}_{R_D_av} \;.\!\times conj\left(\Delta\varpi^{1}_{R_D_av}\right)} \\ &= \frac{1}{2\pi}\frac{\sqrt{\left(i \;\; HH^{1}_{R}\right).\!\times conj\left(i \;\; HH^{1}_{R}\right)}}{avg\left(LL^{1}_{R}\right)} \\ &= \frac{1}{2\pi}\left|HH^{1}_{R}\right| / \, avg\left(LL^{1}_{R}\right) \end{aligned} \tag{232}$$

or

$$\begin{aligned} \Delta f^{1}_{R_H_av} &= \frac{1}{2\pi} abs\left(\Delta\varpi^{1}_{R_H_av}\right) \\ &= \frac{1}{2\pi}\left|LH^{1}_{R}\right| / \, avg\left(LL^{1}_{R}\right) \end{aligned} \tag{233}$$

$$\begin{aligned} \Delta f^{1}_{R_V_av} &= \frac{1}{2\pi} abs\left(\Delta\varpi^{1}_{R_V_av}\right) \\ &= \frac{1}{2\pi}\left|HL^{1}_{R}\right| / \, avg\left(LL^{1}_{R}\right) \end{aligned} \tag{234}$$

$$\begin{aligned} \Delta f^{1}_{R_D_av} &= \frac{1}{2\pi} abs\left(\Delta\varpi^{1}_{R_D_av}\right) \\ &= \frac{1}{2\pi}\left|HH^{1}_{R}\right| / \, avg\left(LL^{1}_{R}\right) \end{aligned} \tag{235}$$

Finally, none of non-overlapping version needs a padding over the original image, with the consequent simplification in coding this implies.

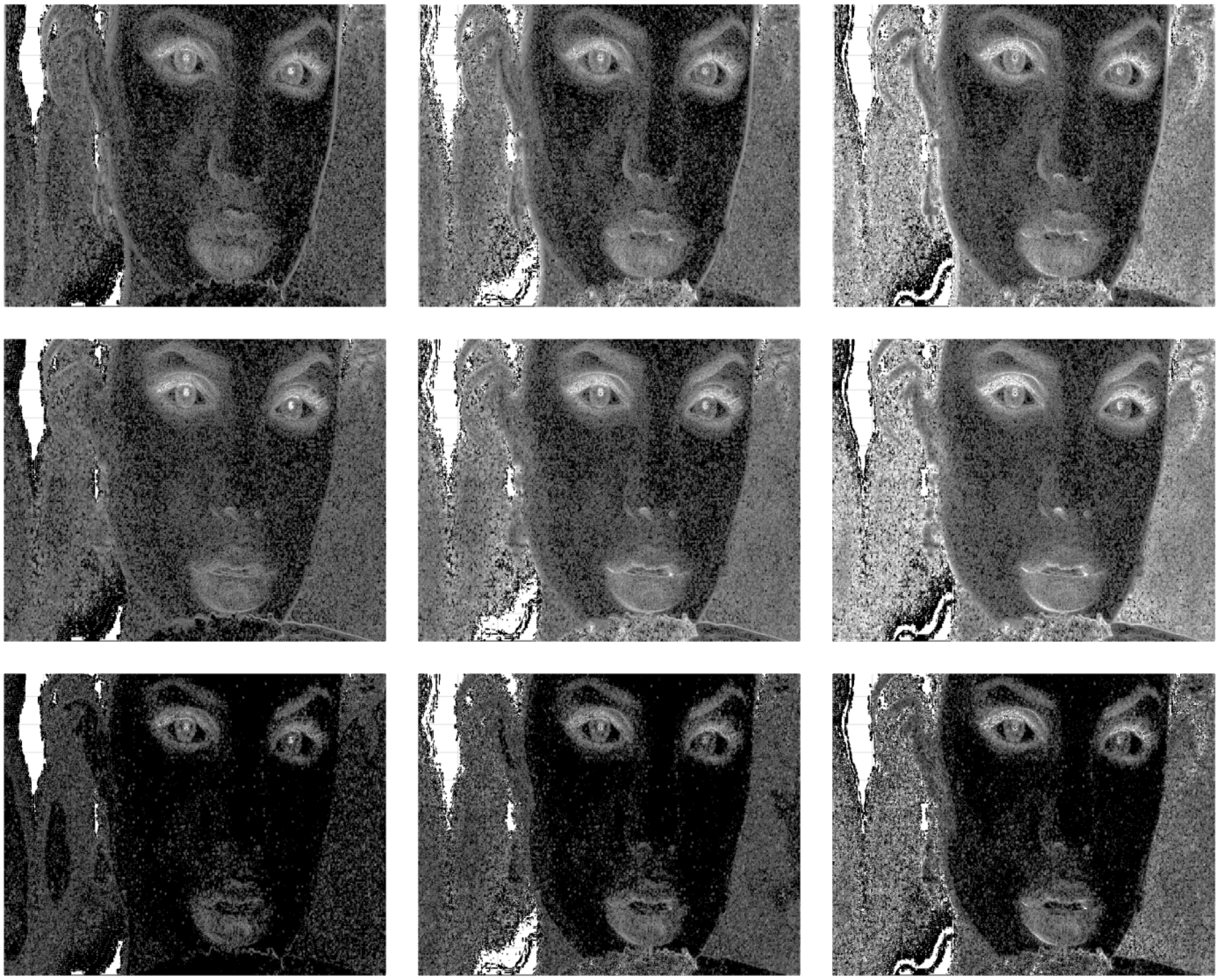

**Fig. 53 FIT over Angelina, raw case without overlap:** first column for red channel, second column for green channel, and third column for blue channel. Besides, first row for horizontal case, second row for vertical case, and third row for diagonal case.

Figure 53 shows us the raw case without overlap, where the first column is for red channel, second column for green channel, and third column for blue channel. Besides, first row represents the horizontal case, second row the vertical case, and third row the diagonal case.

The nine images inside Fig.53 have a resolution equal to the half resolution of original image, that is to say, if the resolution of Angelina is ROW-by-COL pixels with 24 bpp, the images of Fig.53 have a resolution of ROW/2-by-COL/2 pixels with 24 bpp.

Figure 54 shows us the equalized case without overlap, where the first column is for red channel, second column for green channel, and third column for blue channel. Here too, first row represents the horizontal case, second row the vertical case, and third row the diagonal case.

Finally, Fig.55 shows us the averaged case without overlap, where the first column is for red channel, second column for green channel, and third column for blue channel. Newly, first row represents the horizontal case, second row the vertical case, and third row the diagonal case.

Moreover, the next two figures have remarkable similarities to the previous figure. This fact demonstrates the consistency and coherence of the method. The three figures show clearly the spectral particularities of the image regarding to color and direction (i.e., horizontal, vertical, and diagonal).

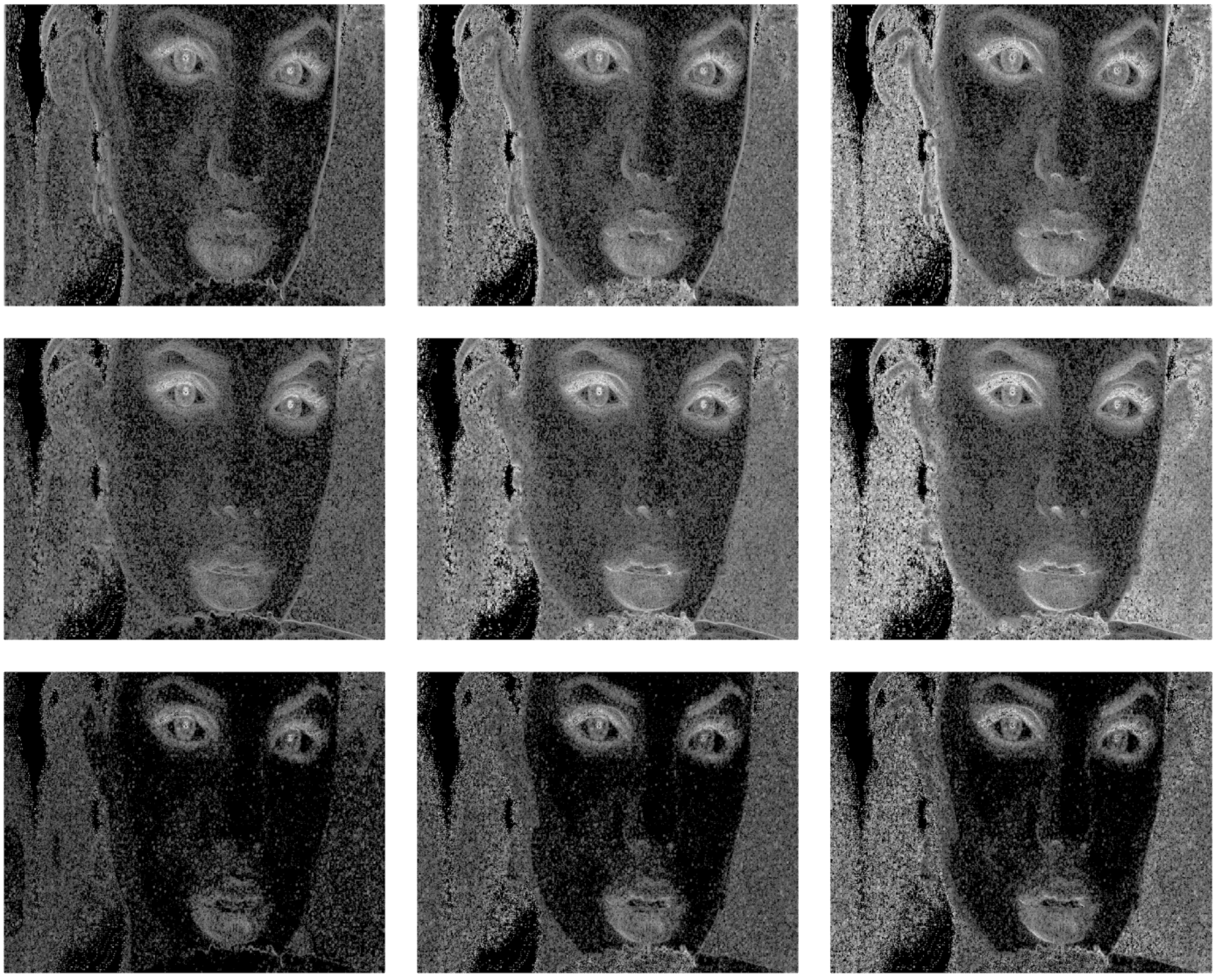

**Fig. 54 FIT over Angelina, equalized case without overlap:** first column for red channel, second column for green channel, and third column for blue channel. Besides, first row for horizontal case, second row for vertical case, and third row for diagonal case.

As we can see from the above examples for signals and images, with and without overlap, the effect of indeterminate FIT when the sample (signal case) or pixel (image case) is a value equal to 0 has solution, through the averaged or equalized versions. Instead, the effect of indeterminate angle (phase) when magnitude = 0 in FFT has no solution [207-210]. Besides, while FFT has no compact support, FIT has it. The latter brings about a lousy treatment of energy by FFT, and an excellent treatment of it by the FIT, to the output of both procedures. Another important comparative aspect between FFT and FIT is the the poor performance of the FFT at the edges (both signals and images), whereby the FFT is replaced by the FCT in applications of compression and filtering. This problem does not exist in FIT. On the other hand, witness bar show that by simple equalization all versions give identical results. Besides, the FIT acts as a detector, which indicates that encode for the case of compression by the witness bar, similar to PPM or nonlinear sampling [206]. In this sense, it is very convenient to use the bars witness both rows and columns on pictures as a new type of profilometry instead of histograms, or complementing these. Moreover, the advantages of nonlinear sampling are obvious in communications and signal compression.

Other relevant advantages of FIT regarding to FFT are:

- FIT give us an instant notion of the spectral components of the signal or image. In other words, FIT demonstrates directly responsibility of flanks on the characteristics and values of such spectral components.
- FIT is responsive to ergodicity, the *regions of interest* (ROIs), textures, noises, flanks or edges tilt and their relationship with Shannon and Nyquist for nonlinear sampling for Communications.

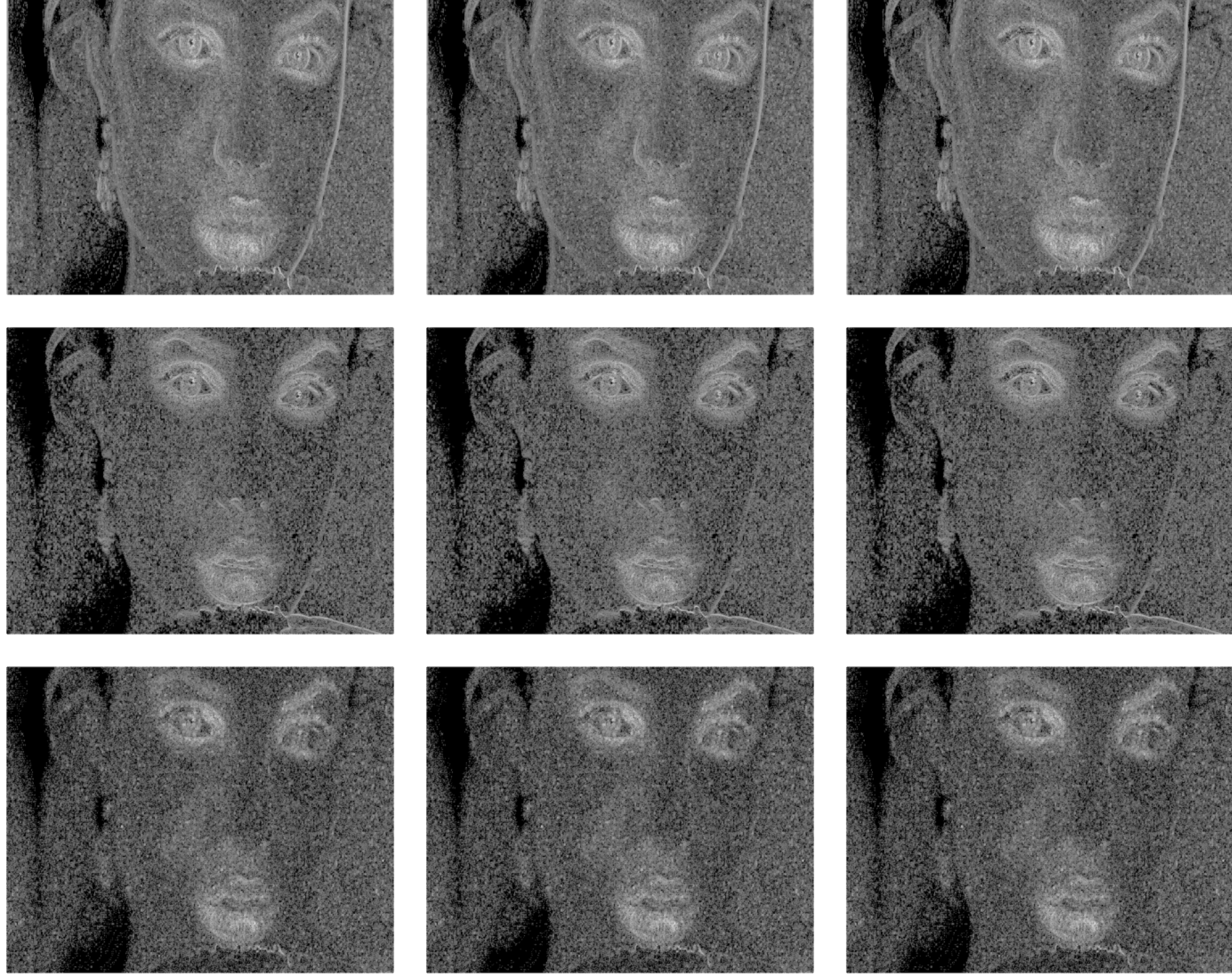

**Fig. 55 FIT over Angelina, averaged case without overlap:** first column for red channel, second column for green channel, and third column for blue channel. Besides, first row for horizontal case, second row for vertical case, and third row for diagonal case.

- FFT loses the link with time.
- FIT can be calibrated and related with FFT, easily.
- FIT gives frequency in terms of time, directly, i.e., $\Delta f\left(t\right)=\Delta\omega\left(t\right)/2\pi$.
- Two-dimensional QSA/FIT is directional, and via Pythagoras it is consistent with the idea of directional QSA for images and N-dimensional arrays.
- In the case of FIT, both the square mask as the segmental mask are (in themselves) direct filtering processes (denoising).
- In FIT, everything is parallelizable: in that case the use of General-purpose graphics processing units (GPGPUs) is recommended [211], and, in fact, FIT is faster than FFT on them.
- In FIT, the Hamiltonian's basal tone [1] is associated with both the spectral bands as the spectral subbands of approximation (in the case of wavelets). This fact makes calibration be considerably easier, as simple as tuning an instrument.
- Flank detection is equivalent to edge detection in visual intelligence. Besides, FIT detects the sign change and texture and thus assess how compress. Otherwise, FIT permits a nonlinear sampling more efficient than the traditional linear sampling regularly employed, all this from the point of view of the Information Theory [1]. In fact. QSA/FIT can perform edge detection equal or better than methods Prewitt, Roberts, Sobel and Canny [45-48]. Although you can easily prove that all of them derive from QSA/FIT.
- Figure 56 shows in symbolic way both complementarity as the perfect linkage between the two theories, i.e., FFT and QSA/FIT, instead, Fig.57 shows us such complementarity and linkage in a rigorous form.

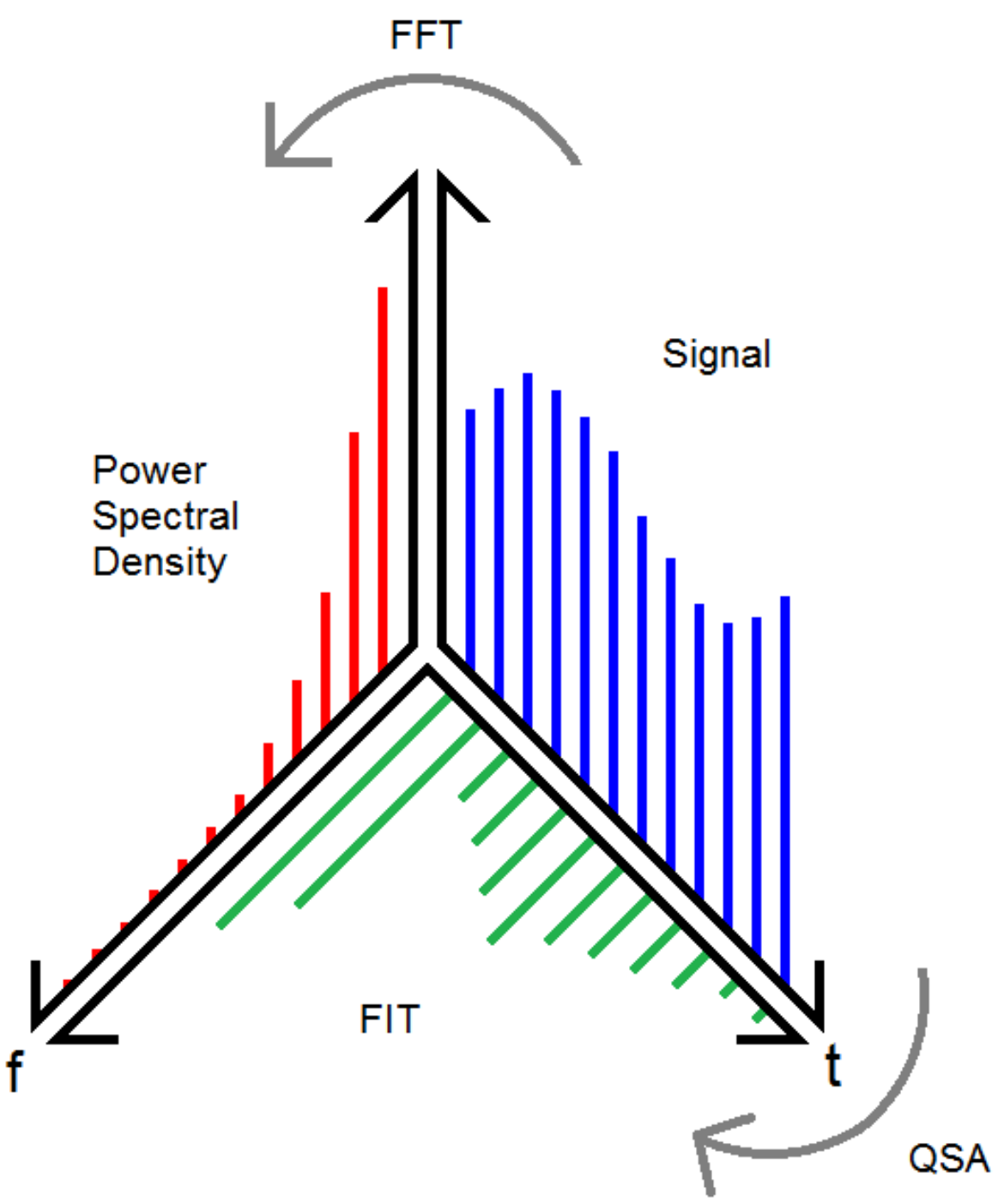

**Fig.56** Symbolic relationship between FIT and FFT (PSD).

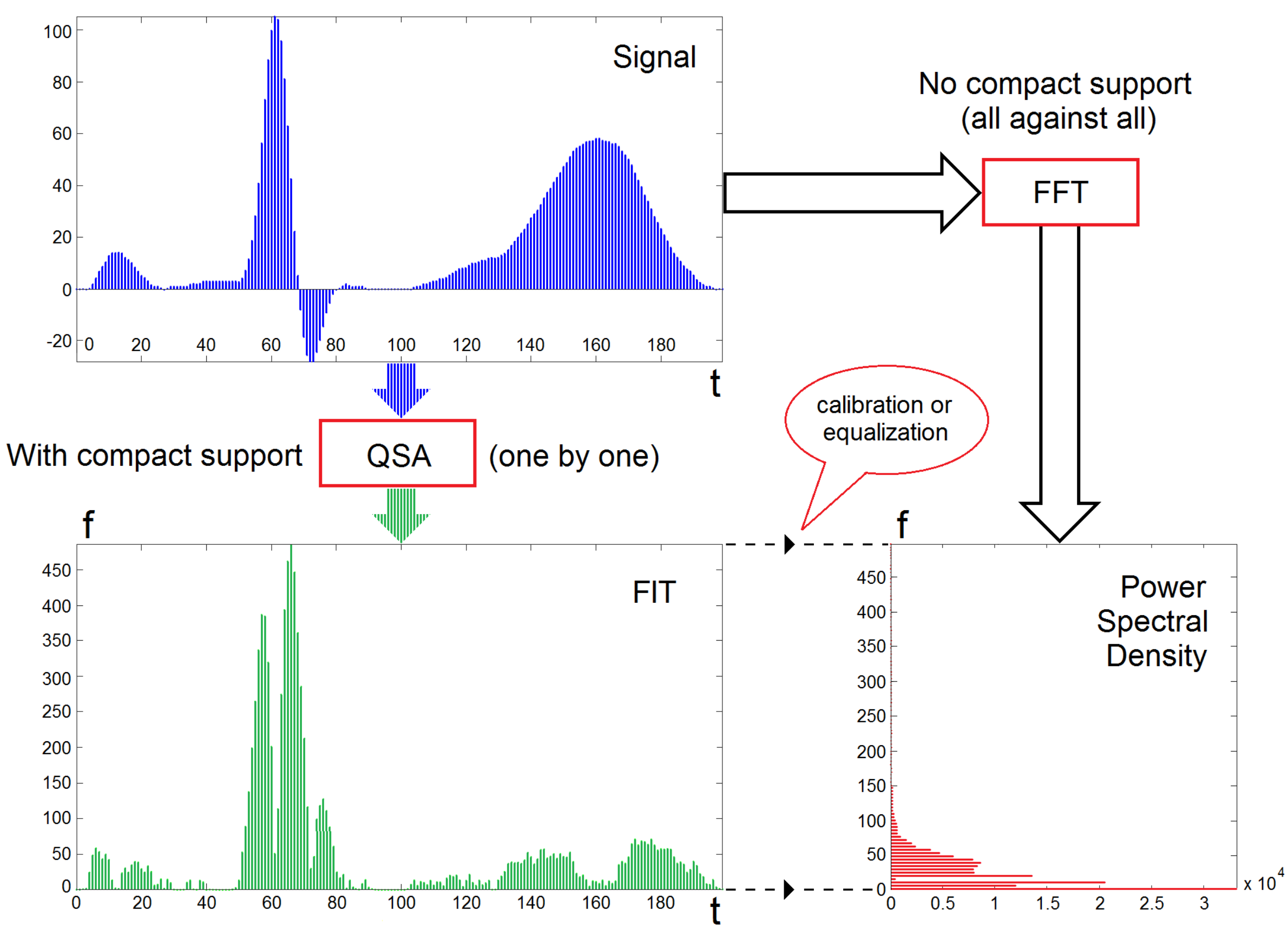

**Fig.57** Rigorous relationship between FIT and FFT (PSD).

Both graphs clearly show a quadrature between FFT and FIT via equalization.

FFT and FIT give information about the same physical element, i.e., the frequency, but in a very different way, in fact, FIT is far superior and accurate (in its ambit) regarding FFT. Besides, unlike FFT, FIT has compact support. However, both are complementary.

Thanks to these two tools (FFT and FIT) we can get the whole universe linked to spectral and temporal analysis (simultaneously) of a signal, image or video. Therefore, we can locate (indirectly) to the FFT at the exact time of the signal by its components. This fact implies a significant advance in the Fourier's theory after almost two and a half centuries.

On the other hand, the distribution of the witness bars is consistent with the possibility of locating a particle by the wave function, or rather, the probability distribution that arises from this function.

Given the signal y = f (t), the witness bars arise as follows:

1. N equidistant lines are distributed along the ordinate axis
2. In those settings where these lines intercept the signal, we identify the projections on the axis of abscissae. At these points we place the witness bars, which (if the signal is nonlinear) shall be separated in a not equidistant way depending on the flanks of the signal at each point. This is a nonlinear sampling itself.

Equations (166-168) for one dimensional case (signals) and (179, 195, 212-214) for two dimensional case (images) represent a type of wavelets known as spectral wavelets, or simply, spectralets. In fact, it is the only case inside wavelets, where high frequency sub-bands (or detail) represent spectral components (unlike traditional wavelet). Such is the case that FIT is measured in Hertz. As we see in the next section, the spectralets are also useful in the case of multislicing (image sequence processing [212], and medical images [213], such as: Magnetic Resonance and Computer Tomography), multi [214-217] and hyperspectral imagery [218-220], and video [201, 202, 221-226]. Besides, they serve to filtering, compression, edge detection and superresolution. For example, for denoising and compression we can use the traditional method of Dohono (thresholding/shrinkage) [125-127], and the smoothing of coefficients via median filter [45-48], directional smoothing [162, 172], or enhanced directional smoothing [171] on detail subbands.

Additionally, spectralets are consistent with the procedure known as splitting of spectral subbands used in the traditional wavelet (see Section 2.2), which continues recursively on approximation and detail subbands ($L^i$ and $H^i$ for signals; $LL^i$ and $LH^i$, $HL^i$, $HH^i$ for images; and $LLL^i$ and $LLH^i$, $LHL^i$, $LHH^i$, $HLL^i$, $HLH^i$, $HHL^i$, $HHH^i$ for image sequence processing, being *i* the level of splitting). In fact, Haar is a particular case of spectralets, see Eq(173) for one-dimensional case, and Eqs(227-229, 233-235) for two-dimensional case. Complementary to this, in the next section, we will see a new concept inside wavelet theory called *cosine wavelets*, or simply, *coslets*, which will be a key in all those applications like superresolution of biomedical signals, still images for mobile, and video for video too.

Some final considerations:

- The transition from QSA to FIT represents the collapse of the wave function, i.e., from vector to scalar at each moment.
- Hamiltonian is real, i.e., it isn't hermitic for a confined single particle
- QSA/FIT is a time filter
- The frequency of a tone (sine) is proportional to its higher slope derivative. If it's a gate, will be infinite. In the latter example, the density of the witness bars is infinite in the flank of the gate. This is very useful for a better understanding of Sampling and Nyquist theorems.
- Like the FFT, the FIT will help in the development of new algorithms for signal, image and video compression, replacing or complementing to FFT or DCT in new versions of, MP3 (audio [228]), JPEG (images [227]) and, H.264 and VP9 (video [221-226]).
- Unlike FFT, FIT does not require decimation in time or frequency.
- For one-dimension FFT has a computational cost of $O(N*log_2(N))$, and FIT of $O(N)$.
- For two-dimensions FFT has a computational cost of $O(N^2*log_2(N)^2)$, and FIT of $O(N^2)$.

# 4 Simulations

## 4.1 Prolegomenous to simulations

In this section we will see only some possible applications of QSA/FIT, given that several of them we have seen them in Section 3, e.g., spectral analysis. Besides, in addition, we also see some additional tools, which will be very useful in this field.

### 4.1.1 Important tools

***Coslets:*** without loss of generality, we present this new tool for two-dimensional case, however, it is possible to make a generalization to one and three dimensions, in an easy and automatic way.

In here, we present a new multirresolution procedure based on DCT using this like a wavelet basis, See Fig.58. For example, for an image I of size ROW-by-COL we have,

1) Choose image $I(r,c) / r \in [1, ROW] \wedge c \in [1, COL]$
2) J = DCT-2D(I) (236)
3) We obtain four subbands from $J = \begin{bmatrix} ll_1 & lh_1 \\ hl_1 & hh_1 \end{bmatrix}$, with (237)

$$ll_1(r,c) / r \in \left[1, \frac{ROW}{2}\right] \wedge c \in \left[1, \frac{COL}{2}\right],$$

$$lh_1(r,c) / r \in \left[1, \frac{ROW}{2}\right] \wedge c \in \left[\frac{COL}{2}+1, COL\right],$$

$$hl_1(r,c) / r \in \left[\frac{ROW}{2}+1, ROW\right] \wedge c \in \left[1, \frac{COL}{2}\right], \text{ and}$$

$$hh_1(r,c) / r \in \left[\frac{ROW}{2}+1, ROW\right] \wedge c \in \left[\frac{COL}{2}+1, COL\right].$$

4) We apply iDCT-2D to each subband of J. Then, we obtain,

$LL_1$ = iDCT-2D($ll_1$), with $LL_1(r,c) / r \in \left[1, \frac{ROW}{2}\right] \wedge c \in \left[1, \frac{COL}{2}\right]$, (238a)

$LH_1$ = iDCT-2D($lh_1$), with $LH_1(r,c) / r \in \left[1, \frac{ROW}{2}\right] \wedge c \in \left[\frac{COL}{2}+1, COL\right]$, (238b)

$HL_1$ = iDCT-2D($hl_1$), with $HL_1(r,c) / r \in \left[\frac{ROW}{2}+1, ROW\right] \wedge c \in \left[1, \frac{COL}{2}\right]$, and (238c)

$HH_1$ = iDCT-2D($hh_1$), with $HH_1(r,c) / r \in \left[\frac{ROW}{2}+1, ROW\right] \wedge c \in \left[\frac{COL}{2}+1, COL\right]$. (238d)

From now on, the treatment is similar to the two-dimensional discrete wavelet transform, e.g., Haar [162-170], see Section 2.2.2.

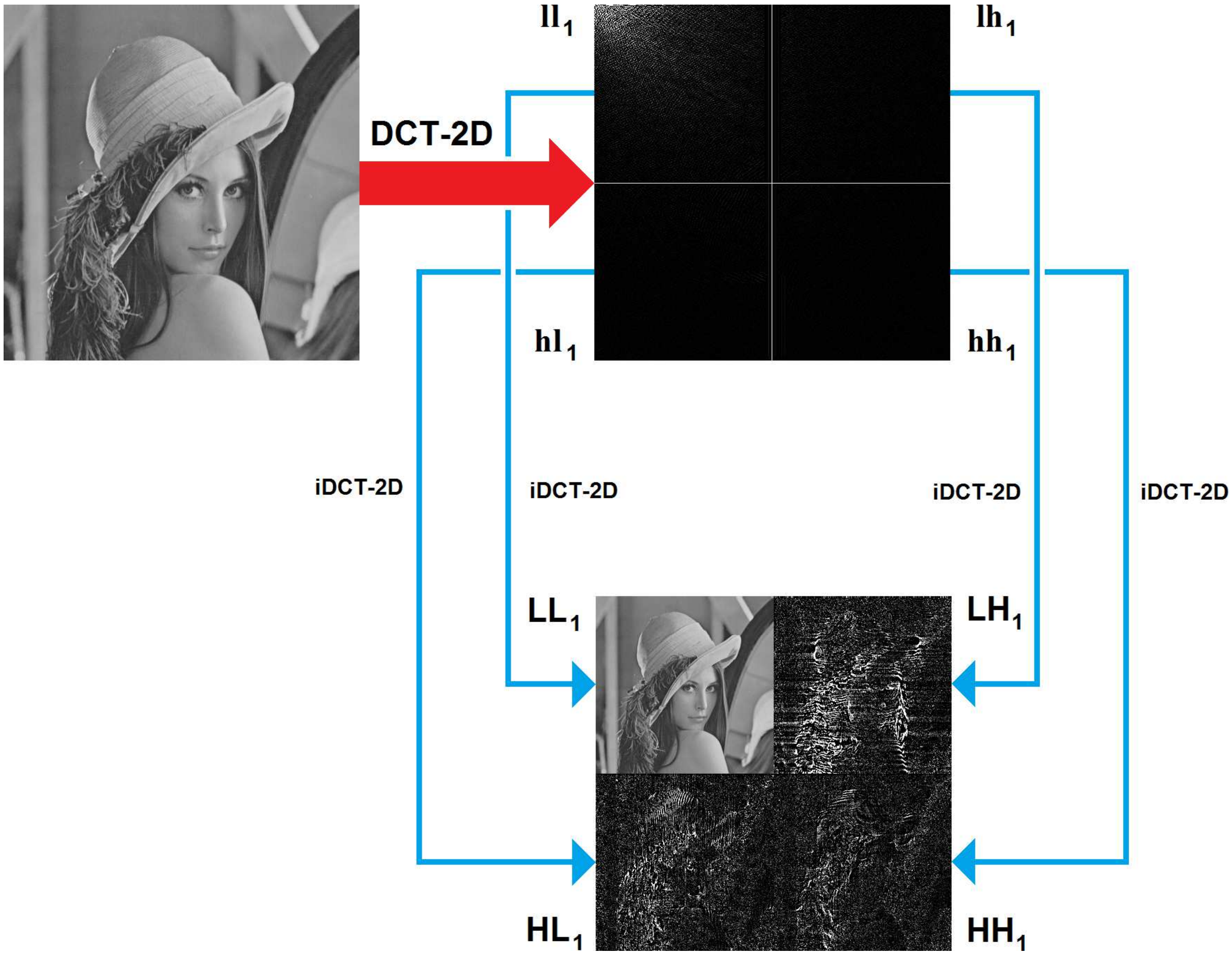

**Fig.58** Coslets-2D on Lena.

*Subband Coefficients and Multirresolution*
The coslets corresponds to multiresolution approximation expressions. In practice, mutiresolution analysis is carried out using 4 channel filter banks (for each level of decomposition) composed of a low-pass and a high-pass filter and each filter bank is then sampled at a half rate (1/2 down sampling) of the previous frequency. By repeating this procedure, it is possible to obtain coslets of any order. The down sampling procedure keeps the scaling parameter constant (equal to ½) throughout successive coslets so that is benefits for simple computer implementation. In the case of an image, the filtering is implemented in a separable way be filtering the lines and columns.

Note that coslets of an image consists of four frequency channels for each level of decomposition (identical to the case of wavelets). For example, for *i*-level of decomposition we have:

$LL_{n,i}$: Noisy Coefficients of Approximation.
$LH_{n,i}$: Noisy Coefficients of Vertical Detail,
$HL_{n,i}$: Noisy Coefficients of Horizontal Detail, and
$HH_{n,i}$: Noisy Coefficients of Diagonal Detail.

The LL part at each scale is decomposed recursively, as illustrated in Fig.12.

To achieve space-scale adaptive noise reduction, we need to prepare the 1-D coefficient data stream which contains the space-scale information of 2-D images. This is somewhat similar to the “zigzag” arrangement of the DCT coefficients in image coding applications [45-48, 227, 229-249]. In this data preparation step, the coslets coefficients are rearranged as a 1-D coefficient series in spatial order so that the adjacent samples represent the same local areas in the original image [248, 249].

Figure 59 shows the interior of the coslets-2D with the four subbands of the transformed image, which will be used in Fig.16. Each output of Fig.59 represents a subband of splitting process of the 2-D coefficient matrix corresponding to Fig.12.

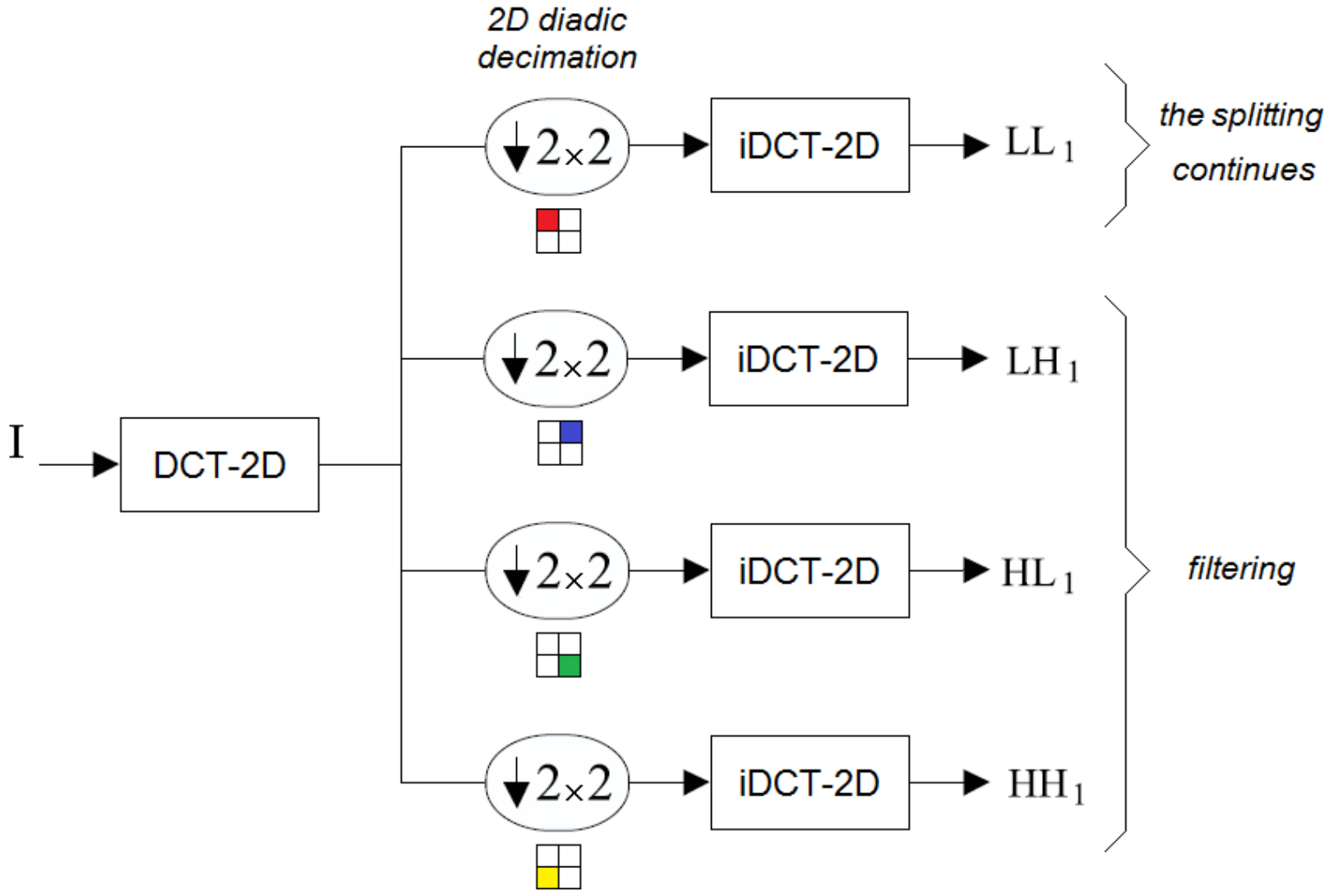

**Figure 59** Two dimensional coslets. A decomposition step. Usual splitting of the subbands.

*Coslet Noise Thresholding*
It is exactly the same of Section 2.2.2, this similarity includes thresholding method of D. L. Donoho and I. M. Johnstone [125-127], and smoothing of coefficients in wavelet domain via directional smoothing [166, 171] and mean filtering [45-48].

***Deblurring:*** these techniques are very useful for those cases in which a downsampling process is followed by an upsampling process. This creates an undesirable effect on the image called blur, in which the edges of said image are degraded. Deblurring consists in the restoration of the edges (in this case) via convolution mask. This technique is known as unsharp masking [45-48].

In this paper, the downsampling and upsampling is done with the techniques of Section 4.1.3, while the deblurring is done by a two-dimensional convolution mask of NxN pixels, which makes a rafter over the upsampled (blurred) image. The parameters of this squared mask (where N is odd) are criticals, therefore, such parameters must be calculated and adjusted with total accuracy.

Coming up next, we will proceed to deduct the mask and set the optimal relationship between its parameters. Later we will proceed to adjust them via a Genetic Algorithm [250].

*Deduction of the mask*
Based on the last section, the single frame is recovered after suffering a pair of processes: downsampling and upsampling, see left side of Fig.60. In this figure:

**X** means original single frame.

**Y** means recovered (blurred) single frame.
$\mathbf{M_b}$ means square mask of NxN pixels (where N is odd).
$\downarrow$ means downsampling.
$\uparrow$ means upsampling.

This mask is known as a blurred mask, smoothing operator or Point Spread Function (PSF) [251].

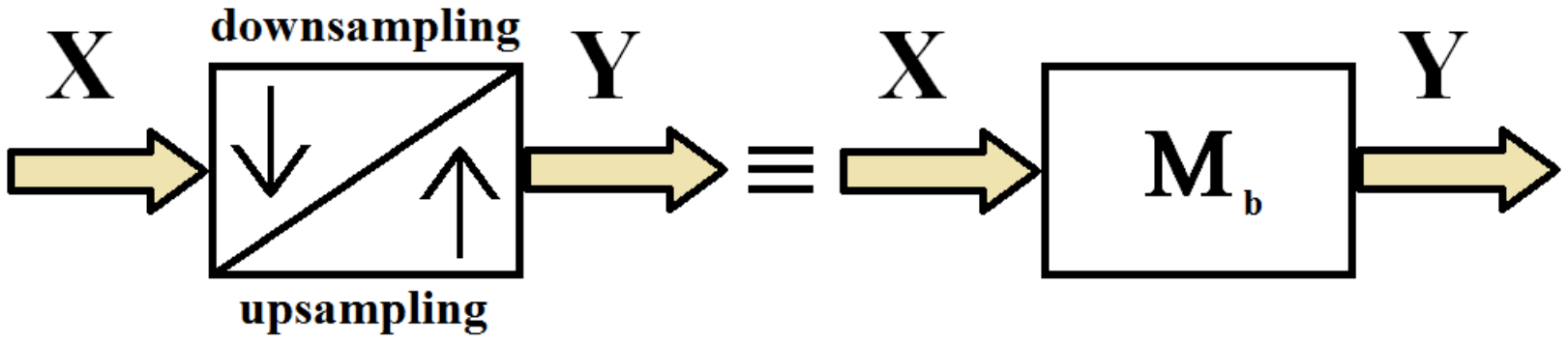

**Figure 60** Downsampling/upsampling as a blurred mask.

In these processes ($\downarrow$ and $\uparrow$), the single frame is affected by a space/time invariant blur, we which interpret as the result of the action of a two-dimensional convolution between the original single frame and a mask known in Digital Image Processing as a mask of mean filtering. The idea of mean filtering is simply to replace each pixel value in an image with the mean (`average') value of its neighbors, including itself. This has the effect of eliminating pixel values which are unrepresentative of their surroundings. Mean filtering is usually thought of as a convolution filter. Like other convolutions it is based around a kernel, which represents the shape and size of the neighborhood to be sampled when calculating the mean. Often a 3×3 square kernel is used, although larger kernels (e.g. 5×5 squares) can be used for more severe smoothing. (Note that a small kernel can be applied more than once in order to produce a similar but not identical effect as a single pass with a large kernel). In Fig.61, we consider the most general case, for N×N kernel, always with odd N, where:

$$\varphi = \frac{1}{N \times N} \tag{239}$$

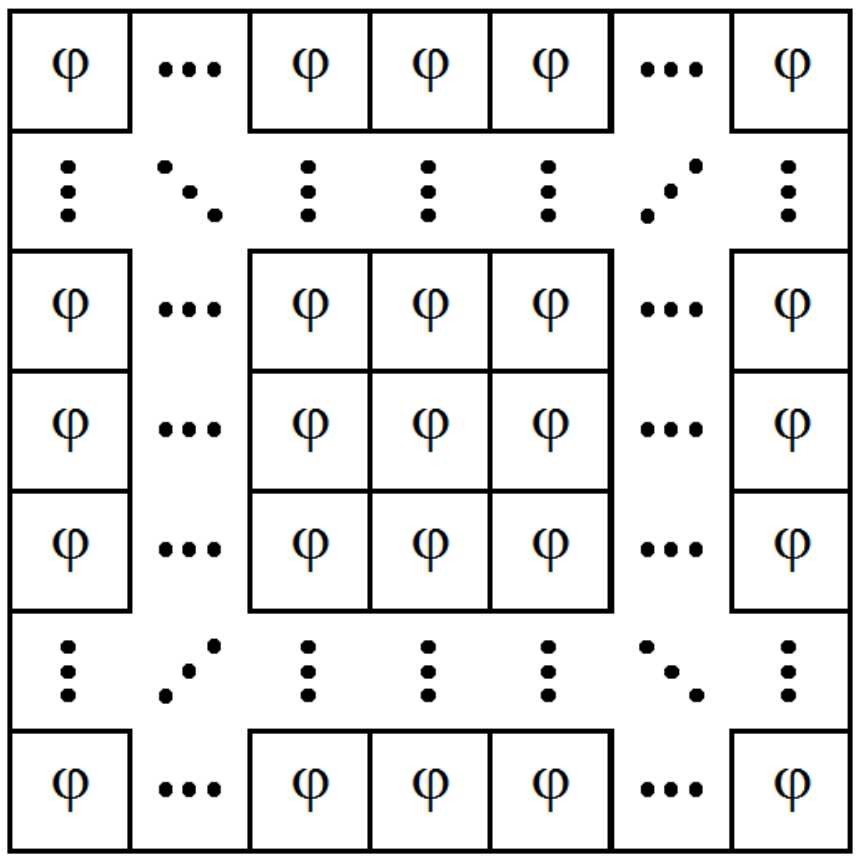

**Figure 61** N×N averaging kernel often used in mean filtering.

Computing the straightforward convolution of an image with this kernel carries out the mean filtering process.

$$Y = M_b \otimes X \tag{240}$$

Where $\otimes$ means two-dimensional convolution.

At this point, we propose a model of deblurring based on low noise, linear space and time invariant blur via convolution, thus,

$$\hat{X} = M_d \otimes Y \tag{241}$$

Where $\boldsymbol{M_d}$ is a mask as shown in Fig.62, and the following relationships to consider are very important,

$$(N^2 - 1) \times \alpha + \beta = 1, \text{ (for deblurring)} \tag{242a}$$

$$(N^2 - 1) \times \alpha + \beta = 0, \text{ (for edge detection)} \tag{242b}$$

Thus, a new and simplified model of deblurring appears on the scene, see Fig.63, where $\alpha < 0$ and $\beta > 1$. We need to establish precisely both parameters, then, there are two possible ways forward:

1. Choose N (integer, positive, odd and small), and $\beta > 1$ (and arbitrarily less than 2), then $\alpha$ is derived from Eq.(242a).
2. Start with arbitrary values of $\alpha$ and $\beta$ (about certain recommendations, e.g., $-0.1 < \alpha < 0$ and $1 < \beta \leq 2$) and generating a random population of the pair [$\alpha$, $\beta$], and deducting $\boldsymbol{N}$ from Eq.(242a). The mentioned pair serves of initial population for the Genetic Algorithm [250] of Fig.64, where the pair is called chromosome, and $\alpha$ and $\beta$ are called genes. The metric for the adjustment is the Mean Squared Error (MSE), which is defined in the next section [45, 46].

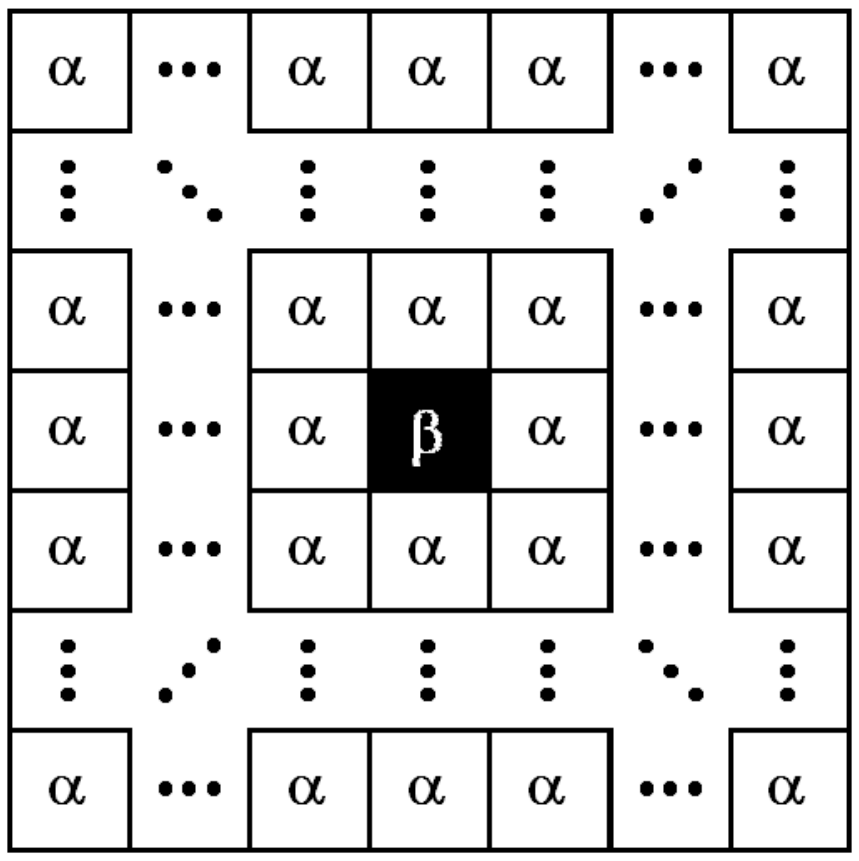

**Figure 62** Deblurring mask $\mathbf{M_d}$ .

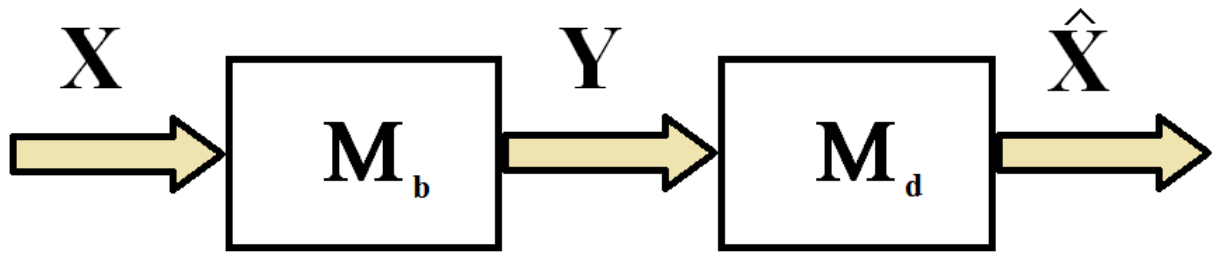

**Figure 63** New and simplified model of deblurring.

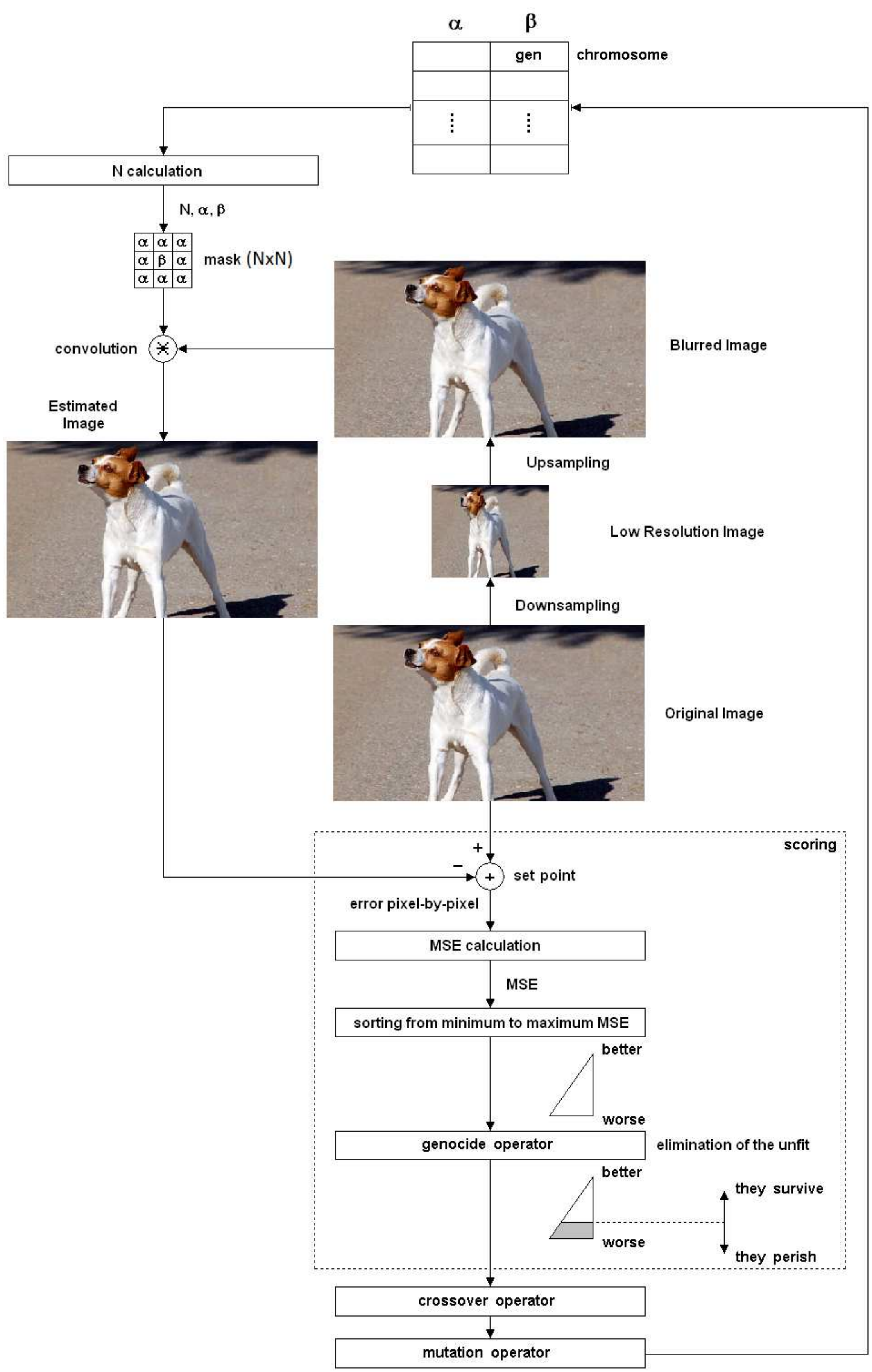

**Figure 64** Genetic Algorithm for calculating the parameters of the mask.

On the other hand, the employed Genetic Algorithm is com-posed of three big modules:
a) Scoring,
b) Crossover Operator, and
c) Mutation Operator

The first consists of the following submodules:
a.1) Set-point where the error pixel-by-pixel arises
a.2) The MSE calculation with the error pixel-by-pixel
a.3) Sorting from minimum to maximum MSE
a.4) Genocide Operator eliminates the chromosomes with biggest MSE, i.e., there are a fixed number of chromosomes that survive per cycle, the fittest. Such fixed number is a design parameter of the Genetic Algorithm.

The Crossover Operator (or Mating Operator) crosses the parent chromosomes (selected randomly) generating new son chromosomes, which will be better and/or worse than their parents [250].

The Mutation Operator must have a low frequency of action for the purpose of not disturbing the nature of the species, i.e., skip to solve another problem [250].

Finally, parameters were found to be high $\alpha$= -0.0129 and $\beta$= 1.63.

With a similar reasoning, we can apply a deblurring procedure to a signal via convolution masks seen in Section 2.2. In fact, the procedure is very simply, given a signal:

1) obtain (N-1)/2 up-signals and (N-1)/2 down-signals by any of the methods of Figures 28 or 29, such that N is an odd number,
2) build a N×N mask whose values are $\alpha$ = -0.0129 except the central element that must be $\beta$= 1.63, and
3) perform a convolution between the mask and the package of N signals.

The result of this convolution is the deblurring on the original signal.

#### 4.1.2 Complementary tools

***Phase plane and stability:*** *Phase plane analysis* is one of the most important techniques for studying the behavior of dynamic systems, especially in the nonlinear case, where *general* methods for computing analytical solution do not exist [252].

*Some comments:*
1) The response characteristics (relative speed of response) for *unforced systems* depend on the initial conditions
2) Eigenvalue/eigenvector analysis allows us to predict the response characteristics (fast and slow, or stable and unstable) depending on initial conditions
3) Another way of obtaining a feel for the effect of initial conditions (and then of the characteristics of the response) is to use a *phase-plane plot*
4) A phase-plane plot for a two-state variable system consists of curves of one state variable versus the other one ($x_1(t)$ vs. $x_2(t)$), where each curve is based on a *different initial condition*
5) Consider a systems of linear differential equations $\mathbf{x}' = \mathbf{A}\mathbf{x}$. Its *phase portrait* is a representative set of its solutions, plotted as parametric curves (with t as the parameter) on the Cartesian plane tracing the path of each particular solution $(x, y) = (x_1(t), x_2(t))$, $-\infty < t < \infty$. *Similar to a direction field*, a phase portrait is a graphical tool to visualize how the solutions of a given system of differential equations would behave in the long run.

6) In this context, the Cartesian plane where the phase portrait resides is called the *phase plane*. The parametric curves traced by the solutions are sometimes also called their trajectories.
7) *Remark:* It is QITe labor-intensive, but it is possible to sketch the phase portrait by hand without first having to solve the system of equations that it represents. Just like a direction field, a phase portrait can be a tool to predict the behaviors of a system's solutions. To do so, we draw a grid on the phase plane. Then, at each grid point $\mathbf{x} = (\alpha, \beta)$, we can calculate the solution trajectory's instantaneous direction of motion at that point by using the given system of equations to compute the tangent / velocity vector, $\mathbf{x}'$. Namely plug in $\mathbf{x} = (\alpha, \beta)$ to compute $\mathbf{x}' = \boldsymbol{A}\mathbf{x}$.
8) An equilibrium solution of the system $\mathbf{x}' = \boldsymbol{A}\mathbf{x}$ is a point $(x_1, x_2)$ where $\mathbf{x}' = \mathbf{0}$, that is, where $x_1' = x_2' = 0$. An equilibrium solution is a constant solution of the system, and is usually called a *critical point*.
9) For a linear system $\mathbf{x}' = \boldsymbol{A}\mathbf{x}$, an equilibrium solution occurs at each solution of the system (of homogeneous algebraic equations) $\boldsymbol{A}\mathbf{x} = \mathbf{0}$. As we have seen, such a system has exactly one solution, located at the origin, if $det(\boldsymbol{A}) \neq 0$. If $det(\boldsymbol{A}) = 0$, then there are infinitely many solutions.
10) For our purpose, and unless otherwise noted, we will only consider systems of linear differential equations whose coefficient matrix $\boldsymbol{A}$ has nonzero determinant. That is, we will only consider systems where the origin is the only critical point.
11) The critical points of various systems of first order linear differential equations are classified by using their *stability*. In addition, due to the truly two dimensional nature of the parametric curves, we will also classify the *type* of those critical points by the shapes formed by the trajectories about each critical point.
12) *Comment:* The accurate tracing of the parametric curves of the solutions is not an easy task without computers. However, we can obtain very reasonable approximation of a trajectory by using the very same idea behind the direction field, namely *the tangent line* approximation. At each point $\boldsymbol{x} = (x_1, x_2)$ on the plane, the direction of motion of the solution curve that passes through the point is determined by the direction vector (i.e. the tangent vector) $\mathbf{x}'$, the derivative of the solution vector $\boldsymbol{x}$, evaluated at the given point. The tangent vector at each given point can be calculated directly from the given matrix vector equation $\mathbf{x}' = \boldsymbol{A}\mathbf{x}$, using the position vector $\boldsymbol{x} = (x_1, x_2)$. Like working with a direction field, there is no need to find the solution before performing this approximation.

Next, we can see a direct application of this tool to signal and images.

*For signals*
For this experiment, we use the electrocardiographic signal of Fig.43 (top). As we can see, such signal is cyclic, i.e., it is repeated periodically.

*a) Case with overlap*
From Eq.(140) we have,

$\dot{s}_n = \frac{s_{n+1} - s_{n-1}}{2}$, $\forall n \in [0, N-1]$, being $n$ the discrete time. The signal must be padded.

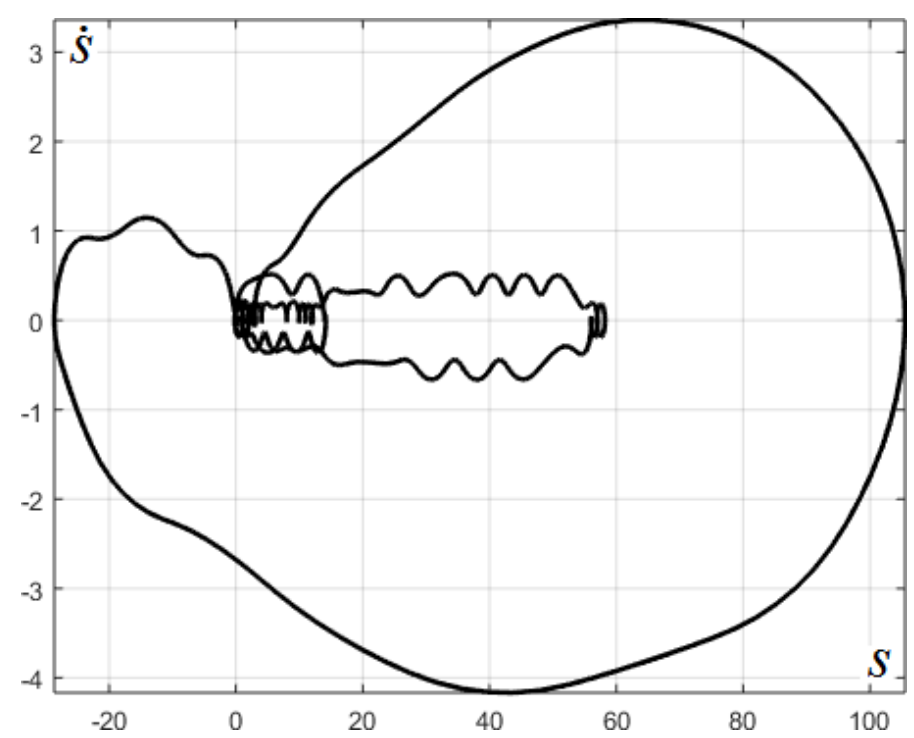

**Figure 65** Phase plane for case with overlap.

Figure 65 shows us phase plane for this case with the corresponding evolutionary trajectory.

*b) Case without overlap*
From Equations (163) and (164) we have

$l_k = \frac{s_n + s_{n+1}}{2}$ and $h_k = \frac{-s_n + s_{n+1}}{2}$, $\forall k \in \left[0, \frac{N}{2} - 1\right]$. The signal should not be padded.

$\dot{l}_k = h_k$

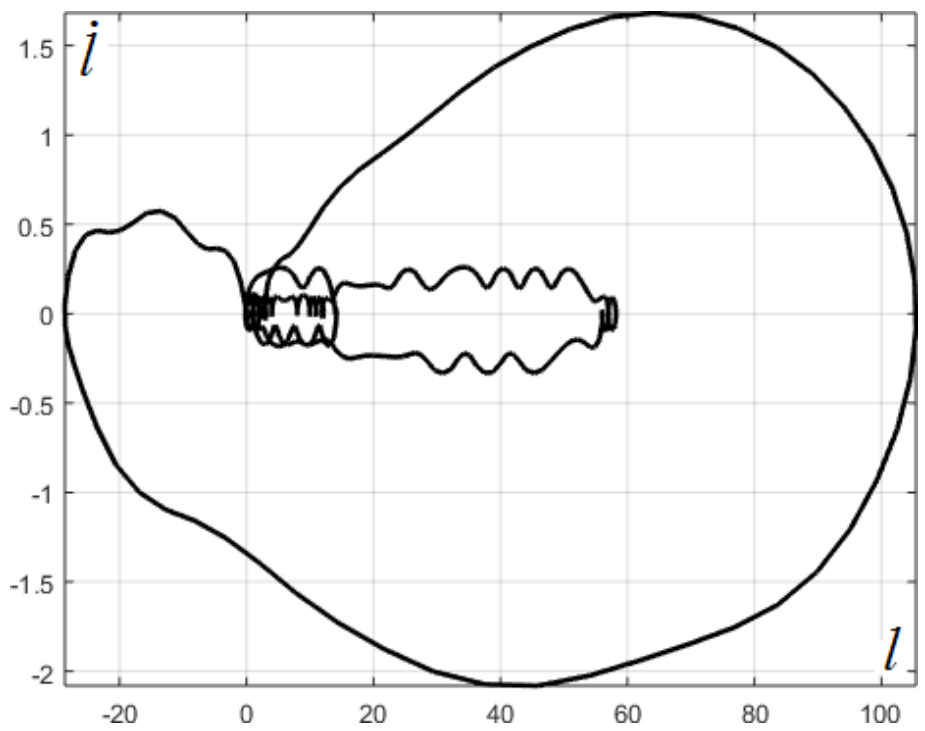

**Figure 66** Phase plane for case without overlap.

Figure 66 shows us phase plane for this case with the corresponding evolutionary trajectory. As we see, both figures are identical except for the scale of the vertical axis. This shows the consistency of QSA/FIT method, which gives similar phase planes for the same signal by different calculation.

*For images*
For this experiment, we use the image of *Angelina* of Fig.48.

*a) Case with overlap*
Given an image *I*, we obtain $\dot{l}_h$ from Eq.(176), $\dot{l}_v$ from Eq.(177), and $\dot{l}_d = \dot{l}$ from Eq.(178), where the subscripts {*h*, *v*, *d*} means {*horizontal*, *vertical*, *diagonal*} respectively. Besides, in this case, image must be padded. Figure 67 shows us the phase plane for case with overlap, by color channel and orientation.

*b) Case without overlap*
Given an image $I = LL^1_{R_r_1,c_1}$ from Eq.(208), we obtain $\dot{l}_h = LH^1_{R_r_1,c_1}$ from Eq.(209), $\dot{l}_v = HL^1_{R_r_1,c_1}$ from Eq.(210), and $\dot{l}_d = HH^1_{R_r_1,c_1}$ from Eq.(211), where here too, the subscripts {*h*, *v*, *d*} means {*horizontal*, *vertical*, *diagonal*} respectively. In all cases, $\forall r_1 \in [1, ROW/2], \wedge c_1 \in [1, COL/2]$, being *ROW-by-COL* the resolution of the original image named ***I***. Besides, in this case, the image should not be padded. Figure 68 shows us the phase plane for case without overlap, by color channel and orientation.

***Histogram:*** distribution of pixels graylevel values. A graph of number of pixels at each graylevel possible in an image. A histogram is a probability distribution of pixels values and may be processed using statistical techniques. These processes result in changes to the brightness and contrast in an image, but are independent of the spatial distribution of the pixels [253].

It plots the number of pixels for each tonal value. By looking at the histogram for a specific image a viewer will be able to judge the entire tonal distribution at a glance [254]. Image histograms are present on many modern digital cameras. Photographers can use them as an aid to show the distribution of tones captured, and whether image detail has been lost to blown-out highlights or blacked-out shadows. This is less useful when using a raw image format, as the dynamic range of the displayed image may only be an approximation to that in the raw file. The horizontal axis of the graph represents the tonal variations, while the vertical axis represents the number of pixels in that particular tone. The left side of the horizontal axis represents the black and dark areas, the middle represents medium grey and the right hand side represents light and pure white areas. The vertical axis represents the size of the area that is captured in each one of these zones. Thus, the histogram for a very dark image will have the majority of its data points on the left side and center of the graph. Conversely, the histogram for a very bright image with few dark areas and/or shadows will have most of its data points on the right side and center of the graph. Finally, Fig.69 shows an image of Angelina and its histogram.

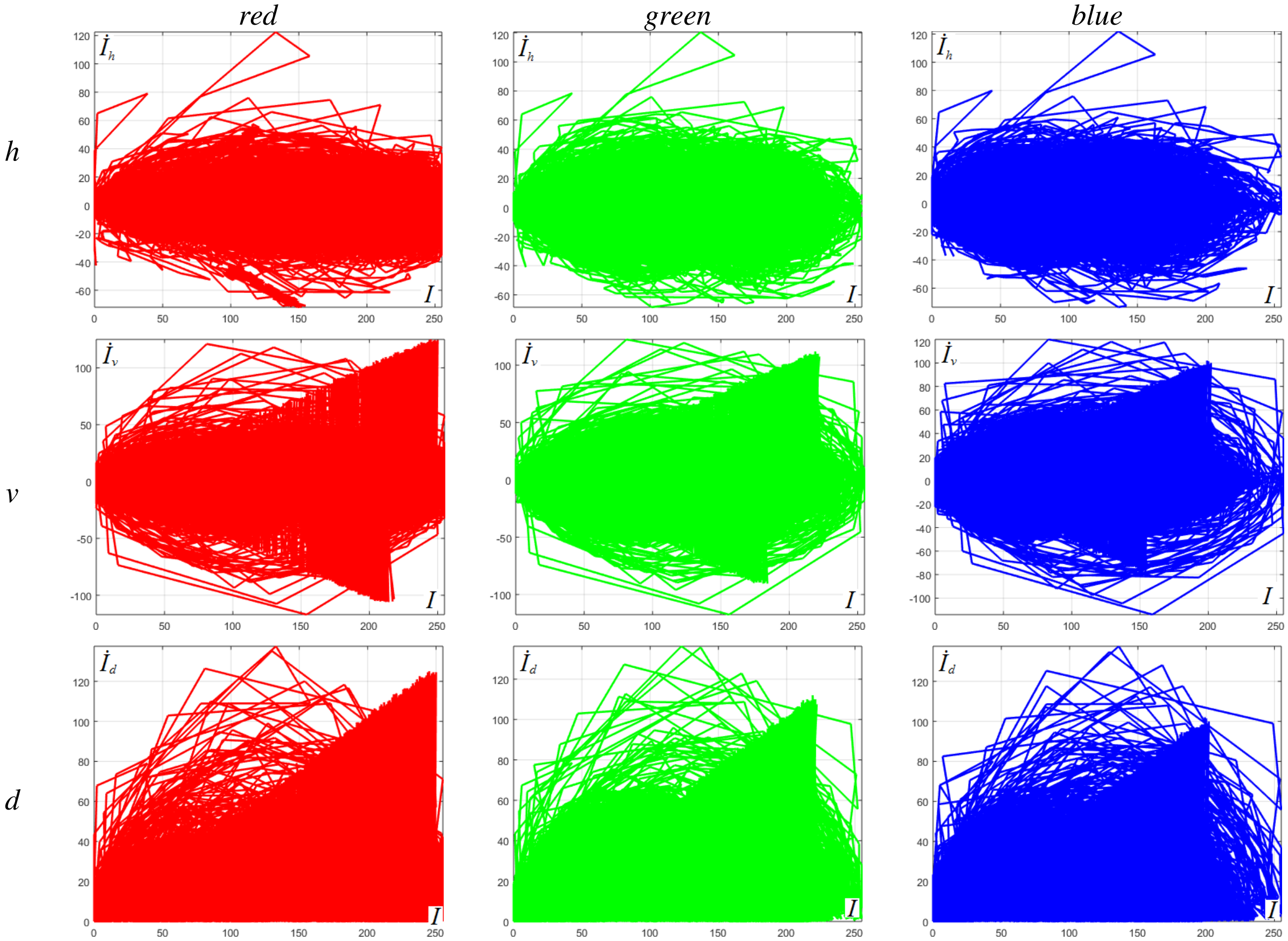

**Figure 67** Phase plane for case with overlap, by color channel and orientation.

As Prof. Emmanuel Agu says [255]: histograms is an important tool, because

- They help detect image acquisition issues
- Problems with image can be identified on histogram, such as:
  - Over and under exposure
  - Brightness
  - Contrast
  - Dynamic Range
- Point operations can be used to alter histogram, e.g.:
  - Addition
  - Multiplication
  - Exponential and Logarithmic
  - Intensity windowing (contrast modification)

With regard to the interpretation of the histograms, Logarithmic scale makes low values more visible. This allows a marked difference between darkest and lightest. See Fig.70.

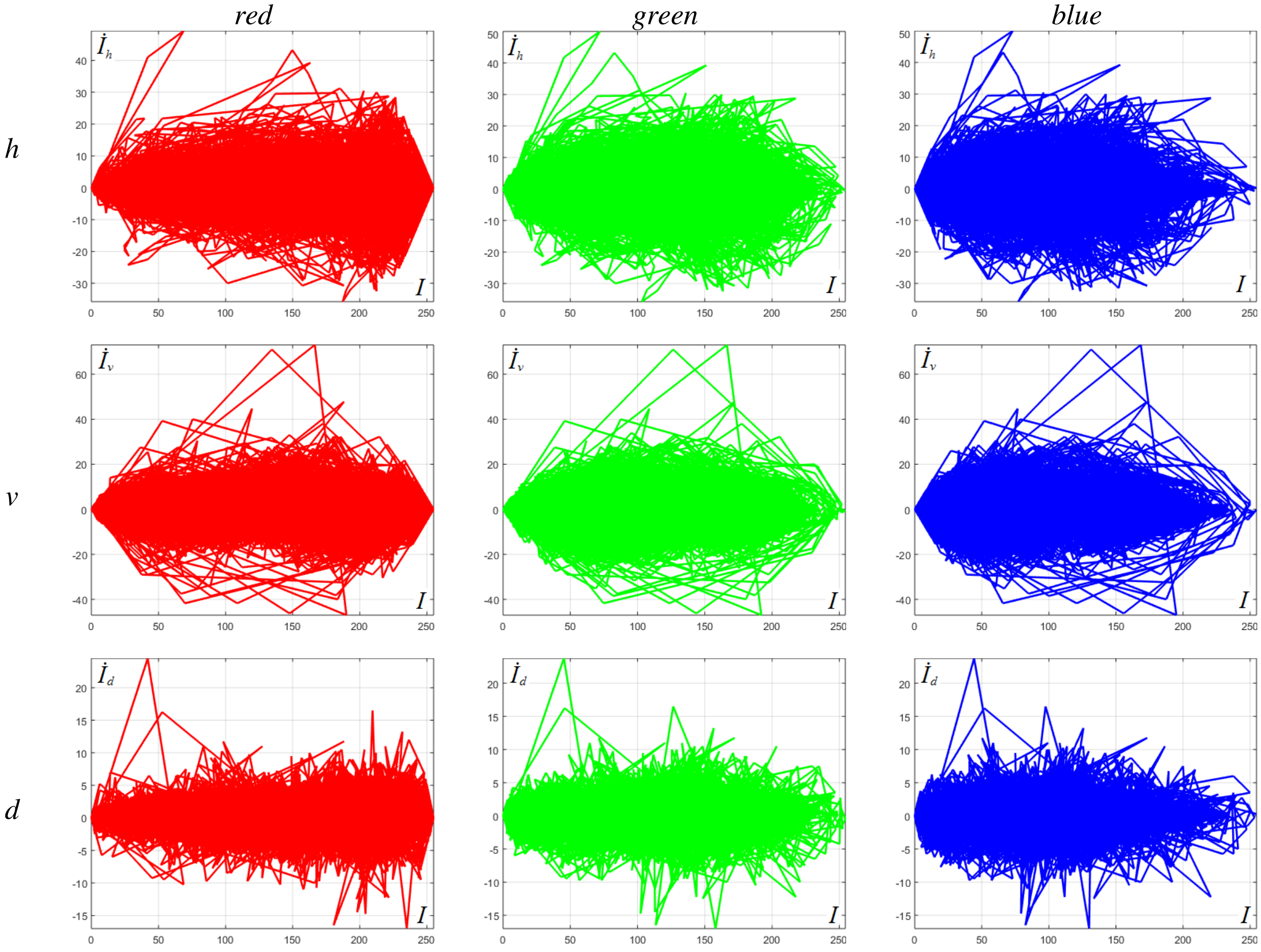

**Figure 68** Phase plane for case without overlap, by color channel and orientation.

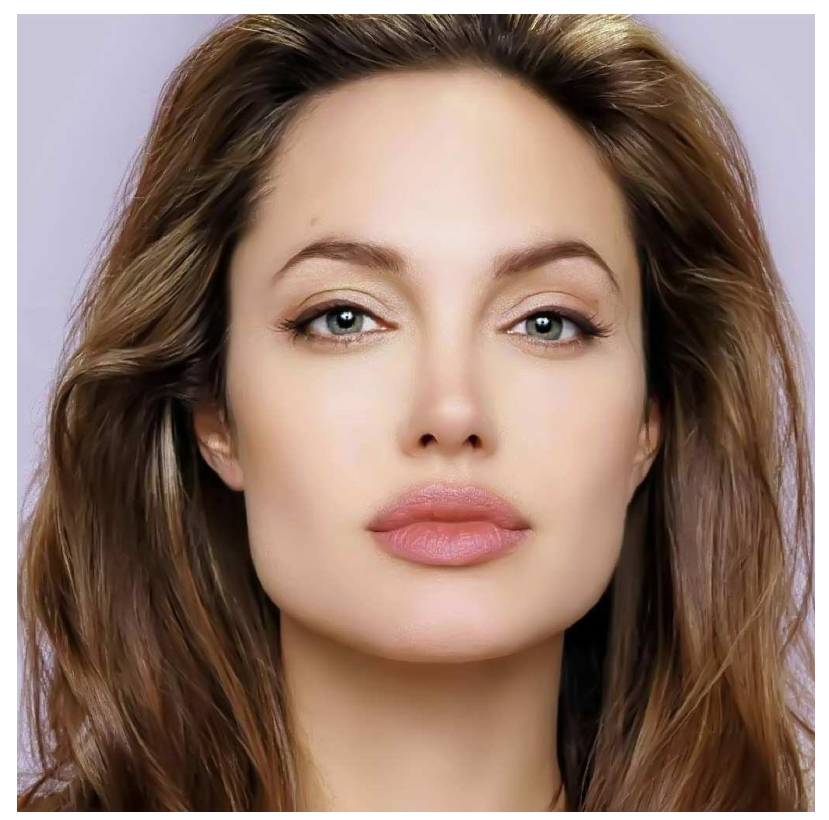

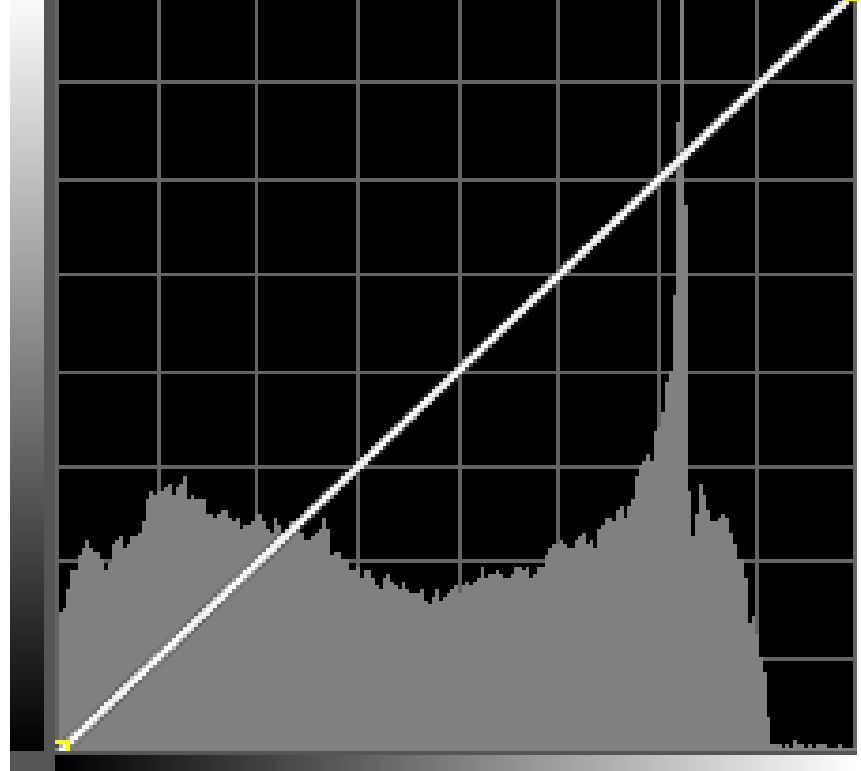

**Figure 69** Left: Angelina, right: its histogram.

On the other hand, histograms allows detect bad exposure, i.e., if the intensity values are spread (good) out or bunched up (bad). See Fig.71.

Besides, image contrast tells us:

- In a grayscale image it indicates how easily objects in the image can be distinguished
- High contrast image: many distinct intensity values
- Low contrast: image uses few intensity values

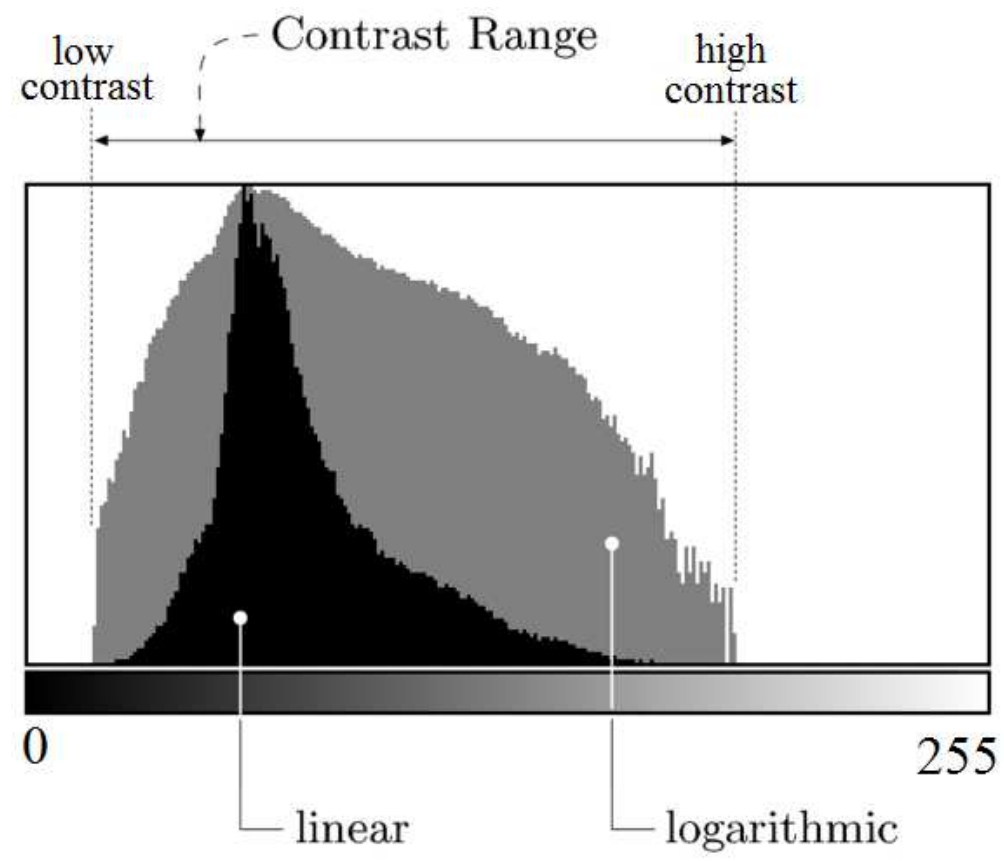

**Figure 70** Linear and logarithmic scale in histogram with explicit contrast range.

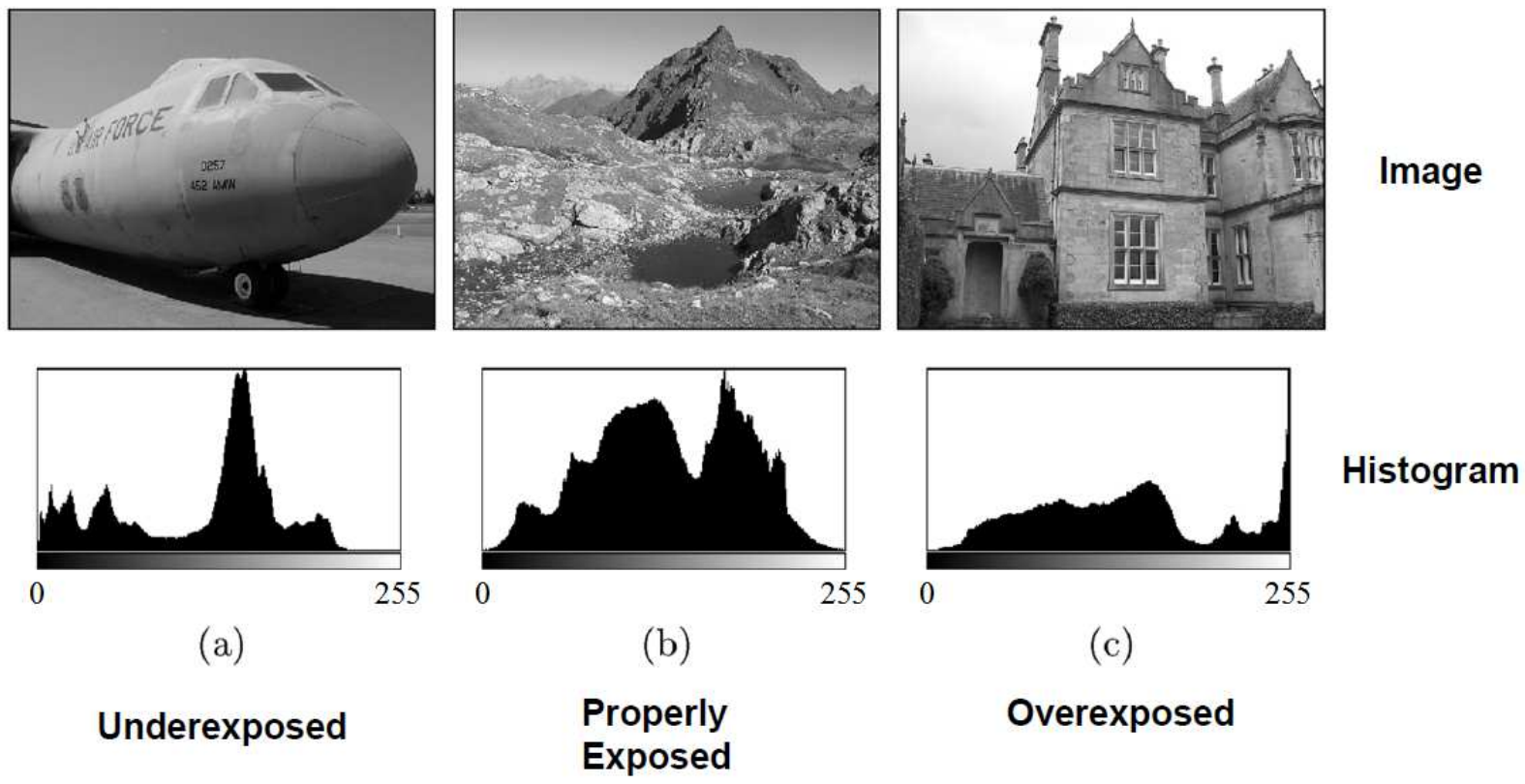

**Figure 71** Histograms in terms of exposure.

Moreover, regarding the relationship between contrast and histograms, we can say that a good contrast is a widely spread intensity values more a large difference between minima and maxima intensity values. See Fig.72.

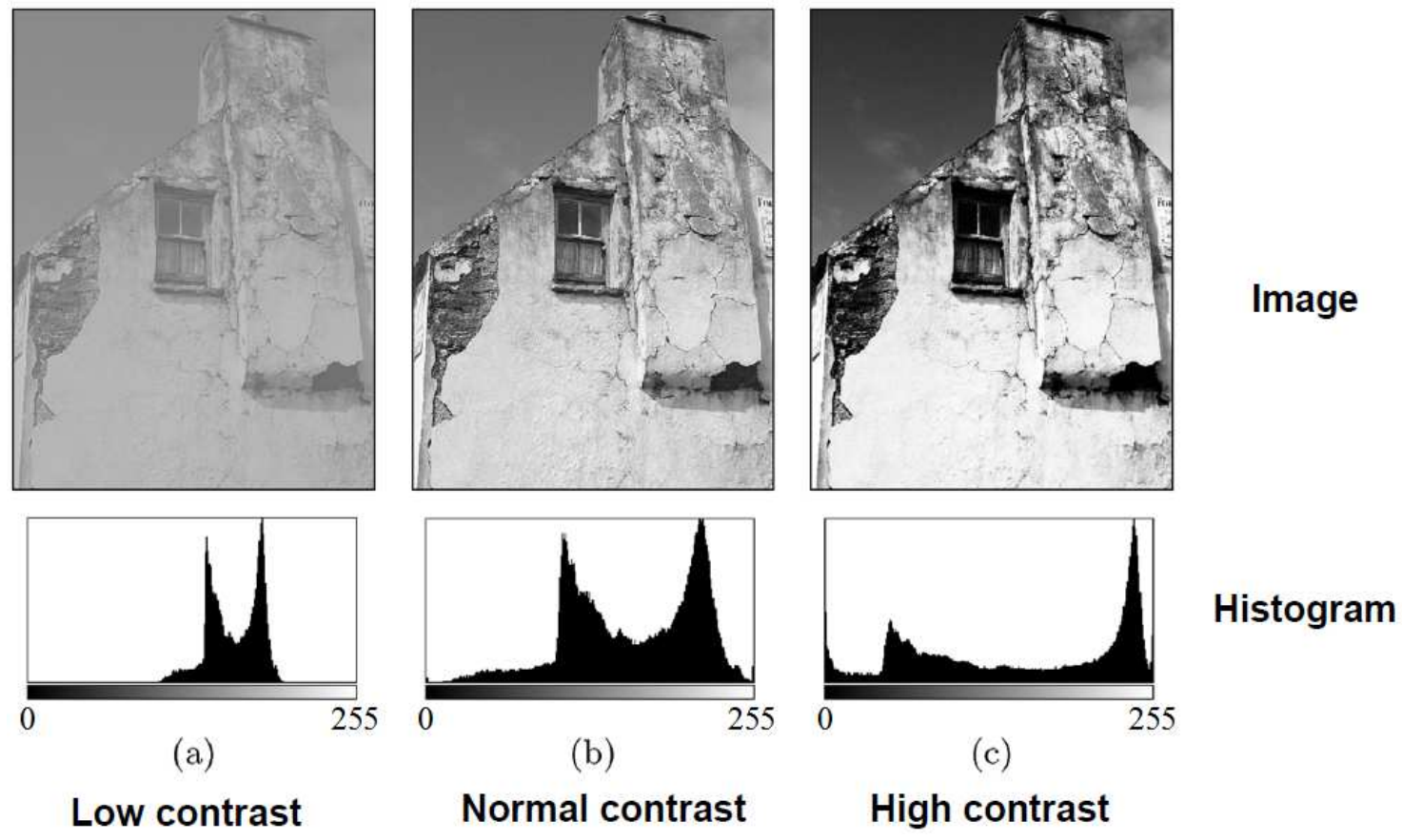

**Figure 72** Histograms in terms of exposure.

Finally, histograms are complementary to the witness bars seen in Section 3.3.1. Finally, they are very useful for calculating mutual information between frames of a video, as we can see in Section 4.1.5.

### 4.1.3 Superresolution of still images and its consequences

Next, we will see three methods based on QSA/FIT, which will allow us to perform super-resolution, compression, and filtering (denoising and despeckling) on signals, images and video. In the three methods we may use (indistinctly) *wavelets* (in particular, *Haar* basis, with compact support, see Section 2.2.2) and *coslets* (without compact support, see Section 4.1.1). On the other hand, we are going to explain it in images, although the method is automatically extended to the case of signals and video, without major complications.

The experiment consists of the following, given an image *I*, we apply wavelet (in this case, *Haar* basis) and get four subbands by each color, see Equations 207 to 211, so that, we will have, e.g., for the red channel,

$$LL_R^0 = I_R \text{, and} \tag{243}$$

$$\left[LL_R^1, LH_R^1, HL_R^1, HH_R^1\right] = haar_2\left(LL_R^0\right), \tag{244}$$

being $haar_2$ an operator which calculated the first level of two-dimensional *Haar* wavelet transform. Section 2.2.2. Besides, if $LL_R^0$ and $I_R \in [ROW, COL]$, instead, $LL_R^1, LH_R^1, HL_R^1$, and $HH_R^1 \in [ROW/2, COL/2]$, i.e., each subband of a second level of splitting with Haar would have a size of $[ROW/4, COL/4]$, and so.

The main idea is to get rid of detail subbands, that is to say, $\{LH_R^1, HL_R^1, HH_R^1\}$ and only transmit and/or store the approximation subband, i.e., $\{LL_R^1\}$. Thus, the data to transmit or store is reduced by 75%. Moreover, and in order to reconstruct the original image, we must restore the missing subbands (detail) from the approximation subband, in such a way that the distortion be minimal, in fact, the three methods (based on QSA/FIT) we will see below comply with this premise. Besides, and in order to organize what we see then, we'll target methods taking into account communications of mobile type, fundamentally. Hence, we will separate the three procedures in encoder and decoder. Subsequently, we synthesize everything is a single table that will be enlightening. Therefore, in the encoder we will have a downsampling process (in fact, a compression), while in the decoder we will have an upsampling (or decompression) which it is also a super-resolution of still image.

***First one***

*Encoder:*
We begin with Eq.(244)

$$\left[LL_R^1, LH_R^1, HL_R^1, HH_R^1\right] = haar_2\left(LL_R^0\right).$$

By tautology, every FIT is equal to itself, e.g.,

$$LH_R^1./LL_R^1 = LH_R^1./LL_R^1, \tag{245}$$

then, and without alteration

$$LH_R^1./LL_R^1 = LH_R^1./LL_R^1\left(LL_R^1./LL_R^1\right). \tag{246}$$

Now, if the size of $LH_R^1$ and $LL_R^1 \to \infty$, i.e., $\mathrm{ROW}/2 \wedge \mathrm{COL}/2 \to \infty$, and clearing $LH_R^1$, we will have

$$\lim_{ROW/2 \wedge COL/2 \to \infty} LH_R^1 \approx \frac{\left\langle LH_R^1 \middle| LL_R^1 \right\rangle}{\left\langle LL_R^1 \middle| LL_R^1 \right\rangle} LL_R^1 = cLH_R^1\ LL_R^1 \tag{247}$$

where

$$cLH_R^1 = \frac{\left\langle LH_R^1 \middle| LL_R^1 \right\rangle}{\left\langle LL_R^1 \middle| LL_R^1 \right\rangle} \tag{248}$$

Identical case for $HL_R^1$ and $HH_R^1$.

Finally, we eliminate to $\left\{LH_R^1, HL_R^1, HH_R^1\right\}$, and therefore, we transmit $\left\{LL_R^1, cLH_R^1, cHL_R^1, cHH_R^1\right\}$, being $\left\{cLH_R^1, cHL_R^1, cHH_R^1\right\}$ simple scalars.

Equation (247) is equivalent to say *mutual information* between $LH_R^1$ and $cLH_R^1\ LL_R^1$ is the maximum possible for a technique that touch resolution. See Section 4.1.5.

*Decoder:*
From $\left\{LL_R^1, cLH_R^1, cHL_R^1, cHH_R^1\right\}$ we rebuild $\left\{LH_R^1, HL_R^1, HH_R^1\right\}$, that is to say

$$LH_R^1 \simeq cLH_R^1\ LL_R^1 \tag{249a}$$
$$HL_R^1 \simeq cHL_R^1\ LL_R^1 \tag{249b}$$
$$HH_R^1 \simeq cHH_R^1\ LL_R^1. \tag{249c}$$

Then, we apply

$$LL_R^0 = ihaar_2\left(LL_R^1, LH_R^1, HL_R^1, HH_R^1\right), \tag{250}$$

where $ihaar_2$ is an operator, which calculated the first level of inverse two-dimensional *Haar* wavelet transform. Identical procedure for green and blue channels. Finally,

$$I_R = LL_R^0 \tag{251}$$

If: a) the original image is small, or b) we use several levels of downsampling/upsampling, or c) the original image comes from a previous filtering process, then, we need the deblurring technique of Section 4.1.1 for a reconstruction of the edges and texture.

***Second one***

*Encoder:*
We begin with Eq.(244)

$$\left[LL_R^1, LH_R^1, HL_R^1, HH_R^1\right] = haar_2\left(LL_R^0\right).$$

Then, we eliminate to $\left\{LH_R^1, HL_R^1, HH_R^1\right\}$, and we transmit only $\left\{LL_R^1\right\}$. This method is known in literature as *Rule of Three* (Ro3), see [257].

*Decoder:*
First, we define something called *replicated element of subband* (RES), which is repeated 4 times the value of each element of a matrix forming a macroelement. We can see this in a very simple example in Eq,(252). In such equation, each element of the matrix is repeated in square shaped.

$$\begin{bmatrix} a & b \\ c & d \end{bmatrix} \rightarrow \begin{bmatrix} a & a & b & b \\ a & a & b & b \\ c & c & d & d \\ c & c & d & d \end{bmatrix} \tag{252}$$

Then and always for red channel (as example), we apply,

$$\left[ LL_R^2, LH_R^2, HL_R^2, HH_R^2 \right] = haar_2\left( LL_R^1 \right). \tag{253}$$

Now, we use operator $res_2(\bullet)$ to each subband of second level of $haar_2$ splitting,

$$LL_{R_res}^2 = res_2\left( LL_R^2 \right), \tag{254a}$$

$$LH_{R_res}^2 = res_2\left( LH_R^2 \right), \tag{254b}$$

$$HL_{R_res}^2 = res_2\left( HL_R^2 \right), \text{ and} \tag{254c}$$

$$HH_{R_res}^2 = res_2\left( HH_R^2 \right). \tag{254d}$$

Therefore, if we make

$$\lim_{ROW/2 \wedge COL/2 \to \infty} LH_R^1 ./ LL_R^1 = LH_{R_res}^2 ./ LL_{R_res}^2 \tag{255}$$

where superscript $2$ means second level of splitting thanks to $haar_2$ . This means that FITs of consistent levels of splitting ($n$ and $n+1$) tend to the same value as the dimension of the matrices tends to infinity. This also means that the mutual information between $LL_R^1 \; and \; LL_{R_res}^2$ and $LH_R^1 \; and \; LH_{R_res}^2$ is maximized for the same circumstances. Thus,

$$LH_R^1 ./ LL_R^1 \simeq LH_{R_res}^2 ./ LL_{R_res}^2 \left( LL_{R_res}^2 ./ LL_{R_res}^2 \right) \tag{256}$$

Now, if the size of $LH_R^1$ , $LL_R^1$ , $LH_{R_res}^2$ , and $LL_{R_res}^2 \to \infty$ , i.e., $\text{ROW}/2 \wedge \text{COL}/2 \to \infty$ , and clearing $LH_R^1$, we will have

$$\lim_{ROW/2 \wedge COL/2 \to \infty} LH_R^1 \approx \frac{\left\langle LH_{R_res}^2 \middle| LL_{R_res}^2 \right\rangle}{\left\langle LL_{R_res}^2 \middle| LL_{R_res}^2 \right\rangle} LL_R^1 \tag{257}$$

Therefore,

$$LH_R^1 \simeq \frac{\left\langle LH_{R_res}^2 \middle| LL_{R_res}^2 \right\rangle}{\left\langle LL_{R_res}^2 \middle| LL_{R_res}^2 \right\rangle} LL_R^1 \tag{258a}$$

$$HL_R^1 \simeq \frac{\left\langle HL_{R_res}^2 \middle| LL_{R_res}^2 \right\rangle}{\left\langle LL_{R_res}^2 \middle| LL_{R_res}^2 \right\rangle} LL_R^1 \tag{258b}$$

$$HH_R^1 \simeq \frac{\left\langle HH_{R_res}^2 \middle| LL_{R_res}^2 \right\rangle}{\left\langle LL_{R_res}^2 \middle| LL_{R_res}^2 \right\rangle} LL_R^1 \tag{258c}$$

Another version element-by-element presented in [257] is

$$LH^1_{R_r,c} \simeq \frac{LH^2_{R_res_r,c}}{LL^2_{R_res_r,c}} LL^1_{R_r,c} \tag{259a}$$

$$HL^1_{R_r,c} \simeq \frac{HL^2_{R_res_r,c}}{LL^2_{R_res_r,c}} LL^1_{R_r,c} \tag{259b}$$

$$HH^1_{R_r,c} \simeq \frac{HH^2_{R_res_r,c}}{LL^2_{R_res_r,c}} LL^1_{R_r,c} \tag{259c}$$

$\forall r \in [1, ROW/2] \wedge \forall c \in [1, COL/2]$.

Under the same circumstances, we can replace $LH^2_R$ by $LH^2_{R_res}$, and $LL^2_R$ by $LL^2_{R_res}$ in Eq.(257),

$$LH^1_R \simeq \frac{\langle LH^2_R | LL^2_R \rangle}{\langle LL^2_R | LL^2_R \rangle} LL^1_R \tag{260a}$$

$$HL^1_R \simeq \frac{\langle HL^2_R | LL^2_R \rangle}{\langle LL^2_R | LL^2_R \rangle} LL^1_R \tag{260b}$$

$$HH^1_R \simeq \frac{\langle HH^2_R | LL^2_R \rangle}{\langle LL^2_R | LL^2_R \rangle} LL^1_R \tag{260c}$$

Therefore, this version has three alternatives for its implementation, i.e., Equations (258, 259, and 260).

Then, we apply

$$LL^0_R = ihaar_2\left(LL^1_R, LH^1_R, HL^1_R, HH^1_R\right), \tag{261}$$

Finally,

$$I_R = LL^0_R \tag{262}$$

Here too, if: a) the original image is small, or b) we use several levels of downsampling/upsampling, or c) the original image comes from a previous filtering process, then, we need the deblurring technique of Section 4.1.1 for a reconstruction of the edges and texture.

***Third one***

*Encoder:*
We begin with Eq.(244)

$$\left[LL^1_R, LH^1_R, HL^1_R, HH^1_R\right] = haar_2\left(LL^0_R\right).$$

Then, we eliminate to $\{LH^1_R, HL^1_R, HH^1_R\}$, and we transmit only $\{LL^1_R\}$. Hitherto, this version is identical to the previous one.

*Decoder:*
From $\{LL^1_R\}$ we rebuild $\{LH^1_R, HL^1_R, HH^1_R\}$, based on directional FITs of Section 3.3.2, that is to say, the developed case with overlap, in particular, we will use a segmental convolution mask with $M$ = 3, based on Equations (from 174 to 178, 182, and 191). Thus, we define an operator called $FIT_2(\bullet)$ which allow us to obtain $\{LH_R^1, HL_R^1, HH_R^1\}$ from $LL_R^1$ from the mentioned equations, then

$$\lim_{ROW/2 \wedge COL/2 \to \infty} \{LH_R^1, HL_R^1, HH_R^1\} \approx bat_2\left(LL_R^1\right), \tag{263}$$

with

$$LH_R^1 \simeq bat_2^h\left(LL_R^1\right), \tag{264a}$$

$$HL_R^1 \simeq bat_2^v\left(LL_R^1\right), \text{ and} \tag{264b}$$

$$HH_R^1 \simeq bat_2^d\left(LL_R^1\right). \tag{264c}$$

Where superscripts {*h, v, d*} means {*horizontal, vertical, diagonal*} respectively. Then, we apply

$$LL_R^0 = ihaar_2\left(LL_R^1, LH_R^1, HL_R^1, HH_R^1\right), \tag{265}$$

Finally,

$$I_R = LL_R^0 \tag{266}$$

Here, we'll make identical considerations to previous versions regarding the deblurring.

***Comparation between them***

Next, in Table III we can see a comparison between three versions, mainly and indirectly in relation to computational cost, and channel usage.

TABLE III
COMPARISON BETWEEN THREE VERSIONS.

| Version | Encoder | Decoder |
|---|---|---|
| 1 | We calculate $\{LL_R^1, LH_R^1, HL_R^1, HH_R^1, cLH_R^1, cHL_R^1, cHH_R^1\}$<br>We eliminate $\{LH_R^1, HL_R^1, HH_R^1\}$<br>We transmit $\{LL_R^1, cLH_R^1, cHL_R^1, cHH_R^1\}$ | We receive $\{LL_R^1, cLH_R^1, cHL_R^1, cHH_R^1\}$<br>We rebuild $\{LH_R^1, HL_R^1, HH_R^1\}$<br>We calculate $\{I\}$<br>We eliminate $\{LL_R^1, LH_R^1, HL_R^1, HH_R^1\}$ |
| 2 | We calculate $\{LL_R^1\}$<br>We transmit $\{LL_R^1\}$ | We receive $\{LL_R^1\}$<br>We calculate $\{LL_R^2, LH_R^2, HL_R^2, HH_R^2\}$<br>We calculate $\{LH_R^1, HL_R^1, HH_R^1\}$<br>We calculate $\{I\}$<br>We eliminate $\{LL_R^1, LH_R^1, HL_R^1, HH_R^1, LL_R^2, LH_R^2, HL_R^2, HH_R^2\}$ |
| 3 | We calculate $\{LL_R^1\}$<br>We transmit $\{LL_R^1\}$ | We receive $\{LL_R^1\}$<br>We rebuild $\{LH_R^1, HL_R^1, HH_R^1\}$<br>We calculate $\{I\}$<br>We eliminate $\{LL_R^1, LH_R^1, HL_R^1, HH_R^1\}$ |

Clearly, the third method has the best of two previous versions. In Section 4.2.2 we simulate the three seen.

On the other hand, exists a fourth version which consists to suppress the bands of detail in the encoder and replace them with zeros in the decoder, and we call *Blacking*, which will not be seen in this work and significantly lower quality than the said.

### 4.1.4 Edge detection

We can obtain edge detection of images via QSA/FIT, e.g., if we use Lena (in gray level) of Fig.13 and 34,

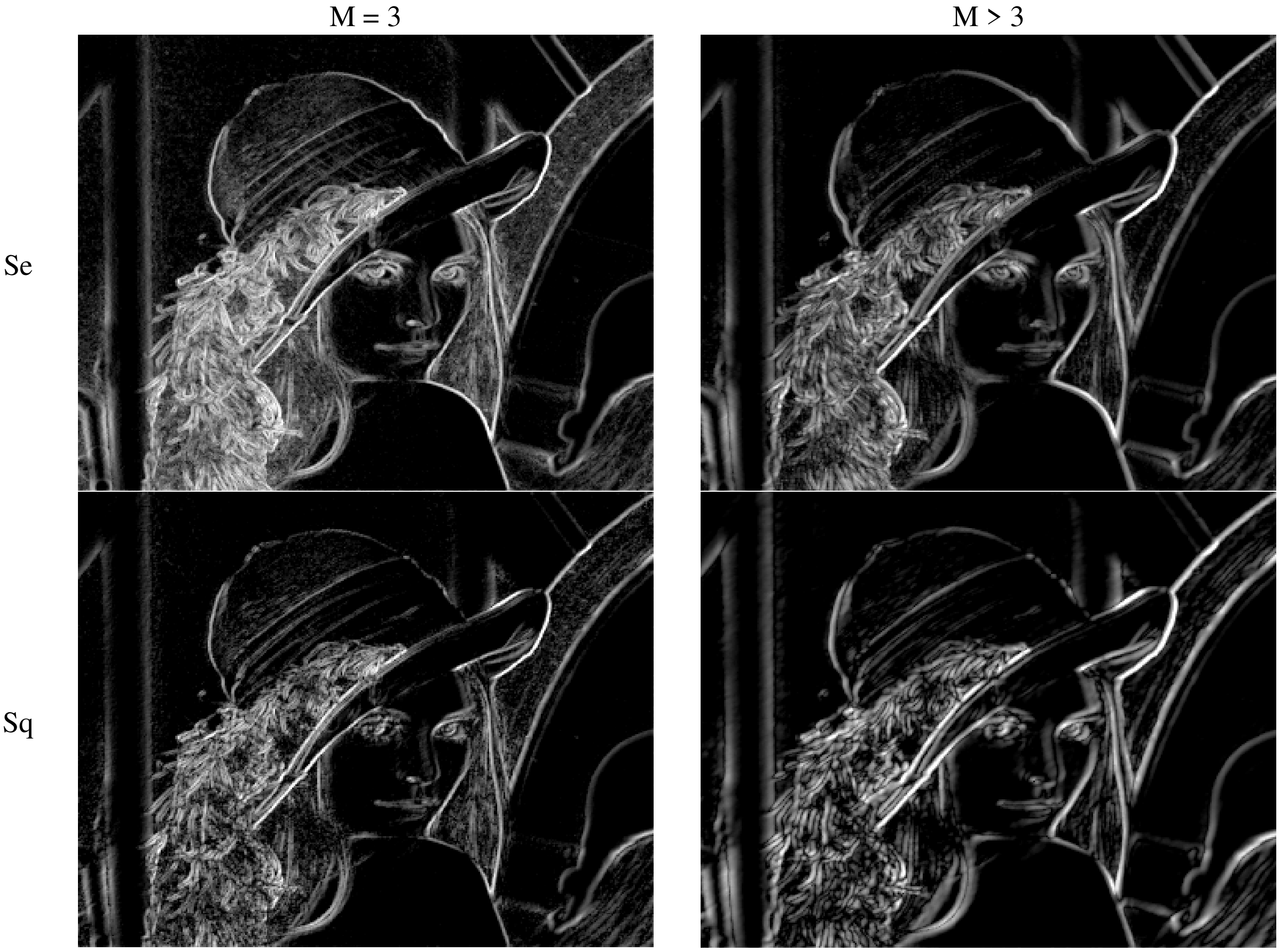

**Fig.73** Edge detection by type and size of the convolution mask.

in particular for the *case with overlap*, we can plot $f_{eq}$ for segmental mask with $M = 3$ from Eq.(190), square mask with $M = 3$ from Eq.(200), segmental mask with $M > 3$ ($M = 7$), and square mask with $M > 3$ ($M = 7$), all in Fig.73, where {Se, Sq} means {segmental, square} respectively. As we can see in such figure, the results are similar to the traditional techniques of edge detection based on Sobel, Prewitt, Roberts, and Canny [45-48, 256]. Finally, Fig.73 was post- produced for better viewing.

### 4.1.5 Metrics

Below, we present metrics that will be used in the simulations.

***Data Compression Ratio (CR)***

Data compression ratio, also known as compression power, is a computer-science term used to quantify the reduction in data-representation size produced by a data compression algorithm. The data compression ratio is analogous to the physical compression ratio used to measure physical compression of substances, and is defined in the same way, as the ratio between the *uncompressed size* and the *compressed size* [45]:

$$CR = \frac{Uncompressed\ Size}{Compressed\ Size} \tag{267}$$

Thus a representation that compresses a 10MB file to 2MB has a compression ratio of 10/2 = 5, often notated as an explicit ratio, 5:1 (read "five to one"), or as an implicit ratio, 5X. Note that this formulation applies equally for compression, where the uncompressed size is that of the original; and for decompression, where the uncompressed size is that of the reproduction.

***Percent Space Savings (PSS)***

Sometimes the space savings is given instead, which is defined as the reduction in size relative to the uncompressed size:

$$PSS = \left(1 - \frac{1}{CR}\right) * 100\% \tag{268}$$

Thus a representation that compresses 10MB file to 2MB would yield a space savings of 1-2/10 = 0.8, often notated as a percentage, 80%.

***Peak Signal-To-Noise Ratio (PSNR)***

The phrase peak signal-to-noise ratio, often abbreviated PSNR, is an engineering term for the ratio between the maximum possible power of a signal and the power of corrupting noise that affects the fidelity of its representation. Because many signals have a very wide dynamic range, PSNR is usually expressed in terms of the logarithmic decibel scale.

The PSNR is most commonly used as a measure of quality of reconstruction in image compression etc [45]. It is most easily defined via the mean squared error (MSE) which for two $NR \times NC$ (rows-by-columns) monochrome images $\boldsymbol{I}$ and $\boldsymbol{I_d}$ , where the second one of the images is considered a decompressed/denoised approximation of the other is defined as:

$$MSE = \frac{1}{NRxNC} \sum_{nr=0}^{NR-1} \sum_{nc-0}^{NC-1} \left[ I(nr,nc) - I_d(nr,nc) \right]^2 \tag{269}$$

The PSNR is defined as [258]:

$$PSNR = 10 \log_{10}\left(\frac{MAX_I^2}{MSE}\right) = 20 \log_{10}\left(\frac{MAX_I}{\sqrt{MSE}}\right) \tag{270}$$

Here, $MAX_i$ is the maximum pixel value of the image. When the pixels are represented using 8 bits per sample, this is 255. More generally, when samples are represented using linear pulse code modulation (PCM) with B bits per sample, maximum possible value of $MAX_i$ is $2^B$-1.

For color images with three red-green-blue (RGB) values per pixel, the definition of PSNR is the same except the MSE is the sum over all squared value differences divided by image size and by three [45]. A similar situation occurs with the mean absolute error (MAE), which is discussed below.

Typical values for the PSNR in lossy image and video compression are between 30 and 50 dB, where higher is better.

Finally, a conspicuous metric for these cases is the mean absolute error, which is a quantity used to measure how close forecasts or predictions are to the eventual outcomes. The mean absolute error (MAE) is given by

$$MAE = \frac{1}{NRxNC} \sum_{nr=0}^{NR-1} \sum_{nc=0}^{NC-1} \left| I(nr,nc) - I_d(nr,nc) \right| \tag{271}$$

which for two $NR \times NC$ (rows-by-columns) monochrome images $\boldsymbol{I}$ and $\boldsymbol{I_d}$ , where the second one of the images is considered a decompressed/denoised approximation of the other of the first one.

In the case of signals, MSE and MAE will be

$$MSE = \frac{1}{N} \sum_{n=0}^{N-1} \left[ S(n) - S_d(n) \right]^2 \tag{272}$$

$$MAE = \frac{1}{N} \sum_{n=0}^{N} \left| S(n) - S_d(n) \right| \tag{273}$$

while PSNR be

$$PSNR = 10 \log_{10} \left( \frac{MAX_S^2}{MSE} \right) = 20 \log_{10} \left( \frac{MAX_S}{\sqrt{MSE}} \right) \tag{274}$$

It's easy to generalize {MAE, MSE, and PSNR} for images (2D) and video (3D).

***Mutual Information between two matrices***

Here, we are going to calculate the Mutual Information between two images, frames or matrices thanks to marginal entropies and the joint entropy [259]. Therefore, Mutual information can be expressed as

$$I(X,Y) = H(X) + H(Y) - H(X,Y), \tag{275}$$

where $I(X;Y)$ is the mutual information between the two images, $H(X)$ is the entropy of image $X$, $H(Y)$ is the entropy of image $Y$, and $H(X,Y)$ is the joint entropy between images.

*Entropy of an image*

If $X$ is the image, we begin with the normalization of $X$, i.e.,

$$\widetilde{X} = \frac{X}{\sum_{r=1}^{ROW} \sum_{c=1}^{COL} X(r,c)}, \quad \forall\, r \in [1, ROW] \wedge c \in [1, COL], \tag{276}$$

such that,

$$\sum_{r=1}^{ROW} \sum_{c=1}^{COL} \widetilde{X}(r,c) = 1, \tag{277}$$

Thus,

$$H(X) = -\sum_{r=1}^{ROW} \sum_{c=1}^{COL} \widetilde{X}(r,c) \log_2 \left( \widetilde{X}(r,c) \right), \tag{278}$$

*Mutual Information between two images, frames or matrices*
However, it is more practical to calculate the entropies (including the joint entropy, and hence the Mutual Information) based on histograms as seen in Section 4.1.2. To this end, we recommend the MATLAB code [49] developed by Jose Delpiano [260, 261].

### 4.2 Simulations properly speaking

For all experiments, we are going to use metrics of the last section. Besides, the employed signal will be that of Fig.43, i.e., *electrocardiographic* (ECG) signal of 80 pulses per second, and 1024 samples per second, while, for the image case, we will use to *Angelina* of Fig.48, with 1920-by-1080 pixels, and 24 bpp.

Finally, we only simulate those techniques that we consider conspicuous in order to test the strength of QSA/FIT, or those that we have not simulated in the previous sections.

#### 4.2.1 Denoising (and despeckling)

The general schema of the denoising procedure can be seen in Fig.74, where the signal (in fact, signal or image) is corrupted by additive noise (or multiplicative noise of Gamma distribution known like speckle [50, 162-171, 199]). After the filter, we compare denoised and original image with a set of metrics like defined in the last section, in order to test the filtering (denoising for additive noise and despeckling for multiplicative noise or speckle) performance.

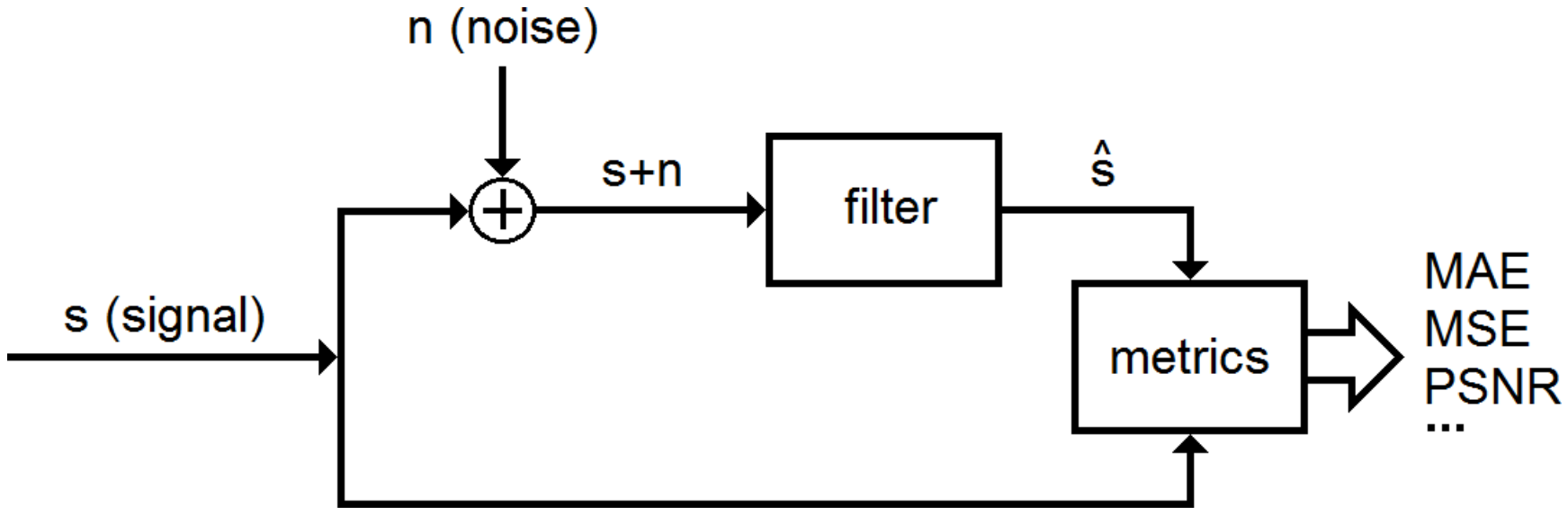

**Fig.74** Filtering: denoising/despeckling.

***Signals***
In all cases, the signal is injected with a 5% of white noise with Gaussian distribution and zero mean value.

On the other hand, Table IV shows us denoising techniques vs metrics, where acronyms means:

**SG:** Savitzky-Golay
**MF:** Mean filter. In all cases $M = 3$, i.e., it is a square mask of 3x3.
**DS:** Directional Smoothing. In all cases $M = 3$, i.e., it is a square mask of 3x3.
**ST:** Soft-thresholding
**HT:** Hard-thresholding
**SR-*i*:** Super-resolution, version *i*

Clearly, **Coslet** is better than **Haar** basis in all applications. Moreover, although Savitzky-Golay has better metric results, it is very slow regarding to the rest of techniques. Besides, Savitzky-Golay is used at its best version, while the other at its worst version.

Figure 75 shows us corresponding simulations with noisy signal in red and denoised signal in blue. Clearly it improved in all cases.

TABLE IV

DENOISING TECHNIQUES VS METRICS FOR ELECTROCARDIOGRAPHIC SIGNAL.

| Technique | Metrics | | |
|---|---|---|---|
| | MAE | MSE | PSNR |
| SG | 0.6127 | 0.6187 | 42.5526 |
| MF | 0.6918 | 0.7976 | 41.4496 |
| DS | 1.7021 | 4.8446 | 33.6149 |
| Haar/ST | 1.2983 | 4.4874 | 33.9474 |
| Haar/HT | 1.3232 | 3.5289 | 34.9910 |
| Haar/MF | 1.2810 | 2.6042 | 36.3107 |
| Haar/DS | 1.0979 | 2.3168 | 36.8185 |
| Coslet/ST | 0.8298 | 1.0913 | 40.0879 |
| Coslet/HT | 1.0184 | 1.5812 | 38.4773 |
| Coslet/MF | 1.2224 | 2.3359 | 36.7828 |
| Coslet/DS | 0.8087 | 1.0250 | 40.3599 |
| SR-1/Haar | 1.4409 | 5.3277 | 33.2020 |
| SR-1/Coslet | 0.8618 | 1.1830 | 39.7375 |
| SR-2/Haar | 1.3447 | 5.0144 | 33.4652 |
| SR-2/Coslet | 0.9315 | 1.3227 | 39.2527 |
| SR-3/Haar | 1.4428 | 5.0915 | 33.3989 |
| SR-3/Coslet | 0.8121 | 1.0195 | 40.3833 |

Next, we define some important aspects of the simulations:

**SG:** Smooth the signal by applying a cubic Savitzky-Golay filter to data frames of length 41.
**MF:** it was applied 26 times
**DS:** it was applied 2 times
**Haar/DS:** DS was applied 10 times
**Haar/MF:** MF was applied 5 times
**Coslet/DS:** DS was applied 20 times
**Coslet/MF:** MF was applied 5 times
Always, **Haar** and **Coslet** were used with 3 level of splitting
Finally, for the three versions of **SR**, we have the following **MI** (mutual information) values between subbands of consecutive levels of splitting:

*Haar*:
MI ($L^0 = S$ and $L^1$) = 5.1395
MI ($L^1$ and $L^2$) = 4.7895
MI ($L^2$ and $L^3$) = 4.5368
MI ($L^3$ and $L^4$) = 4.2153
MI ($H^1$ and $H^2$) = 3.4076
MI ($H^2$ and $H^3$) = 3.2214
MI ($H^3$ and $H^4$) = 3.1905

*Coslet*:
MI ($L^0 = S$ and $L^1$) = 5.1577
MI ($L^1$ and $L^2$) = 4.7741
MI ($L^2$ and $L^3$) = 4.1564
MI ($L^3$ and $L^4$) = 3.9807
MI ($H^1$ and $H^2$) = 1.1155
MI ($H^2$ and $H^3$) = 3.1654
MI ($H^3$ and $H^4$) = 4.1600

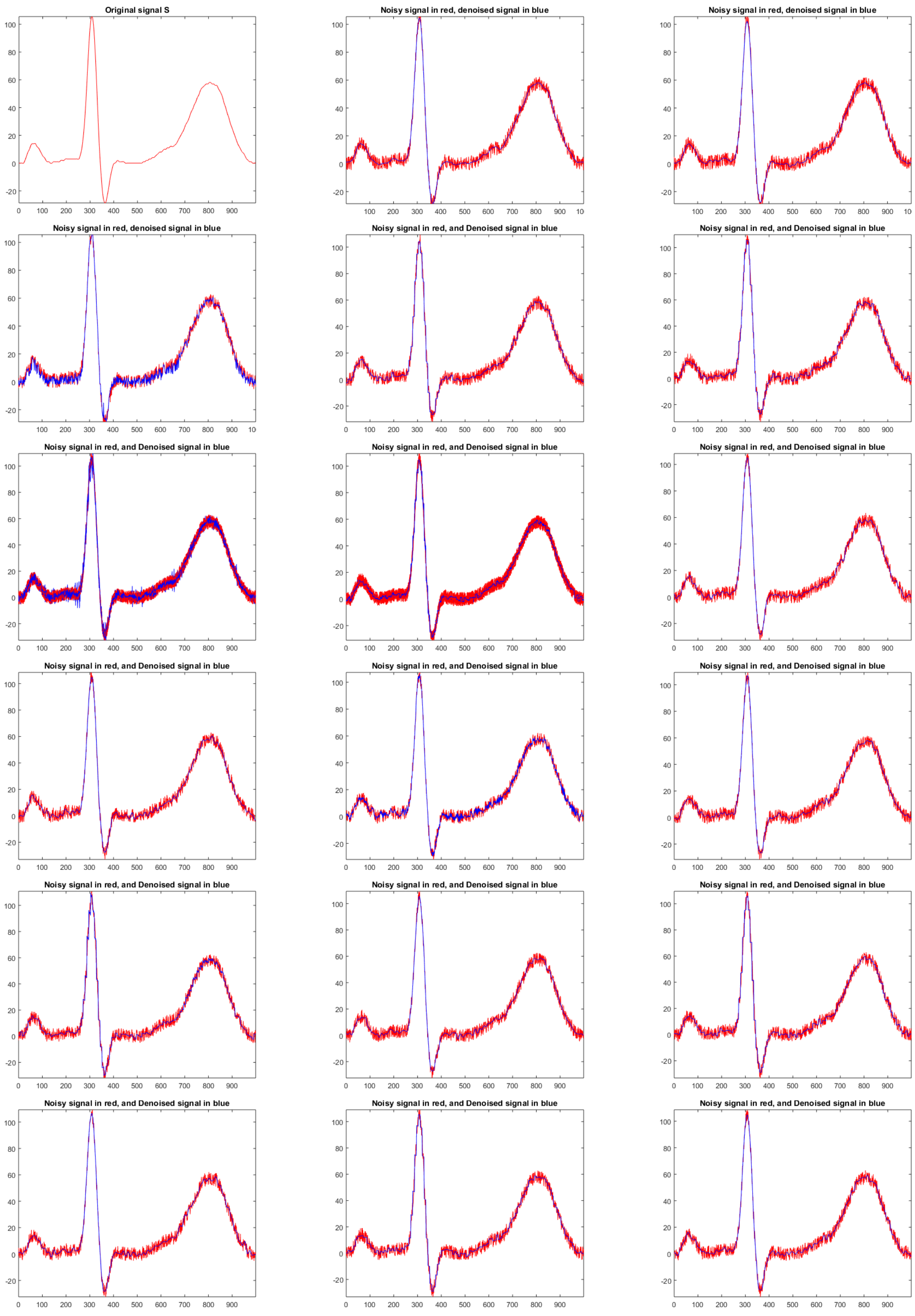

**Fig.75** (1,1) Original signal, for the rest, noisy signal is in red and denoised signal in blue. In order: (1,2) SG, (1,3) MF, (2,1) DS, (2,2) Haar/ST, (2,3) Haar/HT, (3,1) Haar/MF, (3,2) Haar/DS, (3,3) Coslet/ST, (4,1) Coslet/HT, (4,2) Coslet/MF, (4,3) Coslet/DS, (5,1) SR-1/Haar, (5,2) SR-1/Coslet, (5,3) SR-2/Haar, (6.1) SR-2/Coslet, (6,2) SR-3/Haar, (6,3) SR-3/Coslet.

***Images***

Here too, in all cases, the signal is injected with a 5% of white noise with Gaussian distribution and zero mean value. On the other hand, Table V shows us denoising techniques vs metrics. As in the case of signals, **Coslet** has a better performance than **Haar**. Next, we define some important aspects of the simulations:

TABLE V
DENOISING TECHNICS VS METRICS FOR ANGELINA.

| Technique | Metrics | | |
|---|---|---|---|
| | MAE | MSE | PSNR |
| SG | 12.6555 | 128.0198 | 27.0580 |
| MF | 11.0919 | 130.4895 | 26.9750 |
| DS | 11.3643 | 117.4117 | 27.4336 |
| Haar/ST | 13.1756 | 210.3154 | 24.9020 |
| Haar/HT | 13.1386 | 207.4902 | 24.9608 |
| Haar/MF | 13.1870 | 224.6231 | 24.6162 |
| Haar/DS | 15.1179 | 338.7261 | 22.8323 |
| Coslet/ST | 12.9394 | 193.1937 | 25.2708 |
| Coslet/HT | 12.9352 | 192.7052 | 25.2818 |
| Coslet/MF | 13.0036 | 212.7928 | 24.8512 |
| Coslet/DS | 15.0539 | 334.7423 | 22.8836 |
| SR-1/Haar | 13.2233 | 213.9069 | 24.8285 |
| SR-1/Coslet | 12.9421 | 193.9325 | 25.2542 |
| SR-2/Haar | 13.2347 | 214.1865 | 24.8228 |
| SR-2/Coslet | 12.9647 | 194.4172 | 25.2434 |
| SR-3/Haar | 13.2226 | 213.7166 | 24.8324 |
| SR-3/Coslet | 12.9641 | 194.2723 | 25.2466 |

**SG:** Smooth the image by applying a degree = 7 Savitzky-Golay filter to data frames of length 7.
**MF:** it was applied 3 times
**DS:** it was applied 3 times
**Haar/DS:** DS was applied 3 times
**Haar/MF:** MF was applied 3 times
**Coslet/DS:** DS was applied 3 times
**Coslet/MF:** MF was applied 3 times
Always, **Haar** and **Coslet** were used with 3 level of splitting
Finally, for the three versions of **SR**, we have the following **MI** (mutual information) values between subbands of consecutive levels of splitting:

*Haar*:
MI ($LL^0$ = I and $LL^1$) = 4.6510
MI ($LL^1$ and $LL^2$) = 4.1177
MI ($LL^2$ and $LL^3$) = 3.8293
MI ($LL^3$ and $LL^4$) = 3.7332
MI ($LH^1$ and $LH^2$) = 0.7555
MI ($LH^2$ and $LH^3$) = 0.8528
MI ($LH^3$ and $LH^4$) = 1.2026
MI ($HL^1$ and $HL^2$) = 0.6563
MI ($HL^2$ and $HL^3$) = 0.7721
MI ($HL^3$ and $HL^4$) = 1.0706
MI ($HH^1$ and $HH^2$) = 0.4938
MI ($HH^2$ and $HH^3$) = 0.5381
MI ($HH^3$ and $HH^4$) = 0.7983

*Coslet*:
MI ($LL^0$ = I and $LL^1$) = 4.4390
MI ($LL^1$ and $LL^2$) = 3.7376
MI ($LL^2$ and $LL^3$) = 3.4205
MI ($LL^3$ and $LL^4$) = 3.3838
MI ($LH^1$ and $LH^2$) = 0.2762
MI ($LH^2$ and $LH^3$) = 0.3908
MI ($LH^3$ and $LH^4$) = 0.7539
MI ($HL^1$ and $HL^2$) = 0.2614
MI ($HL^2$ and $HL^3$) = 0.3248
MI ($HL^3$ and $HL^4$) = 0.3963
MI ($HH^1$ and $HH^2$) = 0.1762
MI ($HH^2$ and $HH^3$) = 0.3144
MI ($HH^3$ and $HH^4$) = 0.4838

Figure 76a shows us the noisy image, while Figures from 76b to 76r shows us denoised images. Clearly it improved in all cases.

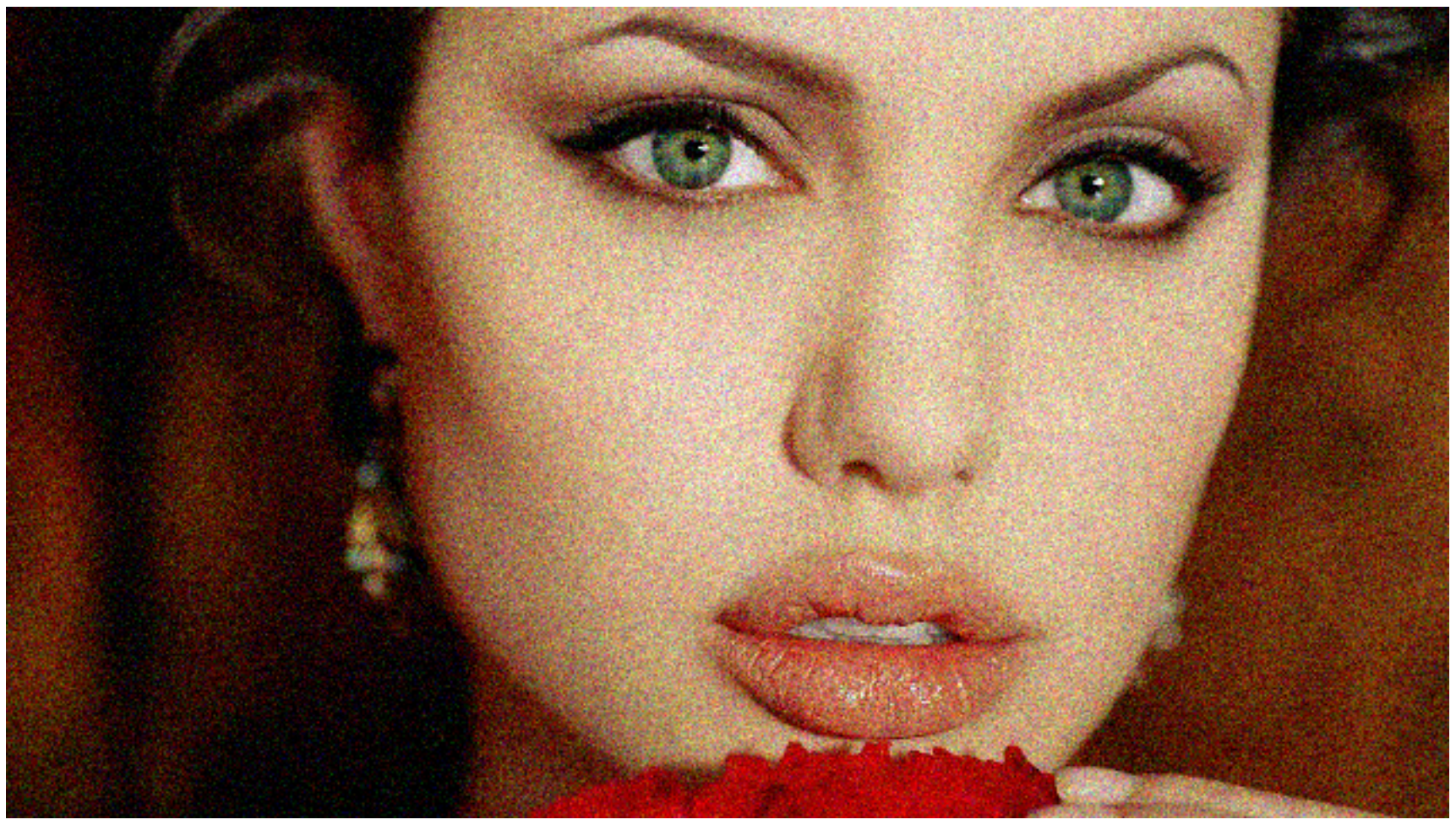

**Fig.76a** Angelina's noisy image.

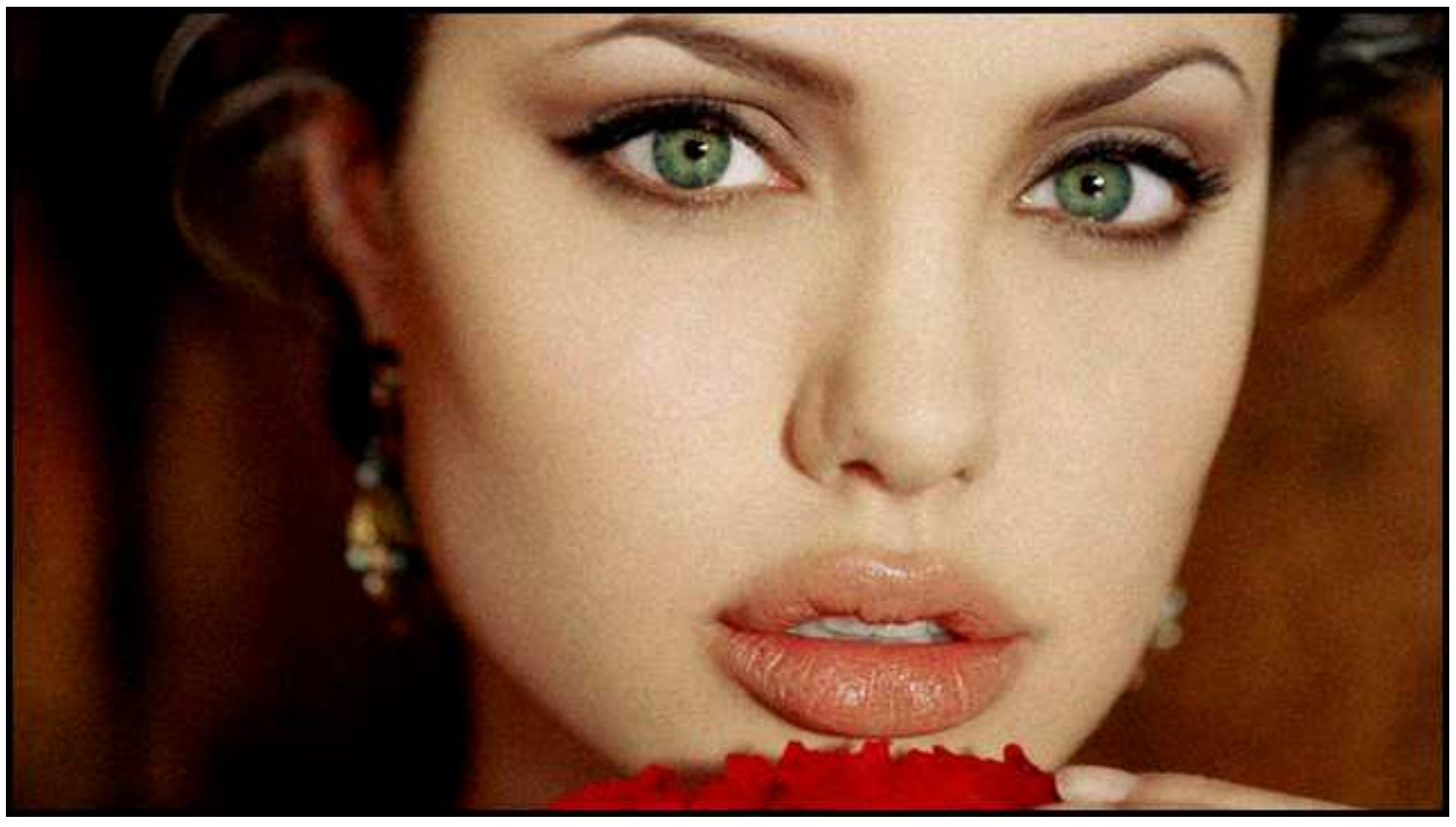

**Fig.76b** Angelina's denoised image via SG.

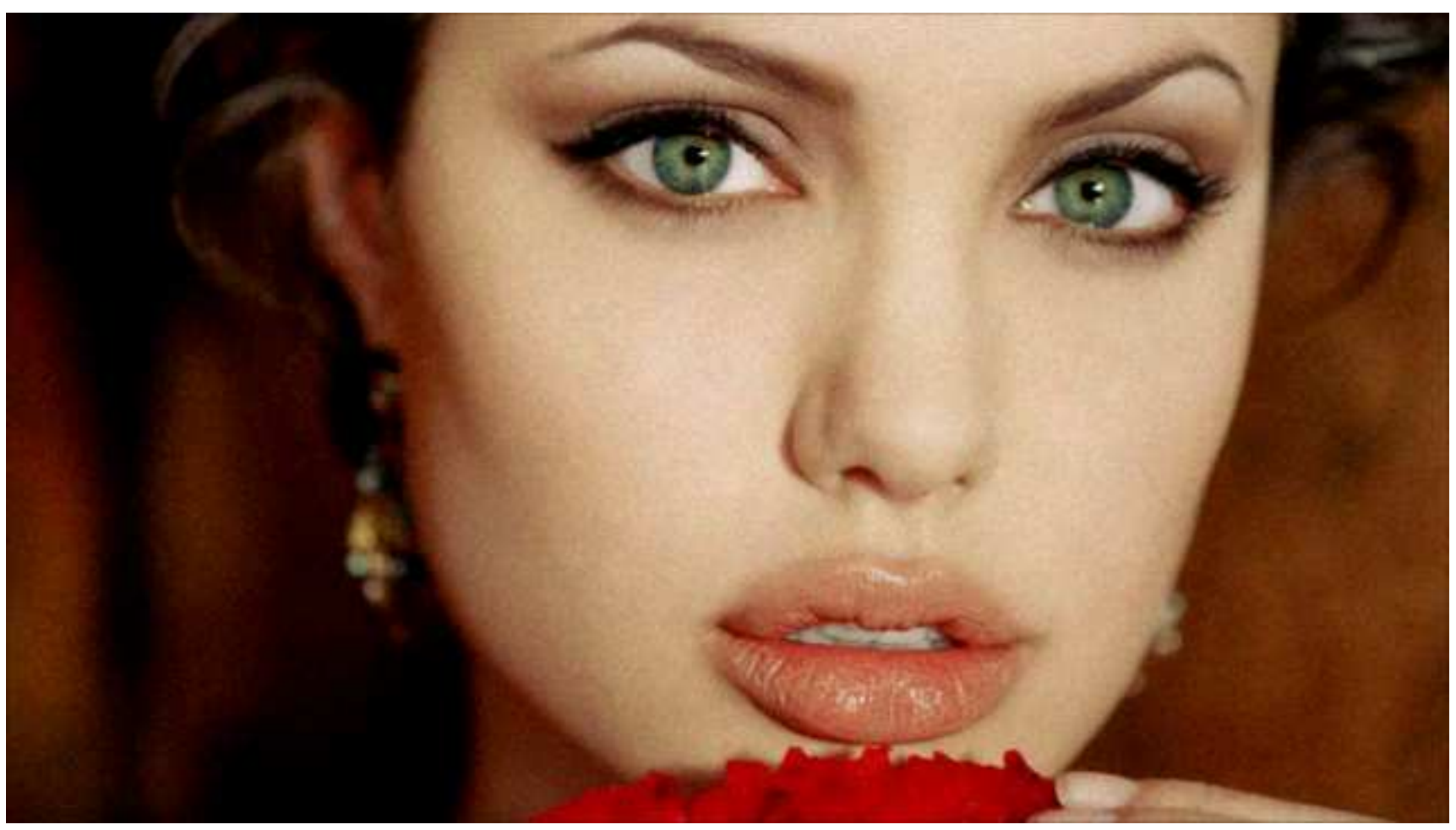

**Fig.76c** Angelina's denoised image via MF.

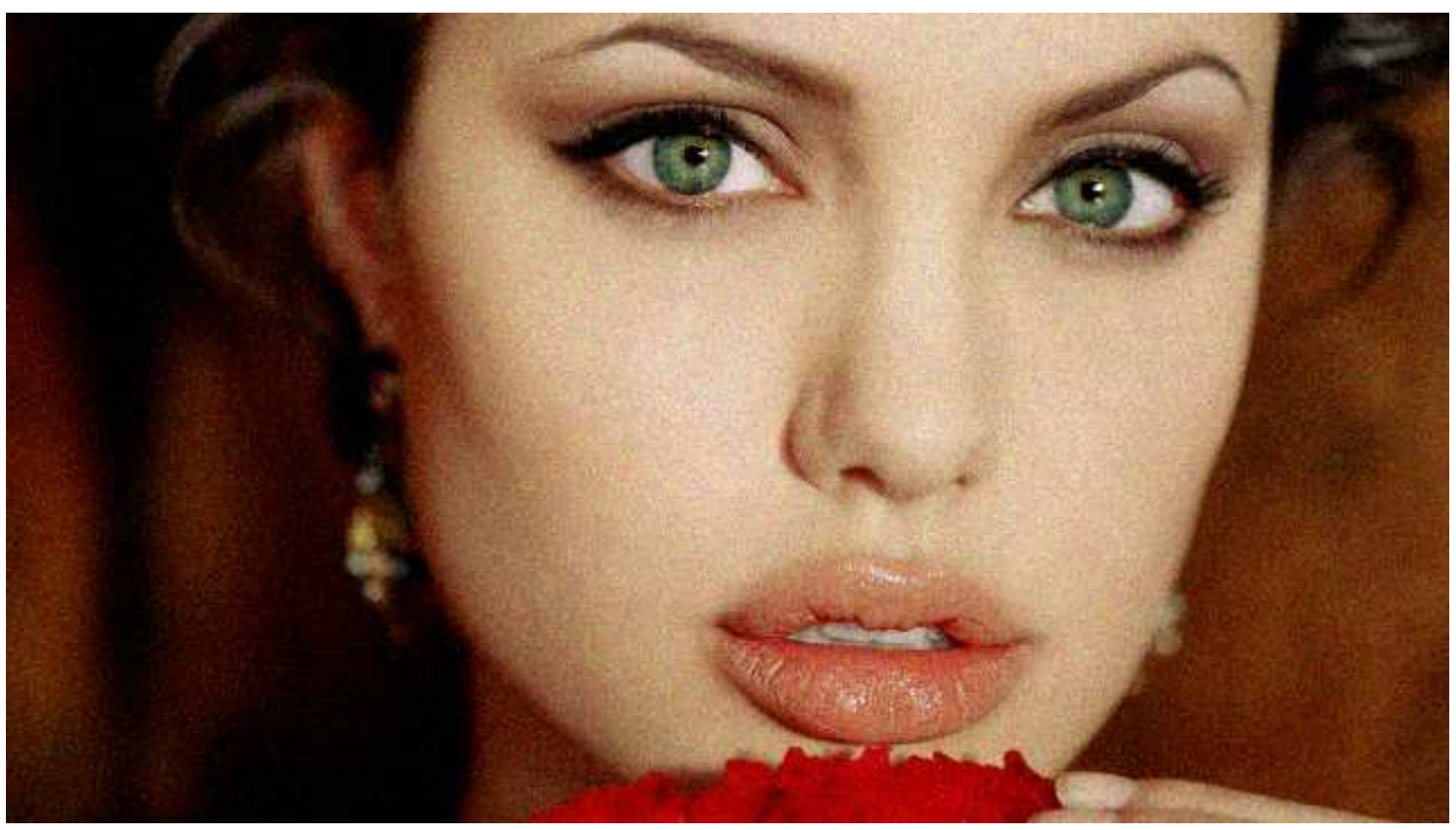

**Fig.76d** Angelina's denoised image via DS.

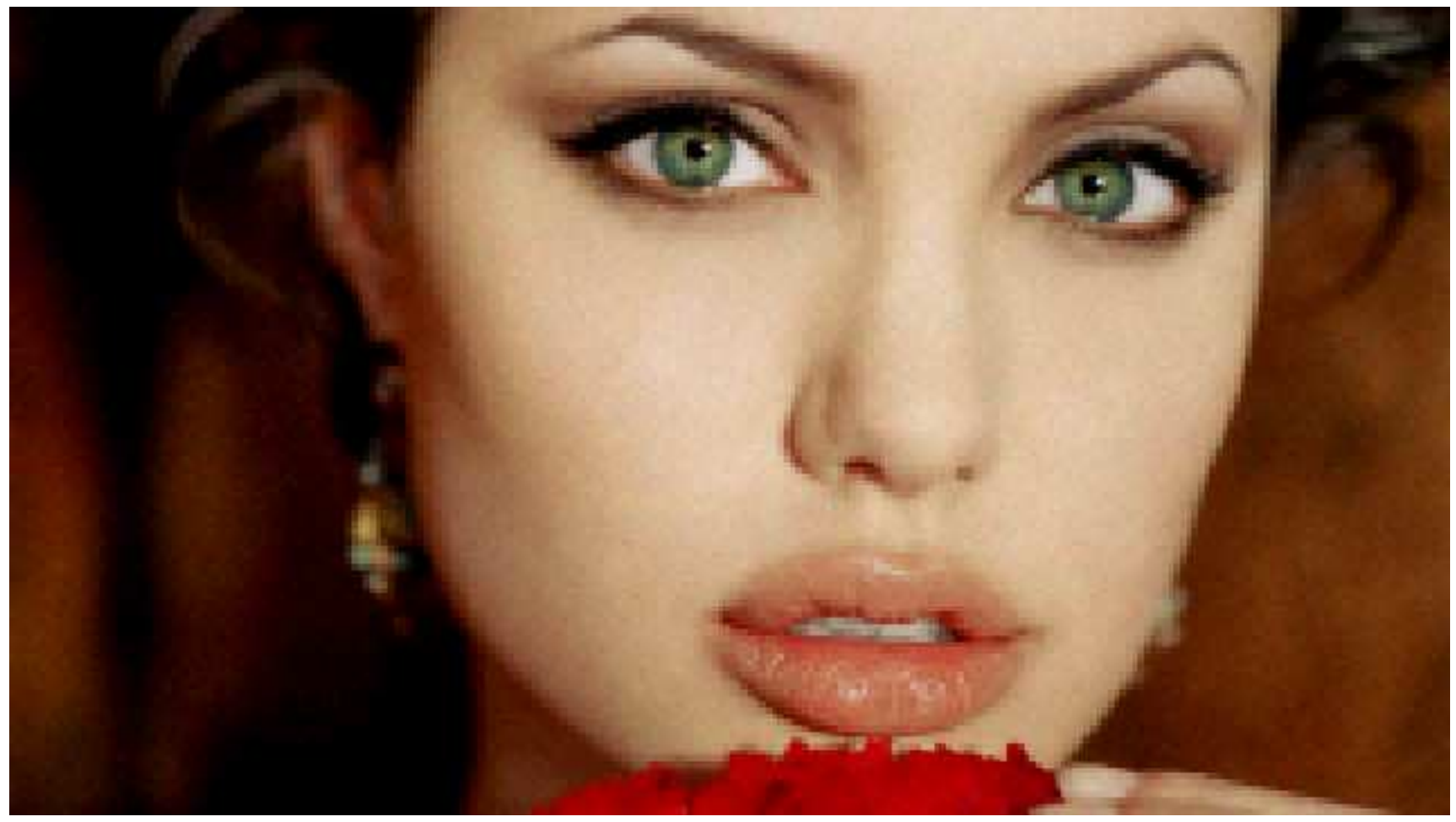

**Fig.76e** Angelina's denoised image via Haar/ST.

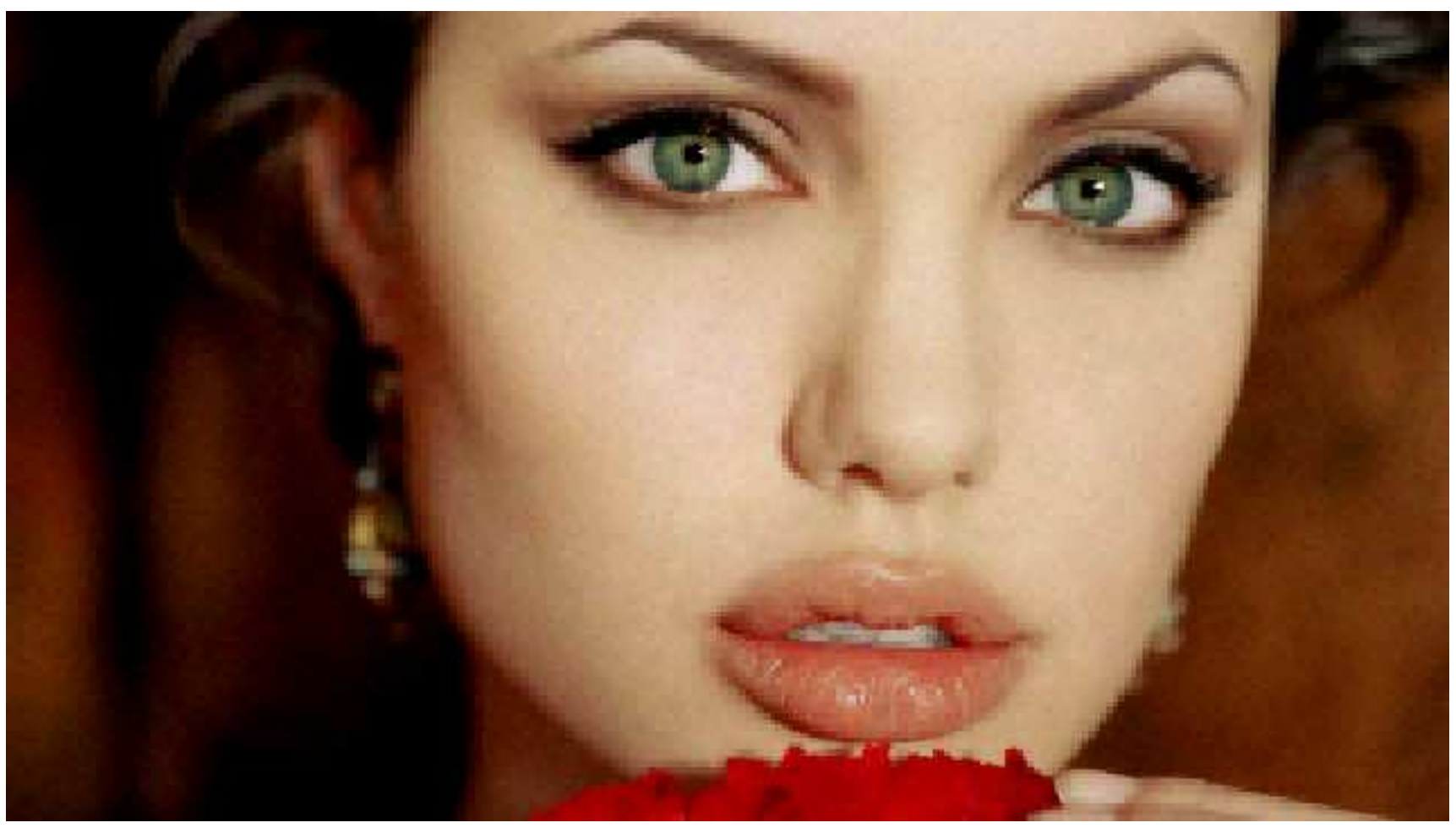

**Fig.76f** Angelina's denoised image via Haar/HT.

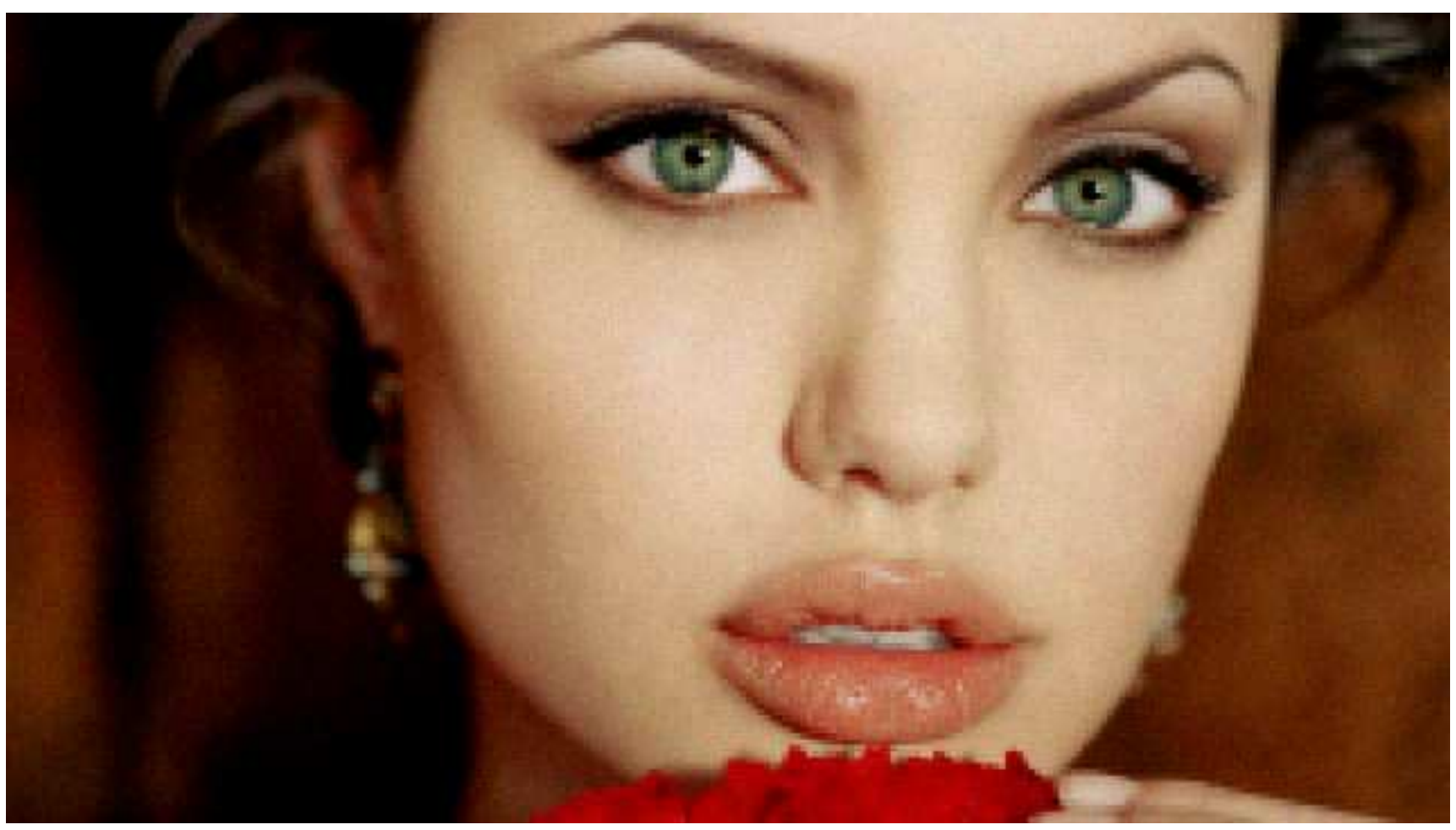

**Fig.76g** Angelina's denoised image via Haar/MF.

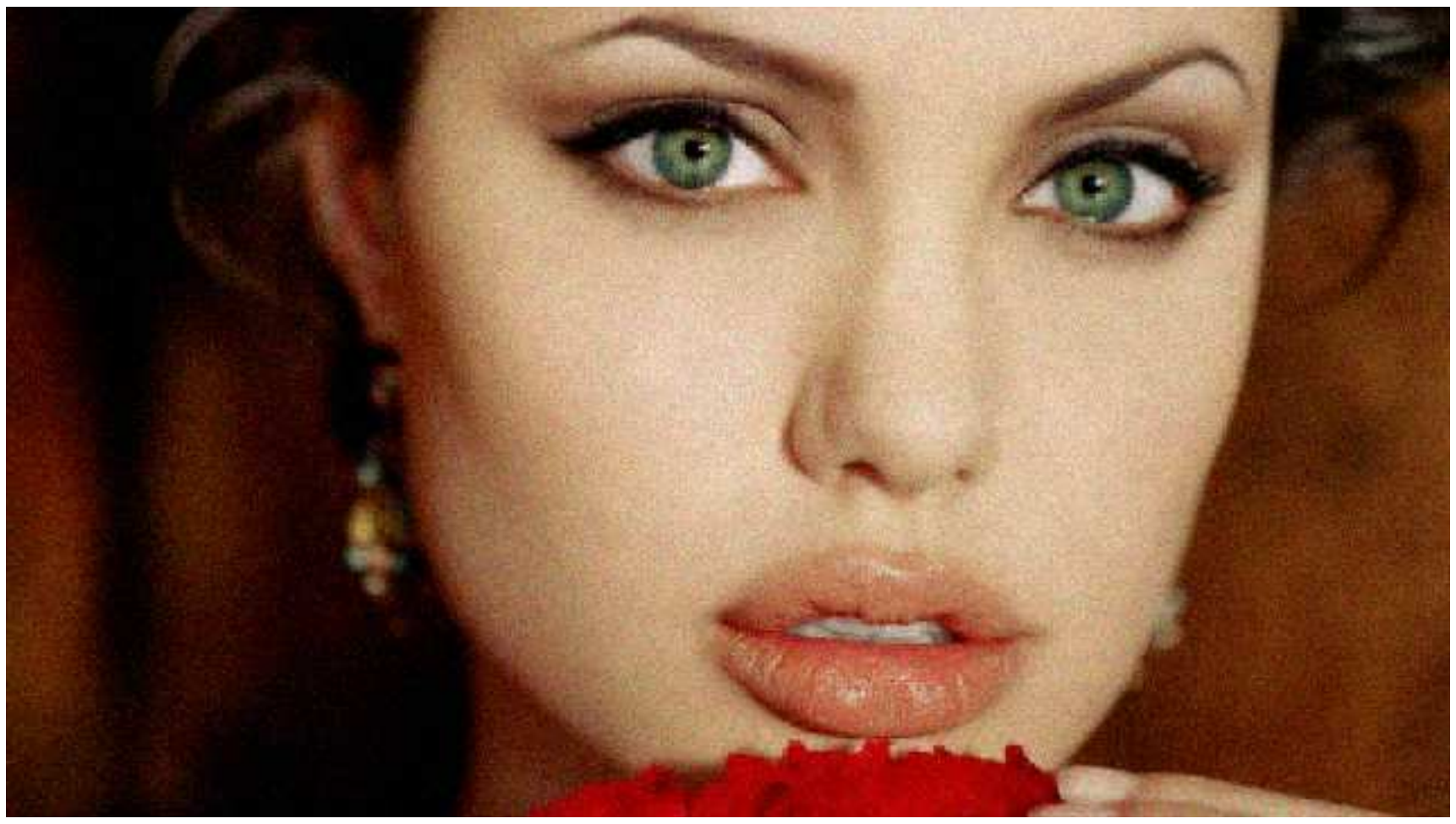

**Fig.76h** Angelina's denoised image via Haar/DS.

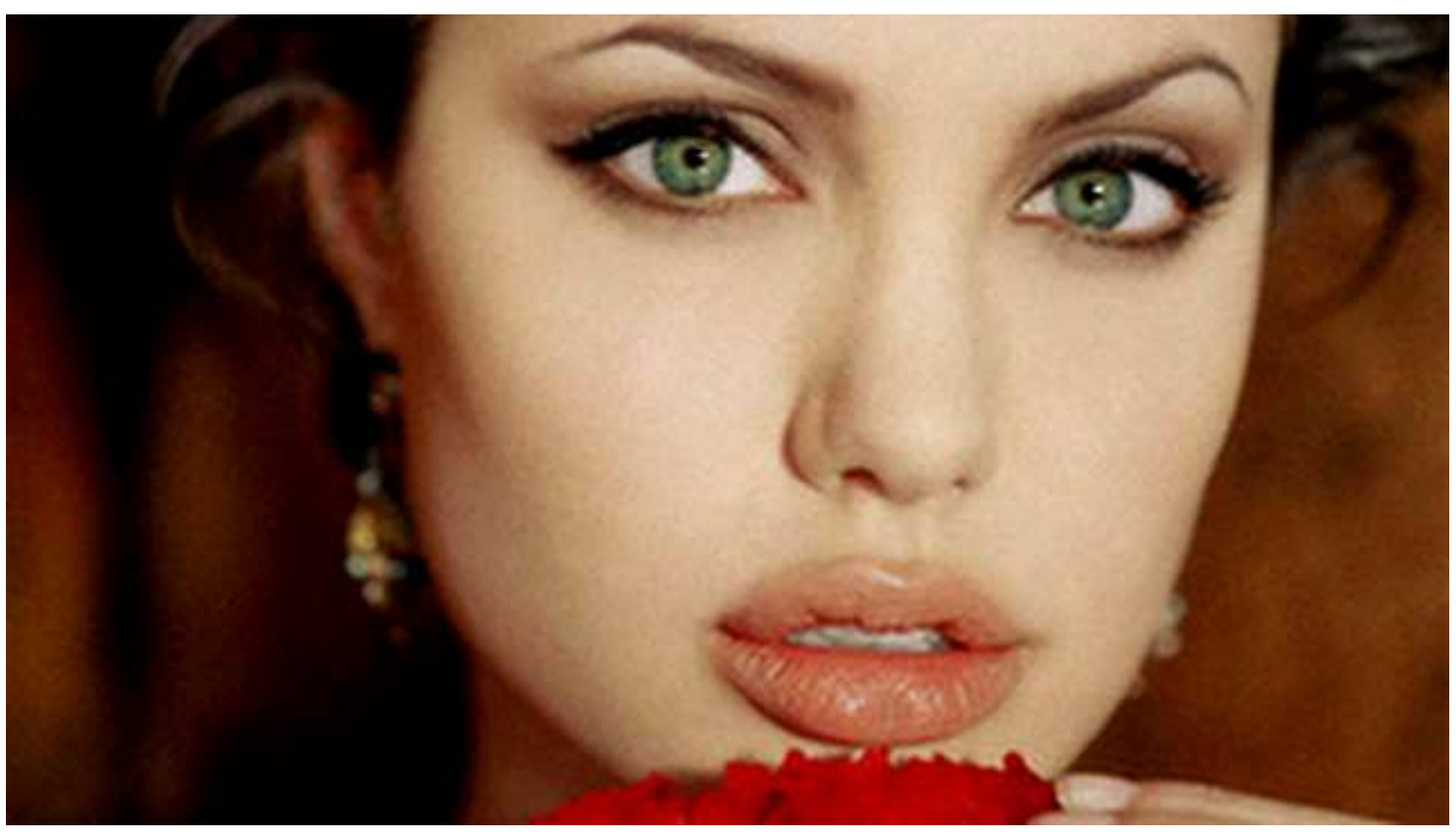

**Fig.76i** Angelina's denoised image via Coslet/ST.

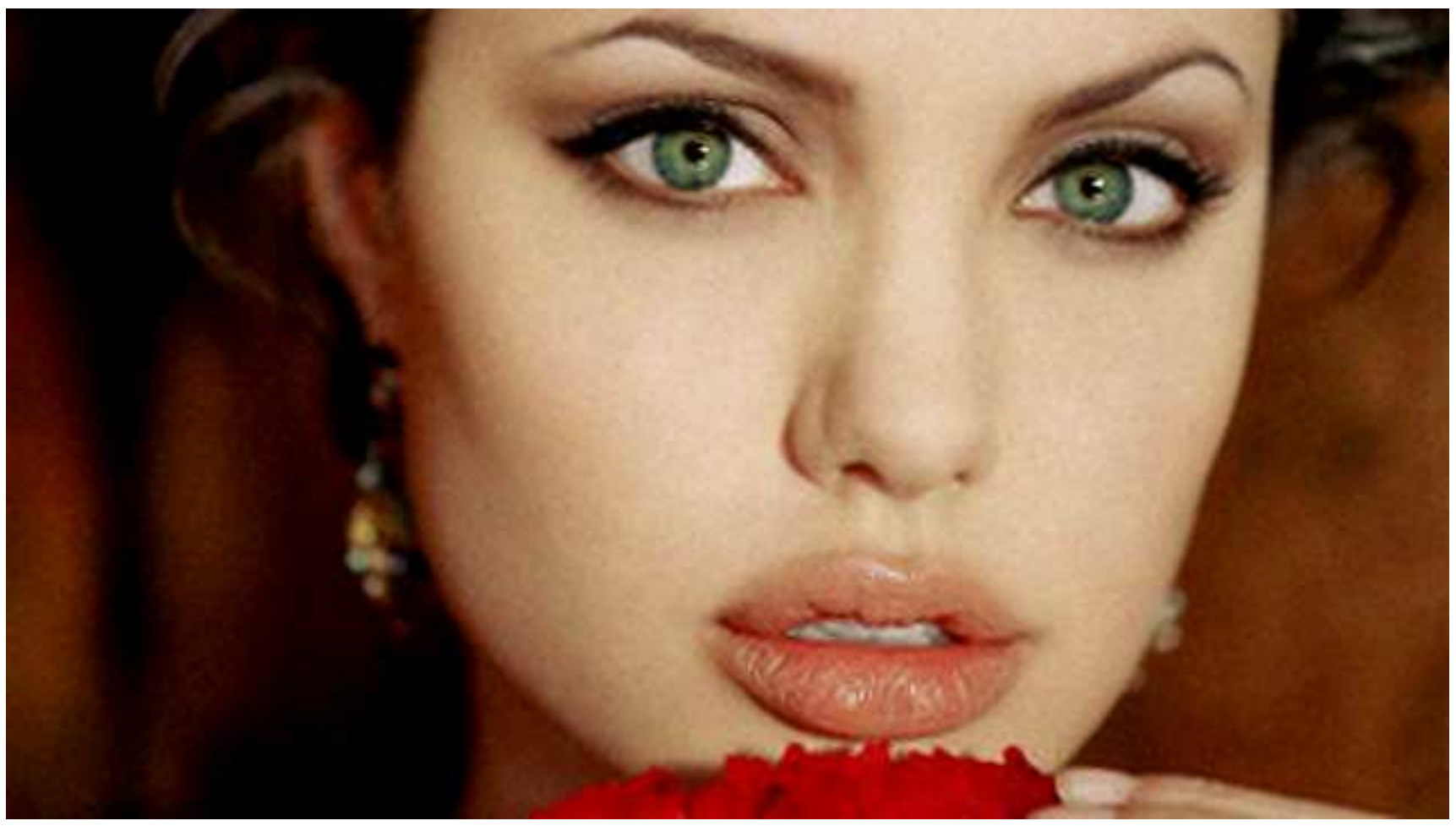

**Fig.76j** Angelina's denoised image via Coslet/HT.

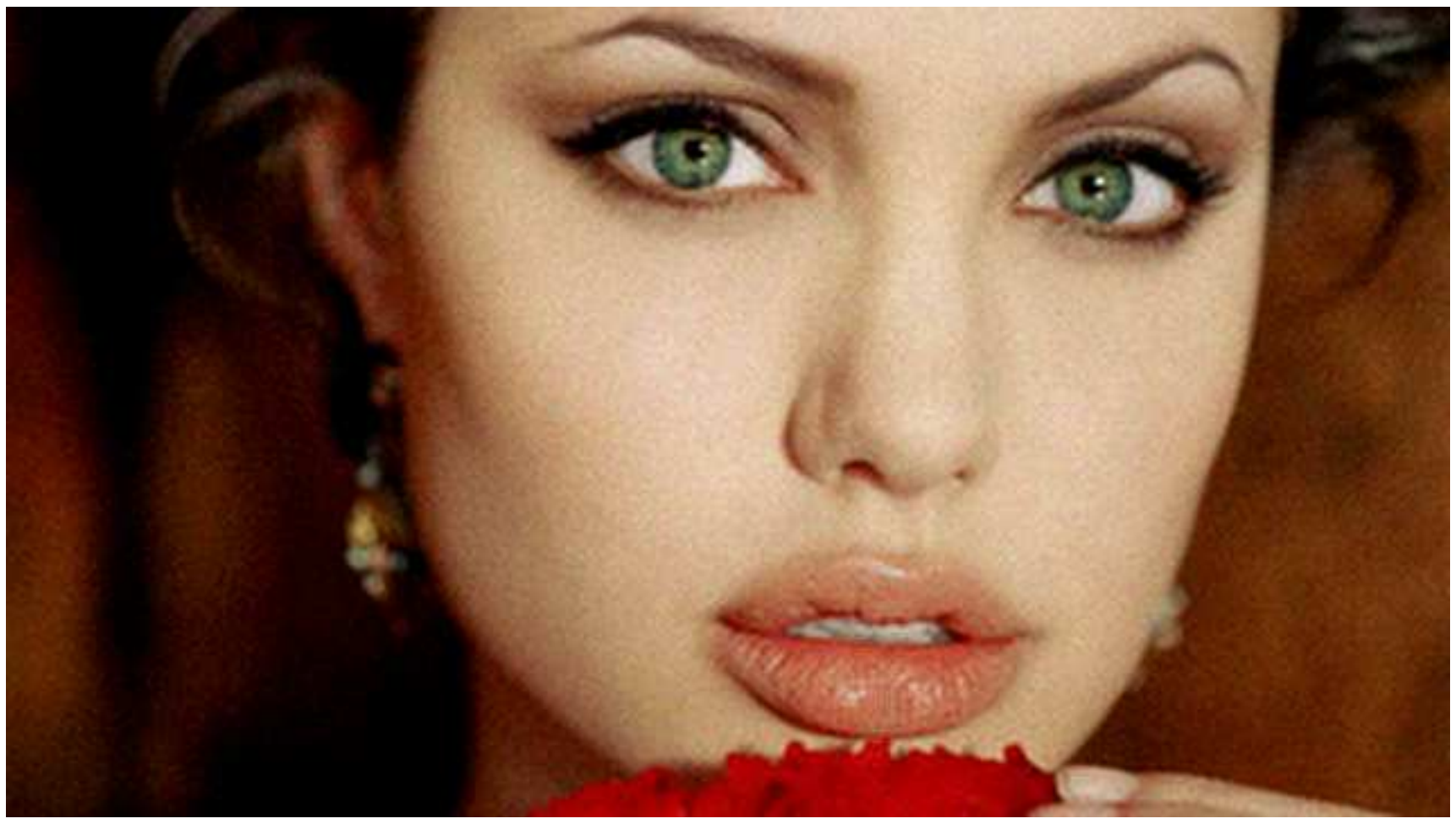

**Fig.76k** Angelina's denoised image via Coslet/MF.

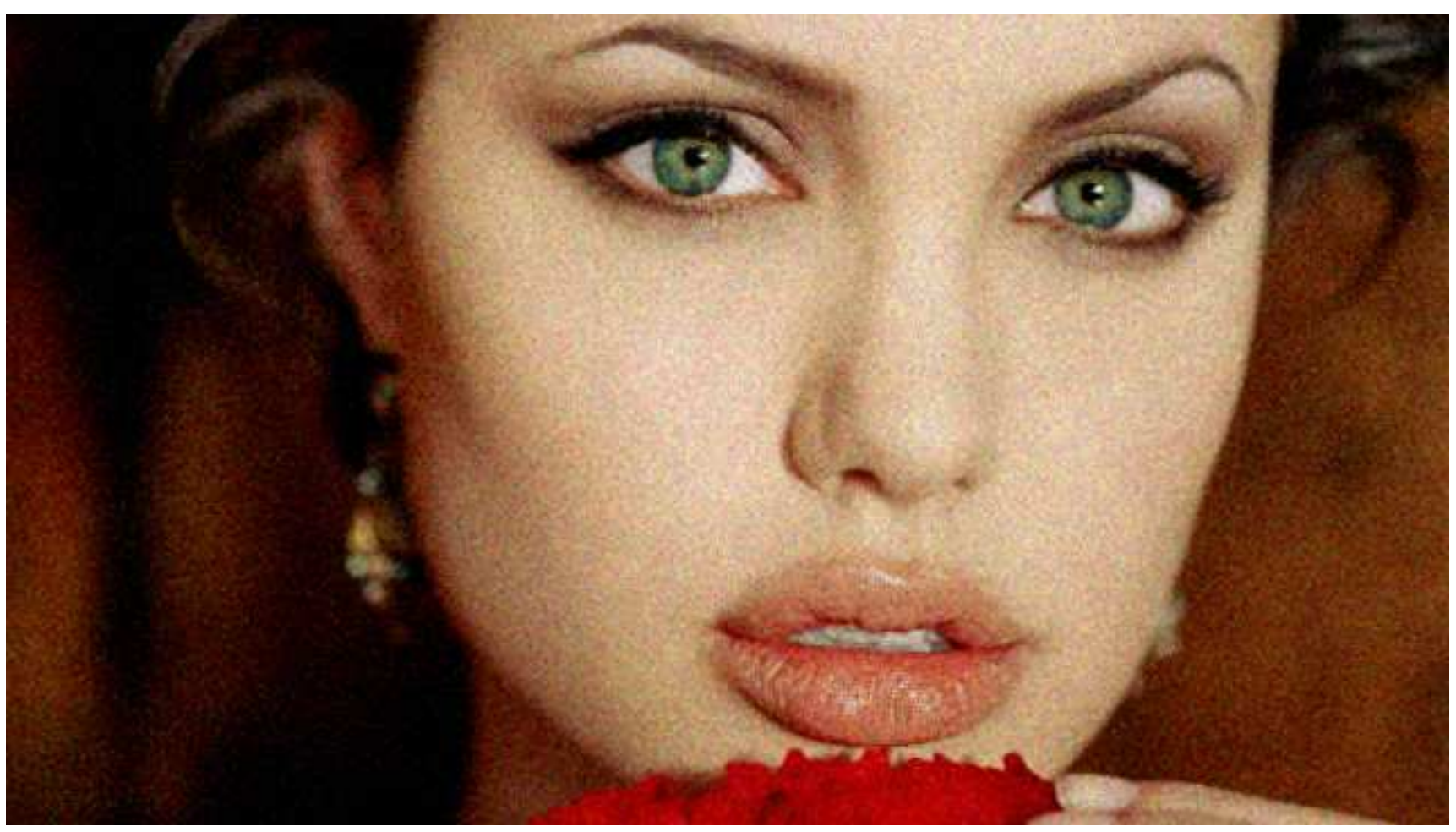

**Fig.76l** Angelina's denoised image via Coslet/DS.

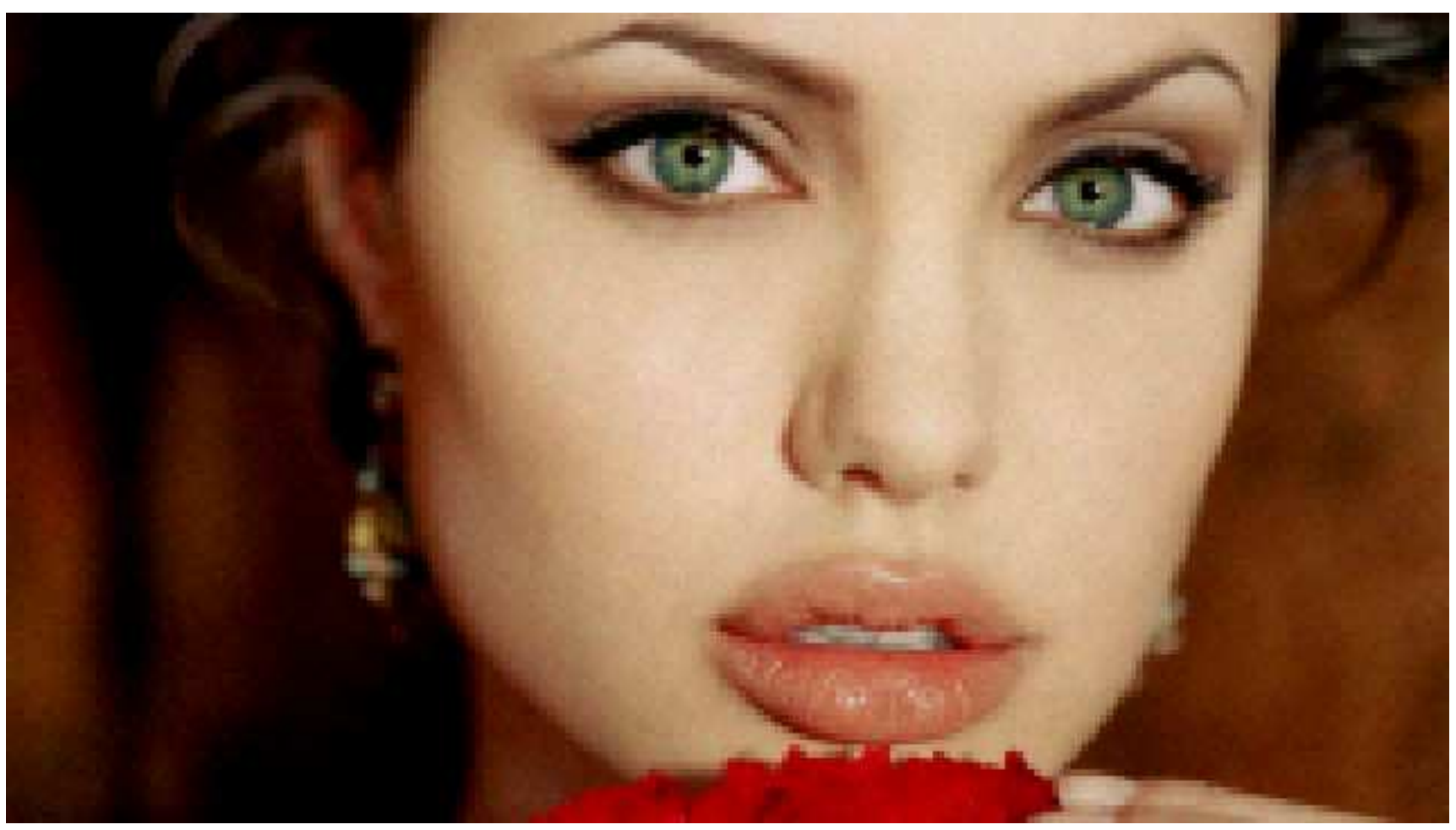

**Fig.76m** Angelina's denoised image via SR-1/Haar.

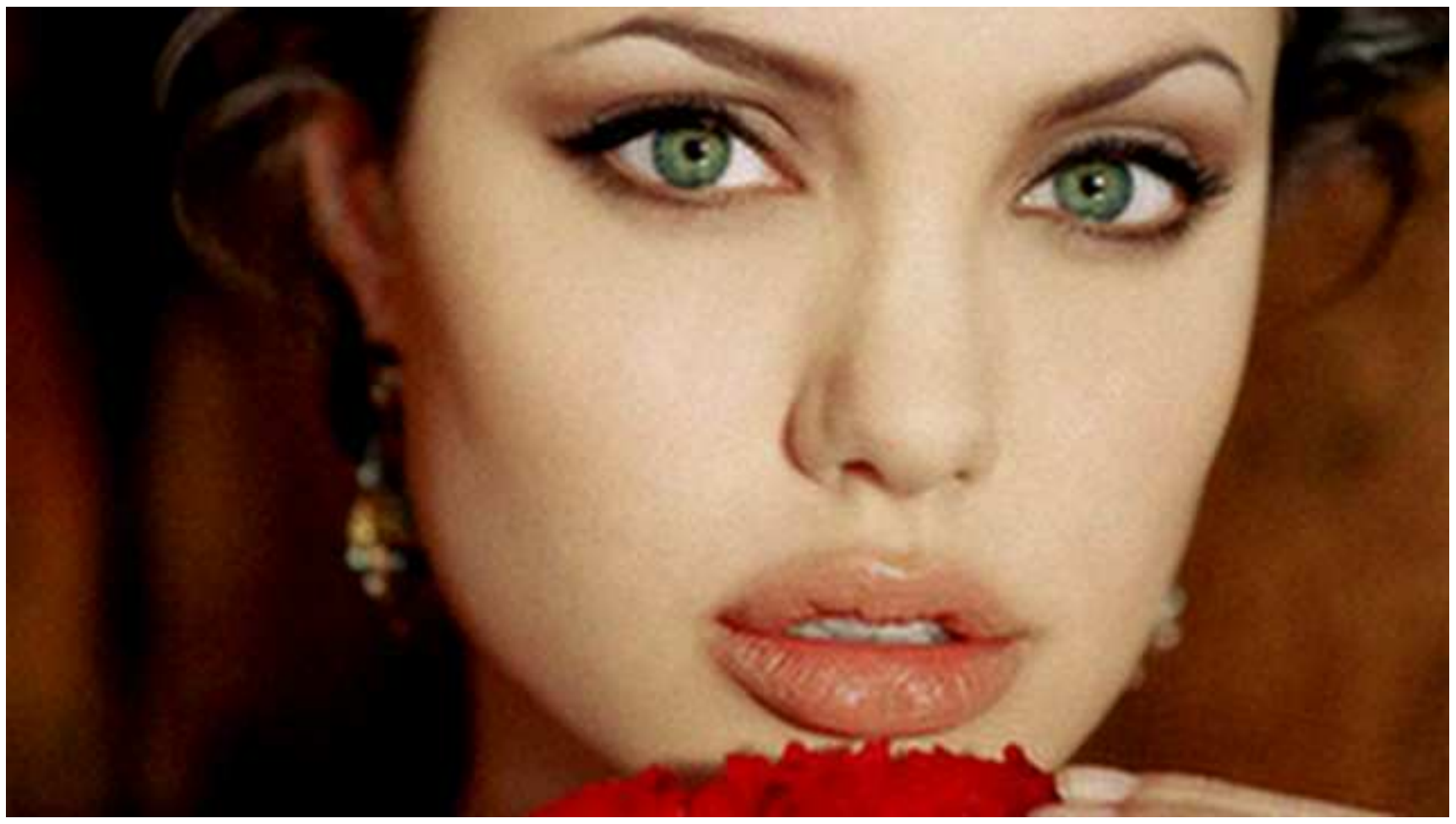

**Fig.76n** Angelina's denoised image via SR-1/Coslet.

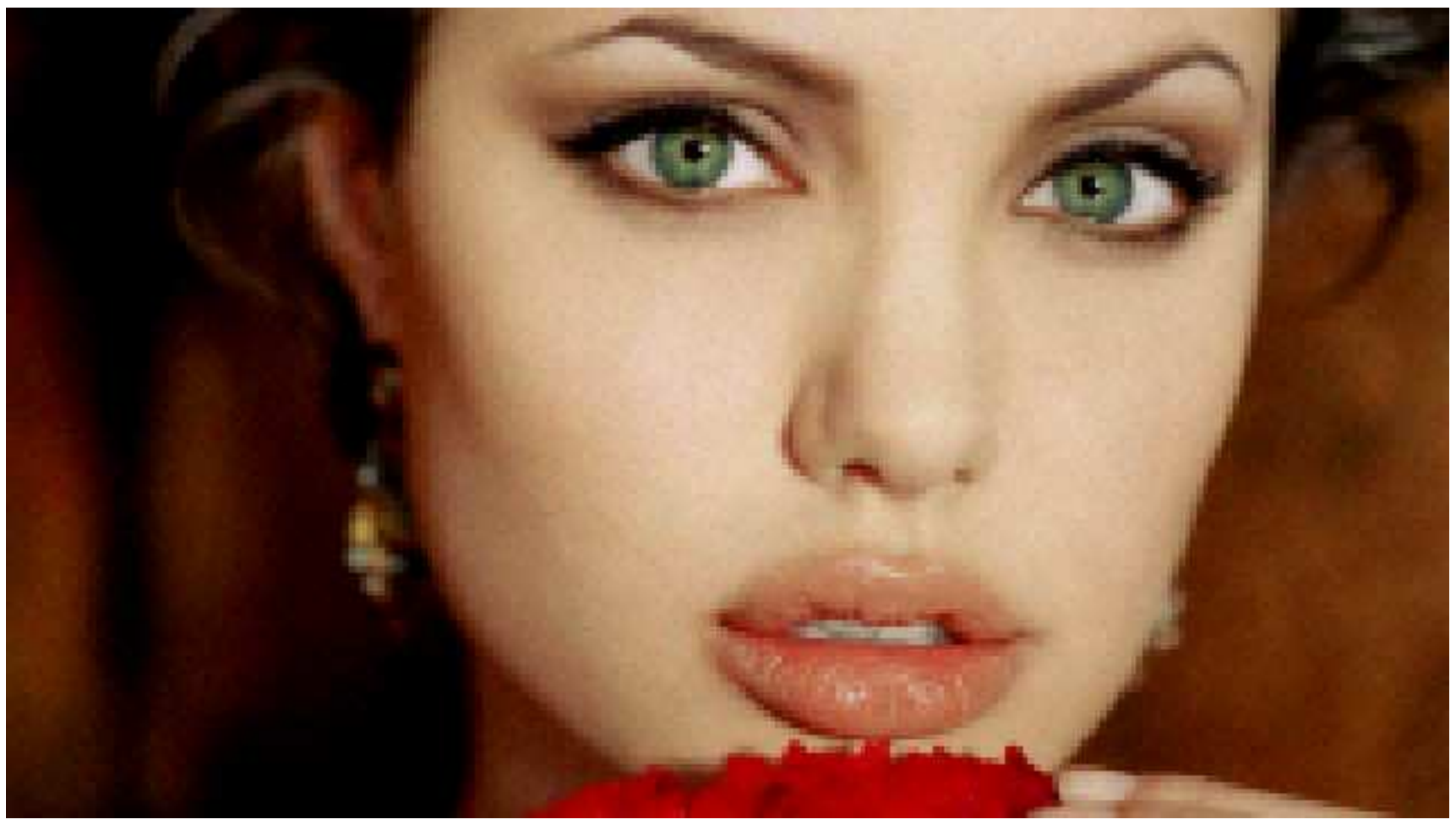

**Fig.76o** Angelina's denoised image via SR-2/Haar.

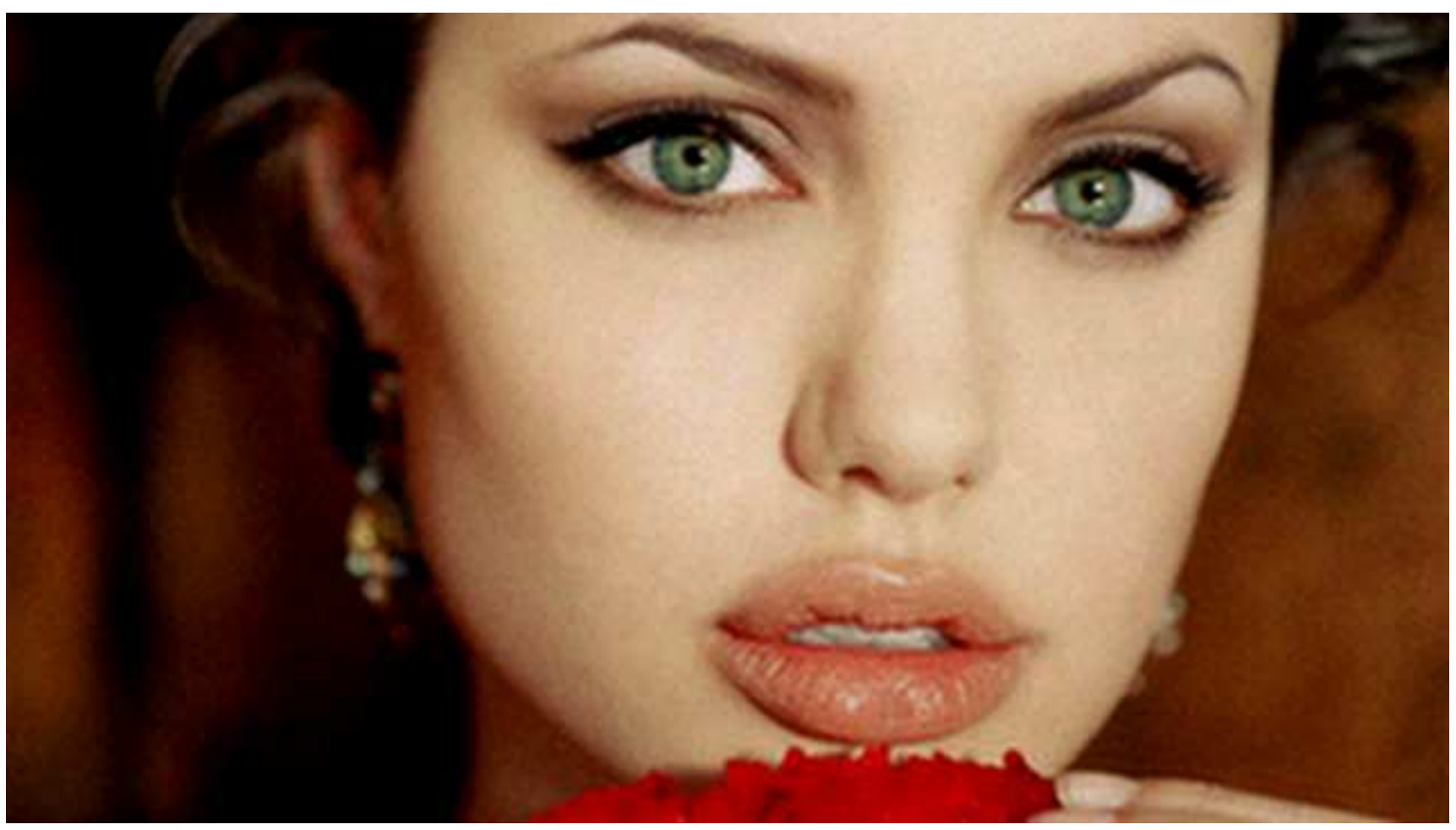

**Fig.76p** Angelina's denoised image via SR-2/Coslet.

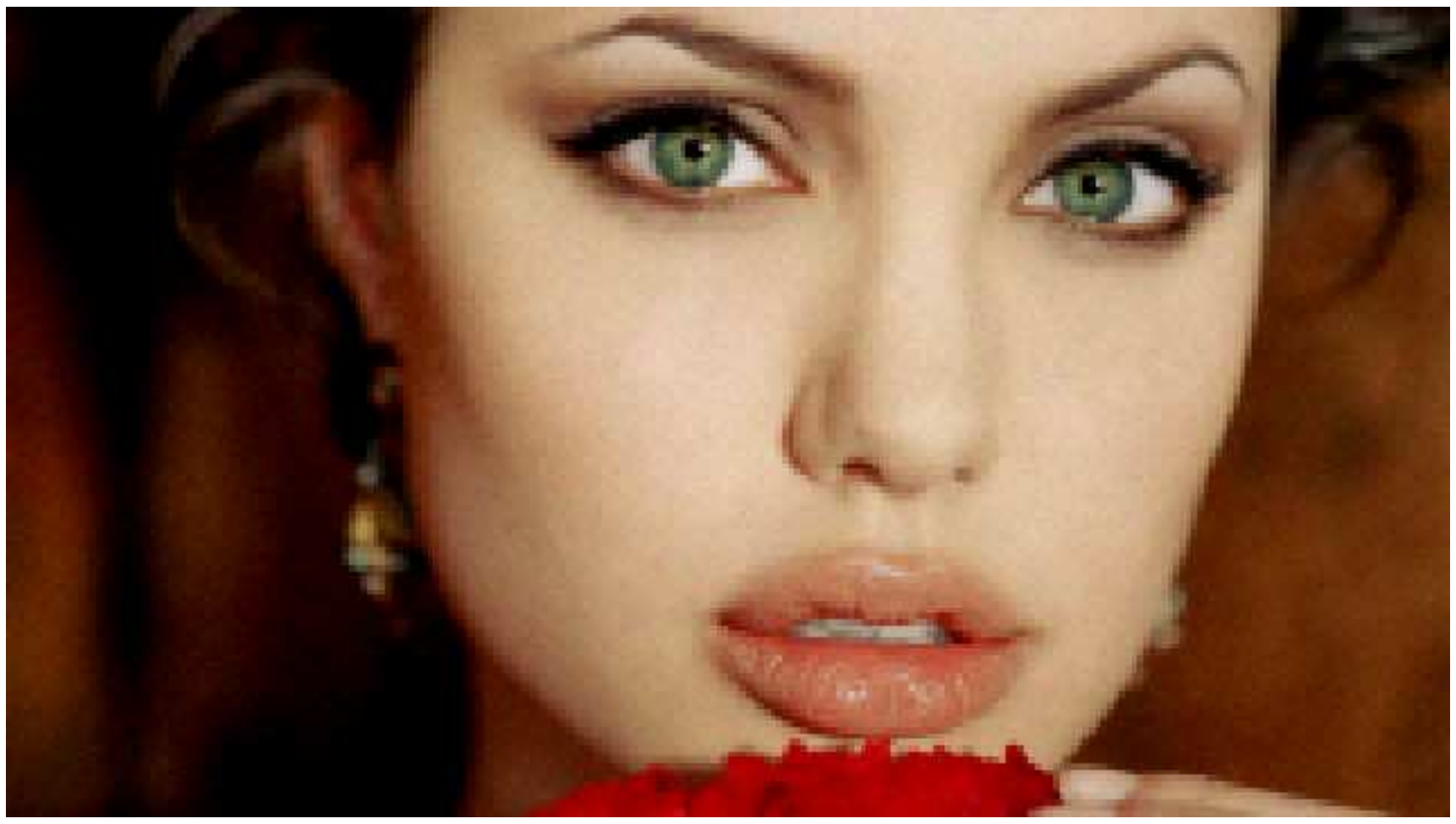

**Fig.76q** Angelina's denoised image via SR-3/Haar.

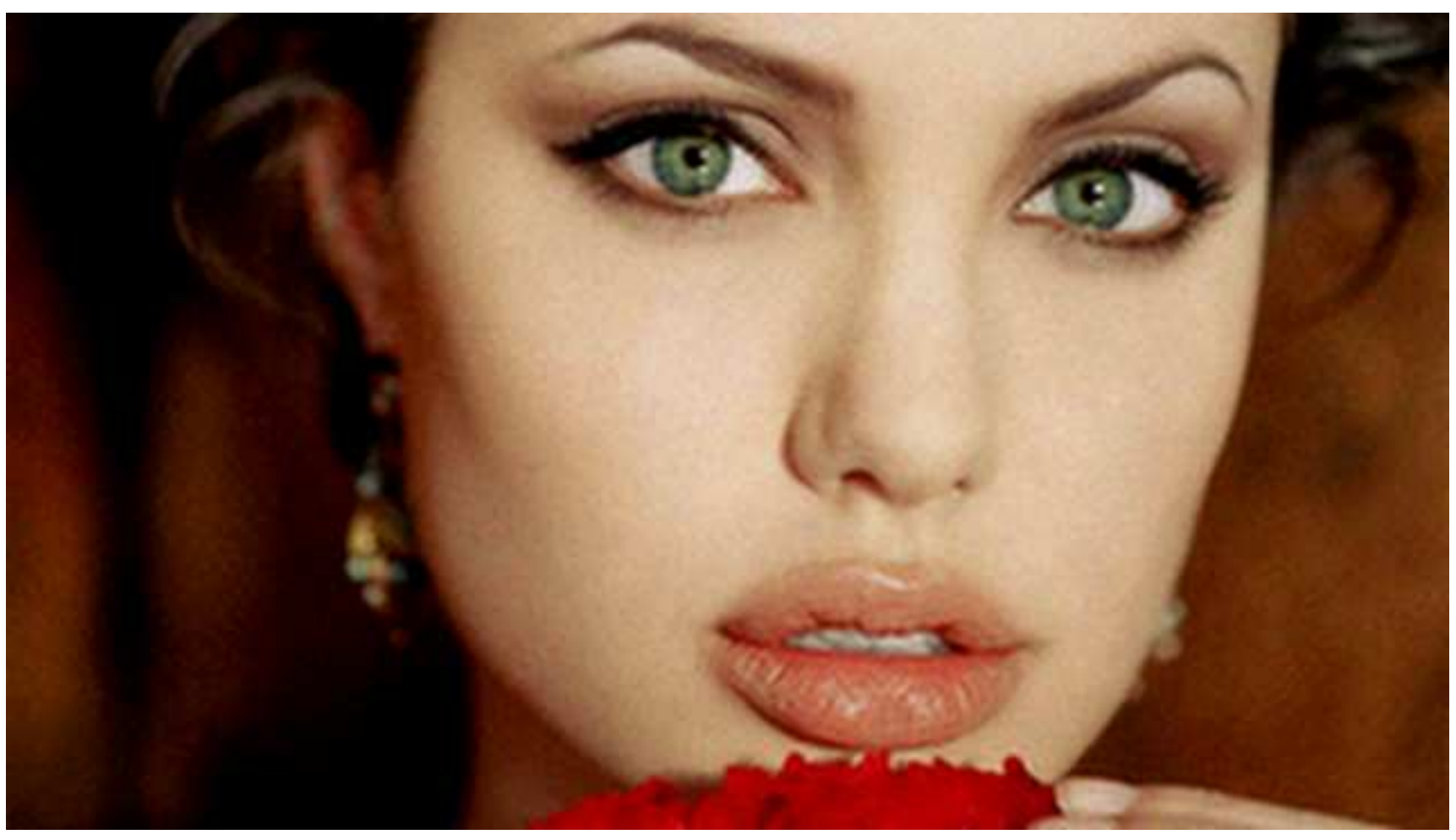

**Fig.76r** Angelina's denoised image via SR-3/Coslet.

### 4.2.2 Compression

In fact, it is an over or super-compression, since this is an additional compression at which for example would do any signal (or image) compression algorithm used originally.

***Signals***
In here, we compare the three techniques of super-compression based in super-resolution of signals.

TABLE VI
COMPRESSION TECHNICS VS METRICS FOR ELECTROCARDIOGRAPHIC SIGNAL.

| Technique | Metrics | | |
|---|---|---|---|
| | MAE | MSE | PSNR |
| SR-1/Haar | 0.2121 | 0.1864 | 47.7612 |
| SR-1/Coslet | 0.0004 | 7.0095 e-07 | 102.0105 |
| SR-2/Haar | 0.2121 | 0.1864 | 47.7612 |
| SR-2/Coslet | 0.0003 | 6.8737e-07 | 102.0955 |
| SR-3/Haar | 0.2121 | 0.1839 | 47.8201 |
| SR-3/Coslet | 0.0073 | 0.0001 | 79.1020 |

Table VI shows us compression techniques vs metrics for an electrocardiographic signal with:
**CR** = 2 (over the compression of any signal compression algorithm)
**PSS** = 50
**MI:** it's the same of Section 4.2.1 for the electrocardiographic signal.

Besides, Table VI shows the absolute superiority of Coslet against Haar in all versions views.

In Fig.77 we can see compressed electrocardiographic signal for all versions. At first glance, they all look the same. However, Table VI shows a marked predominance of versions 1 and 2 with Coslets.

Simulations with audio, i.e., conversations, music and even the sound of gravitational waves have been conducted with identical results, too. However, the results in Table VI clearly show that electrocardiographic signals for versions 1 and 2 Coslets behaves as lossless.

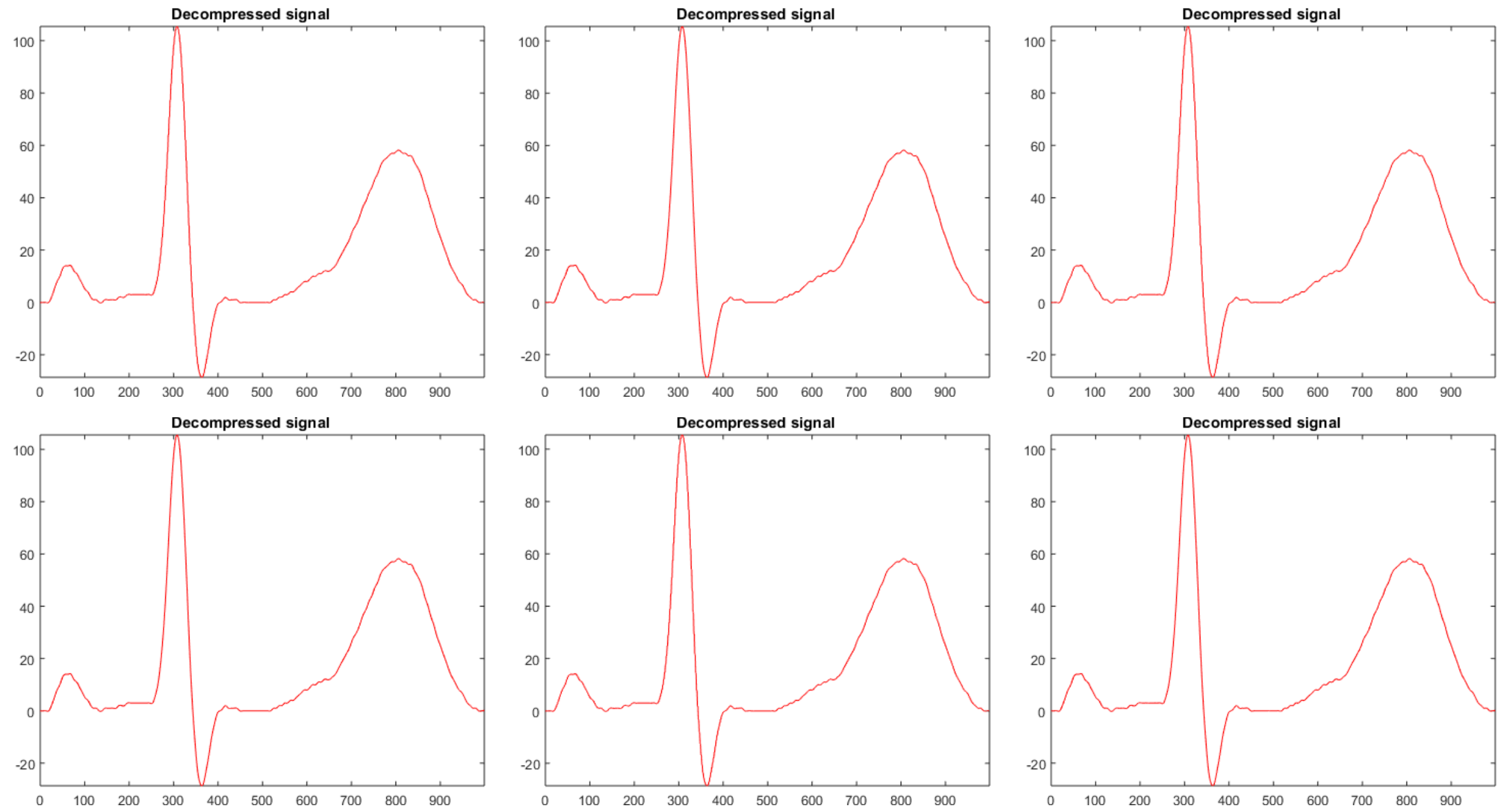

**Fig.77** Compressed electrocardiographic signal via: (1,1) SR-1/Haar, (1,2) SR-1/Coslet, (1,3) SR-2/Haar, (2,1) SR-2/Coslet, (2,2) SR-3/Haar, (2,3) SR-3/Coslet.

***Images***

The same techniques of the previous case are applied here to the two-dimensional case on Angelina.

Here too, Table VII shows a superiority of Coslet against Haar in all versions views, however, this superiority is not as overwhelming.

In Fig.78 we can see compressed image of Angelina for all versions. At first glance, they all look the same. However, Table VII shows a marked predominance of versions 1 and 2 with Coslets.

Simulations with medical and satellite images have been conducted with identical results, too. However, the results in Table VII clearly does not indicate a behavior of lossless type in any version.

TABLE VII

COMPRESSION TECHNIQUES VS METRICS FOR ANGELINA.

| Technique | Metrics | | |
|---|---|---|---|
| | MAE | MSE | PSNR |
| SR-1/Haar | 1.4306 | 6.3992 | 40.0695 |
| SR-1/Coslet | 1.0777 | 2.9354 | 43.4540 |
| SR-2/Haar | 1.4288 | 6.3973 | 40.0707 |
| SR-2/Coslet | 1.0774 | 2.9343 | 43.4556 |
| SR-3/Haar | 1.4283 | 6.3986 | 40.0699 |
| SR-3/Coslet | 1.0774 | 2.9342 | 43.4558 |

Table VII shows us compression techniques vs metrics for Angelina with:

**CR** = 4 (over the compression of any image compression algorithm, e.g., JPEG [227])

**PSS** = 75

**MI:** it's the same of Section 4.2.1 for Angelina.

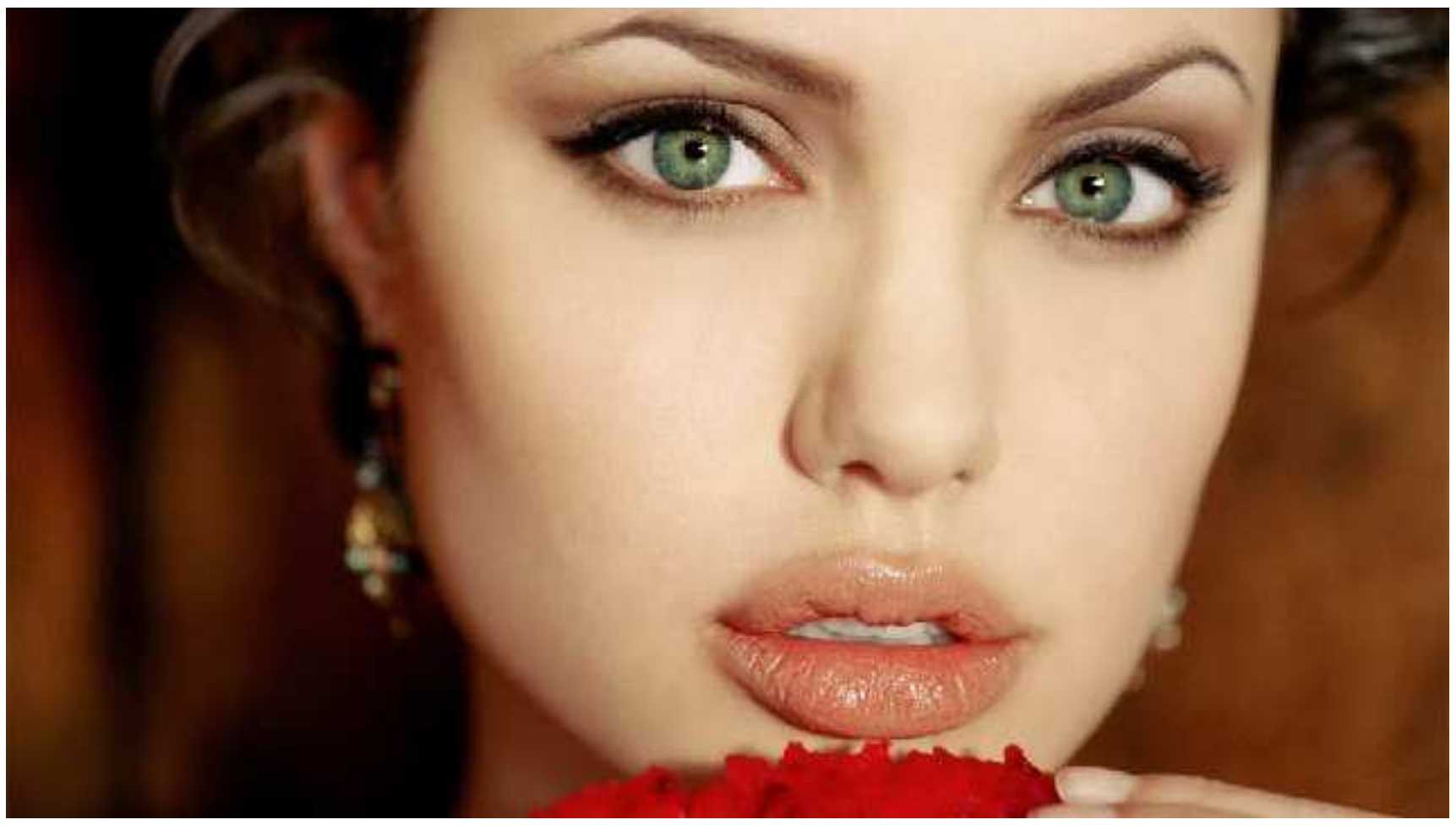

**Fig.78a** Angelina's compressed image via SR-1/Haar.

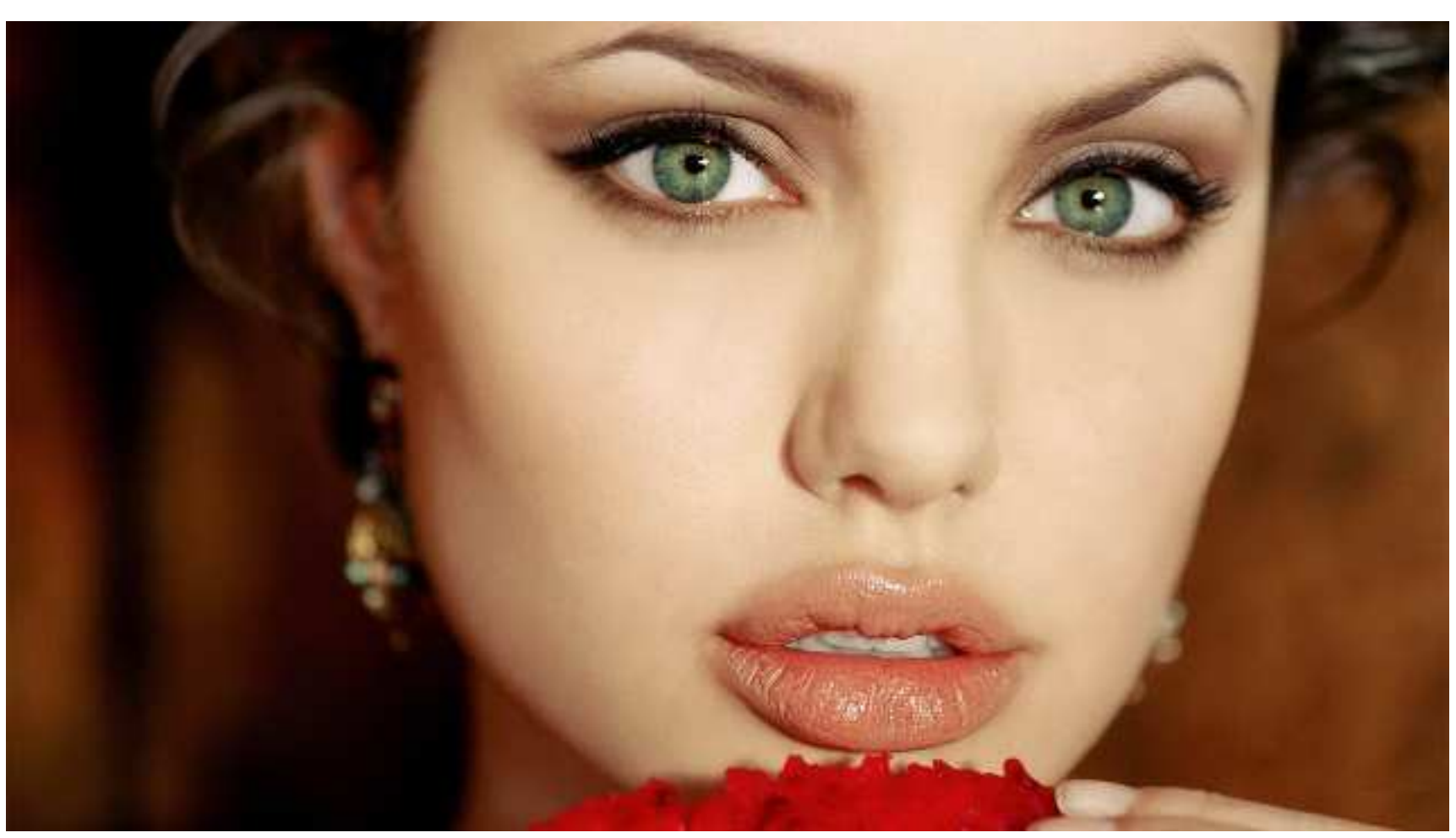

**Fig.78b** Angelina's compressed image via SR-1/Coslet.

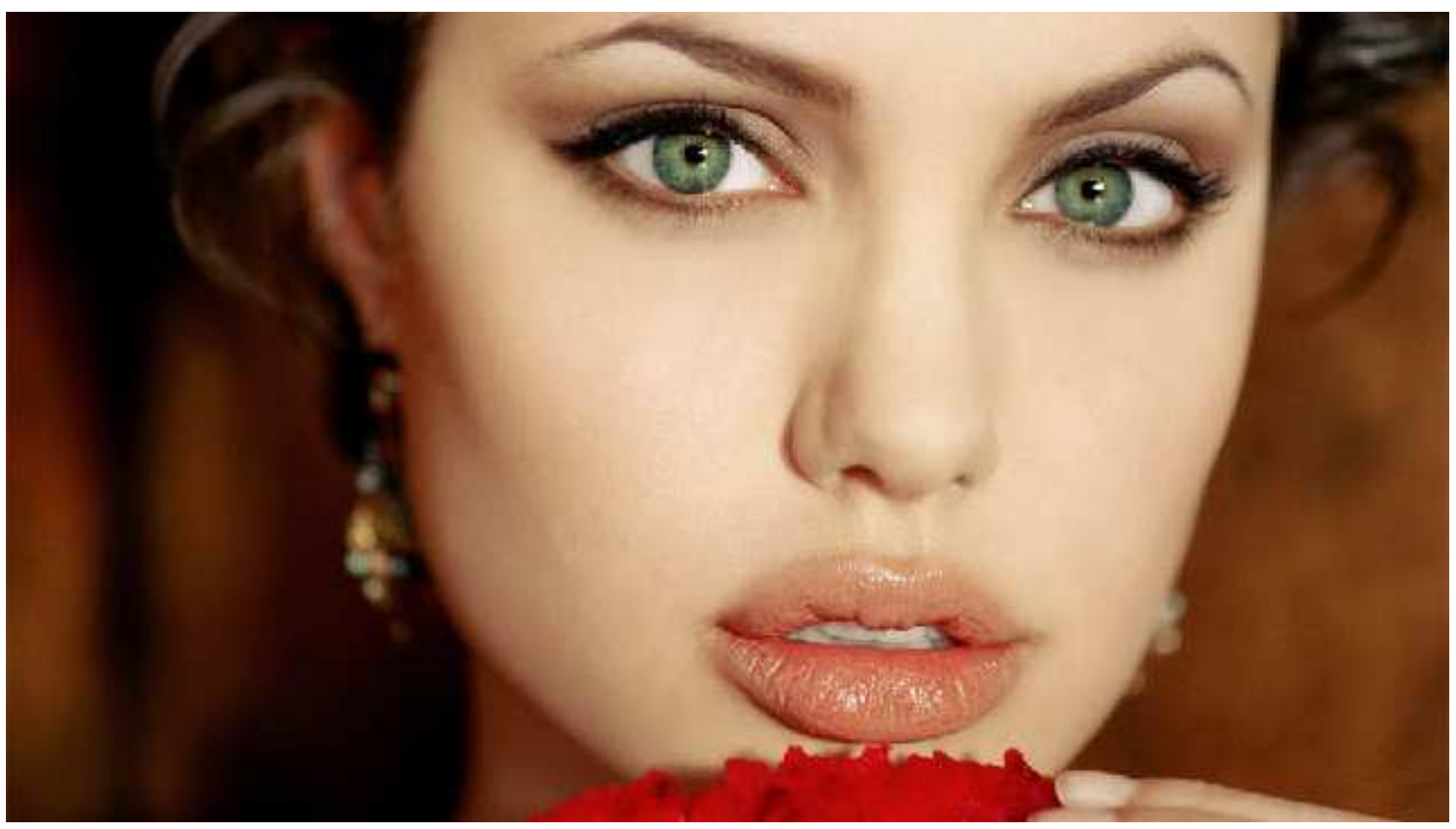

**Fig.78c** Angelina's compressed image via SR-2/Haar.

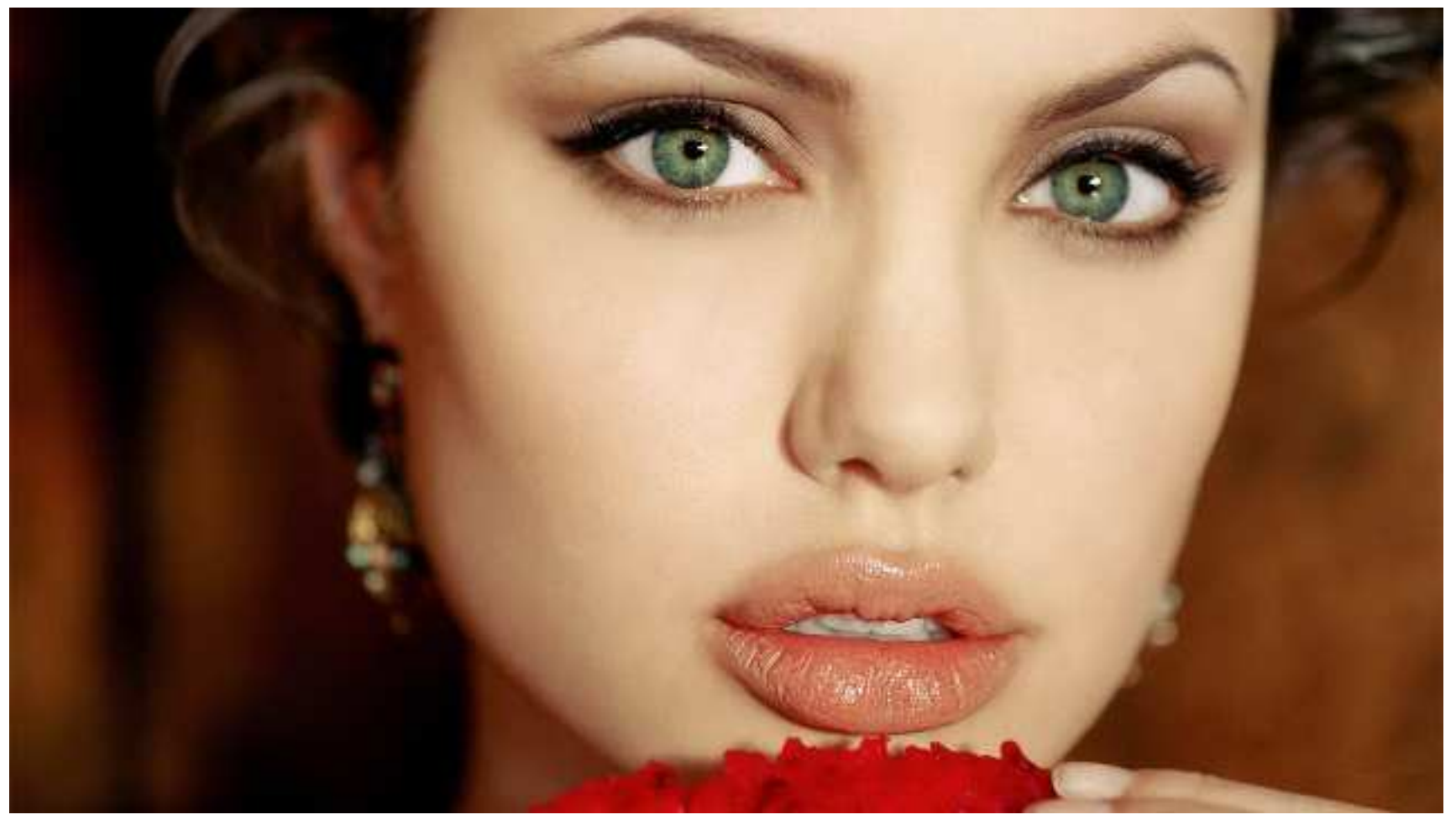

**Fig.78d** Angelina's compressed image via SR-2/Coslet.

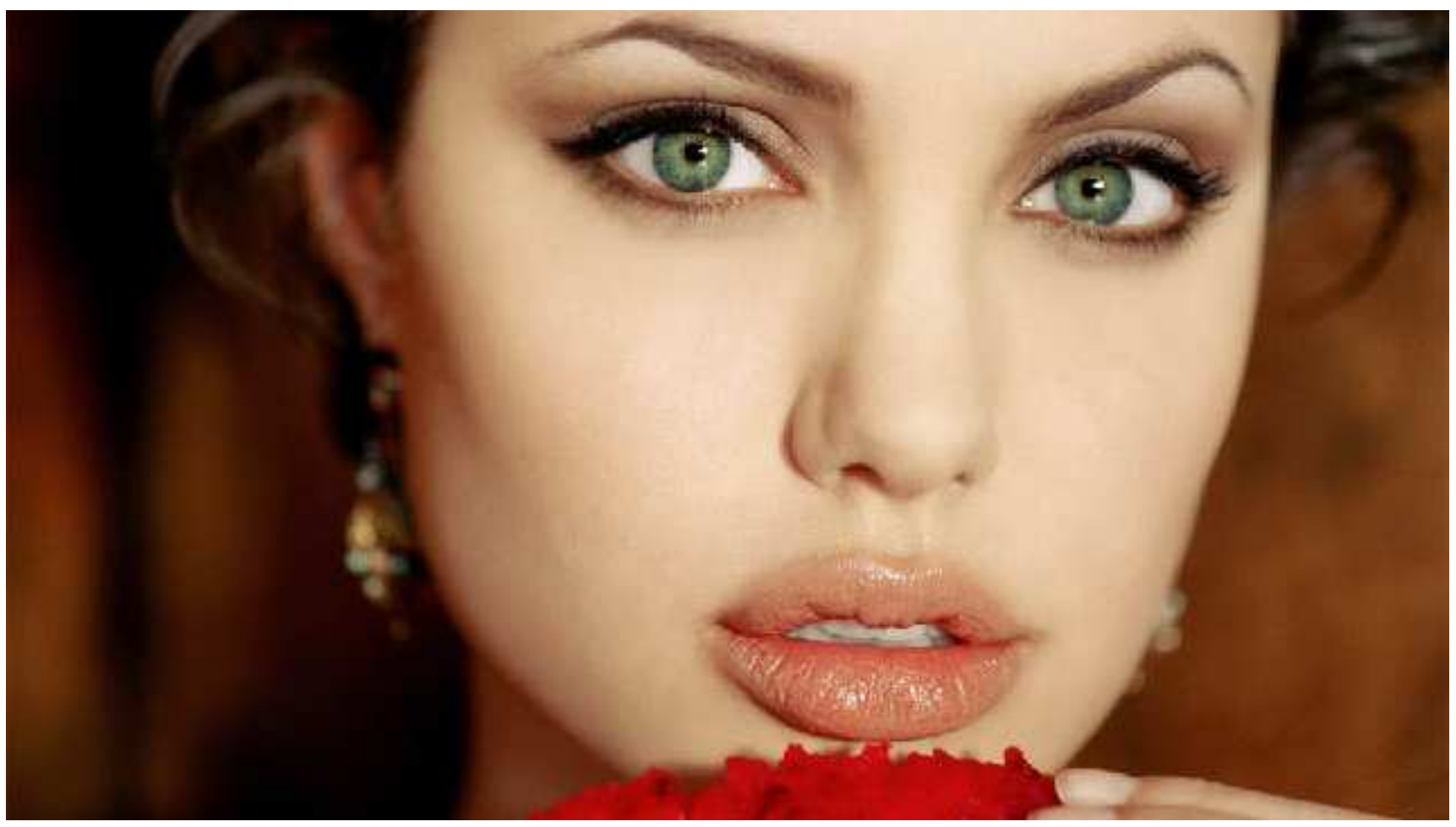

**Fig.78e** Angelina's compressed image via SR-3/Haar.

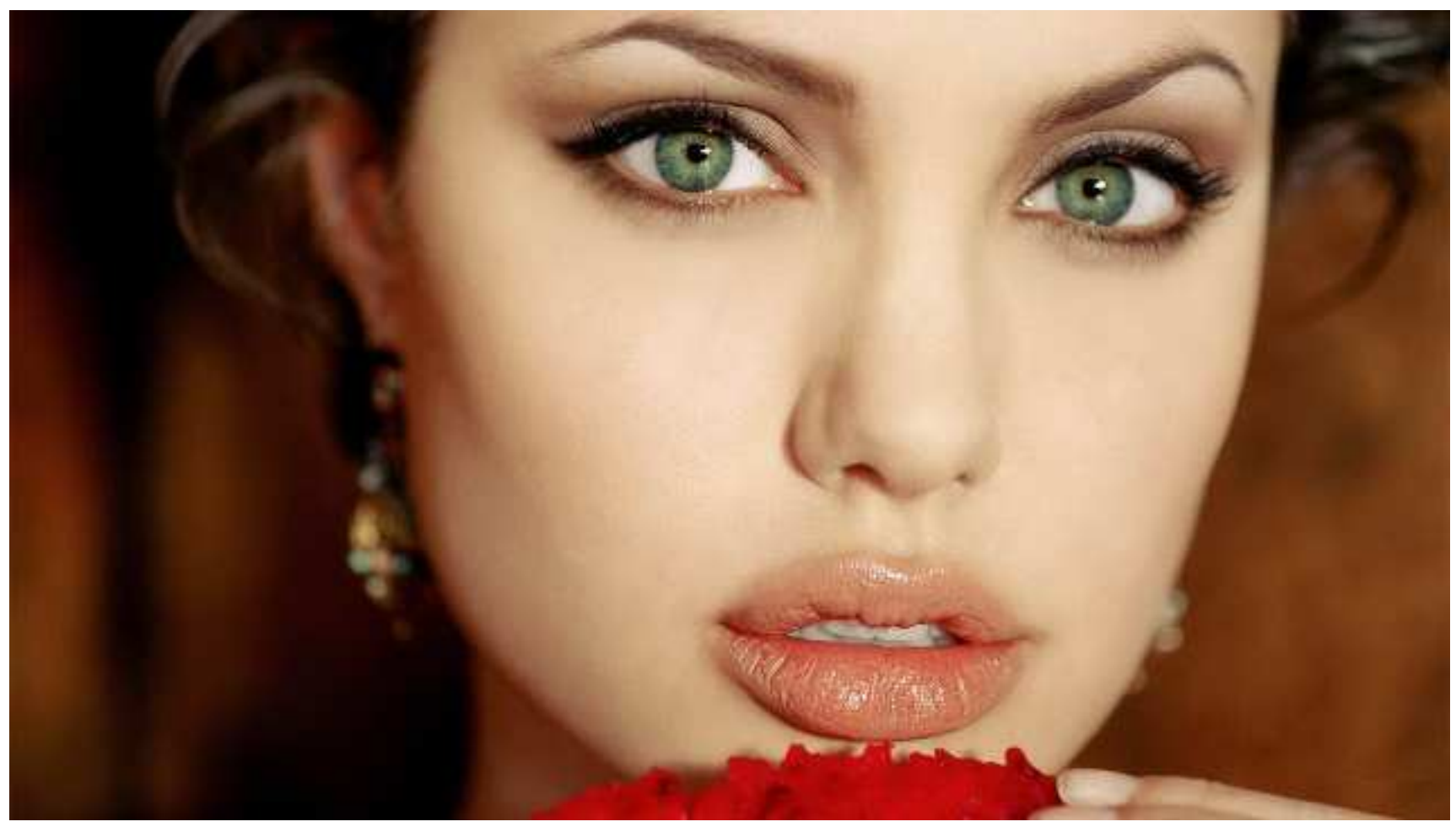

**Fig.78f** Angelina's compressed image via SR-3/Coslet.

#### 4.2.3 Super-resolution

This technique is the decoder of the compression/decompression process of the previous section, both signals to images.

***Signals***
**MI:** it's the same of Section 4.2.1 for the electrocardiographic signal.

***Images***
**MI:** it's the same of Section 4.2.1 for Angelina.

#### 4.2.4 Edge detection

This technique was developed and simulated during this work.

***Signals***
In this case, this technique, i.e. FIT, it is known as flank detection, and it was seen in Section 3.3.1. There is no equivalent to FIT in the case of signals to detect edges as in the case of images based on Roberts, Prewitt, Sobel and Canny.

***Images***
Techniques of edge detection base don Roberts, Prewitt, Canny and Sobel was seen in Section 2.5, see Fig.39. On the other hand, attributes for edge detection of FIT were seen in Sections 3.3.2 (Figures 49 to 52), and 4.1.4 (Fig.73).

#### 4.2.5 Hardware and Software employed

Finally, FIT was implemented in MATLAB® R2015a (Mathworks, Natick, MA) [49] on a notebook with Intel® Core(TM) i7-4702MQ CPU @ 2.20 GHz x 2 and 8 GB RAM on Microsoft® Windows 7© Ultimate 64 bits. Besides, since all algorithms are parallelizable, distributed implementations were done on a GPU cluster, NVIDIA® Tesla© 2050 GPU [211] with a peak performance of approximately 500 GFLOPS, with an achieved performance of approximately 250 GFLOPS in OpenCL. The GPU needed approximately 2.5 GB of bandwidth with InfiniBand connectivity at quad data rate (QDR) QLogic® [262] or 40 Gb speeds.

## 5 Conclusions and future works

This work began with an extensive tour on traditional spectral techniques based on Fourier's Theory, without compact support and completely disconnected from the link between time and frequency, and the responsibility of each flank with respect to final spectral components of a signal, image or video. For that reason QSA/FIT was created, i.e., to cover such space and also as a complement to the aforementioned Fourier's Theory, in particular, FFT. A simply comparison between QSA/FIT and FFT sheds some initial conclusions, which can be seen in Table VIII.

Moreover, when the wave function collapses, we pass from QSA to FIT. This point is essential, because of this begins to be necessary to use the Hadamard's quotient of vectors [80].

Other contributions of this work are:
1) One-dimension directional smoothing
2) One-dimension median filter
3) Coslet (which turned out to be far superior to Haar)
4) Super-resolution of signal and still images
5) Super-compression of signal and images (in fact. three versions)
6) Flank and edge detection via FIT. Etc.

TABLE VIII
COMPARISON BETWEEN FFT AND FIT

| **Characteristics** | **FFT** | **FIT** |
|---|---|---|
| Separability | Yes | Yes |
| Compact support | No | Yes |
| Instantaneous spectral attributes | No | Yes |
| 1D computational cost | $O(N*\log_2(N))$ | $O(N)$ |
| 2D computational cost | $O(N^2*\log_2(N)^2)$ | $O(N^2)$ |
| Energy treatment | Disastrous | Excellent |
| Decimation | In time or frequency | It does not require |
| Parallelization | No | Yes |

Regarding one-dimensional mask mentioned before, we can say:

a) a large mask loses detail, i.e., it loses the texture of signal, it is of low frequency or approximation
b) a small mask emphasizes the details, i.e., highlights the texture signal is high frequency or detail

Otherwise, an interesting aspect is the following, there is a great affinity between the notion of contrast [255] and FIT , in fact, many different equations for contrast exist, e.g.:

$$Contrast = \frac{Change\ in\ Luminance}{Average\ Luminance} \tag{279}$$

The Michalson's equation for contrast is:

$$C_M(I) = \frac{max(I) - min(I)}{max(I) - min(I)} \tag{280}$$

although:
1) These equations work well for simple images with 2 luminances (i.e. uniform foreground and background)
2) Does not work well for complex scenes with many luminances or if min and max intensities are small.

In fact, FIT is a kind of contrast for each considered sample or pixel. This is fundamental, since a non-contrast image is an image without intelligibility.

Regarding simulations:

a) Histograms were used in the calculation of *mutual information*, for this reason, they were not used explicitly.
b) No signal or image used in the simulations has an unstable behavior, which the phase diagrams are not simulated, they only mentioned it in character complementary to the techniques seen in this work.

Finally, the applications of FIT are obvious, e.g.:

- It's a support and it allows a better understanding of the Information Theory and Quantum Information Theory aimed at improving current signal, image a video compression algorithms, and develop new.
- Its applications range from filter design and signal analysis to phase retrieval and pattern recognition.
- It's an excellent complement to Spectrogram in speech processing
- It's very useful in radar signals analysis, analysis of phase migration in Synthetic Aperture Radar (SAR) raw data, Radioastronomy, sonar, etc.
- It's particularly useful in analysis of time-varying spectral characteristics
- It represents a major contribution in Signals Intelligence (SIGINT), Imagery Intelligence (IMINT), and Communications Intelligence (COMINT)
- It retains a direct relationship with compressed sensing

- Besides, its applications are obvious in a fine processing signal, namely: power spectral density (with a strict sense of time); frequency-hopping spread spectrum; analysis of stationarity; time series analysis; study of seismic signals, in general, and, earthquakes, in particular; nonlinear spectral analysis; nonlinear sampling for a most efficient compression schema instead of linear sampling; biomedical signal and image analysis: electrocardiograph, electroencephalography, evoked potential, brain computer interface; analysis, synthesis, and speech recognition, Bioinformatics: Signal Processing for DNA Sequence Analysis; conditioning of acoustic spaces; and Quantum Chaos.

Finally, as we have already said, it is an extraordinary tool to assess the importance of the flanks (or edges) in a compression process weighting in real time and sample by sample (or pixel by pixel) the importance of temporal spectral components in the final result.